\documentstyle[12pt,times,epsfig,twoside]{report}

\makeatletter
\let\@internalcite\cite
\def\cite{\def\@commapen{-1000}\def\citename##1{##1}\@internalcite}
\def\shortcite{\def\@commapen{1000}\def\citename##1{}\@internalcite}
\def\@biblabel#1{\def\citename##1{##1}[#1]\hfill}
\def\@citex[#1]#2{\if@filesw\immediate\write\@auxout{\string\citation{#2}}\fi
  \def\@citea{}\@cite{\@for\@citeb:=#2\do
    {\@citea\def\@citea{;\penalty\@commapen\ }\@ifundefined
       {b@\@citeb}{{\bf ?}\@warning
       {Citation `\@citeb' on page \thepage \space undefined}}%
{\csname b@\@citeb\endcsname}}}{#1}}
\def\footnoterule{\kern 2pt
\kern-3\p@\hrule width .4\columnwidth\kern 2.6\p@}
\newlength{\chaptitlewidth}
\newlength{\chapnowidth}
\newcommand{\BibTeX}{{B\kern-.0833em{\sc ib}\kern-.1667em%
T\kern-.1667em\lower.5ex\hbox{E}\kern-.125emX}}
\newcommand{\SliTeX}{{\rm S\kern-.06em{\sc l\kern-.035emi}\kern-.06em
T\kern -.1667em\lower.7ex\hbox{E}\kern-.125emX}}
\def\etal{{\em et al.}}

\def\weekdaystring#1#2#3{{
\count10=#3%
\multiply\count10 by 365%
\count11=#3%
\divide\count11 by 4%
\advance\count10 by \count11
\count12=\count11%
\multiply\count12 by 4
\count13=#3%
\advance\count13 by -\count12%
\ifcase#2%
\or\ifcase\count13\advance\count10 by -366
\else\advance\count10 by -365
\fi%
\or\ifcase\count13\advance\count10 by -335
\else\advance\count10 by -334
\fi%
\or\advance\count10 by -306
\or\advance\count10 by -275
\or\advance\count10 by -245
\or\advance\count10 by -214
\or\advance\count10 by -184
\or\advance\count10 by -153
\or\advance\count10 by -122
\or\advance\count10 by -92
\or\advance\count10 by -61
\or\advance\count10 by -31
\fi%
\advance\count10 by #1%
\advance\count10 by 5
\count14=\count10%
\divide\count14 by 7%
\multiply\count14 by 7
\count15=\count10%
\advance\count15 by -\count14
\ifcase\count15 Monday\or Tuesday\or Wednesday%
\or Thursday\or Friday\or Saturday\or Sunday\fi%
}}
\def\monthstring#1{\ifcase#1\or January\or February\or March\or April\or May%
\or June\or July\or August\or September\or October\or November\or December\fi}
\def\dateY{\def\today{\number\year}}
\renewcommand{\baselinestretch}{1.5}

\dateY
\def\thesisby#1{\gdef\@thesisby{#1}}
\def\@thesisby{\vskip 2ex by\vskip 2ex}
\def\thesissubmit#1{\gdef\@thesissubmit{#1}}
\def\@thesissubmit{\vskip 6ex Submitted in Partial Fulfillment\\~\\of the\\~\\
Requirements for the Degree\vskip 2ex}
\def\thesisdegree#1{\gdef\@thesisdegree{#1}}
\def\@thesisdegree{{Doctor of Philosophy}}
\def\thesissupervisor#1{\gdef\@thesissupervisor{#1}}
\def\@thesissupervisor{UNKNOWN}
\def\thesissupervise#1{\gdef\@thesissupervise{#1}} 
\def\@thesissupervise{\vskip 6ex Supervised by\\~\\
  \@thesissupervisor\vskip 2ex}
\def\thesisdepartment#1{\gdef\@thesisdepartment{#1}}
\def\@thesisdepartment{Department of Computer Science\\The College\\Arts
  and Sciences}
\def\thesisuniversity#1{\gdef\@thesisuniversity{#1}}
\def\@thesisuniversity{\vskip 3ex
University of Rochester\\Rochester, New York}
\def\maketitle{\par
 \begingroup
 \def\thefootnote{\fnsymbol{footnote}}
 \def\@makefnmark{\hbox
   to 0pt{$^{\@thefnmark}$\hss}}
 \newpage
   \global\@topnum\z@        
   \@maketitle\thispagestyle{empty}\@thanks
 \endgroup
 \setcounter{footnote}{0}
 \let\maketitle\relax
 \let\@maketitle\relax
 \gdef\@thanks{}\gdef\@author{}\gdef\@title{}\let\thanks\relax}
\def\@maketitle{
  \renewcommand{\baselinestretch}{1}
  \newpage
  \thispagestyle{empty}        
  \null
  \vspace{0.25in}              
  \begin{center}
    {\huge \@title \par}        
    {\large \@thesisby          
      \@author \par}
    {\large \@thesissubmit \@thesisdegree}
    {\large \@thesissupervise}
    {\large \@thesisdepartment}
    {\large \@thesisuniversity \vskip 1.5em \@date}
  \end{center}
  \pagenumbering{roman}
  \clearpage
}
\def\curriculumvitaehead#1{\gdef\@curriculumvitaehead{#1}}
\def\@curriculumvitaehead{\thispagestyle{plain}
\chapter*{Curriculum Vitae\@mkboth{CURRICULUM VITAE}{CURRICULUM VITAE}}
\addcontentsline{toc}{chapter}{Curriculum Vitae}}
\newenvironment{curriculumvitae}{\@curriculumvitaehead}{\clearpage}
\def\acknowledgmentshead#1{\gdef\@acknowledgmentshead{#1}}
\def\@acknowledgmentshead{\thispagestyle{plain}
\chapter*{Acknowledgments\@mkboth{ACKNOWLEDGEMENTS}{ACKNOWLEDGEMENTS}}
\addcontentsline{toc}{chapter}{Acknowledgments}}
\newenvironment{acknowledgments}{\@acknowledgmentshead}{\clearpage}
\def\abstracthead#1{\gdef\@abstracthead{#1}}
\def\@abstracthead{\thispagestyle{plain}
\chapter*{Abstract\@mkboth{ABSTRACT}{ABSTRACT}}
\addcontentsline{toc}{chapter}{Abstract}}
\renewenvironment{abstract}{\@abstracthead}{\clearpage}
\def\ps@plain{\let\@mkboth\@gobbletwo
\def\@oddhead{\rm\hfil\thepage}\def\@oddfoot{}%
\def\@evenhead{\rm\thepage\hfil}\def\@evenfoot{}}
\def\@chapapp{Chapter}
\def\thechapter{\arabic{chapter}}
\def\firstchapter{0} 
\def\@makechapterhead#1{     
  \cleardoublepage
  \ifcase\firstchapter
    \pagenumbering{arabic}
    \def\firstchapter{1} 
  \fi
  \vspace*{50pt}             
  { \parindent 0pt \raggedright
    \par
    \huge\bf\thechapter\hspace{1em}%
\settowidth{\chapnowidth}{\huge\bf\thechapter\hspace{1em}}%
\setlength{\chaptitlewidth}{\textwidth}%
\addtolength{\chaptitlewidth}{-\chapnowidth}%
\parbox[t]{\chaptitlewidth}{\raggedright#1}
  \nobreak                   
  \vskip 40pt                
  }
}
\def\tableofcontents{\@restonecolfalse\if@twocolumn\@restonecoltrue\onecolumn
  \fi\chapter*{Table of Contents\@mkboth{TABLE OF CONTENTS}{TABLE OF CONTENTS}}
  \@starttoc{toc}\if@restonecol\twocolumn\fi}
\def\listoffigures{\@restonecolfalse\if@twocolumn\@restonecoltrue\onecolumn
  \fi\chapter*{List of Figures\@mkboth
   {LIST OF FIGURES}{LIST OF FIGURES}}
   \addcontentsline{toc}{chapter}{List of Figures}\@starttoc{lof}\if@restonecol
    \twocolumn\fi}
\def\listoftables{\@restonecolfalse\if@twocolumn\@restonecoltrue\onecolumn
  \fi\chapter*{List of Tables\@mkboth
   {LIST OF TABLES}{LIST OF TABLES}}
   \addcontentsline{toc}{chapter}{List of Tables}\@starttoc{lot}\if@restonecol
  \twocolumn\fi}
\def\@makeschapterhead#1{    
  \vspace*{50pt}             
  \begin{center}
  {\huge #1} 
  \end{center}
  \nobreak                   
  \vskip 40pt                
  }
\def\thebibliography#1{\cleardoublepage\chapter*{Bibliography\@mkboth
  {BIBLIOGRAPHY}{BIBLIOGRAPHY}}\addcontentsline{toc}{chapter}{Bibliography}\list
  {[\arabic{enumi}]}{\settowidth\labelwidth{[#1]}\leftmargin\labelwidth
    \advance\leftmargin\labelsep
    \usecounter{enumi}}
    \def\newblock{\hskip .11em plus .33em minus .07em}
    \sloppy\clubpenalty4000\widowpenalty4000
    \sfcode`\.=1000\relax}
\def\appendix{\par
  \setcounter{chapter}{0}
  \setcounter{section}{0}
  \setcounter{subsection}{0}
  \def\@chapapp{Appendix}
  \def\thechapter{\Alph{chapter}}}
\def\section{\@startsection {section}{1}{\z@}{3.5ex plus 1ex minus
    .2ex}{2.3ex plus .2ex}{\Large\bf}}
\def\subsection{\@startsection{subsection}{2}{\z@}{3.25ex plus 1ex minus
   .2ex}{1.5ex plus .2ex}{\large\bf}}
\def\subsubsection{\@startsection{subsubsection}{3}{\z@}{3.25ex plus
1ex minus .2ex}{1.5ex plus .2ex}{\normalsize\bf}}
\def\paragraph{\@startsection
     {paragraph}{4}{\z@}{3.25ex plus 1ex minus .2ex}{-1em}{\normalsize\bf}}
\def\subparagraph{\@startsection
     {subparagraph}{4}{\parindent}{3.25ex plus 1ex minus
     .2ex}{-1em}{\normalsize\bf}}
\def\section{\@startsection {section}{1}{\z@}{3.5ex plus 1ex minus
    .2ex}{2.3ex plus .2ex}{\Large\bf}}
\def\subsection{\@startsection{subsection}{2}{\z@}{3.25ex plus 1ex minus
   .2ex}{1.5ex plus .2ex}{\large\bf}}
\def\subsubsection{\@startsection{subsubsection}{3}{\z@}{3.25ex plus
1ex minus .2ex}{1.5ex plus .2ex}{\normalsize\bf}}
\def\paragraph{\@startsection
     {paragraph}{4}{\z@}{3.25ex plus 1ex minus .2ex}{-1em}{\normalsize\bf}}
\def\subparagraph{\@startsection
     {subparagraph}{4}{\parindent}{3.25ex plus 1ex minus
     .2ex}{-1em}{\normalsize\bf}}
\makeatother

\renewcommand{\arraystretch}{0.8}

\newcommand{\mc}[1]{\multicolumn{1}{c|}{#1}}

\newcommand{\fnote}[1]{\footnote{{\bf Future:} #1}}

\renewcommand{\fnote}[1]{}
\newcommand{\ignore}[1]{}

\newtheorem{example1}{Example}
\newenvironment{example}[1]{\begin{example1}[#1]\samepage \mbox{} \\ \rm}{\end{example1}}

\newcommand{\trans}[1]{
\parbox[t]{0.5em}{\begin{center}
\vspace*{-1.7em}\mbox{} \\
\makebox[0.2em]{\Large$\uparrow$} \vspace*{-0.5em} \\
\makebox[0.5em][c]{\em #1} 
\end{center}}}
\newcommand{\ip}{\trans{ip}}

\newcommand{\interruptionpoint}{
\parbox[t]{1.0em}{\begin{center}
\vspace*{-1.5em}\mbox{} \\
\LARGE$\uparrow$ \normalsize \vspace*{-2.1em} \\
\makebox[1em]{\em interruption} \vspace*{-2.5em}\\ 
\makebox[1em]{\em point}
\end{center}}}

\newcommand{\reparandum}[1]{$\underbrace{\makebox{#1}}_{\makebox{\em reparandum}}$}
\newcommand{\editingterm}[1]{$\underbrace{\makebox{#1}}_{\makebox{\em editing terms}}$}
\newcommand{\et}[1]{$\underbrace{\makebox{#1}}_{\makebox{\em et}}$}
\newcommand{\alteration}[1]{$\underbrace{\makebox{#1}}_{\makebox{\em
alteration}}$}

\newcommand{\turn}{\verb+<+turn\verb+>+ }
\newcommand{\tone}{\verb+<+tone\verb+>+ }
\newcommand{\frag}{\verb+<+frag\verb+>+ }

\newcommand{\ri}{{1,i}}
\newcommand{\rim}{{1,i{\dash}1}}
\newcommand{\rimm}{{1,i{\dash}2}}
\newcommand{\im}{{i{\dash}1}}
\newcommand{\imm}{{i{\dash}2}}
\newcommand{\immm}{{i{\dash}3}}
\newcommand{\dash}{\mbox{-}}

\begin{document}
\title{\mbox{} \vspace*{-4em} \\ Speech Repairs, Intonational Boundaries \\ and Discourse Markers: \\ Modeling Speakers' Utterances \\ in Spoken Dialog}
\author{Peter Anthony Heeman}
\thesissupervisor{Professor James F. Allen}

\maketitle
\cleardoublepage
\begin{curriculumvitae}

Peter Heeman was born October 22, 1963, and much to his dismay his
parents had already moved away from Toronto.  Instead he was born in
London Ontario, where he grew up on a strawberry farm.  He attended
the University of Waterloo where he received a Bachelors of
Mathematics with a joint degree in Pure Mathematics and Computer
Science in the spring of 1987.

After working two years for a software engineering company, which 
supposedly used artificial intelligence techniques to automate COBOL
and CICS programming, Peter was ready for a change.  What better way
to wipe the slate clear than by going to graduate school at the
University of Toronto, but not without first spending the summer in
Europe.  After spending two months in countries where he couldn't
speak the language, Peter became fascinated by language, and so
decided to give computational linguistics a try.

In the fall of 1989, Peter started his Masters degree in Computer
Science at the University of Toronto, supported by the National
Science and Engineering Research Council (NSERC) of Canada and by
working as a consultant at ManuLife.  Peter took the introductory
course in computational linguistics given by Professor Graeme Hirst,
who later became his advisor.  In searching for a Masters thesis
topic, Graeme got Peter interested in Herbert Clark's work on
collaborating in discourse.  With the guidance of Graeme and Visiting
Professor Janyce Wiebe, Peter made a computational model of Clark's
work on collaborating upon referring expressions.

With the guidance of Professor Hirst, Peter decided to attend the
University of Rochester in the fall of 1991 to do his Ph.D. with
Professor James Allen, with two years of funding supplied by NSERC.
The first few months were a bit difficult since Peter was still
finishing up his Masters thesis.  But he did manage to graduate from
Toronto in that fall with a Masters of Science.

Rochester was of course a major culture shock to Peter; but he
survived.  He even survived the first year and a half in Rochester
without an automobile, relying on a bicycle to get him around
Rochester and to and from school.  Luckily Peter lived close to his
favorite bar, the Avenue Pub, which is where he met Charles on a
fateful evening in the summer after his first year.

As a sign of encouragement (or funding regulations), Peter received a
Masters of Science, again in Computer Science, from the University of
Rochester in the spring of 1993.  It was around this time that James
um like got Peter interested in computationally understanding
disfluencies, which of course is a major theme in this thesis.

Having had a taste of the fast pace of Toronto, one could image that
five and a half years in Rochester would take their toll.  Luckily in
the fall of 1996, Peter was invited to spend four months in Japan at
ATR in the Interpreting Telecommunications Research laboratory working
with Dr.~Lokem-Kim, an offer that he quickly accepted.  Peter's second
chance to escape occurred immediately after his oral defense of this
thesis.  This time the location was in France, where he did a post-doc
at CNET, France T\'el\'ecom.  Although located far from Paris, it did
give Peter a chance to become a true Canadian by forcing him to
improve his French.  It was at CNET Lannion that final revisions to
this thesis were completed.

\end{curriculumvitae}

\cleardoublepage
\begin{acknowledgments}

To begin with, I would like to thank my advisor, James Allen, for his
support and encouragement during my stay at the University of
Rochester.  I would also like to thank the other members of my
committee: Len Schubert and Michael Tanenhaus.  Their feedback helped
shaped this thesis, especially Mike's encouragement to use machine
learning techniques to avoid using ad-hoc rules that happen to fit the
training data.

I also want to thank my co-advisors from my Masters degree at the
University of Toronto: Graeme Hirst and Janyce Weibe.  Their
involvement and encouragement did not stop once I had left Toronto.

I wish to thank the Trains group.  I wish to thank the original Trains
group, especially George Ferguson, Chung Hee Hwang, Marc Light,
Massimo Poesio, and David Traum.  I also wish to thank the current
Trains group, especially George, Donna Bryon, Mark Core, Eric Ringger,
Amon Seagull and Teresa Sikorski.  A special thanks to Mark and Amon
for helping me proofread this thesis.

I also wish to thank the other members of my entering class,
especially Hannah Blau, Ramesh Sarukkai and Ed Yampratoon.  I also
wish to thank everyone else in the department for making it such a
great place, especially Chris Brown, Polly Pook, the administrative
staff---Jill Forster, Peggy Franz, Pat Marshall, and Peg Meeker---and
the support staff---Tim Becker, Liud Bukys, Ray Frank, Brad Miller and
Jim Roche.

I would also like to thank Lin Li, Greg Mitchell, Mia Stern, Andrew
Simchik, and Eva Bero, who helped in annotating the Trains corpus over
the last three years and helped in refining the annotation schemes.

I also wish to thank members of the research community for their
insightful questions and conversations, especially Ellen Bard, John
Dowding, Julia Hirschberg, Lynette Hirschman, Robin Lickley, Marie
Meteer, Mari Ostendorf, Liz Shriberg and Gregory Ward.

I wish to thank Kyung-ho Loken-Kim for providing me with the
opportunity to work on this thesis at ATR in Japan. In addition to a
welcome change in environment (and being able to avoid several major
snow storms), I had the opportunity to present my work there, from
which I received valuable comments, especially from Alan Black, Nick
Cambell, Laurie Fais, Andrew Hunt, Kyung-ho Loken-Kim, and Tsuyoushi
Morimoto.

I also wish to thank David Sadek for providing me the opportunity to
work at CNET, France T\'el\'ecom, where I made the final revisions to
this thesis. I also want to thank the many people at CNET who made my
stay enjoyable and gave me valuable feedback. I would especially like
to thank Alain Cozannet, Geraldine Damnati, Alex Ferrieux, Denis
Jouvet, David Sadek, Jacque Simonin and Christel Sorin and the
administrative support of Janine Denmat.

This material is based upon work supported by the NSF under grant
IRI-9623665, DARPA---Rome Laboratory under research contract
F30602-95-1-0025, ONR/DARPA under grant N00014-92-J-1512, and ONR
under grant N0014-95-1-1088. Funding was also received from the
Natural Science and Engineering Research Council of Canada, from the
Interpreting Telecommunications Laboratory at ATR in Japan, and from
the Centre National d'Etudes des T\'el\'ecommunications, France
T\'el\'ecom.

Finally, I wish to thank the people who are dearest to me.  I wish to
thank my parents and siblings who have always been there for me.  I
wish to thank my friends in Toronto and elsewhere, especially Greg and
Randy, for letting me escape from Rochester.  I also wish to thank my
departed friend Andr\'e.  Finally, I wish to thank Charles Buckner, who
has patiently put up with me while I have worked away on this thesis,
and accompanied me on the occasional escape away from it.

\end{acknowledgments}

\cleardoublepage
\begin{abstract}

\noindent
Interactive spoken dialog provides many new challenges for natural
language understanding systems.  One of the most critical challenges
is simply determining the speaker's intended utterances: both
segmenting a speaker's turn into utterances and determining the
intended words in each utterance.  Even assuming perfect word
recognition, the latter problem is complicated by the occurrence of
speech repairs, which occur where the speaker goes back and changes
(or repeats) something she just said.  The words that are replaced or
repeated are no longer part of the intended utterance, and so need to
be identified.  The two problems of segmenting the turn into
utterances and resolving speech repairs are strongly intertwined with
a third problem: identifying discourse markers.  Lexical items that
can function as discourse markers, such as ``well'' and ``okay,'' are
ambiguous as to whether they are introducing an utterance unit,
signaling a speech repair, or are simply part of the context of an
utterance, as in ``that's okay.''  Spoken dialog systems need to
address these three issues together and early on in the processing
stream.  In fact, just as these three issues are closely intertwined
with each other, they are also intertwined with identifying the
syntactic role or part-of-speech (POS) of each word and the speech
recognition problem of predicting the next word given the previous
words.

In this thesis, we present a statistical language model for resolving
these issues.  Rather than finding the best word interpretation for an
acoustic signal, we redefine the speech recognition problem to so that
it also identifies the POS tags, discourse markers, speech repairs and
intonational phrase endings (a major cue in determining utterance
units).  Adding these extra elements to the speech recognition problem
actually allows it to better predict the words involved, since we are
able to make use of the predictions of boundary tones, discourse
markers and speech repairs to better account for what word will occur
next.  Furthermore, we can take advantage of acoustic information,
such as silence information, which tends to co-occur with speech
repairs and intonational phrase endings, that current language models
can only regard as noise in the acoustic signal.  The output of this
language model is a much fuller account of the speaker's turn, with
part-of-speech assigned to each word, intonation phrase endings and
discourse markers identified, and speech repairs detected and
corrected.  In fact, the identification of the intonational phrase
endings, discourse markers, and resolution of the speech repairs
allows the speech recognizer to model the speaker's {\em utterances},
rather than simply the words involved, and thus it can return a more
meaningful analysis of the speaker's turn for later processing.

\end{abstract}

\tableofcontents
\cleardoublepage
\listoftables
\cleardoublepage
\listoffigures
\cleardoublepage

\chapter{Introduction}

One of the goals of natural language processing and speech recognition
is to build a computer system that can engage in spoken dialog.  We
still have a long ways to go to achieve this goal, but even with today's
technology it is already possible to build limited spoken dialog
systems, as exemplified by the ATIS project \cite{Madcow92:snlp}.  Here
users can query the system to find out air travel information, as the
following example illustrates.
\begin{example}{ATIS}
I would like a flight that leaves after noon in San Francisco and
arrives before 7 p.m. Dallas time
\end{example}
With ATIS, users are allowed to think off-line, with turn-taking
``negotiated'' by the user pressing a button when she\footnote{We use
the pronoun {\em she} to refer to speaker, and {\em he} to refer to
the hearer.}  wants to speak.  The result is that their speech looks
very text-like.

Finding ways of staying within the realms of the current state of
technology is not limited to the ATIS project.  Oviatt
\shortcite{Oviatt95:csl} has investigated other means of structuring
the user's interactions so as to reduce the complexity of the user's
speech, thus making it easier to understand.  However, this takes the
form of structuring the user's actions.  Although such restrictions
might be ideal for applications such as automated census polls
\cite{Cole-etal94:icslp}, we doubt that this strategy will be effective for
tasks in which the human and spoken dialog system need to collaborate
in achieving a given goal.  Rather, both human and computer must be
able to freely contribute to the dialog \cite {HeemanHirst95:cl}.

In order to better understand how people naturally engage in
task-oriented dialogs, we have collected a corpus of human-human
problem-solving dialogs: {\em The Trains Corpus} \cite
{HeemanAllen95:tn-dialogs}.\footnote {Unless otherwise noted, all
examples are drawn from the Trains corpus.  The corpus is available
from the Linguistics Data Consortium on CD-ROM \cite
{HeemanAllen95:cdrom}.}  The Trains corpus differs from the
Switchboard corpus \cite{Godfrey-etal92:icassp} in that it is
task-oriented and has a limited domain, making it a more realistic
domain for studying the types of conversations that people would want
to have with a computer.  From examining the Trains corpus, it becomes
evident that in natural dialog speakers' turns tend to be more complex
than what is seen in the ATIS corpus.  We need to determine how we can
make a spoken dialog system cope with these added complexities in
order to make it more conversationally proficient.

\section{Utterances in Spoken Dialog}

A speaker's turn, as many people have argued, is perhaps of too coarse
a granularity to be a viable unit of spoken dialog processing.
Speakers will often use their turn of the conversation to make several
distinct contributions.  If we were given the words involved in a
speaker's turn, we would undoubtedly need to segment it into a number
of sentence-like entities, {\em utterances}, in order to determine
what the speaker was talking about.  Consider the following example
taken from the Trains corpus.
\begin{example}{d93-13.3 utt63}
um it'll be there it'll get to Dansville at three a.m. and then you
wanna do you take tho- want to take those back to Elmira so
engine E two with three boxcars will be back in Elmira at six a.m. is
that what you wanna do
\end{example}
Understanding what the speaker was trying to say in this turn is not
straightforward, and it probably takes a reader several passes in order to determine how to segment it into smaller units, most likely
into a segmentation similar to the one below.
\addtocounter{example1}{-1}
\begin{example}{Revisited}
um it'll be there it'll get to Dansville at three a.m. \\
and then you wanna do you take tho- want to take those back to Elmira \\
so engine E two with three boxcars will be back in Elmira at six a.m. \\
is that what you wanna do 
\end{example}
Even this segmentation does not fully capture the message that the
speaker intended to convey to the hearer.  The first and second
segments both contain {\em speech repairs}, a repair where the speaker
goes back and changes (or repeats) something she just said.  In the
first utterance, the speaker went back and replaced ``it'll be there''
with ``it'll get to \dots''; and in the second, she replaced ``you
wanna'' with ``do you take tho-'', which is then revised to ``do you
want to take those back to Elmira''.  The reader's resulting
understanding of the speaker's turn is thus as follows.
\addtocounter{example1}{-1}
\begin{example}{Revisited again}
um it'll get to Dansville at three a.m. \\ 
and then do you want to take those back to Elmira \\ 
so engine E two with three boxcars will be back in Elmira at six a.m. \\
is that what you wanna do
\end{example}
The problems that the reader faces are also faced by the hearer,
except that the hearer needs to be doing these tasks online as the
speaker is speaking.  He needs to determine how to segment the
speaker's turn into more manageable sized units, which we will refer
to as {\em utterance units}, and he needs to resolve any speech
repairs.

These two problems are strongly intertwined with a third problem:
identify discourse markers.  Many utterances start with a discourse
marker, a word that signals how the new utterance relates to what was
just said.  For instance, the second utterance began with ``and
then'', and the third with ``so''.  Discourse markers also co-occur
with speech repairs, perhaps to mark the relationship between what the
speaker just said and her correction or to simply help signal that a
speech repair is occurring.  Thus in determining the utterance
segmentation and resolving speech repairs, the hearer will undoubtedly
need to identify the discourse markers.  However, because of
interactions between all three of these issues, all must be resolved
together.  In the next three sections, we introduce each of these
issues.

\subsection{Utterance Units}

Brown and Yule \shortcite{BrownYule83:book} discuss a number of ways
in which speech differs from text.  The syntax of spoken dialog is
typically much less structured than that of text: it contains
fragments, there is little subordination, and it lacks the
meta-lingual markers between clauses.  It also tends to come in
installments and refinements, and makes use of topic-comment sentence
structure.\footnote{Crystal \shortcite{Crystal80} presents some
additional problems with viewing speech as sentences and clauses.}
The following example illustrates what could be taken as an example of
two fragments, or as an example of a topic-comment sentence.
\begin{example}{d92-1 utt4}
so from Corning to Bath \\
how far is that 
\end{example}

Although speech cannot always be mapped onto sentences, there is wide
agreement that speech does come in sentence-like packages, which are
referred to as {\em utterances}.  Following Bloomfield \shortcite
{Bloomfield26}, the term {\em utterance} has often been vaguely
defined as ``an act of speech.''  Utterances are a building block in
dialog for they are the means that speakers use to add to the {\em
common ground} of the conversation---the set of mutual beliefs that
conversants build up during a dialog \cite{Clark96:book}.  Hence,
utterance boundaries define appropriate places for the hearer to
ensure that he is understanding what the speaker is saying
\cite{TraumHeeman97:chapter}.  Although researchers have problems defining 
what an utterance is, hearers do not seem to have this problem as
evidenced by the experiment of Grosjean \shortcite {Grosjean83:l}, in
which he found that subjects listening to read speech could predict at
the potentially last word whether it was in fact the end of an utterance.

Although there is not a consensus as to what defines an utterance unit,
most attempts make use of one or more of the following factors.
\begin{itemize}
\item Has syntactic and/or semantic completion (e.g.~\cite{FordThompson91,NakajimaAllen93:phonetica,MeteerIyer96:emnl,StolckeShriberg96:icslp}).  
\item Defines a single speech act (e.g.~\cite 
{NakajimaAllen93:phonetica,Mast-etal96:icslp,Lavie-etal97}).  Here,
one appeals to the work of Grice \shortcite {Grice57}, Austin \cite
{Austin62:book} and Searle \cite {Searle69:book} in defining language
as action.  Speakers act by way of their utterances to accomplish
various effects, such as promising, informing, and requesting.  This
viewpoint has attracted a strong following in natural language
understanding, starting with the work of Cohen and Perrault \shortcite
{CohenPerrault79:cs} and Allen and Perrault \shortcite
{AllenPerrault80:ai} in formulating a computation model of speech
actions.
\item Is an intonational phrase (e.g.~\cite{Halliday67:jl,GeeGrosjean83:cp,FordThompson91,Gross-etal93:tr,TraumHeeman97:chapter}).
\item Separated by a pause (e.g.~\cite{NakajimaAllen93:phonetica,Gross-etal93:tr,SeligmanHosakaSinger97,TakagiItahashi96:icslp}).  
The use of this factor is probably results from how salient this
feature is and how easy it is to detect automatically, and that it has
been found to correlate with intonational phrases \cite
{GeeGrosjean83:cp}.
\end{itemize}

\subsubsection{Intonation}
\label{sec:intro:intonation}

When people speak, they tend not to speak in a monotone.  Rather, the
pitch of their voice, as well as other characteristics, such as speech
rate and loudness, varies as they speak.\footnote{Pitch is also
referred to as the fundamental frequency, or F0 for short.} The study
of intonation is concerned with describing this phenomenon and
determining its communicative meaning.  For instance, as most speakers
of English implicitly know, a statement can be turned into a question
by ending it with a rising pitch.

Pierrehumbert \shortcite{Pierrehumbert80:thesis} presented a model of
intonation patterns.  Her model describes English intonation as a
series of highs ({\bf H}) and lows ({\bf L}) in the fundamental
frequency contour.  (The formulation that we use is a slight variant
on this, and is described by Pierrehumbert and Hirschberg \shortcite
{PierrehumbertHirschberg90:iic}.)  The lowest level of analysis is at
the word level, in which stressed words are marked with either a high
or low {\em pitch accent}, marked as ${\bf H}^{\bf *}$ and ${\bf
L}^{\bf *}$, respectively.\footnote{There are also some complex pitch
accents, composed of a high and low tone.} The next level is the {\em
intermediate phrase}, which consists of at least one stressed word,
plus a high or low {\em phrase accent} at the end of the phrase,
which is marked as {\bf H-} and {\bf L-}, respectively.  The phrase
accent controls the pitch contour between the last pitch accent and
the end of the phrase.  The highest level of analysis is the {\em
intonational phrase}, which is made up of one or more intermediate
phrases and ends with an additional high or low {\em boundary tone},
which is marked as {\bf H\%} and {\bf L\%}, respectively.  The
boundary tone controls how the pitch contour ends.  Since each
intonational phrase also ends an intermediate phrase, the intonational
phrase ending consists of a phrase accent and a boundary tone,
leading to four different ways the intonational phrase can end: {\bf
H-H\%}, {\bf H-L\%}, {\bf L-H\%}, and {\bf L-L\%}.

Not only does intonation probably play an important role in segmenting
speech, but it is also important for syntactic understanding.  Beach
\shortcite{Beach91:jml} demonstrated that hearers can use
intonational information early on in sentence processing to help
resolve ambiguous attachment questions.  Price \etal~\shortcite
{Price-etal91:jasa} found that hearers can resolve most syntacticly
ambiguous utterances based on prosodic information, and Bear and
Price~\shortcite {BearPrice90:acl} explored how to make a parser use
automatically extracted prosodic features to rule out extraneous
parses.  The prosodic information was represented as a numeric score
between each pair of consecutive words, ranging from zero to five,
depending on the amount of preboundary lengthening (normalized
duration of the final consonants) and the pause duration between the
words.  Ostendorf, Wightman, and Veilleux \shortcite
{Ostendorf-etal93:csl} reported using automatically detected prosodic
phrasing to do syntactic disambiguation and achieved performance
approaching that of human listeners.  Their method utilizes prosodic
phrasing that is automatically labeled by an algorithm developed by
Wightman and Ostendorf \shortcite {WightmanOstendorf94:ieee}. Marcus
and Hindle \shortcite {MarcusHindle90:cmsp} and Steedman \shortcite
{Steedman90:cmsp} also examined the role that intonational phrases
play in parsing; but in their cases, they focused on how to represent
the content of a phrase, which is often incomplete from a syntactic
standpoint.

Pierrehumbert and Hirschberg \shortcite
{HirschbergPierrehumbert86:acl,PierrehumbertHirschberg90:iic}
looked at the role that intonation plays in discourse interpretation.
They claimed that the choice of tune ``[conveys] a particular
relationship between an utterance, currently perceived beliefs of a
hearer or hearers, \dots and anticipated contributions of subsequent
utterances \dots [and] that these relationships are compositional
---composed from the {\em pitch accents}, {\em phrase accents}, and
{\em boundary tones} that make up tunes''
\cite[pg.~271]{PierrehumbertHirschberg90:iic}.  In their theory,
pitch accents contain information about the status of discourse
referents, phrase accents about the relatedness of intermediate
phrases, and boundary tones about whether the phrase is
``forward-looking'' or not.  Intonation has also been found useful in
giving information about discourse structure
\cite{GroszHirschberg92:icslp,NakajimaAllen93:phonetica}, as well as
for turn taking \cite{FordThompson91}.

Since intonational phrasing undoubtedly plays a major role in how
utterance units are defined and is useful in interpreting utterances,
we will focus on detecting these units in this thesis.  Since
intonational phrases end with a boundary tone \cite
{Pierrehumbert80:thesis}, we also refer to the problem of identifying
the intonational phrase boundaries as identifying the boundary tones.

\subsection{Speech Repairs}
\label{sec:intro:repairs}

In spoken dialog, conversants do not have the luxury of producing
perfect utterances as they would if they were writing.  Rather, the
online nature of dialog forces them to sometimes act before they are
sure of what they want to say.  This could lead the speaker to decide
to change what she is saying.  So, she might stop during the middle of
an utterance, and go back and repeat or modify what she just said.  Or
she might completely abandon the utterance and start over.  Of course
there are many different reasons why the speaker does this sort of
thing (e.g. to get the hearer's attention \cite {Goodwin81:book}, or
convey uncertainty \cite {GoodButterworth80}).  But whatever the
reason, the point remains that {\em speech repairs\/}, disfluencies in
which the speaker repairs what she just said, are a normal occurrence
in spoken dialog.

Fortunately for the hearer, speech repairs tend to have a standard
form.  As illustrated by the following example, they can be divided
into three intervals, or stretches of speech: the {\em reparandum\/},
{\em editing term\/}, and {\em alteration}.\footnote {Our notation is
adapted from Levelt \shortcite {Levelt83:cog}.  We follow Shriberg
\shortcite {Shriberg94:thesis} and Nakatani and Hirschberg \shortcite
{NakataniHirschberg94:jasa}, however, in using {\em reparandum} to
refer to the entire interval being replaced, rather than just the non
repeated words.  We have made the same change in defining {\em
alteration}.}
\begin{example}{d92a-2.1 utt29}
\label{ex:d92a-2.1:utt29}
that's the one \reparandum{with the bananas}\interruptionpoint
\editingterm{I mean} \alteration{that's taking the bananas}
\end{example}
The reparandum is the stretch of speech that the speaker intends to
replace, and this could end with a {\em word fragment}, where the
speaker interrupts herself during the middle of the current word.  The
end of the reparandum is called the {\em interruption point} and is
often accompanied by a disruption in the intonational contour.  This
is then followed by the editing term, which can consist of filled
pauses, such as ``um'' or ``uh'' or cue phrases, such as ``I mean'',
``well'', or ``let's see''.  The last part is the alteration, which is
the speech that the speaker intends as the replacement for the
reparandum.  In order for the hearer to determine the speaker's
intended utterance, he must detect the speech repair and then solve
the {\em continuation} problem \cite {Levelt83:cog}, which is
identifying the extent of the reparandum and editing term.\footnote
{The reparandum and the editing terms cannot simply be removed, since
they might contain information, such as the identify of an anaphoric
reference, as the following contrived example displays, ``Peter was
\dots well \dots he was fired.''\label{ft:intro:reparandum}}  We will
 refer to this latter process as {\em correcting} the speech repair.
In the example above, the speaker's intended utterance is ``that's the
one that's taking the bananas''.

Hearers seem to be able to process such disfluent speech without
problem, even when multiple speech repairs occur in a row.  In
laboratory experiments, Martin and Strange \shortcite{MartinStrange68}
found that attending to speech repairs and attending to the content of
the utterance are mutually inhibitory.  To gauge the extent to which
prosodic cues can be used by hearers, Lickley, Shillcock and Bard
\shortcite {Lickley-etal91:eurospeech} asked subjects to attend to
speech repairs in low-pass filtered speech, which removes segmental
information, leaving what amounts to the intonation contour.  They had
subjects judge on a scale of 1 to 5 whether they though a speech
repair occurred in an utterance.  They found that utterances with a
speech repair received an average score of 3.36, while control
utterances without a repair only received an average score of 1.90.
In later work, Lickley and Bard \shortcite {LickleyBard92:icslp} used
a gating paradigm to determine when subjects were able to detect a
speech repair.  In the gating paradigm, subjects were successively
played more and more of the speech, in increments of 35 ms.  They
found that subjects were able to recognize speech repairs after (and
not before) the onset of the first word following the interruption
point, and for 66.5\% of the repairs before they were able to
recognize the word.  These results show that there are prosodic cues
present across the interruption point that can allow hearers to detect
a speech repair without recourse to lexical or syntactic knowledge.

Other researchers have been more specific in terms of which prosodic
cues are useful.  O'Shaughnessy \shortcite {Oshaughnessy92:icslp}
suggests that duration and pitch can be used.  Bear \etal~\shortcite
{Bear-etal92:acl} discuss acoustic cues for filtering potential repair
patterns, for identifying potential cue words of a repair, and for
identifying fragments.  Nakatani and Hirschberg \shortcite
{NakataniHirschberg94:jasa} suggest that speech repairs can be
detected by small but reliable differences in pitch and amplitude and
by the length of pause at a potential interruption point.  However, no
one has been able to find a reliable acoustic indicator of the
interruption point.

Speech repairs are a very natural part of spontaneous speech.  In the
Trains corpus, we find that 23\% of speaker turns contain at least one
repair.\footnote{These rates are comparable to the results reported by
Shriberg~\shortcite{Shriberg94:thesis} for the Switchboard corpus.} As
the length of a turn increases, so does the chance of finding such a
repair.  For turns of at least ten words, 54\% have at least one
speech repair, and for turns of at least twenty words, 70\% have at
least one.\footnote{Oviatt~\shortcite {Oviatt95:csl} found that the
rate of speech repairs per 100 words varies with the length of the
utterance.} In fact, 10.1\% of the words in the corpus are in the
reparandum or are part of the editing term of a speech repair.
Furthermore, 35.6\% of non-abridged repairs overlap, i.e.~two repairs
share some words in common between the reparandum and alteration.

\subsubsection{Classification of Speech Repairs}
\label{sec:intro:classification}

Psycholinguistic work in speech repairs and in understanding the
implications that they pose on theories of speech production
(e.g.~\cite{Levelt83:cog,BlackmerMitton91:cog,Shriberg94:thesis}) have
come up with a number of classification systems.  Categories are based
on how the reparandum and alteration differ, for instance whether the
alteration repeats the reparandum, makes it more appropriate, inserts
new material, or fixes an error in the reparandum.  Such an analysis
can shed information on where in the production system the error and
its repair originated.

Our concern, however, is in computationally detecting and correcting
speech repairs.  The features that are relevant are the ones that the
hearer has access to and can make use of in detecting and correcting a
repair.  Following loosely in the footsteps of the work of Hindle
\shortcite {Hindle83:acl} in correcting speech repairs, we divide
speech repairs into the following categories: {\em fresh starts}, {\em
modification repairs}, and {\em abridged repairs}.

Fresh starts occur where the speaker abandons the current utterance
and starts again, where the abandonment seems to be acoustically
signaled either in the editing term or at the onset of the
alteration.\footnote {Hindle referred to this type of repair as a {\em
restart}.}  Example~\ref {ex:d93-14.3:utt2} illustrates a fresh start
where the speaker abandons the partial utterance ``I need to send'',
and replaces it by the question ``how many boxcars can one engine
take''.
\begin{example}{d93-14.3 utt2}
\label{ex:d93-14.3:utt2}
\reparandum{I need to send}\interruptionpoint \editingterm{let's see} \alteration{how many boxcars can one engine take}
\end{example}
For fresh starts, there can sometimes be very little or even no
correlation between the reparandum and the alteration.\footnote{When
there is little or no correlation between the reparandum and
alteration, labeling the extent of the alteration is somewhat arbitrary.}
Although it is usually easy to determine the onset of the reparandum,
since it is the beginning of the utterance, determining if initial
discourse markers such as ``so'' and ``and'' and preceding
intonational phrases are part of the reparandum can be problematic and
awaits a better understanding of utterance units in spoken dialog
\cite{TraumHeeman97:chapter}.

The second type are modification repairs.  This class comprises the
remainder of speech repairs that have a non-empty reparandum.  The
example below illustrates this type of repair.
\begin{example}{d92a-1.2 utt40}
\label{ex:d92a-1.2:utt40}
you can \reparandum{carry them both on}\interruptionpoint \alteration{tow both on} the same engine
\end{example}
In contrast to the fresh starts, which are defined in terms of a
strong acoustic signal marking the abandonment of the current
utterance, modification repairs tend to have strong word
correspondences between the reparandum and alteration, which can help
the hearer determine the extent of the reparandum as well as help
signal that a modification repair occurred.  In the example above, the
speaker replaced ``carry them both on'' by ``tow both on'', thus
resulting in word matches on the instances of ``both'' and ``on'', and
a replacement of the verb ``carry'' by ``tow''.  Modification repairs
can in fact consist solely of the reparandum being repeated by the
alteration.\footnote{Other classifications tend to distinguish repairs
based on whether any content has changed.  Levelt refers to repairs
with no changed content as {\em covert} repairs, which also includes
repairs consisting solely of an editing term.}  For some repairs, it
is difficult to classify them as either a fresh start or as a
modification repair, especially for repairs whose reparandum onset is
the beginning of the utterance and that have strong word
correspondences.  Hence, our classification scheme allows this
ambiguity to be captured, as explained in Section~\ref
{sec:corpus:repairs:annotations}.

Modification repairs and fresh starts are further differentiated by
the types of editing terms that co-occur with them.  For instance, cue
phrases such as ``sorry'' tend to indicate fresh starts, whereas the
filled pause ``uh'' more strongly signals a modification repair
(cf.~\cite{Levelt83:cog}).

The third type of speech repair is the abridged repair.  These repairs
consist of an editing term, but with no reparandum, as the following
example illustrates.\footnote{In previous work \cite
{HeemanAllen94:acl}, we defined abridged repairs to also include
repairs whose reparandum consists solely of a word fragment.  Such
repairs are now categorized as modification repairs or as fresh starts
(cf.~\cite[pg.~11] {Shriberg94:thesis}).}\fnote{Len says: aren't lots
of these just hesitations?  (i.e.~no intention to repair)}
\begin{example}{d93-14.3 utt42}
we need to\interruptionpoint \editingterm{um} manage to get the bananas to Dansville more quickly
\end{example}
For these repairs, the hearer has to determine that an editing term
has occurred, which can be difficult for phrases like ``let's see'' or
``well'' since they can also have a sentential interpretation.  The
hearer also has to determine that the reparandum is empty.  As the
above example illustrates, this is not necessarily a trivial task
because of the spurious word correspondences between ``need to'' and
``manage to''.

Not all filled pauses are marked as the editing term of an abridged
repair, nor are all cue phrases such as ``let's see''.  Only when
these phrases occur mid-utterance and are not intended as part of the
utterance are they treated as abridged repairs
(cf.~\cite{ShribergLickley93:phonetica}).  In fact, deciding if a
filled pause is a part of an abridged repair can only be done in
conjunction with deciding the utterance boundaries.

\subsection{Discourse Markers}

Phrases such as ``so'', ``now'', ``firstly,'' ``moreover'', and
``anyways'' are referred to as discourse markers \cite
{Schiffrin87:book}.  They are conjectured to give the hearer
information about the discourse structure, and so aid the hearer in
understanding how the new speech or text relates to what was
previously said and for resolving anaphoric references \cite
{RCohen84:coling,Reichmanadar84:ai,Sidner85:ci,GroszSidner86:cl,LitmanAllen87,HirschbergLitman93:cl}.

Although some discourse markers, such as ``firstly'', and
``moreover'', are not commonly used in spoken dialog \cite
{BrownYule83:book}, there are a lot of other discourse markers that
are employed.  These discourse markers are used to achieve a variety
of effects: such as signal an acknowledgment or acceptance, hold a
turn, stall for time, signal a speech repair, or to signal an
interruption in the discourse structure or the return from one.  These
uses are concerned with the interactional aspects of discourse rather
than adding to the content.  

Although Schiffrin defines discourse markers as bracketing units of
speech, she explicitly avoids defining what the unit is.  In this
thesis, we feel that utterance units are the building blocks of spoken
dialog and that discourse markers operate at this level to either set
up expectations for future utterances \cite
{ByronHeeman97:eurospeech}, relate the current utterance to the
previous discourse context, or to signal a repair to the utterance.
In fact, deciding if a lexical item such as ``and'' is being used as a
discourse marker can only be done in conjunction with deciding the
utterance unit boundaries.  Consider the following example, where the
symbol `{\bf \%}' is used to denote the intonational boundary tones.
\newlength{\system}
\settowidth{\system}{systemss}
\begin{example}{d92-1 utt33-35}
\makebox[\system][l]{\bf user:} so how far is it from Avon to
Dansville \% \\
\makebox[\system][l]{\bf system:} three hours \% \\ 
\makebox[\system][l]{\bf user:} three hours \% \\
\makebox[\system][l]{\mbox{}} then from Dansville to Corning \%
\end{example}
The first part of the last turn, ``three hours,'' is repeating what
the other conversant just said, which was a response to the question
``how far is it from Avon to Dansville''.  After repeating ``three
hours'', the speaker then asks the next question, ``from Dansville to
Corning''.  To understand the user's turn, the system must realize
that the user's turn consists of two utterances.  This realization is
facilitated by the recognition that ``then'' is being used as a
discourse marker to introduce the second utterance.

The example above hints at the difficulty that can be encountered in
labeling discourse markers.  For some discourse markers, the discourse
marker meaning is closely associated with the sentential meaning.  For
instance, the discourse markers ``and then'' can simply indicate a
temporal coordination of two events, as the following example
illustrates.
\begin{example}{d92a-2.1 utt137}
making the orange juice \% \\
and then going to Corning \% \\
and then to Bath \% 
\end{example}
The two instances of ``and then'' are marked as discourse markers
because the annotator felt that they were being used to introduce the
subsequent utterance, and hence they have a discourse purpose in
addition to their sentential role of indicating temporal coordination.

\section{Interactions}
\label{sec:intro:interactions}

The problems of identifying boundary tones, resolving speech repairs,
and identifying discourse markers are highly intertwined.  In this
section we argue that each of these problems depends on the solution
of the other two.  Hence, in order to model speaker utterances in
spontaneous speech, we need to resolve all three together in a model
that can evaluate competing hypotheses.

\subsection{Speech Repairs and Boundary Tones}

The problems of resolving speech repairs and detecting boundary tones
are interrelated.  ``When we consider spontaneous speech (particularly
conversation) any clear and obvious division into intonational-groups
is not so apparent because of the broken nature of much spontaneous
speech, including as it does hesitation, repetitions, false starts,
incomplete sentences, and sentences involving a grammatical caesura in
their middle'' \cite[pg.~36] {Cruttenden86:book}.  The work of Wang
and Hirschberg \shortcite {WangHirschberg92:csl} also suggests the
difficulty in distinguishing these two types of events.  They found
the best recall rate of intonational phrase endings occurred when they
counted disfluencies as intonational phrase endings, while the best
precision rate was obtained by not including them.

Confusion between boundary tones and the interruption points of speech
repairs can occur because both types of events share a number of
features.  Pauses often co-occur with both interruption points and
boundary tones, as does lengthening of the last syllable of the
preceding word.  Even the cue of strong word correspondences,
traditionally associated with the interruption point of modification
repairs, can also occur with boundary tones.  Consider the following
example.
\begin{example}{d93-8.3 utt73}
that's all you need \% \\ 
you only need one boxcar
\end{example}
Here the speaker was rephrasing what she just said for emphasis, but
this has the effect of creating strong word correspondences across the
boundary tone, thus giving the allusion that it is a speech repair in
which the speaker is changing ``you need'' to ``you only
need''.\footnote {See Walker \shortcite{Walker93:thesis} for a
discussion of the role of informationally redundant utterances in
spoken dialog.}

There are also interactions between speech repair correction and
boundary tone identification.  Fresh starts tend to cancel the current
utterance; hence to correct such repairs, one needs to know where
the current utterance begins, especially since fresh starts often do
not employ the strong word correspondences that modification repairs
rely on to delimit the extent of the reparandum.  Since intonational
boundaries are a key ingredient in signaling utterances, speech repair
correction cannot happen before intonational phrase detection.

Intonational boundary tone detection is also needed to help
distinguish between speech repairs and other types of first-person
repairs.  Speakers often repair what they have just said \cite
{Schegloff-etal77:lang}, but this does not mean that the repair is a
speech repair.  Just as we do not include repairs that cross speaker
turns as speech repairs, we also do not include repairs where the
speaker corrects or changes the semantic content after a complete
utterance, or is simply voicing uncertainty with her last complete
utterance.  Consider the following example, with each line being a
complete intonational phrase.
\begin{example}{d93-26.2 utt41}
oh that wouldn't work apparently \% \\
wait wait \% \\
let's see \% \\
maybe it would \% \\
yeah it would \% \\
right \% \\
nineteen hours did you say \%
\end{example}
In this example, there is not a speech repair after the first phrase,
nor is the fifth phrase ``yeah it would'' a replacement for ``maybe it
would''.  Rather each line is a complete intonational phrase and is
acting as a contribution to the discourse state.  Any revision that is
happening is simply the type of revision that often happens in
collaborative dialog.

\subsection{Boundary Tones and Discourse Markers}
\label{sec:intro:tones_dm}

Identifying boundary tones and discourse markers is also highly
interrelated.  Discourse marker usages of ambiguous words tend to
occur at the beginning of an utterance unit, while sentential usages
tend to occur mid-utterance.  Example~\ref{ex:d92-1:utt32} below
illustrates a speaker's turn, which consists of three intonational
phrases, each beginning with a discourse marker.
\begin{example}{d92-1 utt32-33}
\label{ex:d92-1:utt32}
okay \% \\
so we have the three boxcars at Dansville \% \\
so how far is it from Avon to Dansville \% \\
\end{example}
Example~\ref{ex:d93-15.2:utt9} illustrates ``so'' being used in its
sentential form, as a subordinating conjunction, but not at the
beginning of an utterance.
\begin{example}{d93-15.2 utt9}
\label{ex:d93-15.2:utt9}
it takes an hour to load them \% \\
just so you know \%
\end{example}
Hence, we see a tendency that discourse marker usage strongly
correlates with the ambiguous words being used at the beginning of an
utterance.  

As further evidence, consider the following example.
\begin{example}{d93-11.1 utt109-111}
\makebox[\system][l]{\bf system:} so so we have three boxcars of oranges 
at Corning \\
\makebox[\system][l]{\bf user:}   three boxcars of orange juice at Corning \\
\makebox[\system][l]{\bf system:} no um oranges
\end{example}
In the third turn of this example, the system is not using ``no'' as a
quantifier to mean that there are not any oranges available; rather,
she is using ``no'' as a discourse marker to signal that the system
is rejecting the user's prior utterance and is indicating that the
user misrecognized ``oranges'' as ``orange juice''. This reading is
made clear in the spoken version by a clear intonational boundary
between the words ``no'' and ``oranges''.  In fact, the recognition of
the intonational boundary facilitates the identification of ``no'' as
a discourse marker, since the determiner reading of ``no'' is unlikely
to have an intonational boundary separating it from the noun it
modifies.  Likewise, the recognition of ``no'' as a discourse marker,
and in fact an acknowledgment, makes it more likely that there will be
an intonational boundary tone following it.  Hirschberg and Litman
\shortcite {HirschbergLitman93:cl} propose further constraints on how
discourse marker disambiguation interacts with intonational cues.

\subsection{Speech Repairs and Discourse Markers}
\label{sec:intro:repairs_dm}

Speech repair detection and correction is also highly intertwined with
discourse marker identification.  Discourse markers are often used in
the editing term to help signal that a repair occurred, and can be
used to help determine if it is a fresh start (cf.~\cite
{Hindle83:acl}).  The following example illustrates ``okay'' being
used as a discourse marker to signal a speech repair.
\begin{example}{d92a-4.2 utt62}
\label{ex:d92a-4.2:utt62}
\reparandum{I don't know if the}\ip \et{okay} that'll be three hours right
\end{example}
For this example, recognizing that ``okay'' is being used as a
discourse marker following the word ``the'' facilitates the detection
of the repair.  Likewise, recognizing that the interruption point of a
repair follows the word ``the'' gives evidence that ``okay'' is being
used as a discourse marker.  Discourse markers can also be used in
determining the start of the reparandum for fresh starts, since they are
often utterance initial.

\section{POS Tagging and Speech Recognition}
\label{sec:intro:tagging}

Not only are the problems of resolving speech repairs,
identifying boundary tones, and identifying discourse markers highly
intertwined, but these three problems are also intertwined with two
additional problems: identifying the lexical category or
part-of-speech (POS) of each word, and the speech recognition problem
of predicting the next word given the previous context.

\subsection{POS Tagging}

Just as POS taggers for written text take advantage of sentence
boundaries, it is natural to assume that in tagging spontaneous speech
we would benefit from taking into account the occurrence of
intonational phrase boundary tones and interruption points of speech
repairs.  This is especially true for speech repairs, since the
occurrence of these events disrupts the local context that is needed to
determine the POS tags \cite{Hindle83:acl}.  To illustrate the
dependence of POS tagging on speech repair identification, consider
the following example.
\begin{example}{d93-12.4 utt44}
\label{ex:d93-12.4:utt44}
by the time we \reparandum{load in}\ip load the bananas
\end{example}
Here, the second instance of ``load'' is being used as a present tense
verb, exactly as the first instance of ``load'' is being used.
However, in the Trains corpus, ``load'' is also commonly used as a
noun, as in ``a load of oranges''.  Since the second instance of
``load'' follows a preposition, it could easily be mistaken as a noun.
Only by realizing that it follows the interruption point of a speech
repair and it corresponds to the first instance of ``load'' will it be
properly tagged as a present tense verb.  Conversely, since speech
repairs disrupt the local syntactic context, this disruption, as
captured by the POS tags, can be used as evidence that a speech repair
occurred.  In fact, for the above example of a preposition followed by
a present tense verb, no fluent examples were observed in the Trains
corpus.

Just as POS tagging is intertwined with speech repair modeling, the
same applies to boundary tones.  Since speakers have flexibility as to
how they segment speech into intonational phrases, it is difficult to
find examples as illuminating as Example~\ref {ex:d93-12.4:utt44}
above.  The clearest examples deal with distinguishing between
discourse marker usage and a sentential interpretation, as we
illustrated in Section~\ref{sec:intro:tones_dm}.  Deciding whether
``so'' is being used as a subordinating conjunct, an adverb, or a
discourse conjunct is clearly related to identifying the intonational
boundaries.

\subsection{Speech Recognition}

Modeling the occurrences of boundary tones, speech repairs and
discourse markers also has strong interactions with the speech
recognition task of predicting the next word given the previous
words.\footnote{Section~\ref {sec:related:wordmodel} explains the
speech recognition problem of predicting the next word given the
previous words.}  Obviously, the word identities are an integral part
of predicting boundary tones, speech repairs and discourse markers.
However, the converse is also true.  The occurrence of a boundary tone
or interruption point of a speech repair affects what word will occur
next.  After a speech repair, the speaker is likely to retrace some of
the prior words, and hence modeling speech repairs will allow this
retracing to be used in predicting the words that follow a repair.
After a boundary tone, she is likely to use words that can introduce a
new utterance, such as a discourse marker.  Already, some preliminary
work has indicated the fruitfulness of modeling speech repairs
\cite{StolckeShriberg96:icassp} and utterance boundaries
\cite{MeteerIyer96:emnl} as part of the speech recognition problem.

\section{Thesis}

In this thesis, we address the problem of modeling speakers'
utterances in spoken dialog.  This involves identifying intonational
phrase boundary tones, identifying discourse markers, and detecting
and correcting speech repairs.  Our thesis is that this can be done
using local context and early in the processing stream.  Hearers are
able to resolve speech repairs and boundary tones very early on, and
hence there must be enough cues in the local context that make this
feasible.  Second, we claim that all three tasks need be done together
in a framework in which competing hypotheses for the speaker's turn
can be evaluated.  In this way, the interactions between these three
problems can be modeled.  Third, these tasks are highly intertwined
with determining the syntactic role or POS tag of each word, as well
as the speech recognition task of predicting the next word given the
context of the preceding words.  Hence, in this thesis, we propose a
statistical language model suitable for speech recognition that not
only predicts the next word, but also assigns the POS tag, identifies
boundary tones and discourse markers, and detects and corrects speech
repairs.  Since all of the tasks are being done in a unified framework
that can evaluate alternative hypotheses, the model can account for
the interactions between these tasks.  Not only does this allow us to
model the speaker's utterance, but it also results in an improved
language model, evidenced by both improved POS tagging and in better
estimating the probability of the next word.  Thus, this model can be
incorporated into a speech recognizer to even help improve the
recognition of spoken dialog.  Furthermore, speech repairs and
boundary tones have acoustic correlates, such as pauses between words.
By resolving speech repairs and boundary tones during speech
recognition, these acoustic cues, which otherwise would be treated as
noise, can give evidence as to the occurrence of these events.

By resolving the speaker's utterances early on, this will not only
help a speech recognizer determine what was said, but it will also
help later processing, such as syntactic and semantic analysis.  The
literature (e.g.~\cite {BearPrice90:acl,Ostendorf-etal93:csl}) already
indicates the usefulness of intonational information for syntactic
processing.  Speech repair resolution will also prove useful for later
syntactic processing.  Previous methods for syntactic analysis of
spontaneous speech have focused on robust parsing techniques that try
to parse as much of the input as possible and simply skip over the
rest (e.g.~\cite{Ward91:icassp}), perhaps with the aid of pragmatic
and semantic information (e.g.~\cite{YoungMatessa91:eurospeech}).  By
modeling speech repairs, the apparent ill-formedness that these cause
can now be made sense of, allowing richer syntactic and semantic
processing to be done on the input.  This will also make it easier for
later processing to cope with the added syntactic and semantic
variance that spoken dialog seems to license.\footnote {One way for
the syntactic and semantic processes to take into account the
occurrence and correction of speech repairs is for them to skip over
the reparandum and editing terms.  However, as Footnote~\ref
{ft:intro:reparandum} shows, this is not always advisable.}

Like all work in spoken language processing, top-down information is
important.  Although POS information only provides a shallow
syntactic analysis of the words in a turn, richer syntactic and
semantic analysis would be helpful.  Our model could operate in
lockstep with a statistical parser and provide the base probabilities
that it needs \cite {Charniak93:book}.  Another approach is to use a
richer tagset that captures higher level syntactic information as is
done by \cite {JoshiSrinivas94:coling}.

\section{Related Work}

\subsection{Utterance Units and Boundary Tones}

There have been a number of attempts to automatically identify
utterance unit boundaries and boundary tones.  For detecting
boundary tones, one source of information is the presence of
preboundary lengthening and pausal durations, which strongly correlate
with boundary tones in read speech \cite{Wightman-etal92:jasa}.
Wightman and Ostendorf \shortcite {WightmanOstendorf94:ieee} use
these cues as well as other cues to automatically detect boundary
tones as well as pitch accents in read speech. Wang and Hirschberg
\shortcite {WangHirschberg92:csl} take a different approach.  They
make use of knowledge inferable from its text representation as well
as some intonational features to predict boundary tones.  Like
Wightman and Ostendorf, they use a decision tree \cite
{Breiman-etal84:book} to automatically learn how to combine these
cues.  Kompe \etal~\shortcite{Kompe-etal94:icassp} present an
algorithm for automatically detecting prosodic boundaries that
incorporates both an acoustic model and a word-based language model.
Mast \etal~\shortcite {Mast-etal96:icslp} investigate finding speech
acts segments using a method that also combines acoustic modeling with
language modeling.  Meteer and Iyer \shortcite{MeteerIyer96:emnl},
working on the Switchboard corpus, incorporate the detection of
linguistic segments into the language model of a speech recognizer,
and find that this improves the ability of the language model to
predict the next word.  Expanding on the work of Meteer and Iyer,
Stolcke and Shriberg \shortcite{StolckeShriberg96:icslp} found that to
add linguistic segment prediction to a language model, it is best to
include POS information and discourse markers.  However, they treat
the POS tags and discourse marker usage as part of their input.

\subsection{Speech Repairs}

Previous work in speech repairs has examined different approaches to
detecting and correcting speech repairs.  One of the first was Hindle
\shortcite {Hindle83:acl}, who added grammar rules to a deterministic
parser to handle speech repairs.  This work was based on research that
indicated that the alteration of a speech repair replaces speech of
the same category.  However, Hindle assumed an edit signal would mark
the interruption point, a signal that has yet to be found.  Another
approach, taken by Bear \etal~\shortcite {Bear-etal92:acl}, uses a
pattern matcher to look for patterns of matching words.  In related
work, Dowding \etal~\shortcite {Dowding-etal93:acl} employed a
parser-first approach.  If the parser and semantic analyzer are unable
to make sense of an utterance, they look for speech repairs using the
pattern matcher just mentioned.  Nakatani and Hirschberg \shortcite
{NakataniHirschberg94:jasa} have tried using intonational features to
detect speech repairs.  They used hand-transcribed features, including
duration, presence of fragments, presence of filled pauses, and
lexical matching.

Recent work has focused on modeling speech repairs in the language
model \cite
{Rosenfeld-etal96:icslp,StolckeShriberg96:icassp,SiuOstendorf96:icslp}.
However, the speech repair models proposed so far have been limited to
abridged repairs with filled pauses, simple repair patterns, and
modeling some of the editing terms.  With such limited models, only
small improvements in speech recognition rates have been observed.

\subsection{Discourse Markers}

Although numerous researchers
(e.g.~\cite{RCohen84:coling,Reichmanadar84:ai,Sidner85:ci,GroszSidner86:cl,LitmanAllen87,HirschbergLitman93:cl})
have noted the importance of discourse markers in determining
discourse structure, there has not been a lot of work in actually
identifying them.  Two exceptions are the work done by Hirschberg and Litman
\shortcite {HirschbergLitman93:cl}, who looked at how intonational
information can disambiguate lexical items that can either be a
discourse marker or have a sentential reading, and the work of Litman
\shortcite{Litman96:jair}, who used machine learning techniques to
improve on the earlier results.

\section{Contribution}

This thesis makes a number of contributions to the field of spoken
dialog understanding.  The first contribution is that it shows that
the problems in modeling speakers' utterances---segmenting a
speaker's turn into intonational phrases, detecting and correcting
speech repairs and identifying discourse markers---are all highly
intertwined.  Hence a uniform model is needed in which various
hypotheses can be evaluated and compared.  Our statistical language
model provides such a solution.  Since the model allows these issues
to be resolved online, rather than waiting for the end of the
speaker's turn, it can be used in a spoken dialog system that uses a
natural turn-taking mechanism and allows the user to engage in a
collaborative dialog.

The second contribution of this thesis is that it explicitly accounts
for the interdependencies between modeling speakers' utterances, local
syntactic disambiguation (POS tagging) and the speech recognition task
of predicting the next word.  Furthermore, incorporating acoustic cues
that give evidence as to the occurrence of boundary tones and speech
repairs translates into improved speech recognition and POS tagging
results.  We find that by accounting for speakers' utterances we are
able to improve POS tagging by 8.1\% and perplexity (defined in
Section~\ref {sec:related:perplexity}) by 7.0\%

Third, we present a new model for detecting and correcting speech
repairs, which uses local context.  We find that we can detect and
correct 65.9\% of all speech repairs with a precision of 74.3\%.  This
shows that this phenomenon can be resolved for the most part before
syntactic and semantic processing, and hence should simplify those
processes.  Departing from most previous work, this thesis shows that
these two problems, that of detection and correction, should not be
treated separately.  Rather, the presence of a good correction should
be used as evidence that a repair occurs.  Furthermore, we flip the
problem of finding a correction around.  Rather than searching for the
best correction after identifying a repair, we instead choose the
repair interpretation that is most helpful in predicting the words
that follow the interruption point of the repair.  Thus our model for
correcting speech repairs can be used in the speech recognition task.

This work also shows the importance of modeling discourse markers in
spoken dialog.  Discourse markers, as we argue, are an intrinsic part
of modeling speakers' utterances, both the segmentation of turns into
utterances and the detection and correction of speech repairs.  Our
work shows that they can be incorporated into a statistical language
model of the speaker's utterances, and doing so improves the
performance of the model.  Additionally, by accounting for the
interactions with modeling intonational phrases and speech repairs, we
are able to improve the identification of discourse markers by 15.4\%.

Finally, this thesis presents a new way of doing language
modeling. Rather than choosing either a strict POS tagging model or a
word-based language model, this thesis presents a language model that
views word information as simply a refinement of POS information.
This offers the advantage of being able to access syntactic
information in the language model, while still being able to make use
of lexical information.  Furthermore, since POS tagging is viewed as
part of the speech recognition problem, the POS tags can be used by
later processing.

There is still more work that needs to be done.  With the exception of
silence durations, we do not consider acoustic cues.  This is
undoubtedly a rich source of information for detecting intonational
boundaries, the interruption point of speech repairs, and even
discourse markers.  Second, we do not make use of higher level
syntactic or semantic knowledge.  Having access to partial syntactic
and even semantic interpretation would give a richer model of
syntactic well-formedness, and so would help in detecting speech
repairs, which are often accompanied by a syntactic anomaly.  A richer
model would also help in correcting speech repairs since there are
sometimes higher level correspondences between the reparandum and
alteration.  Third, we still need to incorporate our work into a
speech recognizer.  Because of poor word error rates of speech
recognizers on spontaneous speech, all of our experiments have been
conducted using a written transcript of the dialog, with word
fragments marked.  Speech with disfluencies will prove problematic for
speech recognizers, since there is often an effect on the quality of
the words pronounced as well as the problem of detecting word
fragments.  There is also the problem of supplying an appropriate
language model for spoken language, one that can account for the
presence of speech repairs and the other phenomena, such as filled
pauses and editing terms, that often accompany them.  It is in this
last area that our work should prove relevant for research being done
in speech recognition.

\section{Overview of the Thesis}

Chapter~\ref {chapter:related} discusses the relevant previous work in
statistical language modeling, speech repair detection and correction,
and boundary tone and discourse marker identification. Chapter~\ref
{chapter:corpus} describes the Trains corpus, which is a corpus of
human-human task oriented dialogs set in a limited domain. This corpus
provides both a snapshot into an ideal human-computer interface that
is conversationally proficient, and a domain that is limited enough to
be of practical consideration for a natural language interface. We
also describe the annotation of speech repairs, boundary tones, POS
tags, and discourse markers.

Chapter~\ref {chapter:model} introduces our POS-based language model,
which also includes discourse marker identification.  POS tags (and
discourse markers) are introduced in a speech recognition language
model since they provide syntactic information that is needed for the
detection and correction of speech repairs and identification of
boundary tones. In this chapter, however, we argue that the
incorporation of POS tagging into a speech recognition language model
leads to better language modeling, as well as paves the way for the
eventual incorporation of higher level understanding in the speech
recognition process. In order to effectively incorporate POS tagging,
we make use of a decision tree learning algorithm and word clustering
techniques. These techniques are also needed in order to augment the
model to account for speech repairs and boundary tones.  This chapter
concludes with an extensive evaluation of the model and a comparison
to word-based and class-based approaches.  We also evaluate the effect
of incorporating discourse markers into our POS tagset.

The next three chapters augment the POS-based model.  Chapter~\ref
{chapter:detection} describes how the language model is augmented so
as to detect speech repairs and identify boundary tones. Chapter~\ref
{chapter:correction} adds the correction of speech repairs into the
language model. Chapter~\ref {chapter:acoustic} adds silence
information for detecting speech repairs and boundary tones.

Chapter~\ref {chapter:examples} presents sample runs of the full
statistical language model in order to better illustrate how it makes
use of the probability distributions to find the best interpretation.
Chapter~\ref {chapter:results} presents an extensive evaluation of the
model by contrasting the effects that modeling boundary tones, speech
repairs, discourse markers, and POS tagging have on each other and on
the speech recognition problem of predicting the next word.  The
chapter concludes with a comparison of the model with previous work.
Finally, Chapter~\ref {chapter:conclusion} presents the conclusions
and future work.

\cleardoublepage
\chapter{Related Work}
\label{chapter:related}

We start the literature review with an overview of statistical
language modeling.  We then review the literature on identifying
utterance unit boundaries and intonational phrase boundaries.  Next,
we review the literature on detecting and correcting speech
repairs. We conclude with a review of the literature on identifying
discourse markers.

For the literature on detecting and correcting speech repairs, and
identifying boundary tones and discourse markers, we standardize all
reported results so that they use {\em recall} and {\em precision}
rates.  The easiest way to define these terms is by looking at a
confusion matrix, as illustrated in Table~\ref{tab:related:confusion}.
\begin{table}
\begin{center}
\begin{tabular}{|l|l|c|c|} \hline
\multicolumn{2}{|c|}{}& \multicolumn{2}{c|}{Algorithm} \\ \cline{3-4}
\multicolumn{2}{|c|}{}& \multicolumn{1}{c|}{$x$} & \multicolumn{1}{c|}{$\bar{x}$} \\ \hline
Actual & $x$       & hits            & misses             \\ \cline{2-4}
       & $\bar{x}$ & false positives & correct rejections \\ \hline
\end{tabular}
\end{center}
\caption{Confusion Table for Defining Recall, Precision, and Error Rates}
\label{tab:related:confusion}
\end{table}
Confusion matrices contrast the performance of an algorithm in
identifying an event, say $x$, against the actual occurrences of the
event.  The recall rate of identifying an event is the number of times
that the algorithm correctly identifies it over the total number of
times that it actually occurred.
\[ \mbox{recall} = \frac{\mbox{hits}}
                        {\mbox{hits}+\mbox{misses}} \]
The precision rate is the number of times the algorithm correctly
identifies it over the total number of times it identifies it.
\[ \mbox{precision} = \frac{\mbox{hits}}
                           {\mbox{hits}+\mbox{false positives}} \]
For most algorithms, recall and precision trade off against each
other.  So, we also use a third metric, the {\em error rate}, which we
define as the number of errors in identifying an event over the number
of times that the event occurred.\footnote{The error rate is typically
defined as the number of errors over the total number of events.
However, for low occurring events, this gives a misleading impression
of the performance of an algorithm.}
\[ \mbox{error} = \frac{\mbox{misses}+\mbox{false positives}}
		       {\mbox{hits}+\mbox{misses}} \]
We standardize all reported results (where possible) to use recall and
precision so that the low occurrence of boundary tones, speech repairs
and discourse markers does not hide the performance or lack of
performance in doing these tasks.  For instance, if intonational
phrases occur once every ten words, an algorithm that always guesses
``no'' would be right 90\% of the time, but its recall rate would be
zero.

\section{Statistical Language Modeling}
\label{sec:related:slm}

The first area that we explore is statistical language modeling.  We
start with word-based language models, which are used extensively in
the speech recognition community to help prune out alternatives
proposed by acoustic models.  Statistical language models have also
been used for the task of POS tagging, in which each word in an
utterance is assigned its part-of-speech tag, or syntactic category.
Statistical language models, of course, require probability
distributions.  Hence, we next explore different methods that have
been used for estimating the probabilities involved.  We conclude this
section with a brief discussion of how these probabilities can be used
to find the best interpretation.
 
\subsection{Word-based Language Models}
\label{sec:related:wordmodel}
\label{sec:related:perplexity}

From the standpoint of speech recognition, the goal of a language
model is to find the {\em best} sequence of words $\hat{W}$ given the
acoustic signal $A$.  Using a probabilistic interpretation, we define
`best' as {\em most probable}, which gives us the following equation
\cite {RabinerJuang93}.
\begin{eqnarray}
\hat{W} = \arg\max_{W} \Pr(W|A)
\label{eqn:related:wordmodel:1}
\end{eqnarray}
Using Bayes' rule, we rewrite the above equation in the following
manner.
\begin{eqnarray}
\hat{W} & = & \arg\max_{W} \frac{\Pr(A|W)\Pr(W)}{\Pr(A)} 
\label{eqn:related:wordmodel:2}
\end{eqnarray}
Since $\Pr(A)$ is independent of the choice of $W$, we simplify the
above as follows.
\begin{eqnarray}
\hat{W} & = & \arg\max_{W} \Pr(A|W)\Pr(W)
\label{eqn:related:wordmodel:3}
\end{eqnarray}
The first term, $\Pr(A|W)$, is the probability attributable to the
acoustic model and the second term, $\Pr(W)$, is the probability
attributable to the language model, which assigns a probability to the
sequence of words $W$.  We can rewrite $W$ explicitly as a sequence of
words $W_1W_2W_3
\dots W_N$, where $N$ is the number of words in the sequence.  For
expository ease, we use the notation $W_{i,j}$ to refer to the
sequence of words from $W_i$ through to $W_j$.  We can now use the
definition of conditional probabilities to rewrite $\Pr(W_{1,N})$ as
follows.
\begin{eqnarray}
\Pr(W_{1,N}) &=& \prod_{i=1}^N \Pr(W_i|W_{\rim})
\label{eqn:related:wordmodel:4}
\end{eqnarray}

The above equation gives us the probability of the word sequence as
the product of the probability of each word given its previous lexical
context.  Of course, there is no way to know the actual probabilities.
The best we can do is to come up with an estimated probability
distribution $\hat{\Pr}(W_i|W_{\rim})$.  Different techniques for
estimating the probabilities will affect how well the model performs.
Since the probability distribution is intended to be used by a speech
recognizer, one could measure the effectiveness of the probability
distribution by measuring the speech recognition word error rate.
However, this makes the evaluation specific to the implementation of
the speech recognizer and the interaction between the acoustic model
and the language model.

A second alternative for measuring the effectiveness of the estimated
probability distribution is to measure the {\em perplexity} that it
assigns to a test corpus \cite {Bahl-etal77:asa}.  Perplexity is an
estimate of how well the language model is able to predict the next
word of a test corpus in terms of the number of alternatives that need
to be considered at each point.  For word-based language models, the
perplexity of a test set of $N$ words $w_{1,N}$ is calculated as
$2^H$, where $H$ is the entropy, which is defined as follows.
\begin{eqnarray}
H & = & -\frac 1 N \sum_{i=1}^N \log_2 \hat{\Pr}(w_i|w_\rim)
\end{eqnarray}
The best approach to measure the effectiveness of a language model
intended for speech recognition is to measure both the word error rate
and the perplexity.

\subsection{POS Tagging}
\label{sec:related:tagger}

Before examining techniques for estimating the probabilities, we first
review POS tagging.  POS tagging is the process of finding the best
sequence of category assignments $P_{1,N}$ for the sequence of words
$w_{1,N}$.\footnote{We use lower case letters to refer to the word
sequence to denote that the word has a given value.}  Consider the
sequence of words ``hello can I help you''.  Here we want to determine
that ``hello'' is being used as an acknowledgment, ``can'' as a modal
verb, ``I'' as a pronoun, ``help'' as an untensed verb, and ``you'' as
a pronoun.\footnote{Section~\ref {sec:corpus:pos} presents the tagset
that we use.}
  
As with word-based language models, one typically adopts a
probabilistic approach and defines the problem as finding the category
assignment that is most probable given the sequence of words \cite
{DeRose88:cl,Church88:anlp,Charniak-etal93:aaai}.
\begin{eqnarray}
\hat{P}_{1,N} = \arg\max_{P_{1,N}} \Pr(P_{1,N} | w_{1,N})
\label{eqn:related:tagger:1}
\end{eqnarray}
Using the definition of conditional probabilities, we can rewrite this
as follows.
\begin{eqnarray}
\hat{P}_{1,N} = \arg\max_{P_{1,N}} \frac{\Pr(w_{1,N}P_{1,N})}{\Pr(w_{1,N})}
\label{eqn:related:tagger:2}
\end{eqnarray}
Since $\Pr(w_{1,N})$ is independent of the choice of the category
assignment, we can ignore it and thus equivalently find
the following.
\begin{eqnarray}
\hat{P}_{1,N} = \arg\max_{P_{1,N}} \Pr(w_{1,N} P_{1,N})
\label{eqn:related:tagger:3}
\end{eqnarray}
Again, using the definition of conditional probabilities, we rewrite
the probability as a product, as we did above in
Equation~\ref{eqn:related:wordmodel:4} for the word-based language
model.  Here, however, we have two probability distributions: the
lexical probability and the POS probability.
\begin{eqnarray}
\Pr(w_{1,N} P_{1,N}) = \prod_{i=1}^N \Pr(w_i|w_\rim P_\ri)
\Pr(P_i|w_\rim P_\rim) \label{eqn:related:tagger:4}
\end{eqnarray}

For POS taggers, the common practice is to just have the lexical
probability be conditioned on the POS category of the word, and the
POS probability conditioned on the preceding POS tags, which leads to
the following two assumptions.\footnote {Two notable exceptions are
the work of Black \etal~\shortcite{Black-etal92:darpa:pos} and
Brill~\shortcite {Brill95:cl}.  Black \etal~used the POS tags of the
previous words and the words that follow to predict the POS tag.  They
used a decision tree algorithm to estimate the probability
distribution.  Brill learned a set of symbolic rules to apply to the
output of a probabilistic tagger.  These rules could look at the local
context, namely the POS tags and words that precede and follow the POS
tag under consideration.}
\begin{eqnarray}
\Pr(w_i|w_\rim P_\ri)  &\approx& \Pr(w_i|P_i)    
\label{eqn:related:tagger:a1}\\
\Pr(P_i|w_\rim P_\rim) &\approx& \Pr(P_i|P_\rim) 
\label{eqn:related:tagger:a2}
\end{eqnarray}
This leads to the following approximation of Equation~\ref
{eqn:related:tagger:4}.
\begin{eqnarray}
\Pr(w_{1,N} P_{1,N}) &\approx& \prod_{i=1}^N \Pr(w_i|P_i) \Pr(P_i|P_\rim)
\label{eqn:related:tagger:f}
\end{eqnarray}

For POS tagging, rather than use perplexity, the usual approach for
measuring the quality of the probability estimates is to actually use
them in a POS tagger and measure the POS error rate.

\subsection{Sparseness of Data}
\label{sec:related:sparseness}

No matter whether one is doing POS tagging or word-based language
modeling, one needs to estimate the conditional probabilities used in
the above formulae.  The simplest approach to estimating the
probability of an event given a context is to use a training corpus
and simply compute the relative frequency of the event given the
context.  However, no matter how large the training corpus is, there
will always be event-context pairs that have not been seen, or that
have been seen too rarely to accurately estimate the probability.  To
alleviate this problem, one must partition the contexts into a
smaller number of equivalence classes.  For word-based models, a
common technique for estimating the probability $\Pr(W_i|W_\rim)$ is
to partition $W_\rim$ into contexts based on the last few words.  If
we consider the $n$-$1$ previous words then the context for estimating the
probability of $W_i$ is $W_{\mbox{$i$-$n$+1,$i$-1}}$.  The literature
refers to this as an $n$-gram language model.

We can also mix in smaller size language models when there is not
enough data to support the larger context.  Below, we present the two
most common approaches for doing this: interpolated estimation \cite
{JelinekMercer80} and the backoff approach \cite
{Katz87:assp}.\footnote {See Chen and Goodman \shortcite
{ChenGoodman96:acl} for a review and comparison of a number of
smoothing algorithms for word models.}\fnote{Rosenfeld also talks
about using interpolated estimation to mix in other sources of data.}

\subsubsection{Interpolated Estimation}

Consider probability estimates based on unigrams $\Pr_1(W_i)$, bigrams
$\Pr_2(W_i|W_{\im})$ and trigrams $\Pr_3(W_i|W_{\im}W_{\imm})$.
Interpolated estimation of these probability estimates involves mixing
these together, such that $\lambda_3 + \lambda_2 +
\lambda_1 = 1$.
\begin{eqnarray}
\Pr(W_i|W_{\im}W_{\imm})&=&\lambda_3 {\Pr}_3(W_i|W_{\imm}W_{\im}) + 
\lambda_2 {\Pr}_2(W_i|W_{\im}) + 
\lambda_1 {\Pr}_1(W_i) \hspace{3em}
\end{eqnarray}
The forward-backward algorithm can be used to automatically calculate
the values of the lambdas.  This is an iterative algorithm in which
some starting point $\lambda_j^0$ must be specified.  Below, we give
the formula for how $\lambda_j^{k+1}$ is computed from the values of
$\lambda_l^k$, and where $w_{1,N}$ is training data for estimating the
lambdas.
\begin{eqnarray}
\lambda_j^{k+1} &=& 
\sum_{i=1}^N \frac{\lambda_j^k \Pr_j(w_i|w_{\imm}w_{\im})}
                  {\sum_{l=1}^3 \lambda_l^k \Pr_l(w_i|w_{\imm}w_{\im})}
\end{eqnarray}
The training data for estimating the lambdas should not be the same
data that was used for estimating the probability distributions
$\Pr_j$; for otherwise, the lambdas will be biased and not suitable
for estimating unseen data.

One of the strengths of interpolated estimation is that the lambdas
can depend on the context $W_{\imm}W_{\im}$, thus allowing more
specific trigram information to be used where warranted.  Here, one
defines equivalence classes (or buckets) of the contexts, and each
bucket is given its own set of lambdas.  Brown~\etal~\shortcite
{Brown-etal92:cl} advocate bucketing the context based solely on the
counts of $W_{\imm}W_{\im}$ in the training corpus.  If the count is
high, the corresponding trigram estimates should be reliable, and
where they are low, they should be much less reliable.  Another
approach for bucketing the lambdas is to give each context its own
lambda.  For contexts that occur below some minimum number of times in
the training corpus, these can be grouped together to achieve the
minimum number. With this approach, the lambdas can be
context-sensitive.

\subsubsection{Backoff Approach}

The second approach for mixing in smaller order language models is the
backoff approach \cite{Katz87:assp}.  This scheme is based on
computing the probabilities based on relative frequency, except that
the probability of the higher-order $n$-grams is discounted and this
probability mass is redistributed to the lower order $n$-grams.  The
discounting method is based on the Good-Turning formula.  If an event
occurs $r$ times in a training corpus, the corrected frequency is
defined as follows where $n_r$ is the number of events that occur $r$
times in the training data.\footnote{Katz proposes a further
modification of this formula so that just events that occur less than
a specific number of times, for instance 5, are discounted.}
\[ r^* = (r+1) \frac{n_{r+1}}{n_r} \] 
The probability for a seen $n$-gram is now computed as follows, where
$c(X)$ is the number of times that $X$ appears in the training corpus
and $c^*(X)$ is the discounted number, as given above.
\[ \Pr(W_i|W_{i-n+1,i-1}) = 
       \frac{c^*(W_{i-n+1,i})}
            {c(W_{i-n+1,i-1})} \]
The leftover probability is then distributed to the $n$-$1$-grams.
These probabilities are also discounted and the weight distributed
to the $n$-$2$-grams, and so on.

\subsection{Class-Based Models}
\label{sec:related:class}

The choice of equivalence classes for a language model need not be the
previous words. Words can be grouped into classes, and these classes
can be used as the basis of the equivalence classes of the context
rather than the word identities \cite{Jelinek85}. Below we give the
equation that is usually used for a class-based trigram model, where
the function $g$ maps each word to its unambiguous class.  
\begin{eqnarray}
\Pr(W_i|W_\rim) &\approx& \Pr(W_i|g(W_i))\Pr(g(W_i)|g(W_{i-1}) g(W_{i-2}))
\end{eqnarray}
This has the potential of reducing the problem of sparseness of data
by allowing generalizations over similar words, as well as reducing
the size of the language model.

Brown \etal~\shortcite{Brown-etal92:cl} propose a method for
automatically clustering words into classes.  The classes that they
want are the ones that will lead to high mutual information between
the classes of adjacent words.  In other words, for each bigram
$w_{\im}w_i$ in a training corpus, one should choose the classes such
that the classes for adjacent words $g(w_{\im})$ and $g(w_i)$ lose as
little information about each other as possible.  They propose a
greedy algorithm for finding the classes.  They start with each word
in a separate class and iteratively combine classes that lead to the
smallest decrease in mutual information between adjacent words.  They
suggest that once the required number of classes has been achieved,
the greedy assignment can be improved by swapping words between
classes that leads to an increase in mutual information.  For the
Brown corpus, they were able to achieve an decrease in perplexity from
244 for a word-based trigram model to 236, but only by interpolating
the class-based model with the word-based model.  The class-based
model on its own, using 1000 classes, resulted in a perplexity of 271.

The Brown~\etal~algorithm can also be used for constructing a
hierarchy of the words. Rather than stopping at a certain number of
classes, one keeps merging classes until only a single class remains.
However, the order in which classes are merged gives a hierarchical
binary tree with the root corresponding to the entire vocabulary and
each leaf to a single word of the vocabulary.  Intermediate nodes
correspond to groupings of the words that are statistically similar.
We will be further discussing these trees in Sections~\ref
{sec:related:dt:words} and~\ref {sec:model:ctrees}.

The clustering algorithm of Brown~\etal~does not have a mechanism
to decide the optimal number of classes.  Using too few classes will
cause important lexical distinctions to be lost, and probably
partially accounts for the poor performance they report for the
class-based approach on its own.  In fact, as Kneser and
Ney~\shortcite {KneserNey93:eurospeech} point out, each merge
supposedly results in a loss of information.  However, this is only
because the same data is used for choosing which class to merge as is
used for estimating the resulting change in mutual information.
Hence, Kneser and Ney present a clustering algorithm that uses
heldout-data, as simulated by the {\em leaving-one-out} method
\cite{DudaHart73:book}.  The algorithm is thus able to determine an
optimal number of classes, as well as the word to class mapping.  The
actual form of their algorithm differs from that of the
Brown~\etal~algorithm.  The Kneser and Ney algorithm initially assigns
all words to a single class.  The algorithm then iteratively looks for
the best word-class combination (the class can be empty) such that
moving the word to that class results in the best {\em increase} in
mutual information as estimated by the leaving-one-out
cross-validation procedure.  They find that using the cross-validation
technique results in a class assignment that gives equivalent
perplexity results as can be achieved using the optimal results of not
using the cross-validation procedure (as determined empirically with
the test data).  Furthermore, they find that using a class-based
language model, without interpolating a word-based model, results in a
perplexity improvement for the LOB corpus from 541 for a word-based
bigram model to 478 for a class-based bigram model.  Interpolating the
word-based and class-based models resulted in a further improvement to
439.

\subsection{POS-Based Models}
\label{sec:related:aclass}

Classes can also be ambiguous, such as when POS tags are used
\cite{Jelinek85}.  Here, one needs to sum over all of the POS
possibilities.  Below, we give the derivation for the language model
equations based on using trigrams.
\begin{eqnarray}
\lefteqn{\Pr(W_i|W_{\rim})} \nonumber \\
& = & 
\sum_{P_{\imm,i}} \Pr(W_iP_{\imm,i}|W_{\rim}) \nonumber \\
& = & 
\sum_{P_{\imm,i}} 
\Pr(W_i|P_{\imm,i}W_{\rim})
\Pr(P_i|P_{\imm,\im}W_{\rim})
\Pr(P_{\imm,\im}|W_{\rim}) \nonumber \\
& \approx & 
\sum_{P_{\imm,i}} 
\Pr(W_i|P_i) 
\Pr(P_i|P_{\imm,\im}) 
\Pr(P_{\imm,\im}|W_{\rim}) \label{eqn:related:jel}
\end{eqnarray}
The final equation \cite[Equation C.16]{Jelinek85} makes use of the
approximations that are used for POS tagging, in which the lexical
probability is conditioned on just its POS tag and the POS tag
probability of the current word is conditioned on the POS tags of the
previous words.  Note that there is an extra factor
involved: $\Pr(P_{\imm}P_{\im}|W_{\rim})$.  This factor takes into account
the different likelihoods for the POS tags of the previous two words and
can be computed recursively as follows.
\begin{eqnarray}
\lefteqn{\Pr(P_{\imm}P_{\im}|W_{\rim})} \nonumber \\
& = & \sum_{P_{\immm}} 
	\Pr(P_{\immm,\im}|W_{\rim}) \nonumber \\
& = & \sum_{P_{\immm}} 
	\frac{\Pr(W_{\rim}|P_{\immm,\im}) 
	      \Pr(P_{\immm,\im})}
             {\Pr(W_{\rim})} \nonumber \\
& = & \sum_{P_{\immm}} 
	\frac{\Pr(W_{\im}|P_{\immm,\im}W_{\rimm})
	      \Pr(W_{\rimm}|P_{\immm,\im}) 	
	      \Pr(P_{\immm,\im})}
	     {\Pr(W_{\rim})} \nonumber \\
& = & \sum_{P_{\immm}}
	\Pr(W_{\im}|P_{\immm,\im}W_{\rimm})
	\Pr(P_{\immm,\im}|W_{\rimm}) 
	\frac{\Pr(W_{\rimm})}
	     {\Pr(W_{\rim})} \nonumber \\
& = & \sum_{P_{\immm}}
	\Pr(W_{\im}|W_{\rimm}P_{\immm,\im})
	\Pr(P_{\im}|P_{\immm,\imm}W_{\rimm}) 
	\Pr(P_{\immm,\imm}|W_{\rimm}) 
	\frac{\Pr(W_{\rimm})}
	     {\Pr(W_{\rim})} \nonumber \\
& \approx & \sum_{P_{\immm}}
	\Pr(W_{\im}|P_{\im})
	\Pr(P_{\im}|P_{\immm,\imm}) 
	\Pr(P_{\immm,\imm}|W_{\rimm}) 
	\frac{\Pr(W_{\rimm})}
	     {\Pr(W_{\rim})} \label{eqn:related:jel2}
\end{eqnarray}
In Equation~\ref{eqn:related:jel2}, the factor
$\frac{\Pr(W_{\rimm})}{\Pr(W_{\rim})}$ can be viewed as a normalizing
constant that assures that the probability
$\Pr(P_{\imm,\im}|W_{\rim})$ adds to one when summed over all possible
values for $P_{\imm,\im}$ \cite{Jelinek85}.  Note that the extra
probabilities involved in the equation are the same ones used in
Equation~\ref {eqn:related:jel}.

A more direct way of deriving the equations for using POS tags is to
directly change the language model equation given in Equation~\ref
{eqn:related:wordmodel:4}.
\begin{eqnarray}
\Pr(W_{1,N}) 
&=& \sum_{P_{1,N}} \Pr(W_{1,N}P_{1,N}) \nonumber \\
&=& \sum_{P_{1,N}} \prod_{i=1}^N \Pr(W_iP_i|W_{\rim}P_{\rim}) \nonumber\\
&=& \sum_{P_{1,N}} \prod_{i=1}^N \Pr(W_i|W_{\rim}P_{\ri}) 
                                \Pr(P_i|W_{\rim}P_{\rim}) \nonumber \\
&\approx& \sum_{P_{1,N}} \prod_{i=1}^N \Pr(W_i|P_i) 
                                      \Pr(P_i|P_{\imm,\im})
                                      \label{eqn:related:jel3}
\end{eqnarray}
Although Equation~\ref{eqn:related:jel3} works out exactly the same as
Equation~\ref {eqn:related:jel} and Equation~\ref {eqn:related:jel2},
it more readily shows how POS tags can be added to the derivation of
the language model equations.

The above approach for incorporating POS information into a language
model has not been of much success in improving speech recognition
performance.  Srinivas \shortcite {Srinivas96:icslp} reports that such
a model results in a 24.5\% increase in perplexity over a word-based
model on the Wall Street Journal; Niesler and Woodland \shortcite
{NieslerWoodland96:icassp} report an 11.3\% increase (but a 22-fold
decrease in the number of parameters of such a model) for the LOB
corpus; and Kneser and Ney~\shortcite {KneserNey93:eurospeech} report
a 3\% increase on the LOB corpus.  The POS tags remove too much of the
lexical information that is necessary for predicting the next word.
Only by interpolating it with a word-based model is an improvement
seen \cite {Jelinek85}.

Classes containing even richer syntactic information than POS tags can
also be used.  Srinivas \shortcite{Srinivas96:icslp} presents a speech
recognition language model based on disambiguating {\em Supertags},
which are the elementary structures in Lexicalized Tree Adjoining
Grammars \cite {Shabes-etal88:coling}.  Supertags provide more
syntactic information than regular POS tags.  Joshi and Srinivas
\shortcite {JoshiSrinivas94:coling} refer to disambiguating supertags
as ``almost parsing'' since in order to get the full parse these
supertags must be linked together.  Using supertags as the basis of
the ambiguous classes, Srinivas \shortcite {Srinivas96:icslp} reported
a 38\% perplexity reduction on the Wall Street Journal over a trigram
word model.

\subsection{Using Decision Trees}
\label{sec:related:dt}

The above approaches to dealing with sparseness of data require the
language modeler to handcraft a backoff or interpolation strategy and
decide the equivalence classes for each language model involved.  As
Charniak \shortcite[pg.~49] {Charniak93:book} points out, ``one must
be careful not to introduce conditioning events \dots unless one has a
very good reason for doing so, as they can make the data even sparser
than necessary.''  An alternative approach, as advocated by Bahl
\etal~\shortcite {Bahl-etal89:tassp}, is to automatically learn how to
partition the context by using mutual information.  Here, one can use
a decision tree learning algorithm \cite {Breiman-etal84:book}.  The
decision tree learning algorithm starts by having a single equivalence
class (the root node) of all of the contexts.  It then looks for a
binary question to ask about the contexts in the root node in order to
partition the node into two {\em leaves}.  Information theoretic
metrics can be used to decide which question to ask: find the question
that results in the partitioning of the node that is most informative
as to which event occurred.  Brieman~\etal~discuss several measures
that can be used to rank the informativeness of a node, such as
minimizing entropy, which was used by Bahl~\etal~\shortcite
{Bahl-etal89:tassp}.

After a node is split, the resulting leaves should be better
predictors as to which event occurred.  The process of splitting nodes
continues with the new leaves of the tree and hence builds a
hierarchical binary partitioning of the context.  With this approach,
rather than trying to specify stopping criteria, Bahl~\etal\shortcite
{Bahl-etal89:tassp} recommend using heldout data to verify the
effectiveness of a proposed split.  The split is made only if the
heldout data agrees that the proposed split leads to a decrease in
entropy.  If the split is rejected, the node is not further explored.

After having grown a tree, the next step is to use the partitioning of
the context induced by the decision tree to determine the probability
estimates.  Using the relative frequencies in each node will be biased
towards the training data that was used in choosing the questions.
Hence, Bahl~\etal~smooth these probabilities with the probabilities of
the parent node using the interpolated estimation method with a
second heldout dataset, as described in Section~\ref
{sec:related:sparseness}.\footnote {Full details of applying
interpolated estimation to decision trees is given by Magerman
\shortcite {Magerman94:thesis}, as well as a more detailed overview of
the decision tree growing algorithm.}

Using the decision tree algorithm to estimate probabilities is
attractive since the algorithm can choose which parts of the context
are relevant, and in what order \cite{Bahl-etal89:tassp}.  This means
that the decision tree approach lends itself more readily to allowing
extra contextual information to be included.  If the extra information
is not relevant, it will not be used.

\subsubsection{Word Information}
\label{sec:related:dt:words}

In using a decision tree algorithm to estimate a probability
distribution that is conditioned on word information, such as
$W_{i{\dash}j}$, one must deal with the fact that these variables have a
large number of possible values, rather than just two values.  The
simplest approach is to allow the decision tree to ask questions of
the form `is $W_{i{\dash}j} = w$' for each $w$ in the lexicon.  However,
this approach denies the decision trees from forming any
generalizations between similar words.  Two alternative approaches
have been proposed in the literature that allow the decision tree to
ask questions of the form `is $W_{i-j} \in S$', where $S$ is a subset
of the words in the lexicon.

The first approach was used by Bahl \etal~\shortcite
{Bahl-etal89:tassp}, who dealt with word information as categorical
variables.  If $C$ is a categorical variable, the decision tree will
search over all questions of the form `is $C \in S$', where $S$ is a 
subset (or partition) of the values taken by $C$.  Since finding the
best partitioning involves an exponential search, they use a greedy
algorithm.  Start with $S$ being empty.  Search for the insertion into
$S$ that results in the greatest reduction in impurity.  Delete from
$S$ any member which results in a reduction in impurity.  Continue
doing this until no more insertions are possible into $S$.

The second approach \cite
{Black-etal92:darpa:hbg,Black-etal92:darpa:pos,Magerman94:thesis}
alleviates the problem of having the decision tree algorithm search for the
best partition; instead, the partitions are found as a preprocessing
step.  Here, one uses a clustering algorithm, such as the algorithm of
Brown \etal~\shortcite {Brown-etal92:cl} discussed in
Section~\ref{sec:related:class}.  Rather than search for a certain
number of classes of values, one continues merging classes until all
values are in a single class.  However, the order in which classes
were merged gives a hierarchical binary structure to the classes, and
thus an implicit binary encoding for each word, which is used for
representing the words to the decision tree algorithm.  The decision
tree algorithm can ask about which partition a word belongs to by
asking questions about the binary encoding.

Both of these approaches have advantages and limitations.  The first
approach can take into account the context when deciding the word
partitioning.  Depending on how the previous questions have divided up
the context, the optimal word partitioning might be different.
However, this is not without a drawback.  First, with each question
that is asked of the context, the amount of data available for
deciding the next question decreases, and hence there might not be
enough data for the decision tree to construct a good partitioning.
Second, having the decision tree decide the partitioning means that it
is quite limited at what information that it can use; in fact, it can
only make use of correlations with the variable that is being
predicted by the decision tree.  The work of Brown
\etal~in clustering words actually uses both the next word and the
previous word as features in clustering, which we feel yields more
informative classes and might transcend the limits of the local
optimization that the first approach affords.  In fact, any relevant
features can be used, rather than just those that fit into the
decision tree paradigm.  Third, having the decision tree partition the
word space complicates the decision tree algorithm and requires it to
perform much more computation.  With the second method, the word
partitioning is only learned once as a preprocessing step, rather than
being repeatedly learned while growing the decision tree.

\subsubsection{Results}
\label{sec:related:dt:results}

Bahl~\etal~\shortcite{Bahl-etal89:tassp} contrasted using decision
trees based on 21-grams (but only grown to 10,000 leaves) versus a
trigram interpolated language model.  Both models took about the same
amount of storage.  They found that for known words (words that were
in the training corpus), the tree based approach resulted in a
perplexity of 90.7 for a test corpus whereas the trigram model
achieved a perplexity of 94.9.  The tree model assigned 2.81\% of the
test words a probability less than $2^{-15}$, whereas the trigram
model assigned 3.87\% of the words such a probability.  This led the
authors to speculate that this would have a significant impact on the
error rate of a speech recognizer: ``speech recognition errors are
more likely to occur in words given a very low probability by the
language model.''  When they interpolated the decision tree model and
the word trigram model, the combined model achieved a perplexity of
82.5, and only 2.73\% of the words had a perplexity less than
$2^{-15}$.  This led them to speculate that the role of decision tree
language models might be to supplement rather than replace traditional
language models.

\subsection{Markov Assumption versus Pruning}
\label{sec:related:paths}

Once the probability estimates have been computed, the next issue is
how to keep the search for the best interpretation tractable.  To find
the best interpretation, one must search over all word sequences,
which will be an exponential search.  To make this computation
tractable, there are two alternatives.  The first alternative is to
make the Markov assumption.  Here we encode the context as one of a
finite number of states, which in fact is the same as using an
$n$-gram model for dealing with sparseness of data.  Thus the
probability of the next word simply depends on what state we currently
are in.  With this assumption, one can use the Viterbi algorithm to
find the most probable state sequence in time linear with the input
(and polynomial in the number of states).  As output, rather than just
returning the best path, a lattice can be returned, thus allowing
later processing to incorporate additional constraints to re-score the
alternatives.

With an $n$-gram language model, all possible sequences of the last
$n$-$1$ words are used to define the number of states.  For language
models above bigrams, this number becomes quite large.  For instance
for POS taggers, with tagset ${\cal P}$, the number of states in the
model is $|{\cal P}|^{n-1}$.  The Viterbi search then takes $|{\cal
P}|^n$ time.  Many of these alternatives are very unlikely.  Hence,
Chow and Schwartz \shortcite {ChowSchwartz89:darpa} only keep a small
number of alternative paths.  Rather than return a lattice, this
approach can return a set of paths as the final answer, which later
processing can re-score.

Speech recognizers, which must search over many different acoustic
alternatives, tend to make use of a bigram model during an initial or
first pass, in which acoustic alternatives are considered.  The result
of the first pass is usually a word lattice, with low scoring word
alternatives pruned.  The resulting lattice can then be evaluated by
a larger $n$-gram model in which only the language model scores are
re-computed.

\section{Utterance Units and Boundary Tones}
\label{sec:related:tones}

Research work on identifying utterance boundaries has followed several
different paths.  In order to give a rough comparison of their
performances, we will normalize the results so that they report on
turn-internal boundary detection.  Our reason for doing this is that
most approaches use the end-of-turn as evidence as to whether the end
of an utterance has occurred.  However, the end of the speaker's turn
is in fact jointly determined by both participants.  So when building
a system that is designed to participate in a conversation, one cannot
use the end of the user's turn as evidence that a boundary tone has
occurred.  

For utterance units defined by boundary tones, one approach to
detecting them is to make use of durational
cues. Price~\etal~\shortcite{Price-etal91:jasa} created a corpus of
structurally ambiguous sentences, read by professional FM public radio
announcers.  Trained labelers rated the perceived juncture between
words, using a range of 0 to 6 inclusive.  Break indices of 3
correspond to the intermediate phrases discussed in Section~\ref
{sec:intro:intonation}, and indices of 4, 5 and 6 correspond to
intonational phrases.  Wightman~\etal~\shortcite
{Wightman-etal92:jasa} found that preboundary lengthening and pausal
durations correlate with boundary types (no tone, phrase accent, or
boundary tone).  Preboundary lengthening can be measured by
normalizing the duration of the last vowel and the final consonants in
a word to take account of their normal duration.  

Wightman and Ostendorf \shortcite {WightmanOstendorf94:ieee} used the
cue of preboundary lengthening, pausal durations, as well as other
acoustic cues to automatically label intonational phrase endings as
well as word accents.  They trained a decision tree to estimate the
probability of a boundary type given the acoustic context.  These
probabilities were fed into a Markov model whose state is the boundary
type of the previous word.  For training and testing their algorithm,
they used a single-speaker corpus of radio news stories read by a
professional FM public radio announcer.\footnote{They also do
speaker-independent experiments on the ``ambiguous sentence corpus''
developed by Price~\etal~\shortcite{Price-etal91:jasa}.}  With this
speaker-dependent model using professionally read speech, they
achieved a recall rate of 78.1\% and a precision of
76.8\%.\footnote{From Table IV in their paper, we find that their
algorithm achieved 1447 hits (correct boundaries), 405 misses, and 438
false positives.  This gives a recall rate of $1447/(1447+405) =
78.1\%$, a precision rate of $1447/(1447+438) = 76.8\%$, and an error
rate of $(405+438)/(1447+405) = 45.5\%$.  In this experiment, there
was no indication that they used a cue based on end-of-story as a
feature to their decision tree.}  As is the case with the previous
work \cite {Wightman-etal92:jasa}, it is unclear how well this
approach will adapt to spontaneous speech, where speech repairs might
interfere with the cues that they use.

Wang and Hirschberg \shortcite{WangHirschberg92:csl} also looked at
detecting intonational phrase endings, running after syntactic
analysis has been performed.  Using automatically-labeled features,
such as category of the current word, category of the constituent
being built, distance from last boundary, and presence of observed
word accents, they built decision trees using CART
\cite{Breiman-etal84:book} to automatically classify the presence of a
boundary tone.  Rather than use the relative frequencies of the events
in the leaf node to compute a probability distribution as was
explained in Section~\ref{sec:related:dt}, the event that occurs most
often in the leaf is used to classify the test data.  With this
method, they achieved a (cross-validated) recall rate of 79.5\% and a
precision rate of 82.7\% on a subset of the ATIS corpus.  When we
exclude the end-of-turn data, we arrive at a recall rate of 72.2\% and
a precision of 76.2\%.\footnote {The recall and precision rates were
computed from Table 1 in their paper, in which they give the confusion
table for the classification tree that is most successful in
classifying observed boundary tones.  This particular tree uses
observed (hand-transcribed) pitch accents and classifies the 424
disfluencies in their corpus as boundary tones.  This tree identified
895 of the boundaries (hits), incorrectly hypothesized 187 boundaries
(false positives) and missed 231 boundaries.  This gives a recall rate
of $895/(895+231)=79.5\%$ and a precision rate of
$895/(895+187)=82.7\%$.  These results include 298 end-of-turn events.
The first node in their tree queries whether the time to the end of
the utterance of the current word is less then 0.04954 seconds.  This
question separates exactly 298 events, which thus must be all of the
end-of-turn events.  (In the two decision trees that they grew that
did not include the variable that indicates the time to the end of the
utterance, the end-of-turn events were identified by the first node by
querying whether the accent type of the second word was `NA', which
indicates end-of-turn.)  Of the 298 end-of-turn events, 297 have a
boundary tone.  To compute the effectiveness of their algorithm on
turn-internal boundary tones, we ignore the 297 correctly identified
end-of-turn boundary tones, and the one incorrectly hypothesized
boundary tone.  This gives 598 hits, 187 false positives, and 230
misses, giving a recall rate of $598/(598+230)=72.2\%$ and precision
of $598/(598+187)=76.2\%$.}  Note, however, that these results group
disfluencies with boundary tones.

Kompe \etal~\shortcite{Kompe-etal94:icassp}, as part of the Verbmobil
project \cite{Wahlster93}, propose an approach that combines acoustic
cues with a statistical language model in order to predict boundary
tones.  Their acoustic model makes use of normalized syllable
duration, length of interword pauses, pitch contour, and maximum
energy.  These acoustic features were combined by finding a polynomial
function made up of linear, quadratic and cubic terms of the features.
They also tried a Gaussian distribution classifier.  The acoustic
scores were combined with scores from a statistical language model,
which determined the probability of the word sequence with the
predicted boundary tones inserted into the word sequence.  They have
also extended this approach to work on word graphs as well \cite
{Kompe-etal95:eurospeech}.

In work related to the above, Mast \etal~\shortcite{Mast-etal96:icslp}
aim to segment speech by dialog acts as the first step in
automatically classifying them.  Again, a combination of an acoustic
model and language model is used.  The acoustic model is a multi-layer
perceptron that estimates the probability $\Pr(v_i|c_i)$, where $v_i$
is a variable indicating if there is a boundary after the current word
and $c_i$ is a set of acoustic features of the neighboring six
syllables and takes into account duration, pause, F0-contour and
energy.  The language model gives the probability of the occurrence of
a boundary (or not) and the neighboring words.  This probability is
estimated using a backoff strategy.  These two probabilities are
combined (with the language model score being weighted by the
optimized weight $\xi$) in the following formula to give a score for
the case in which $v_i$ is a boundary and for when it is not.
\[ \Pr(v_i|c_i)P^{\xi}(\dots w_{i-1}w_iv_iw_{i+1}w_{i+2} \dots) \]
Using this method, they were able to achieve a recognition accuracy of
92.5\% on turn internal boundaries.  Translated into recall and
precision, they achieved a recall rate of 85.0\% and a precision of
53.1\% for turn-internal boundaries.\footnote {We calculated their
recall and precision rates from Table~1 in their paper, which gave the
results of their model for turn-internal boundary tones.  The table
reported that they classified 85.0\% of the 662 boundaries (562.7)
while mistaking 6.8\% of the 7317 non-boundaries (497.6) as
boundaries.  This gives a precision rate of $562.7 / (562.7 + 497.6) =
53.1\%$.  Their error rate is $(662 - 562.7 + 497.6) / 662 = 90.1\%$}

Meteer and Iyer \shortcite{MeteerIyer96:emnl} investigated whether
having access to linguistic segments improves language
modeling.\footnote {Meteer and Iyer \shortcite{MeteerIyer96:emnl} also
present a brief overview of the conventions for annotating
conversational speech events in the Switchboard corpus.}  Like the
statistical language model used by Kompe \etal~\shortcite
{Kompe-etal94:icassp}, they compute the probability of the sequence of
words with the hypothesized boundary tones inserted into the sequence.
Working on the Switchboard corpus of human-human conversational
speech, they find that if they had access to linguistic boundaries,
they can improve word perplexity from 130 to 78.  In the more
realistic task in which they must predict the boundaries as part of
the speech recognition task, they still achieve a perplexity
reduction, but only from 130 to 127.\footnote {The baseline perplexity
of 130 was obtained from Table 1, under the case of training and
testing a language model with no segmentation.  The perplexity of 78
was obtained from the same table under the case of training and
testing a language model with linguistic segmentation.  The perplexity
of 127 was obtained from Table~2, under the condition of training with
linguistic segments but testing without segments.}  Hence, they find
that predicting linguistic segments improves language modeling.

Stolcke and Shriberg \shortcite{StolckeShriberg96:icslp}, building on
the work of Meteer and Iyer, investigated how well a language-model
can find the linguistic boundaries.  They found that best results were
obtained if they also took into account the POS tags, as well as the
word identities of certain word classes, in particular filled pauses,
conjunctions, and certain discourse markers.  These results were a
recall rate of 79.6\% and a precision of 73.5\% over all linguistic
segment boundaries.\footnote{These results were taken from Table~3 in
their paper under the condition of {\em POS-based II}.}  However, like
speech repairs, segment boundaries disrupt the context that is needed
to determine to predict POS tags.  Hence, once they try to
automatically determine the POS tags and identify the discourse
markers, which their algorithm relies on, their results will
undoubtedly degrade.

\section{Speech Repairs}
\label{sec:related:repairs}

Most of the current work in detecting and correcting speech repairs
starts with the seminal work of Levelt \shortcite
{Levelt83:cog}.\footnote{Recent work \cite{Finkler97:thesis,Finkler97}
has begun exploring the use of speech repairs as a mechanism for
allowing incremental natural language generation.}  Levelt was
primarily interested in speech repairs as evidence for how people
produce language and how they monitor it to ensure that it meets the
goals it was intended for.  From studying task-oriented monologues,
Levelt put forth a number of claims.  The first is that when a speaker
notices a speech error, she will only interrupt the current word if it
is in error.  Second, repairs obey the following well-formedness rule
(except those involving syntactically or phonologically ill-formed
constructions).  The concatenation of the speech before the
interruption point (with some completion to make it well formed)
followed by the conjunction ``and'' followed by the text after the
interruption point must be syntactically well-formed.  For instance,
``did you go right -- go left'' is a well-formed repair since ``did
you go right and go left'' is syntactically well-formed; whereas ``did
you go right -- you go left'' is not since ``did you go right and you
go left'' is not.  Levelt did find exceptions to his well-formedness
rule.  The Trains corpus also contains some exceptions, as illustrated
by the following example.
\begin{example}{d93-10.4 utt81}
\label{ex:d93-10.4:utt81}
the two boxcars \reparandum{of orange juice should} er of oranges
should be made into orange juice
\end{example}

Third, Levelt hypothesized that listeners can use the following rules
for determining the extent of the reparandum (the continuation
problem).
\begin{enumerate}
\item If the last word before the interruption is of the same syntactic
category as the word after, then that word is the reparandum
onset.\footnote {Here we use the definition of reparandum and
alteration given in Section~\ref{sec:intro:repairs}, rather than
Levelt's definitions.}
\item If there is a word prior to the interruption point that is identical
to the word that is the alteration onset and of the same syntactic
category, then that word is the reparandum onset.
\end{enumerate}
Levelt found that this strategy found the
correct reparandum onset for 50\% of all repairs (including fresh
starts), incorrectly identified the reparandum for 2\% of the repairs,
and was unable to propose a reparandum onset for the remaining
48\%.\footnote{These numbers were derived from Table~8 in the paper.
There were 959 repairs.  If the word identity constraint is applied
first, it would correctly guess 328 of the repairs, incorrectly guess
17, and have no comment for the remaining 614.  Of the 614, the
category identity constraint would correctly guess 149, incorrectly
guess 7 and have no comment for the remaining 458 repairs.  Thus, the
two constraints would correctly guess 477 repairs (49.7\%),
incorrectly guess 24 repairs (2.5\%), and have no comment about the
remaining 458 repairs (47.8\%).}  For Example~\ref{ex:d92a-2.1:utt29},
repeated below, Levelt's strategy would incorrectly guess the
reparandum onset as being the first word.
\begin{example}{d92a-2.1 utt29}
that's the one \reparandum{with the bananas}\interruptionpoint
\editingterm{I mean} \alteration{that's taking the bananas}
\end{example}

Fourth, Levelt showed that different editing terms make different
predictions about the repair the speaker is about to make.  For
instance, ``uh'' strongly signals an abridged repair, whereas a word
like ``sorry'' strongly signals a repair in which ``the speaker
neither instantly replaces a trouble word, nor retraces to an earlier
word \dots, but restarts with fresh material'' (pg.~85), as
Example~\ref{ex:d93-14.3:utt2}, repeated below, illustrates.\footnote
{Levelt refers to such repairs as {\em fresh starts}.  As explained in
Section~\ref{sec:intro:repairs}, we use {\em fresh starts} to refers
to repairs that abandon the current utterance.}
\begin{example}{d93-14.3 utt2}
\reparandum{I need to send}\ip \et{let's see} 
how many boxcars can one engine take
\end{example}

One of the first computational approaches was by Hindle \shortcite
{Hindle83:acl}, who addressed the problem of correcting self-repairs
by adding rules to a deterministic parser that would remove the
necessary text.  Hindle assumed the presence of an edit signal that
marks the interruption point, the POS assignment of the input words,
and sentence boundaries.  With these three assumptions, he was able to
achieve a recall rate of 97\% in finding the correct repair.  For
modification repairs, Hindle used three rules for expunging text.  The
first rule ``is essentially a non-syntactic rule'' that matches
repetitions (of any length); the second matches repeated constituents,
both complete; and the third matches repeated constituents, in which
the first is not complete, but the second is.  Note that Example~\ref
{ex:d93-10.4:utt81}, which failed Levelt's well-formedness rule, also
fails to be accounted for by these rules.  For fresh starts, Hindle
assumed that they would be explicitly marked by a lexical item such as
``well'', ``okay'', ``see'', and ``you know''.\footnote{From Table~1
in his paper, it seems clear that Hindle does account for abridged
repairs, in which only the editing term needs to be removed.  However,
not enough details are given in his paper to ascertain how these are
handled.}

Kikui and Morimoto \shortcite{KikuiMorimoto94:icslp}, working with a
Japanese corpus, employed two techniques to determine the extent of
reparanda of modification repairs. First, they find all possible
onsets for the reparandum that cause the resulting correction to be
well-formed.  They do this by using local syntactic knowledge in the
form of an adjacency matrix, that states whether a given category can
follow another category.  Second, they used a similarity based
analyzer \cite {KurohashiNagao92:coling} that finds the best path
through the possible repair structures.  They assigned scores for
types of syntactic category matches and word matches.  They then altered
this path to take into account the well-formedness information from
the first step.  Like Hindle, they were able to achieve high
correction rates, in their case 94\%, but they also had to assume
their input includes the location of the interruption point and
the POS assignments of the words involved.

The results of Hindle and Kikui and Morimoto are difficult
to translate into actual performance.  Both strategies depend
upon the ``successful disambiguation of the syntactic categories''
\cite{Hindle83:acl}.  Although syntactic categories can be determined
quite well by their local context (as is needed by a deterministic
parser), Hindle admits that ``[self-repair], by its nature, disrupts
the local context.''  A second problem is that both algorithms depend
on the presence of an edit signal and one that can distinguish between
the three types of repairs.  So far, the abrupt cut-off that some have
suggested signals the repair (cf.~\cite {Labov66}) has been difficult
to find.  Rather, there are a number of difficult sources that give
evidence as to the occurrence of a repair, including the presence of a
suitable correction.

Bear \etal~\shortcite {Bear-etal92:acl} investigated the use of
pattern matching of the word correspondences, global and local
syntactic and semantic ill-formedness, and acoustic cues as evidence
for detecting speech repairs.  They tested their pattern matcher on a
subset of the ATIS corpus from which they removed all {\em trivial}
repairs, repairs that involve only the removal of a word fragment or a
filled pause.  For their pattern matching results, they were able to
achieve a detection recall rate of 76\%, and a precision of 62\%, and
they were able to find the correct repair 57\% of the time, leading to
an overall correction recall of 43\% and correction precision of 50\%.
They also tried combining syntactic and semantic knowledge in a
``parser-first'' approach---first try to parse the input and if that
fails, invoke repair strategies based on word patterns in the input.
In a test set containing 26 repairs \cite {Dowding-etal93:acl}, they
obtained a detection recall rate of 42\% and a precision of 84.6\%;
for correction, they obtained a recall rate of 30\% and a precision of 62\%.

Nakatani and Hirschberg \shortcite {NakataniHirschberg94:jasa} take a
different approach by proposing that speech repairs be detected in a
{\em speech-first} model using acoustic-prosodic cues, without having
to rely on a word transcription.  In their corpus, 73.3\% of all repairs
are marked by a word fragment. Using hand-transcribed prosodic
annotations, they built a decision tree using CART
\cite{Breiman-etal84:book} on a 148 utterance training set to identify
the interruption point (each utterance contained at least one repair)
using such acoustic features as silence duration, energy, and pitch,
as well as some traditional text-first cues such as presence of word
fragments, filled pauses, word matches, word replacements, POS tags,
and position of the word in the turn.  On a test set of 202 utterances
containing 223 repairs, they obtained a recall rate of 86.1\% and a
precision of 91.2\% in detecting speech repairs.  The cues that they
found relevant were duration of pauses between words (greater than
0.000129), presence of fragments, and lexical matching within a window
of three words.

Stolcke and Shriberg \shortcite {StolckeShriberg96:icassp} incorporate
speech repair detection and correction into a word-based language
model.  They limit the types of repairs to single and double word
repetitions, single and double word deletions, deletions from the
beginning of the sentence, and occurrences of filled pauses.  In
predicting a word, they treat the type of disfluency (including no
disfluency at all) as a hidden variable, and sum over the probability
distributions for each type.  For a hypothesis that includes a speech
repair, the prediction of the next word is based upon a cleaned-up
representation of the context, as well as taking into account if they
are predicting a single or double word repetition.  Surprisingly, they
found that this model actually results in worse performance, in terms
of both perplexity and word error rate.  In analyzing the results,
they found that the problem was attributed to their treatment of
filled pauses.  In experiments performed on linguistically segmented
utterances, they found that utterance-medial filled pauses should be
cleaned up before predicting the next word, whereas utterance-initial
filled pauses should be left intact and used to predict the next word.

Siu and Ostendorf \shortcite{SiuOstendorf96:icslp} extended the work
of Stolcke and Shriberg \shortcite{StolckeShriberg96:icassp} in
differentiating utterance-internal filled-pauses from
utterance-initial filled-pauses.  Here, they differentiated three
roles that words such as filled-pauses can play in an utterance.  They
can be utterance initial, involved in a non-abridged speech repair, or
involved in an abridged speech repair.  They found that by using training
data with these roles marked, and by using a function-specific
variable $n$-gram model (i.e.~use different context for the
probability estimates depending on the function of the word), they
could achieve a perplexity reduction from 82.9 to 81.1 on a test
corpus.  Here, the role of the words is treated as an unseen condition
and the probability estimate is achieved by summing over each possible
role.

\section{Discourse Markers}
\label{sec:related:dm}

Many researchers have noted the importance of discourse markers \cite
{RCohen84:coling,Reichmanadar84:ai,Sidner85:ci,GroszSidner86:cl,LitmanAllen87}.
These markers serve to inform the reader about the structure of the
discourse---how the current part relates to the rest.  For instance,
words such as ``now'', ``anyways'' signal a return from a digression.
Words such as ``firstly'' and ``secondly'' signal that the speaker is
giving a list of options.  The structure of the text is also important
because in most theories of discourse, it helps the listener resolve
anaphoric references.

Spoken dialog also employs a number of other discourse markers that
are not as closely tied to the discourse structure.  Words such as
``mm-hm'' and ``okay'' function as acknowledgments.  Words such as
``well'', ``like'', ``you know'', ``um'', and ``uh'' can act as a
part of the editing term of a filled paused, as well as help signal
the beginning of an utterance.  Because of their lack of sentential
content, and their relevance to the discourse process (including
preventing someone from stealing the turn), they are also regarded as
discourse markers.

Hirschberg and Litman \shortcite {HirschbergLitman93:cl} examined how
intonational information can distinguish between the discourse and
sentential interpretation for a set of ambiguous lexical items.  This
work was based on hand-transcribed intonational features and only
examined discourse markers that were one word long.\footnote{As will
be explained in Section~\ref{sec:corpus:dm}, we also restrict
ourselves to single word discourse markers.}  In an initial study
\cite{HirschbergLitman87:acl} of the discourse marker ``now'' in a
corpus of spoken dialog from the radio call-in show ``The Harry Gross
Show: Speaking of Your Money'' \cite {PollackHirschbergWebber82:tr},
they found that discourse usages of the word ``now'' were either an
intermediate phrase by themselves (or in a phrase consisting
entirely of ambiguous tokens), or they are first in an intermediate
phrase (or preceded by other ambiguous tokens) and are either
de-accented or have a ${\bf L}^{\bf *}$ word accent.  Sentential uses were
either non-initial in a phrase or, if first, bore a ${\bf H}^{\bf *}$
or complex accent (i.e.~not a ${\bf L}^{\bf *}$ accent).

In a second study, Hirschberg and Litman \shortcite
{HirschbergLitman93:cl} used a corpus consisting of a speech given by
Ronald Brachman from prepared notes, which contained approximately
12,500 words.  From previous work on discourse markers, the authors
assembled a list of words that have a discourse marker interpretation.
This list gave rise to 953 tokens in their corpus that needed to be
disambiguated.  Each author then hand-annotated these tokens as having
a discourse or sentential interpretation, or as being ambiguous.  The
authors were able to agree on 878 of the tokens as having a discourse
or as having a sentential interpretation.  They found that the
intonational model that they had proposed for the discourse marker
``now'' in their previous study
\cite{HirschbergLitman87:acl} was able to predict 75.4\% (or 662) of
the 878 tokens.  This translates into a discourse marker recall rate
of 63.1\% and a precision of 88.3\%.\footnote{From Table 7 of their
paper, they report that there model obtained 301 hits, 176 misses, 40
false positives and 361 correct rejections.  This gives a recall rate
of $301/(301+176)=63.1\%$, a precision rate of $301/(301+40) =
88.3\%$, and an error rate of $(176+40)/(301+176)=45.3\%$.}
Hirschberg and Litman found that many of the errors occurred on
coordinate conjuncts, such as ``and'', ``or'' and ``but'', and report
that these proved problematic for annotating as discourse markers as
well, since ``the discourse meanings of conjunction as described in
the literature
\dots seem to be quite similar to the meanings of sentential
conjunction'' \cite[pg.~518] {HirschbergLitman93:cl}.  From this, they
conclude that this ``may make the need to classify them less
important''.  Excluding the conjuncts gives them a recall rate of
81.5\% and a precision of 82.7\%.\footnote{Table 8 of their paper
gives the results of classifying the non-conjuncts, where they report
167 hits, 38 misses, 35 false positives, and 255 correct rejections.
This gives a recall rate of $167/(167+38) = 81.5\%$, a precision of
$167/(167+35)=82.7\%$, and an error rate of
$(38+35)/(167+38)=35.6\%$.}

Hirschberg and Litman also looked at the effect of orthographic
markers and POS tags.  For the orthographic markings, they looked at
how well discourse markers can be predicted based on whether they
follow or precede a hand-annotated punctuation mark.  Although of
value for text-to-speech synthesis, these results are of little
interest for speech recognition and understanding since automatically
identifying punctuation marks will probably be more difficult than
identifying prosodic phrasing.  They also examined correlations with
POS tags.  For this experiment, they chose discourse marker versus
sentential interpretation based on whichever is more likely for that
POS tag, where the POS tags were automatically computed using Church's
part-of-speech tagger
\shortcite{Church88:anlp}.  This gives them a recall rate of 39.0\%
and a precision of 55.2\%.\footnote {Recall and precision results were
computed from Table~12 in their paper.  From this table, we see that
the majority of singular or mass nouns, singular proper nouns, and
adverbs have a discourse interpretation, while the rest favor a
sentential interpretation.  The strategy of classifying potential
discourse markers based on whichever is more likely for that POS tag
thus results in $10+5+118=133$ hits (of discourse markers),
$139+43+1+4+21=208$ misses, $7+1+101=109$ false positives, and
$6+244+21+58+3+12+6+78=428$ correct rejections.  This gives 561
correct predictions out of a total of 878 potential discourse markers
leading to a 63.9\% success rate.  When translated into recall and
precision rates for identifying discourse markers, this gives a recall
rate of $133/(133+208) = 39.0\%$, a precision of $133/(133+109) =
55.0\%$, and an error rate of $(208+109)/(133+208) = 93.0\%$.}  Thus,
we see that POS information, even exploited in this fairly simplistic
manner, can give some evidence as to the occurrence of discourse
marker usage.

Litman \shortcite{Litman96:jair} explored using machine learning
techniques to automatically learn classification rules for discourse
markers.  She contrasted the performance of CGRENDEL
\cite{WCohen92:ml,WCohen93:ijcai} with C4.5 \cite {Quinlan93:book}.  CGRENDEL is a learning algorithm that learns an ordered set of
if-then rules that map a condition to its most-likely event (in this
case discourse or sentential interpretation of potential discourse
marker).  C4.5 is a decision tree growing algorithm similar to CART
that learns a hierarchical set of if-then rules in which the leaf
nodes specify the mapping to the most-likely event.  She found that
machine learning techniques could be used to learn a classification
algorithm that was as good as the algorithm manually built by
Hirschberg and Litman \shortcite {HirschbergLitman93:cl}.  Further
improvements were obtained when different sets of features about the
context were explored, such as the identify of the token under
consideration.  The best results (although the differences between
this version and some of the others might not be significant) was
obtained by using CGRENDEL and letting it choose conditions from the
following set: length of intonational phrase, position of token in
intonational phrase, length of intermediate phrase, position of token
in intermediate phrase, composition of intermediate phrase (token is
alone in intermediate phrase, phrase consists entirely of potential
discourse markers, or otherwise), and identity of potential discourse
marker.  The automatically derived classification algorithm achieved a
success rate of 85.5\%, which translates into a discourse marker
error rate of 37.3\%,\footnote {The success rate of 85.5\% is taken
from the row titled ``phrasing+'' in Table~8.  Not enough details are
given to compute the recall and precision rate of the discourse
markers for that experiment.  However, we can compute our standardized
error rate by first computing the number of tokens that were
incorrectly guessed: $14.5\% \times 878 = 127.3$.  We then normalize
this by the number of discourse markers, which is 341.  Hence, their
error rate for discourse markers is $127.3/341 = 37.3\%$.} in
comparison to the error rate of 45.3\% for the algorithm of Hirschberg
and Litman \shortcite {HirschbergLitman93:cl}.  Hence, machine
learning techniques are an effective way in which a number of
different sources of information can be combined to identify discourse
markers.

\cleardoublepage
\chapter{The Trains Corpus}
\label{chapter:corpus}

One of the goals that we are pursuing at the University of Rochester
is the development of a conversationally proficient planning
assistant, which assists a user in constructing a plan to achieve some
task involving the manufacturing and shipment of goods in a railroad
freight system (the Trains domain) \cite
{Allen-etal95:jetai,Allen-etal96:acl}.  In order to do this, we need
to know what kinds of phenomena occur in such dialogs, and how to deal
with them.  To provide empirical data, we have collected a corpus of
dialogs in this domain with a person playing the role of the system
(full details of the collection procedure are given in \cite
{HeemanAllen95:tn-dialogs}).  The collection procedure was designed to
make the setting as close to human-computer interaction as possible,
but was not a {\em wizard} scenario, where one person pretends to be a
computer; rather, both participants know that they are speaking to a
real person.  Thus these dialogs provide a snapshot into an ideal
human-computer interface that is able to engage in fluent
conversations.

In Table~\ref{tab:corpus:occurrences}, we give the size of the Trains
corpus.  The corpus consists of 98 dialogs, totaling six and a half
hours of speech and 6163 speaker turns.
\begin{table}
\begin{center}
\begin{tabular}{|l|r|} \hline
Dialogs                      &    98 \\
Speakers                     &    34 \\
Problem Scenarios            &    20 \\
Turns                        &  6163 \\
Words                        & 58298 \\
Fragments                    &   756 \\
Filled Pauses                &  1498 \\ 
Discourse Markers            &  8278 \\ \hline
Distinct Words               &   859 \\
Distinct Words/POS           &  1101 \\
Singleton Words              &   252 \\
Singleton Words/POS          &   350 \\ \hline
Boundary Tones               & 10947 \\
Turn-Internal Boundary Tones &  5535 \\ 
Abridged Repairs             &   423 \\
Modification Repairs         &  1302 \\
Fresh Starts                 &   671 \\
Editing Terms                &  1128 \\ \hline
\end{tabular}
\end{center}
\caption{Size of the Trains Corpus}
\label{tab:corpus:occurrences}
\end{table}
There are 58298 words of data, of which 756 are word fragments and
1498 are filled pauses (``um'', ``uh'', and ``er'').  Ignoring the
word fragments, there are 859 distinct words and 1101 distinct
combinations of words and POS tags.  Of these, 252 of the words and
350 of the word-POS combinations only occur once.  There are also
10947 boundary tones, 8278 discourse markers (marked as {\bf AC}, {\bf
UH\_D}, {\bf CC\_D}, and {\bf RB\_D}, as explained in Section~\ref
{sec:corpus:dm}), 1128 words involved in an editing term, and 2396
speech repairs.\footnote {In the two years since the Trains corpus was
released on CD-ROM \cite {HeemanAllen95:cdrom}, we have been fixing up
problematic word transcriptions.  The results reported here are based
on the most recent transcriptions of the Trains dialogs, which will be
made available to the general public at a later date, along with the
POS, speech repair and intonation annotations.}

Since the corpus consists of dialogs in which the conversants work
together in solving the task, the corpus is ideal for studying
problem-solving strategies, as well as how conversants collaborate in
solving a task.  The corpus also provides natural examples of dialog
usage that spoken dialog systems will need to handle in order to carry
on a dialog with a user.  For instance, the corpus contains instances
of overlapping speech, back-channel responses, and turn-taking:
phenomena that do not occur in collections of single speaker
utterances, such as ATIS \cite {Madcow92:snlp}.  Also, even for
phenomena that do occur in single speaker utterances, such as speech
repairs, our corpus allows the interactions with other dialog
phenomena to be examined.

The Trains corpus also differs from the Switchboard corpus \cite
{Godfrey-etal92:icassp}.  Switchboard is a collection of human-human
conversations over the telephone on various topics.  Since this corpus
consists of spontaneous speech, it has recently received a large
amount of interest from the speech recognition community.  However,
this corpus is not task-oriented, nor is the domain limited.  Thus, it
is of less interest to those interested in building a spoken dialog
system.

Of all of the corpora that are publicly available, the Trains corpus
is probably most similar to the HCRC Map Task corpus \cite
{Anderson-etal91:ls}. The map task involves one person trying to
explain his route to another person.  The Trains corpus, however,
involves two conversants working together to construct a plan that
solves some stated goal.  So, the conversants must do high-level domain
planning in addition to communicative planning.  Hence, our corpus
allows researchers to examine language usage during collaborative
domain-planning---an area where human-computer dialogs will be very
useful.

In the rest of this chapter, we first describe how the dialogs were
collected, how they were segmented into single-speaker audio files,
and the conventions that were followed for producing the word
transcriptions.  We then discuss the intonation and speech repair
annotations, including the annotation of overlapping repairs.  We then
end the chapter with a description of the POS tagset that we use, and
how discourse markers are annotated.

\section{Dialog Collection}

\begin{figure}
\begin{tabular}{|c|}\hline
\psfig{figure=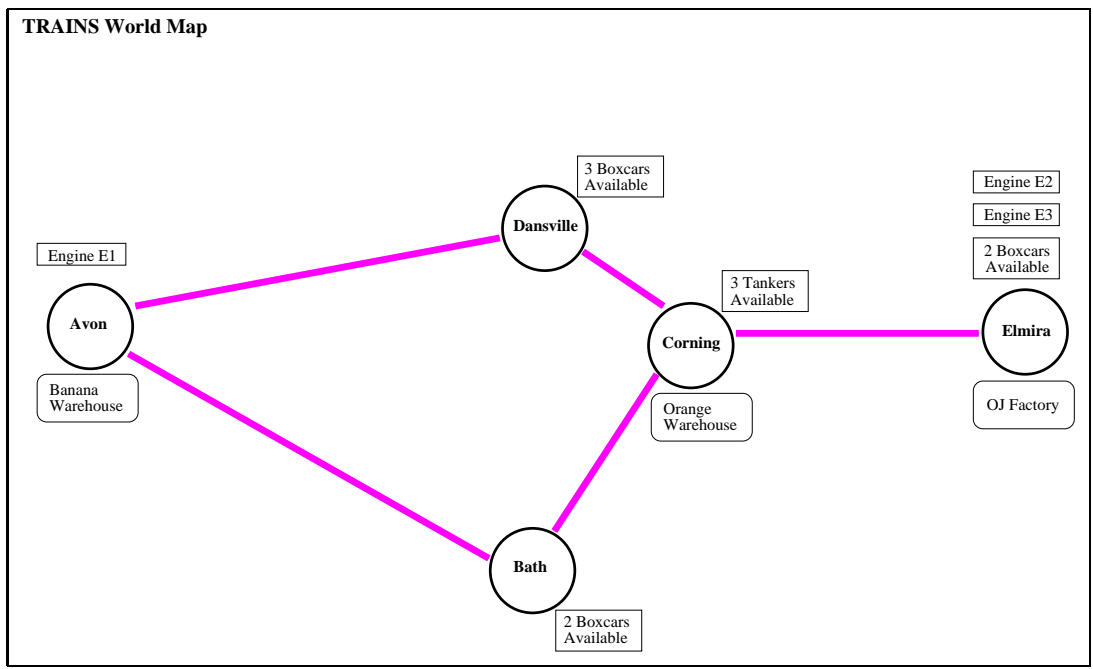,width=5.5in} \\ \hline
\end{tabular}
\caption{Map Used by User in Collecting Trains Corpus}
\label{fig:usermap}
\end{figure}

The corpus that we describe in this chapter, which is formally
known as ``The Trains Spoken Dialog Corpus'' \cite
{HeemanAllen95:cdrom} and as ``The Trains 93 Dialogues'' \cite
{HeemanAllen95:tn-dialogs},\footnote{The ``93'' in the name ``The
Trains 93 Dialogues'' refers to the year when most of the dialogs were
collected.  It does not refer to the implementation of the Trains
spoken dialog system known as ``Trains 93'' (e.g.~\cite
{Allen-etal95:jetai,Traum-etal96:jes}), which was implemented in
1993.} is the third dialog collection done in the Trains domain (the
first was done by Nakajima and Allen \shortcite
{NakajimaAllen93:phonetica}, and the second by Gross, Traum and Allen
\shortcite {Gross-etal93:tr}).  This dialog collection has much in common
with the second collection; for instance, the Trains map used in this
collection, shown in Figure~\ref {fig:usermap}, differs only slightly
from the one used previously.

There are, however, some notable differences between the third dialog
collection and the previous two.  First, more attention was paid to
minimizing outside noise and obtaining high-quality recordings.
Second, the dialogs were transcribed using the Waves software \cite
{Entropic93:waves}, resulting in time-aligned transcriptions.  The
word transcriptions, automatically obtained phonetic transcriptions
\cite {Entropic94:aligner} and audio files are available on CD-ROM
\cite {HeemanAllen95:cdrom} from the Linguistic Data Consortium.  This
allows the corpus to be used for speech analysis purposes, such as
speech recognition and prosodic analysis.  Third, this collection also
expands on the number of different tasks, and the number of different
speaker pairs.  We have 20 different problem scenarios, and 34
speakers arranged in 25 pairs of conversants.  For each pair of
conversants, we have collected up to seven dialogs, each involving a
different task.  Fourth, less attention this time was spent in
segmenting the dialogs into utterance units.  Rather, we used a more
pragmatically oriented approach for segmenting the dialogs into
reasonable sized audio files, suitable for use with Waves.  This
convention is described in Section~\ref {sec:corpus:segmentation}.

\subsection{Setup}

The dialogs were collected in an office that had partitions separating
the two conversants; hence, the conversants had no visual contact.
Dialogs were collected with Sennheiser HMD 414 close-talking
microphones and headphones and recorded using a Panasonic SV-3900
Digital Audio Tape deck at a sampling rate of 48 kHz.  In addition to
a person playing the role of the system and a second person playing
the role of the user, a third person---the coordinator who ran the
experiments---was also present.  All three could communicate with each
other over microphones and headphones, but only the system and user's
speech was recorded, each on a separate channel of a DAT tape.  Both
the user and system knew that the coordinator was overhearing and
would not participate in the dialogs, even if a problem arose.

\begin{figure}
\begin{tabular}{|c|}\hline
\psfig{figure=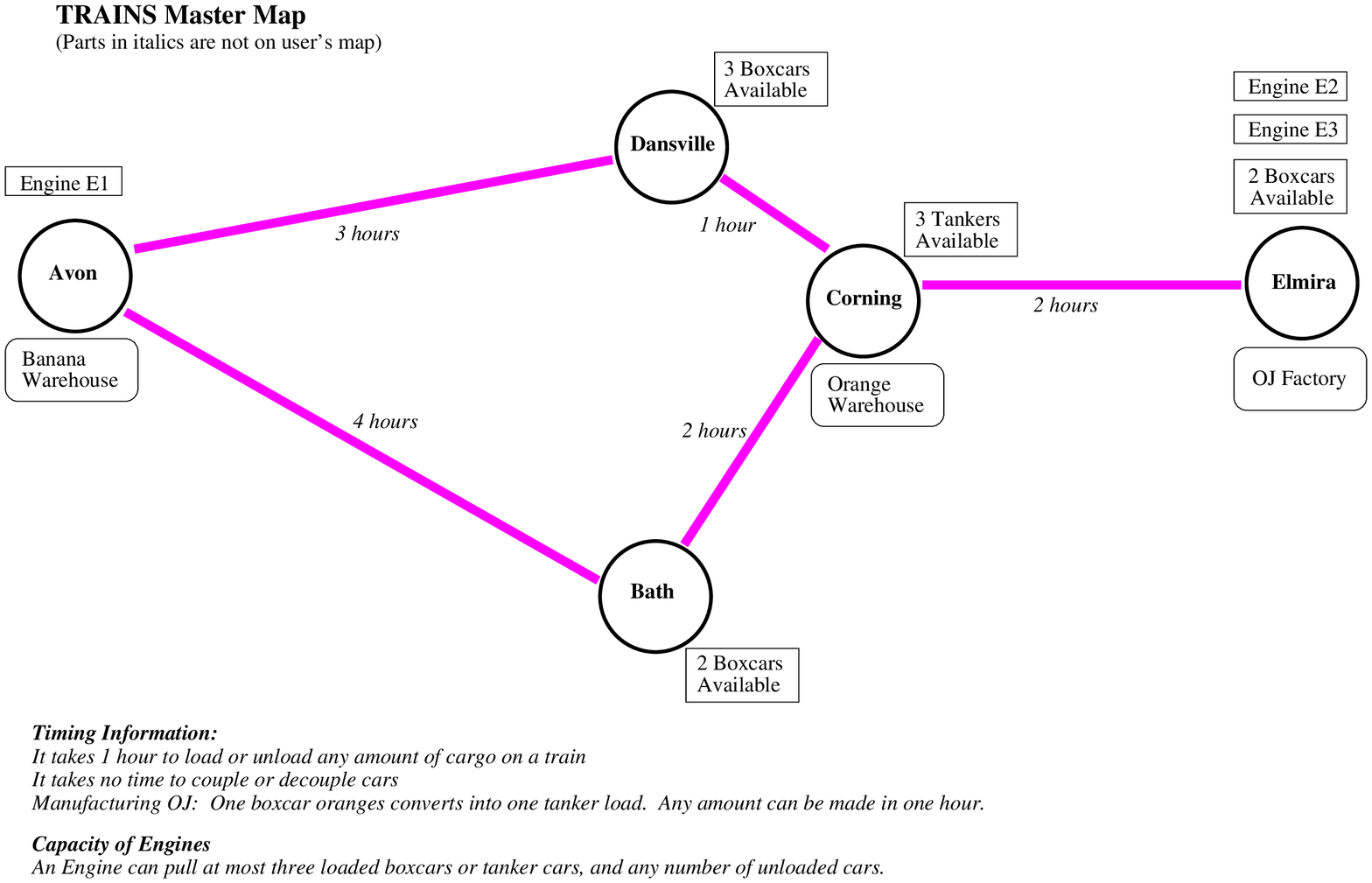,width=5.5in} \\ \hline
\end{tabular}
\caption{Map Used by System in Collecting Trains Corpus}
\label{fig:systemmap}
\end{figure}

At the start of the session, the user was given a copy of a consent
form to read and sign, as well as a copy of the user instructions and
user map (Figure~\ref{fig:usermap}).  The user was not allowed to
write anything down.  The system was given a copy of the system
instructions as well as copies of the system map (Figure~\ref
{fig:systemmap}).
The system map includes information that is not given to the user,
such as the distance between cities and the length of time it takes to
load and unload cargo and make orange juice.  The system was also
given blank paper and a pen, and was encouraged to use these to help
remember the plan and answer the user's queries.  Once the user and
system had read over the instructions, the coordinator had them
practice on the warmup problem given in the user's instructions.

The participants then proceeded to do anywhere between two and seven
more problems, depending on how many they could complete in the thirty
minute session.  The problems were arranged into three piles on the
user's desk, with each pile corresponding to a different level of
difficulty.  When the user and system were ready to begin a dialog,
the coordinator would instruct the user to take a problem from the top
of a certain pile.  The first problem, after the warmup, was always
from the easiest pile.  For later problems, the level of difficulty
was chosen on the basis of how well the participants handled the
previous problem and how much time remained.

After a problem was chosen, the user was given time to read the
problem over (less than a minute).  Once this was done, the user would
signal by saying ``ready''.  The coordinator would then set the DAT deck
into record mode and push a button that would cause a green light to
turn on at the user's and system's desk, which would signal the system
to begin the conversation with the phrase ``hello can I help you.''

The coordinator would record the conversation until it was clear that
the two participants had finished the dialog.  At that point, the user
would hand the problem card to the coordinator, who would write the
problem number (written on the back of the card) in the recording
log.  A sample problem that the user would be given is ``{\em
Transport 2 boxcars of bananas to Corning by 11 AM.  It is now
midnight.}''

\subsection{Subjects}

The role of the system was played primarily by graduate students from
the department of Computer Science and the department of Linguistics.
About half of these people were involved in the Trains project.  As for
the users, almost all of them were naive subjects who did the
experiment as course credit for an introductory cognitive science
course.  All participants were native speakers of North American
English.

\section{Initial Transcription}

After the dialogs were collected, we segmented them into single
speaker audio files and transcribed the words that were spoken.

\subsection{Segmentation}
\label{sec:corpus:segmentation}

We have segmented the dialogs into a sequence of single-speaker
segments that captures the sequential nature of the two speakers'
contributions to the dialog \cite {HeemanAllen94:tn-tools}.  Most
times, turn-taking proceeds in an orderly fashion in a dialog, with no
overlap in speaker turns and each speaker's turn building on the other
conversant's turn, thus making it easy to view a dialog as an orderly
progression of single-speaker stretches of speech.  Sometimes,
however, the hearer might make a back channel response, such as
`mm-hm', while the speaker is still continuing with her turn, or there
might be brief contentions over who gets to talk next.  To deal with
these problems, we use several guidelines for segmenting the speech
into turns.
\begin{description}
\item [A1:\hspace*{.25em}] Each speaker segment should be short 
enough so that it does not include effects attributable to
interactions from the other conversant that occur after the start of
the segment.
\item [A2:\hspace*{.25em}] Each speaker segment should be long enough so that 
local phenomena are not split across segments.  Local phenomena
include speech repairs, intonational phrases and syntactic structures.
\end{description}

The first guideline should ensure that the sequence of single-speaker
audio files captures the sequential nature of the dialog, thus
allowing the flow and development of the dialog to be preserved.  In
other words, the single-speaker audio files should not contain or
overlap a contribution by the other speaker.  The second guideline
ensures that the segments allow local phenomena to be easily studied,
since they will be in a single file suitable for intonation and speech
repair annotation.  There can be conflicts between these two aims.  If
this happens, the first guideline (A1) takes priority.  For instance,
if a speaker restarts her utterance after a contention over the turn,
the restart is transcribed in a separate audio file and is not viewed
as a speech repair.

Now consider the case of a back-channel response.  When the hearer
makes a back-channel response in the middle of the speaker's
utterance, it is usually not the case that the speaker responds to it.
Rather, the speaker simply continues with what she was saying.  Of
course at the end of her utterance, she would probably use the
hearer's acknowledgment to help determine what she is going to say
next.  So, the first guideline (A2) tells us not to segment the
speaker's speech during the middle of her utterance and the second
guideline (A1) tells us to segment it after the utterance.

In order to make the segments easy to use with the Waves software, we
tried to make the segments no longer than twelve seconds in length.
Thus, we typically segment a long speaker turn into several audio
segments as allowed by guideline A2.  A close approximation of the
turns in the dialog can then be captured by simply concatenating
sequential audio files that have the same speaker.

\subsection{Word Transcriptions}

Since we are interested in a time-aligned word transcription, we
transcribed each word at its end-point in the speech signal using the
Waves software \cite {Entropic93:waves}.  Each word is usually
transcribed using its orthographic spelling, unless it is a word
fragment, was mispronounced and the speaker subsequently repairs
the mispronunciation, or is a common contraction, including
``lemme'', ``wanna'', ``gonna'' and ``gotta'', which are written as
a single word.

Word fragments, where the speaker cuts off a word in midstream, were
transcribed by spelling as much of the word as can be heard followed
by a dash.  If it is clear what word the speaker was saying, then the
rest of the word is enclosed in parentheses before the dash.  For
instance, if the speaker was saying ``orange'', but cut it off before
the `g' sound, it would be transcribed as \makebox{``oran(ge)-''.}
Words that have an abrupt cutoff, but the whole word can be heard,
are transcribed as the complete word, followed by parentheses,
followed by a dash, as in ``the()-''.

Other phenomena are also marked with the word annotations, including
silences, breaths, tongue clicking, throat clearing, and miscellaneous
noises.  We used the tokens {\bf $<$sil$>$}, {\bf $<$brth$>$}, {\bf
$<$click$>$}, {\bf $<$clear-throat$>$}, and {\bf $<$noise$>$},
respectively, to mark these events.

\subsection{Sample Dialog}

Table~\ref{tab:dialog} gives a sample dialog.
\begin{table}
\noindent
{\bf Problem 1-B}

\noindent
{\em Transport 2 boxcars of bananas to Corning by 11 AM. It is now
midnight.} \\

\newcommand{\dialine}[3]{#1 & #2 & \parbox[t]{5in}{\setlength{\baselineskip}{1.3em} #3} \\}
\begin{tabular}{lll}
\dialine{utt1}{s:}{ hello $<$sil$>$ can I help you}
\dialine{utt2}{u:}{ I need to take $<$sil$>$ two boxcars of bananas $<$sil$>$ um from $<$sil$>$ Avon to Corning by eleven a.m.}
\dialine{utt3}{s:}{ so two boxcars of what}
\dialine{utt4}{u:}{ bananas}
\dialine{utt5}{s:}{ bananas $<$brth$>$ $<$sil$>$ to where}
\dialine{utt6}{u:}{ Corning}
\dialine{utt7}{s:}{ to Corning $<$sil$>$ okay}
\dialine{utt8}{u:}{ um $<$sil$>$ so the first thing we need to do is to get $<$sil$>$ the uh $<$sil$>$ boxcars $<$sil$>$ to uh $<$sil$>$ Avon}
\dialine{utt9}{s:}{ okay $<$sil$>$ so there's boxcars in Dansville and there's boxcars in Bath}
\dialine{utt10}{u:}{ okay $<$sil$>$ is $<$sil$>$ Dansville $<$sil$>$ the shortest route}
\dialine{utt11}{s:}{ yep}
\dialine{utt12}{u:}{ okay}
\dialine{utt13}{}{ how long will it take from $<$sil$>$ to $<$sil$>$ to have the $<$sil$>$ oh I need it $<$sil$>$ $<$noise$>$ $<$sil$>$ ooh $<$brth$>$ how long will it take to get from $<$sil$>$ Avon to Dansville}
\dialine{utt14}{s:}{ three hours}
\dialine{utt15}{u:}{ okay $<$sil$>$ so $<$sil$>$ I'll need to go $<$sil$>$ from Avon to Dansville with the engine to pick up $<$brth$>$ two boxcars}
\dialine{utt16}{s:}{ okay $<$sil$>$ so we'll g- we'll get to Dansville at three a.m.}
\dialine{utt17}{u:}{ okay I need to return $<$sil$>$ to Avon to load the boxcars}
\dialine{utt18}{s:}{ okay so we'll get back $<$sil$>$ to Avon $<$sil$>$ at six a.m. $<$sil$>$ and we'll load them $<$sil$>$ which takes an hour so that'll be done by seven a.m.}
\dialine{utt19}{u:}{ and then we need to travel $<$sil$>$ to uh $<$sil$>$ Corning}
\dialine{utt20}{s:}{ okay so the quickest way to Corning is through Dansville which will take four hours $<$brth$>$ $<$sil$>$ so we'll get there at + eleven a.m. +}
\dialine{utt21}{u:}{ + eleven + a.m.}
\dialine{utt22}{}{ okay $<$sil$>$ it's doable}
\dialine{utt23}{s:}{ great}
\end{tabular}
\caption{Transcription of Dialog d93-12.2}
\label{tab:dialog}
\end{table}%
As we mentioned earlier, the user is given the problem written on a
card and has a copy of the map given in Figure~\ref{fig:usermap}.  The
system does not know the problem in advance, but has a copy of the
system map (Figure~\ref{fig:systemmap}).  The dialog is shown as it
was segmented into audio files.  Noticeable silences are marked with
`{\bf $<$sil$>$}'.  Overlapping speech, as determined automatically
from the word alignment, is indicated by the `{\bf +}' markings.

\section{Intonation Annotations}
\label{sec:corpus:tobi}

The ToBI (TOnes and Break Indices) annotation scheme \cite
{Silverman-etal92:icslp,BeckmanAyers94:tr,BeckmanHirschberg94:tr,Pitrelli-etal94:icslp}
is a scheme that combines the intonation scheme of Pierrehumbert,
which was introduced in Section~\ref{sec:intro:intonation}, with a
scheme that rates the perceived juncture after each word, as is used
by Price~\etal~\shortcite {Price-etal91:jasa} and described in
Section~\ref{sec:related:tones}.  Just as the word annotations are
done in the {\em word} tier (or file) using Waves, the intonation
scheme is annotated in the {\em tone} tier, and the perceived
junctures in the {\em break} tier.  The annotations in the break and
tone tiers are closely tied together, since the perceived juncture
between two words depends to a large extent on whether the first word
ends an intonational phrase or intermediate phrase.  The ToBI
annotation scheme makes these interdependencies explicit.

Labeling with the full ToBI annotation scheme is very
time-consuming. Hence, we chose to just label intonational boundaries
in the tone tier with the ToBI boundary tone symbol `{\bf
\%}', but without indicating if it is a high or low boundary 
tone and without indicating the phrase accent.\footnote {A small number
of the dialogs have full ToBI annotations.  These were provided by
Gayle Ayers and by Laura Dilley.}

\section{Speech Repair Annotations}
\label{sec:corpus:repairs:annotations}

The speech repairs in the Trains corpus have also been annotated.
Speech repairs, as we discussed in Section~\ref{sec:intro:repairs},
have three parts---the reparandum, editing term, and alteration---and
an interruption point that marks the end of the reparandum.  The
alteration for fresh starts and modification repairs exists only in so
far as there are correspondences between the reparandum and the speech
that replaces it.  We define the alteration in terms of the {\em
resumption}.  The resumption is the speech following the interruption
point and editing term.  The alteration is a contiguous part of this
starting at the beginning of it and ending at the last word
correspondence to the reparandum.\footnote {\label{ft:alteration}For
fresh starts and modification repairs with no word correspondences,
and abridged repairs, we define the alteration as being the first word
of the resumption.}\label{sec:ft:alteration} The correspondences
between the reparandum and alteration give valuable information: they
can be used to shed light on how speakers make repairs and what they
are repairing \cite {Levelt83:cog}, they might help the hearer
determine the onset of the reparandum and help confirm that a repair
occurred \cite {HeemanLokenkimAllen96:icslp}, and they might help the
hearer recognize the words involved in the repair.  An annotation
scheme needs to identify the interruption point, the editing terms,
the reparandum onset and the correspondences between the reparandum
and resumption.

The annotation scheme that we used is based on the one proposed by
Bear \etal~\shortcite {Bear-etal93:tr}, but extends it to better deal
with overlapping repairs and ambiguous repairs.\footnote {Shriberg
\shortcite {Shriberg94:thesis} also extends the annotation scheme of
Bear \etal~\shortcite{Bear-etal93:tr} to deal with overlapping
repairs.  We review her scheme in Section~\ref
{sec:corpus:repairs:shriberg}.}  Like their scheme, ours allows the
annotator to capture the word correspondences that exist between the
reparandum and the alteration. Table~\ref{tab:labels} gives a listing
of the labels in our scheme and their definitions.
\begin{table}
\parbox{\textwidth}{
\setlength{\baselineskip}{1.4em}
\setlength{\parskip}{1.4em}
\begin{list}{}{\settowidth{\labelwidth}{\bf ip$r$:mod+}\setlength{\leftmargin}{\labelwidth}\addtolength{\leftmargin}{\labelsep}\renewcommand{\makelabel}[1]{\makebox[\labelwidth][l]{#1}}}

\item[\bf ip$r$] Interruption point of a speech repair.  The index $r$
is used to distinguish between multiple speech repairs in the same
audio file.  Indices are in multiples of 10 and all word
correspondence for the repair are given a unique index between the
repair index and the next highest repair index.

\item[\bf ip$r$:mod] The {\bf mod} suffix indicates that the repair is a modification repair.

\item[\bf ip$r$:can] The {\bf can} suffix indicates that the repair is a fresh start (or {\em cancel}).

\item[\bf ip$r$:abr] The {\bf abr} suffix indicates that the repair is
an abridged repair. 

\item[\bf ip$r$:mod+] The {\bf mod+} suffix indicates that the
transcriber thinks the repair is a modification repair, but is
uncertain.  For instance, the repair might not have have the strong
acoustic signal associated with a fresh start, but might be confusable
because the reparandum starts at the beginning of the
utterance.

\item[\bf ip$r$:can+] The {\bf can+} suffix indicates that the
transcriber thinks the repair is a fresh start, but is uncertain.  For
instance, the repair might have the acoustic signal of a fresh start,
but also might seem to rely on the strong word correspondences to
signal the repair.

\item[\bf sr$r$$<$] Denotes the onset of the reparandum of a fresh
start.

\item[\bf m$i$] Used to label word correspondences in which the two
words are identical. The index $i$ is used both to co-index the two
words that match and to associate the correspondence with the
appropriate repair.

\item[\bf r$i$] Used to label word correspondences in which one word
replaces another.

\item[\bf x$r$] Word deletion or insertion.  It is indexed by the
repair index.

\item[\bf p$i$] Used to label a multi-word correspondence, such as a
replacement of a pronoun by a longer description.

\item[\bf et] Used to label the editing term (filled pauses and cue
words) that follows the interruption point.

\end{list}
}
\caption{Labels Used for Annotating Speech Repairs}
\label{tab:labels}
\end{table}

Each repair in an audio file is assigned a unique repair index $r$,
which is used in marking all annotations associated with the repair,
and hence separates annotations of different repairs.  All repair
annotations are done in the {\em miscellaneous} tier using Waves.  The
interruption point occurs at the end of the last word (or word
fragment) of the reparandum.  For abridged repairs, we define it as
being at the end of the last word that precedes the editing term.  The
interruption point is marked with the symbol `{\bf ip$r$}'.  To denote
the type of repair, we add the suffix `{\bf :mod}' for modification
repairs, `{\bf :can}' for fresh starts (or {\em cancels}), and `{\bf
:abr}' for abridged repairs.  Since fresh starts and modification
repairs can sometimes be difficult to distinguish, we mark the
ambiguous cases by adding a `{\bf +}' to the end.

Each word of the editing term is marked with the symbol `{\bf et}'.
Since we only consider editing terms that occur immediately after the
interruption point, we dispense with marking the repair
index.\footnote {Section~\ref {sec:corpus:et} discusses editing terms
that occur after the alteration.}

Word correspondences have an additional index for co-indexing the
parts of the correspondence.  Each correspondence is assigned a unique
identifier $i$ starting at $r$+$1$.\footnote{We separate the repair
indices by at least 10, thus allowing us to determine to which repair
a correspondence belongs.  Also, a repair index of 0 is not marked, as
Example~\ref {ex:d93-15.2:utt42} illustrates.}  Word correspondences
for word matches are labeled with `{\bf m$i$}', word replacements with
`{\bf r$i$}', and multi-word replacements with `{\bf p$i$}'.  Any
word in the reparandum not marked by one of these annotations is
marked with `{\bf x$r$}', denoting that it is a deleted word.  As for
the alteration, any word not marked from the alteration onset to the
last marked word (thus defining the end of the alteration) is also
labeled with `{\bf x$r$}', meaning it is an inserted word.  Since
fresh starts often do not have strong word correspondences, we do away
with labeling the deleted and inserted words, and instead mark the
reparandum onset with `{\bf sr$r$$<$}'.

Below, we illustrate how a speech repair is annotated.
\begin{example}{d93-15.2 utt42}
\label{ex:d93-15.2:utt42}
\setlength{\tabcolsep}{0.2em}
\begin{tabular}{cccccccccc}
engine & two & from & Elmi(ra)- & & or & engine  & three & from & Elmira \\
\small\bf m1 & \small\bf r2 & \small\bf m3 & \small\bf m4 & \makebox[0.1em][c]{\Large\bf$\uparrow$} & \small\bf
et & \small\bf m1 & \small\bf r2 & \small\bf m3 & \small\bf m4 \\
       &     & & & \makebox[0.1em][r]{\small\bf ip:mod+}   &   &    &    &         \\
\end{tabular} 
\end{example}
In this example, the reparandum is ``engine two from Elmi(ra)-'', the
editing term is ``or'', and the alteration is ``engine three from
Elmira''.  The word matches on ``engine'' and ``from'' are annotated with
`{\bf m}' and the word replacement of ``two'' by ``three'' is
annotated with `{\bf r}'.  Note that word fragments, indicated by a
`-' at the end of the word annotation, can also be annotated with word
correspondences.

As with Bear \etal~\shortcite{Bear-etal93:tr}, we allow contracted
words to be individually annotated. This is done by conjoining the
annotation of each part of the contraction with `{\bf $\wedge$}'.
Note that if only one of the words is involved with the repair, a
null marking can be used for the other word. For instance, if we want
to denote a replacement of ``can'' by ``won't'', we can annotate
``won't'' with `{\bf r1$\wedge$}'.

Marking the word correspondences can sometimes be problematic.  The
example below illustrates how it is not always clear what should be
marked as the alteration.
\begin{example}{d93-20.2 utt57}
four hours \reparandum{to Corn(ing)-}\ip from Corning to Avon
\end{example}
In this example, we could annotate ``from Corning'' as replacing the
reparandum, or we could annotate ``from Corning'' as inserted words
and ``to Avon'' as the replacement.  The important point, however, is
that the extent of the reparandum is not ambiguous.\footnote{In the
current version of our training algorithm, we use an automatic
algorithm to determine the word correspondences.  This algorithm takes
into account the reparandum onset, the interruption point and the
editing term of each repair.}

Table~\ref{tab:repairs} gives summary statistics on the speech repairs
in the Trains corpus.  
\begin{table}
\begin{center}
\begin{tabular}{|l|r|r|r|}\hline
\multicolumn{1}{|c|}{Repair Pattern} & \mc{Abridged} & \mc{Modification} & \mc{Fresh Start} \\ \hline \hline
Word fragment            &      &  320 &   29 \\
Single word match        &      &  248 &   15 \\
Multiple word match      &      &  124 &   24 \\
Initial word match       &      &  276 &   99 \\
Single word replacement  &      &  138 &   17 \\
Initial word replacement &      &   66 &   18 \\
Other                    &      &  130 &  469 \\ \hline
Total                    &  423 & 1302 &  671 \\ \hline
\end{tabular}
\end{center}
\caption{Occurrences of Speech Repairs}
\label{tab:repairs}
\end{table}
We show the division of speech repairs into abridged, modification and
fresh starts and subdivide the repairs based on the word
correspondences between the reparandum and alteration.  We subdivide
repairs as to whether the reparandum consists solely of a word
fragment, the alteration repeats the reparandum (either single word
repetition or multiple word repetition), the alteration retraces only
an initial part of the reparandum, the reparandum consists of a single
word that is replaced by the alteration, the first word of the
reparandum is replaced by the first word of the alteration, or other
repair patterns.  What we find is that modification repairs exhibit
stronger word correspondences that can be useful for determining the
extent of the repair.  Fresh starts, which are those repairs in which
the speaker abandons the current utterance, tend to lack these
correspondences.  However, as long as the hearer is able to determine
it is a fresh start, he will not need to rely as much on these cues.

\subsection{Branching Stucture}

Before we introduce overlapping speech repairs, we first introduce a
better way of visualizing speech repairs.  So far, when we have
displayed a speech repair, we have been showing it in a linear
fashion.  Consider again Example~\ref {ex:d92a-1.2:utt40}, which we
repeat below.
\begin{example}{d92a-1.2 utt40}
\label{ex:d92a-1.2:utt40:a}
you can \reparandum{carry them both on}\interruptionpoint \alteration{tow both on} the same engine
\end{example}
We display the reparandum, then the editing terms and then the alteration
in a linear order.  However, this is not how speakers or hearers
probably process speech repairs.  The speaker abandons what she was
saying in the reparandum and starts over again.  Hence, to better
understand how speakers and hearers process speech repairs, it is
helpful if we also view speech repairs in this fashion.  Hence we propose
representing speaker's utterance as a {\em branching structure}, in
which the reparandum and resumption are treated as two branches of the
utterance.

\begin{figure}[bthp]
\centerline{\psfig{file=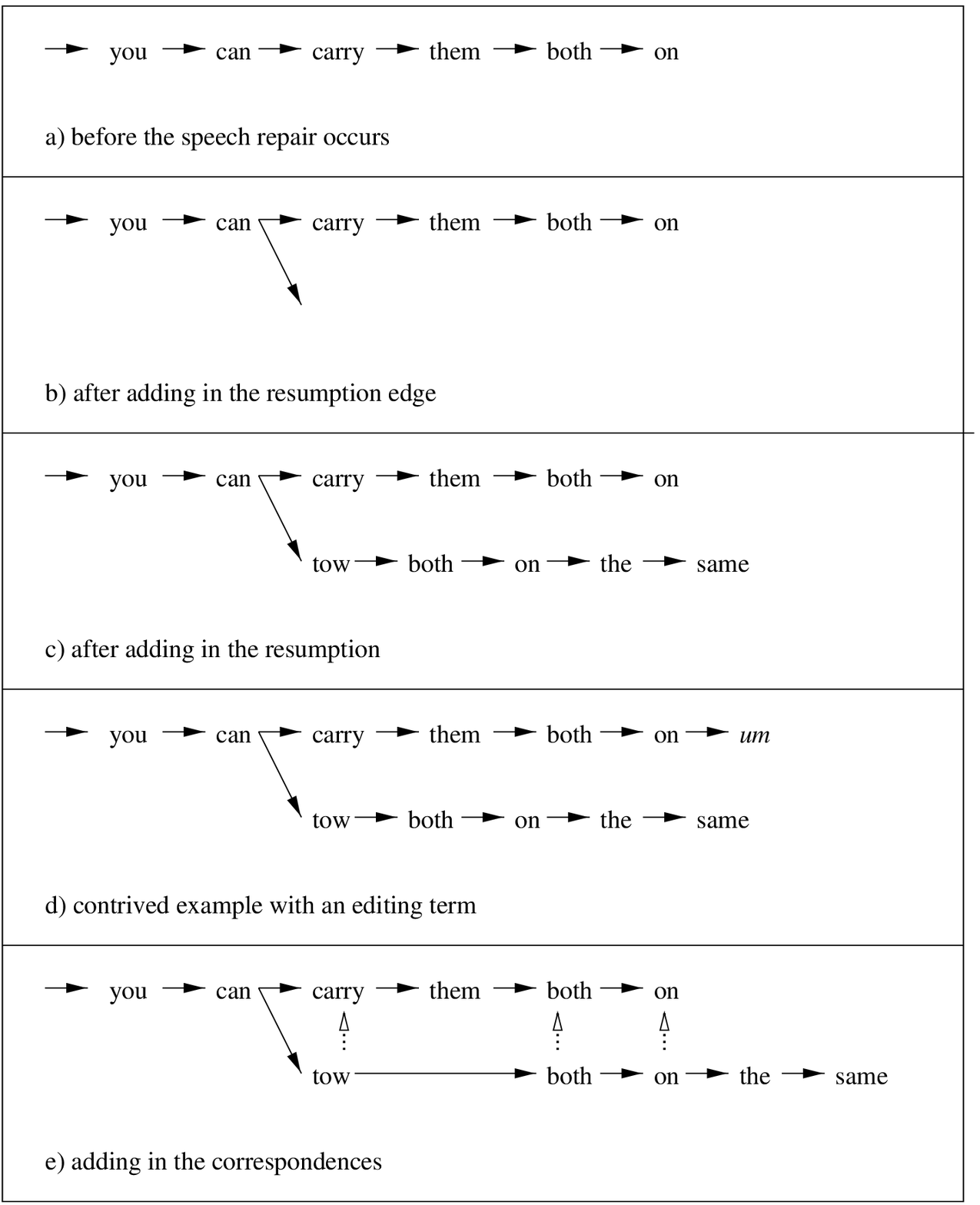}}
\caption{Branching Structure for d92a-1.2 utt40}
\label{fig:d92a-1.2:utt40}
\end{figure}

We start the branching structure with a start node.  With each word
that the speaker utters we add an arc from the last word to a new node
that contains the word that was spoken.  Figure~\ref
{fig:d92a-1.2:utt40} (a), depicts the state of the branching structure
of Example~\ref {ex:d92a-1.2:utt40:a} just before the first speech
repair.
When speakers make a repair, they are backing up in the branching
structure, to the word prior to the reparandum onset.  The speaker
then either changes or repeats the reparandum.  In terms of the
branching structure, we can view this as adding an alternative arc
before the onset of the reparandum, as indicated in Figure~\ref
{fig:d92a-1.2:utt40} (b).  The speaker's resumption is then added on
to this new arc, as illustrated in Figure~\ref {fig:d92a-1.2:utt40}
(c).  We will refer to the node that these two arcs stem from as the
{\em prior} of the repair.  The two alternative nodes from the prior
are the onset of the reparandum and the onset of the resumption.
As we add to the branching structure, we keep track of the order that
we add new edges.  This allows us to determine the {\em current
utterance} by simply starting at the root and following the most
recent arc at each choice point.

The example illustrated does not include an editing term.  Editing
terms are simply added after the end of the reparandum, and before we
backtrack in the branching structure.  However, their role as an
editing term is marked as such in the branching structure, which we
show by marking them in italics.  Figure~\ref{fig:d92a-1.2:utt40} (d)
contains a contrived version of the example that has an editing term.

With speech repairs, there are often word correspondences between the
reparandum and alteration.  These correspondences can be marked with
arcs, as indicated in Figure~\ref{fig:d92a-1.2:utt40} (e).  Here, we
show that ``tow'' is replacing ``carry'', and that the second
instances of ``both'' and ``on'' correspond to the first instances.

\subsection{Overlapping Repairs}
\label{sec:corpus:overlap}

Sometimes a speaker makes several speech repairs in close proximity to
each other.  Two speech repairs are said to {\em overlap} if it is
impossible to identify distinct regions of speech for the reparandum,
editing terms, and alteration of each repair.  Such repairs need to be
annotated.  In this section, we propose a way of annotating these
repairs that will allow us to treat them as a composition of two
individual repairs and that will lend itself to the task of
automatically detecting and correcting them.  For non-overlapping
repairs, the annotation scheme marked the interruption point, 
editing term, reparandum onset, and the correspondences between
the reparandum and the resumption.  We need to do the same for
overlapping repairs.

For overlapping repairs, identifying the interruption point does not
seem to be any more difficult than for non-overlapping repairs.
Consider the example given below.
\begin{example}{d93-16.3:utt4}
\label{ex:d93-16.3:utt4}
what's the shortest route from engine\ip from\ip for engine two at Elmira
\end{example}
When looking at the transcribed words, and more importantly when
listening carefully to the speech, it becomes clear that the above
utterance has two interruption points.  The first interruption point,
as indicated above, occurs after the first instance of ``engine'', and
the second after the second instance of ``from''.

\begin{figure}
\centerline{\psfig{file=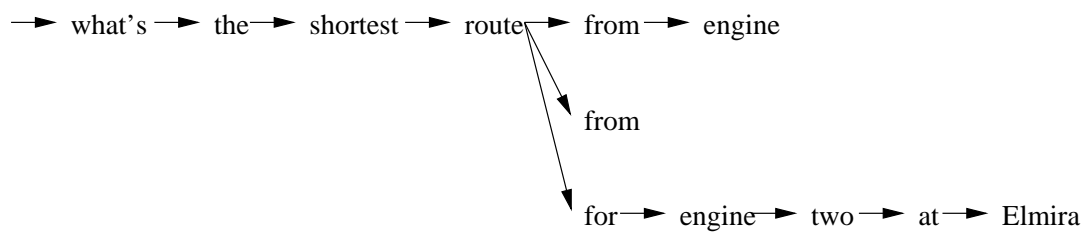}}
\caption{Branching Structure for d93-16.3 utt4}
\label{fig:d93-16.3:utt4}
\end{figure}

The second aspect of the annotation scheme for non-overlapping repairs
is to determine the reparandum, which is the speech that the repair
{\em removes}.  For overlapping repairs, one needs to determine the
overall speech that is removed by the overlapping repairs.  Again,
this task is no more difficult than with non-overlapping repairs.  For
the above example, this would be the stretch of speech corresponding
to ``from engine from''.  Next, one needs to attribute the removed
speech to the individual repairs.  We define the {\em removed speech}
of an overlapping repair as the extent of speech that the repair
removes in the current utterance at the time that the repair occurs;
in other words, it does not include the removed speech of any repair
whose interruption point precedes it, and it ignores the occurrence of
any repairs that occur after it.  For Example~\ref{ex:d93-16.3:utt4},
the removed speech of the first repair is ``from engine'' since it is
clear that the occurrence of ``from'' that is after the interruption
point of the first repair is replacing the first instance of ``from''.
At the interruption point of the second repair, the current utterance
is ``what's the shortest route from'' and the removed speech of this
repair is the word ``from'', which is the second instance of ``from''.
From this analysis, we can construct the branching structure, which is
given in Figure~\ref {fig:d93-16.3:utt4}.
In this example, both repairs have the same prior, and hence there are
three arcs out of the prior node.

\begin{figure}[hbt]
\centerline{\psfig{figure=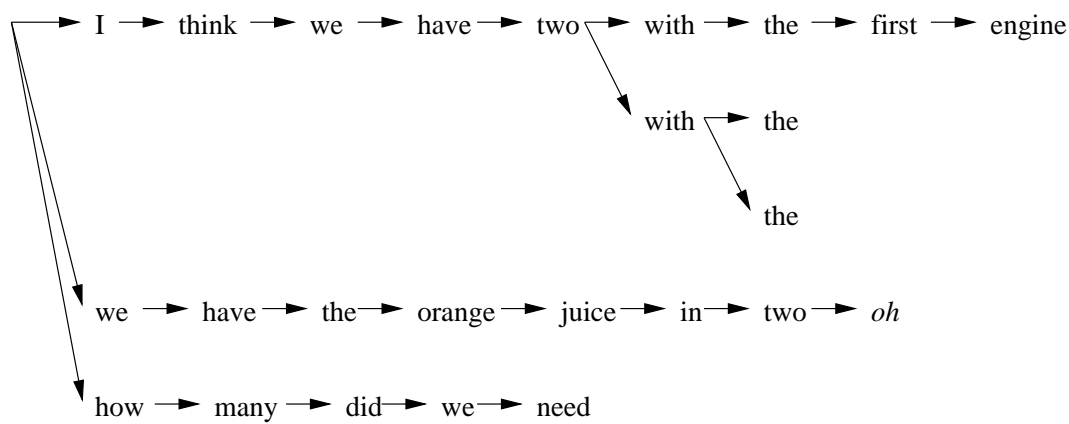}}
\caption{Branching Structure of d92a-1.3 utt75}
\label{fig:d92a-1.3:utt75}
\end{figure}

Overlapping repairs are sometimes more complicated than the one shown
in Figure~\ref{fig:d93-16.3:utt4}.  Consider the example given in
Figure~\ref{fig:d92a-1.3:utt75}.  
Here the speaker started with ``I think we have two with the first
engine''.  She then went back and repeated the words ``with the'',
making the removed speech of the first repair ``with the first
engine''.  The speaker then repeated ``the'', making the removed
speech of the second repair ``the''.  The speaker then abandoned the
current utterance of ``I think we have two with the'' and started over
with ``we have the orange juice in two'' and then uttered ``oh''.  The
speaker then abandoned even this, and replaced it with ``how many did
we need''.

The third aspect of annotating non-overlapping speech repairs is
determining the correspondences between the reparandum and resumption.
In order to treat overlapping repairs as a composition of individual
repairs, we need to determine the correspondences that should be
marked and to which repair they belong.  For non-overlapping repairs,
one annotates all of the suitable word correspondences between the
reparandum and resumption.  However, for overlapping repairs, the
occurrence of the second repair can disrupt the resumption of the
first, and the occurrence of the first repair can disrupt the
reparandum of the second.  Consider the example given in Figure~\ref
{fig:d93-16.3:utt4}.  For the first repair, ``engine'' is part of its
reparandum, but it is unclear if we should include the second instance
of ``engine'' as part of the resumption.  The decision as to whether
we include it or not impacts whether we include the correspondence as
part of the first repair.  Likewise, ``engine'' is part of the
resumption of the second repair, but it is unclear whether it is part
of the reparandum.  Again, whether we include it or not dictates
whether we include it as a correspondence of the second repair.

\subsubsection{Occurrence of Overlapping Repairs}

Before defining the reparandum and resumption of overlapping repairs,
it is worthwhile to look at the occurrence of overlapping repairs.  In
the Trains corpus, there are 1653 non-overlapping speech repairs and
315 instances of overlap made up of 743 repairs (sometimes more than
two repairs overlap).  If we remove the abridged repairs, we get 1271
non-overlapping repairs and 301 instances of overlap made up of 702
repairs.  In these 301 instances of overlap, there are 392 cases in
which two adjacent speech repairs overlap.  One way to classify
overlapping repair instances is by the relationship between the prior
of the second repair and the prior of the first.  Consider again the
example given in Figure~\ref {fig:d92a-1.3:utt75}.  The prior of the
second repair is ``with'', which is after the prior of the first
repair, which is ``two''.  The prior of the third repair is the
beginning of the utterance, and hence it is earlier than the prior of
the second repair.  The prior of the fourth repair is also the
beginning of the utterance, and hence the priors of these two repairs
coincide.  Table~\ref {tab:corpus:overlap} classifies the adjacent
overlapping repairs using this classification.
\begin{table}[tbh]
\begin{center}
\begin{tabular}{|l|r|} \hline
\multicolumn{1}{|c|}{Type}    & \mc{Frequency} \\ \hline \hline
{\em Earlier}           & 42  \\
{\em Coincides}         & 340 \\
{\em Later}             & 10  \\ \hline
\end{tabular}
\end{center}
\caption{Distribution of Overlapping Repairs}
\label{tab:corpus:overlap}
\end{table}
We find that 86\% of overlapping repair instances have priors that
coincide.  Hence, most overlapping repairs are due to the speaker
simply restarting the utterance at the same place she just restarted
from.  Since this class accounts for such a large percentage, it is
worthwhile to further study this class of speech repairs.

\subsubsection{Defining the Reparandum}

As explained above, to determine the correspondences that need to be
annotated for overlapping repairs, one must first define the
reparandum of each repair.  The reparandum of the first repair
involved in an overlap is its removed speech.  However, what are the
possibilities for the reparandum of subsequent repairs?  The answer
that probably comes first to mind is that the reparandum of a repair
is simply its removed speech.  However, consider the example in
Figure~\ref {fig:d93-16.3:utt4} of repairs with co-inciding priors.
Here, after the speaker uttered the words ``what's the
shortest route from engine'', she went back and repeated ``from'',
then changed this to ``for engine'' and then continued on with the
rest of the utterance.  But what is the speaker doing here?  We claim
that in making the second repair, the speaker might not be necessarily
fixing the removed speech of the second repair, which is the second
instance of ``from'', but may in fact have decided to take a second
attempt at the fixing the reparandum of the first repair.

As for the hearer, this is another story.  It is unclear how much of
this that the hearer is aware.  Is the hearer able to recognize the
second instance of ``from'' as such, and is he able to determine that
it corresponds with the first instance of ``from'' and with the
instance of ``for'', especially since the hearer does not have the
context that is often needed to correctly recognize the words involved
\cite {LickleyBard96:icslp}?  However, it really does not matter
whether the hearer is able to recognize all of this.  In order to
understand the speaker's utterance, he simply needs to be able to
detect the second repair and realize its resumption is a continuation
from ``route''.  So, he could even ignore the second instance of
``from'', especially if the removed speech from the first repair is
more informative.  In this case, the annotator would want to view the
reparandum of the second repair as being ``from engine'', which is the
removed speech of the first repair.

Now consider a second example in which the first repair is further
reduced.
\begin{example}{d92a-2.1 utt140}
\label{ex:d92a-2.1:utt140}
they would uh \\
w(e)- \\
we wouldn't want them both to start out at the same time
\end{example}
In this example, the speaker started to replace the reparandum by
``we'', but cut off this word in the middle.  The speaker then made a
second attempt, which was successful.  Again, it is unlikely that the
hearer did much in the way of processing the fragment, but instead
probably concentrated on resolving the second repair with respect to
``they would'', which again is the removed speech of the first repair.

Now consider a third example, an example in which the speaker reverts
back to the original reparandum.
\begin{example}{d92a-3.2 utt92}
\setlength{\tabcolsep}{0.2em}
\begin{tabular}{llllllll}
it & \em uh \\
I  & \\
it & only & takes \\
\end{tabular}
\end{example}
Here the speaker replaced ``it'' by ``I'', and then reverted back to
`it'.  Again, it is unclear how much attention the hearer paid
to ``I'', and so he might have just viewed this repair as a simple
repetition.

The fourth example is a more extensive version of the previous one.
Again, the speaker reverts back to what she originally said, and hence
we might want to capture the parallel between the removed speech of
the first repair and the resumption of the second.
\begin{example}{d93-25.5 utt57}
\setlength{\tabcolsep}{0.2em}
\begin{tabular}{lllllllll}
and & you & can & be & do- \\
    & you & don't & have \\ 
    & you & can & be & doing & things simultaneously \\
\end{tabular}
\end{example}

The examples above illustrated overlapping speech repairs
whose priors coincide.  For these examples, we have argued
that there are two candidates for the reparandum of the second repair:
the removed speech of the first repair and the removed speech of the
second repair.  In fact, we propose that the reparandum alternatives
for a repair can be defined in terms of the branching structure for
the speech that has been uttered so far.
\begin{description}
\item{\bf Reparandum Constraint:} The reparandum of a speech repair can be
any branch of the branching structure of the utterance such that the
resulting reparandum onset has an arc from the prior of the speech
repair (excluding the branch that is being created for the
resumption).
\end{description}
For the second repair in Figure~\ref{fig:d93-16.3:utt4}, this means it
can either be the removed speech ``from'' or the removed speech of the
previous repair ``from engine''.

\begin{figure}
\centerline{\psfig{file=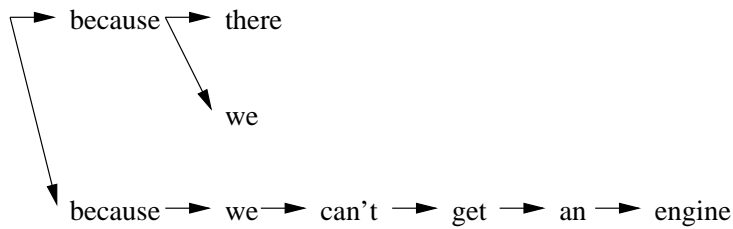}}
\caption{Branching Structure of d92-1 utt30}
\label{fig:d92-1:utt30}
\end{figure}

As we mentioned in the previous section, there are three alternatives
for overlapping repairs.  Our discussion so far has focused on the
most prevalent type: those in which the repairs share the same prior.
The reparandum constraint also accounts for repairs where the prior of the
second repair precedes the prior of the first.  For these cases of overlap,
the second repair removes speech further back along the current
utterance than the resumption of the first repair.  Consider the
example given in Figure~\ref{fig:d92-1:utt30}.
The removed speech of the first repair is ``there'', and this is
replaced with ``we'', making the current utterance ``because we''.
The second repair removes the current utterance and starts back at the
beginning.  The utterance branching shows us that there are two paths
from the root node other than the resumption.  Both alternatives start
with the node containing ``because'', which then splits into the path
containing ``there'' and the path containing ``we''.  Hence the two
possible alternatives are ``because there'' and ``because we''.

The third case is where the prior of the second repair is after the
prior of the first.  In the Trains corpus, there are only eleven
instance of this type of repair.  This type of repair also does not
cause a problem.  Consider the following repair.
\begin{example}{d92a-2.1 utt95}
\label{ex:d92a-2.1:utt95}
\setlength{\tabcolsep}{0.2em}
\begin{tabular}{lllllllllll}
a & total & of & \makebox[1em][l]{\em um let's see} \\ & total & of &
  s- \\ & & of & seven & hours \\
\end{tabular}
\end{example}
Here the removed speech of the second repair is ``of s-'', and this is
the only reparandum alternative.

More restrictions can undoubtedly be placed on the choice of
reparandum.  After a certain amount of time, branches that are not
part of the current utterance should probably be pruned back in order
to model the speaker's and the hearer's limited memory.  For instance,
for the second repair in Figure~\ref{fig:d92-1:utt30}, we might want
to exclude from consideration the branch ``because there''.  Some
branches should probably be immediately pruned; for instance, branches
that consist simply of a word fragment (see Example~\ref
{ex:d92a-2.1:utt140}), branches that simply repeat just the first part
of another path (see Figure~\ref {fig:d93-16.3:utt4}), and branches
that end in an abridged repair.  However, by allowing the annotator to
choose which path to use rather than constraining this choice, we will
be able to gather psycholinguistic data to check for meaningful
restrictions.\footnote {See Section~\ref {sec:correction:active} for
details on which paths are pruned in the current implementation.}

\subsubsection{Defining the Resumption}

The resumption of a speech repair is the second part of the equation
in defining the correspondences that can be associated with a repair.
Consider the example given in Figure~\ref{fig:d93-16.3:utt4}, which we
repeat below.
\begin{example}{d93-16.3 utt4}
\label{ex:d93-16.3:utt4:d}%
\newlength{\xxx}%
\settowidth{\xxx}{what's the shortest route}%
what's the shortest route from engine \\
\hspace*{\xxx}            from \\
\hspace*{\xxx}            for engine two at Elmira
\end{example}
There are two choices for the reparandum of the second repair.  If the
annotator thought that the hearer was not able to use the second
occurrence of ``from'' in detecting the occurrence of the second
repair or in realizing that the prior of the second repair was
``route'', then the annotator would choose ``from engine'' as the
reparandum of the second repair.  In this case, the second repair
would include the word correspondences between the first instance of
``from'' and the instance of ``for'', and between the two instances of
``engine''.  As for the first repair, what correspondences should it
include?  The speaker had intended the second instance of ``from'' to
repeat the first, and this correspondence should be included.  The
first repair also includes the first instance of ``engine'' in its
reparandum.  However, since the second repair already includes the
correspondence between the two instances of ``engine'', we do not
include it as part of the first repair.

In the above example, we considered a case of overlapping repairs in
which the reparandum of the second repair was chosen to be the removed
speech of the first repair.  Now let's consider the following example,
in which the speaker was saying ``one engine'', then changed this to
``the u-'', and then changed this to ``the first engine \dots''.
\begin{example}{d92a-1.4 utt25}
\label{ex:d92a-1.4:utt25}
one engine \\
the u-  \\
the first engine will go back to Dansville 
\end{example}
Let's assume that the annotator decided that the correspondence
between the two instances of ``the'' helps the hearer identify the
second repair.  In this case, the annotator would choose ``the u-'' as
the reparandum of the second repair, and thus only the correspondence
between the two instances of ``the'' would be included in this repair.
Now we need to determine the correspondences that should be included
in the first repair, whose reparandum is ``one engine''.  Here, one
might argue that the hearer is only able to identify the first repair
after resolving the second repair and hearing ``the first engine'' and
its prosodic pattern \cite {Shriberg94:thesis}.  This would imply that
the resumption of the first repair should be ``the first engine''.

However, the problem with this analysis is that one must look forward
to the subsequent repairs before annotating the correspondences with
the previous repairs.  In uttering the first instance of ``the'', the
speaker has undoubtedly decided that this word is replacing ``one''.
As for the hearer, even though he uses the repetition of
``the'' to identify the second repair, this does not preclude him from
using the first instance of ``the'' to help identify the first repair.
It might not be until after he has heard ``the first engine'' that he
resolves the ambiguity, but the ambiguity probably started after
hearing the first instance of ``the''
(cf.~\cite{LickleyBard92:icslp}).  Hence, the resumption of the first
repair should include the first instance of ``the''.  The second
instance of ``the'' should not be included since this is a replacement
for the ``the'' already included in the resumption.  The resumption of
the first repair also needs to include the second instance of
``engine'' since this helps confirm the first repair.

We have used the above two examples to argue that the resumption of a
speech repair should not include the alteration of a subsequent
repair.  For the first repair in Example~\ref {ex:d93-16.3:utt4:d}, we
excluded ``engine'' from the resumption since it was already part of
the alteration of the second repair.  For the first repair in
Example~\ref {ex:d92a-1.4:utt25}, we argued for the exclusion of the
second instance of ``the'', which was already part of the alteration
of the second repair.  In fact, excluding the alteration of subsequent
repairs gives us the resumption of the earlier repairs.
\begin{description}
\item{\bf Resumption Constraint:} The resumption of a repair includes
all speech after its editing term but excluding the alterations of 
subsequent repairs.
\end{description}
One of the implications of this constraint is that it lets us view
overlapping repairs in an incremental manner.  Our annotation of a
speech repair does not need to be revised if we encounter a subsequent
overlapping repair.  It also means that each word following the
interruption point of a speech repair is predicted by at most one word
that precedes it.  These two features allow us to treat overlapping
repairs as a straight-forward extension of non-overlapping repairs in
our model of correction that we propose in Chapter~\ref
{chapter:correction}.

\subsubsection{Annotation Scheme}

In the previous two sections, we presented the constraints on the
reparandum and resumption of an overlapping repair.  Once the
annotator has determined the reparandum and resumption for an
overlapping repair, the repair can be annotated following the rules
for non-overlapping repairs.  Although the reparandum of a repair
might not be its removed speech, we do not need to annotate both the
removed speech of a repair in addition to the reparandum.  By
annotating just the reparandum we can automatically determine the
extent of the removed speech.

To illustrate the annotation scheme, consider the example given in
Figure~\ref {fig:d93-16.3:utt4} and assume that the annotator has
decided that the reparandum of the second repair is ``from engine''.
Hence, the alteration of the second repair is ``for engine'', and the
alteration of the first repair is the second instance of ``from''.
This repair would be annotated as follows.
\begin{example}{d93-16.3 utt4}
\setlength{\tabcolsep}{0.2em}
\begin{tabular}{rrrrrrrrrrrr}
what's & the & shortest & route & from & engine & from & for & engine
& two & at & Elmira  \\
       &     &          &       & \small\bf m11 
				       & \small\bf x10  
						& \small\bf m11 \\
       &     &          &       &      &\small\bf ip10:mod \\ 
       &     &          &       & \small\bf r21 
				       &\small\bf m22
						&      & \small\bf r21
							     & \small\bf m22 \\
       &     &          &       &      &        & \small\bf ip20:mod \\ 
\end{tabular}
\end{example} 

Now consider Example~\ref {ex:d92a-2.1:utt95}.  Here, the first speech
repair involves repeating ``total of'', and the second one involves
replacing ``of s-'' by ``of seven'' hours.  This repair would be
simply annotated as follows.
\begin{example}{d92a-2.1 utt95}
\setlength{\tabcolsep}{0.2em}
\begin{tabular}{rrrrrrrrrrrrr}
a & total & of & um & let's & see & total & of & s- & of & seven & hours \\
  & \small\bf m1 
	  & \small\bf m2
	       & \small\bf et
		    & \small\bf et
			    & \small\bf et
				  & \small\bf m1
					  & \small\bf m2 \\
  &       & {\small\bf ip:mod} \\
  &       &    &    &       &     &       & \small\bf m11 
					       & \small\bf m12
						    & \small\bf m11
							 & \small\bf m12 \\
  &       &    &    &       &     &       &    & \small\bf ip10:mod \\
\end{tabular}
\end{example} 

Some word correspondences cannot be captured by our scheme.  Consider
the following example.
\begin{example}{d93-18.2 utt28}
\setlength{\tabcolsep}{0.2em}
\begin{tabular}{llllllll}
it & just \\
it & picks & up & \\ 
it & just & picks & up two tankers \\
\end{tabular}
\end{example}
In this example, the resumption of the second repair is ``it just
picks up two tankers'', where ``just'' is a repetition from the
removed speech of the first repair and ``picks up'' is a repetition
from the removed speech of the second repair.  However, since the
reparandum of the second repair must be either the removed speech of
the first repair {\em or} the removed speech of the second repair,
both sets of correspondences cannot be annotated.  Such examples of
overlapping repairs are very rare in the Trains corpus.

\subsection{Comparison to Shriberg's Scheme}
\label{sec:corpus:repairs:shriberg}

Shriberg \shortcite{Shriberg94:thesis} also has proposed an annotation
scheme that can account for overlapping repairs.  Like our scheme, it
is an adaption of the scheme proposed by Bear~\etal~\shortcite
{Bear-etal93:tr}.  The goal of Shriberg's scheme is the same as ours:
overlapping repairs should be treated as a composition of individual
repairs.  Unlike our approach in which overlapping repairs can share
the same reparandum but not the same alteration, she advocates the
exact opposite.  The annotator specifies the order in which the
overlapping repairs are resolved.  As each repair is resolved, its
alteration is available to be annotated by the next repair, but not
its reparandum.

To show the order of evaluation, brackets are used to enclose the
reparandum and alteration of each repair.  Although this scheme works
for most overlapping repairs, problems can arise.  Consider
Example~\ref{ex:d92a-2.1:utt95}, repeated below.\footnote{This repair
is similar to her example ``show me the flight the delta flight delta
fare''.  Shriberg calls this a {\em partially chained structure}.}
\begin{example}{d92a-2.1 utt95}
\setlength{\tabcolsep}{0.2em}
\begin{tabular}{lllllllllll}
a & total & of & {\em um let's see} \\
  & total & of & s- \\
  &       & of & seven hours \\
\end{tabular}
\end{example}
Here the first repair involves the words ``total of total of'',
whereas the second involves ``of s- of seven''.  So the alteration of
the first repair overlaps with the reparandum of the second, but
neither is totally embedded in the other.  The annotation for the
first repair would be `[m m.m m]'.\footnote{We have translated her
symbol for word match (repetition) `{\bf r}' to our symbol `{\bf m}'.
Likewise we have translated her symbol for word replacement
(substitution) `{\bf s}' to our symbol `{\bf r}'.}  Since the second
repair needs part of the alteration of the first, the entire
alteration of the first must be included in annotating the reparandum
of the second repair.  But, the word ``total'' is not part of the
second repair.  Hence Shriberg uses the symbol `\#' to indicate that
``total'' ``is merely a word in the fluent portion of the sentence at
the level of the analysis of the second [repair]'' (pg.~72).  The
resulting annotation is as follows.
\begin{example}{d92a-2.1 utt95}
\setlength{\tabcolsep}{0.2em}
\begin{tabular}{cccccccccccccccccc}
a& total& [& OF& [& total& of& .& total& of& ]& s-& .& of& seven &] &hours\\
 &   \# & [& M & [&   m  & m & .&   m  & m & ]& m & .& m &   m   &] \\
\end{tabular}
\end{example}
After the first repair is resolved, its alteration, namely the two
words ``total of'', are passed to the next repair.  But since there is
only one symbol (and not two), the first word is taken to be part of
``the fluent utterance'', which is further indicated by the preceding
`\#'.\footnote{It would seem to make more sense to use the `\#' inside
of the bracket, which would lead to the annotation of `[M \# [m m.m m]
m.m m]'.}

Now consider the example given in Figure~\ref{fig:d93-16.3:utt4},
repeated below.
\begin{example}{d93-16.3 utt4}
\setlength{\tabcolsep}{0.2em}
\begin{tabular}{lllllllllllllllll}
what's & the & shortest & route & from & engine & \\
       &     &          &       & from & \\
       &     &          &       & for  & engine & two & at & Elmira \\
\end{tabular}
\end{example}
Here if the annotator decided that the second repair is resolved
first, the resulting annotation is `[r m.R[r.r] m]'.  If the annotator
decides that the first repair is resolved first and wanted to capture
the correspondence on ``engine'', it is unclear if the `\#' symbol can
be used in her system to pass an unused portion of the reparandum of
one repair to a later one.  If we take a more liberal definition of
`\#' than perhaps Shriberg intended (as described in the preceding
footnote), we could annotate this repair as `[R M [m \#.m].r m]'.
Note that due to the difference in perspective as to whether
overlapping repairs can share alterations or reparanda, neither of the
above two interpretations are equivalent to the two interpretations
that our annotation scheme offers for this example.

\subsection{Editing terms}
\label{sec:corpus:et}

Speakers usually restrict themselves to a small number of editing
terms.  Table~\ref{tab:ets} lists the number of occurrences of the
editing terms found in the Trains corpus that occur at least twice.
\begin{table}[tbh]
\begin{center}
\begin{tabular}{|l|r|} \hline
um & 303 \\
uh & 261 \\
okay & 64 \\
oh & 44 \\
let's see & 36 \\
well & 33 \\
no & 31 \\
or & 29 \\
hm & 25 \\
yeah & 23 \\
alright & 12\\ 
let me see & 11 \\
I mean & 10 \\
actually & 10 \\
like & 10 \\ 
wait & 10 \\ \hline
\end{tabular} \hspace{1em}
\begin{tabular}{|l|r|} \hline
er & 9 \\
mm & 9 \\
I guess & 6 \\
sorry & 5 \\
then & 4 \\
I'm sorry & 3 \\
let me think & 3 \\
ooh & 3 \\
right & 3 \\
yes & 3 \\
you know & 3 \\
boy & 2 \\
excuse me & 2 \\
let's see here & 2 \\
oops & 2 \\ 
 & \\ \hline
\end{tabular}
\end{center}
\caption{Occurrences of Editing Terms in the Trains Corpus}
\label{tab:ets}
\end{table}
Levelt \shortcite{Levelt83:cog} noted that editing terms can give
information as to the type of repair that a speaker is making, and
Hindle \shortcite{Hindle83:acl} used the presence of certain types of
editing terms, such as ``well'', and ``okay'', as evidence of a fresh
start.  Note that some speech repairs have complex editing terms that
consist of a number of these basic ones, as the following example
illustrates.
\begin{example}{d92a-4.2 utt13}
I guess I gotta \et{let's see here alright um uh} I want to take engine two 
\end{example}

Editing terms are almost always uttered before the alteration.
However, in the Trains corpus, there are a few examples that do not
follow this pattern.  The next example illustrates a common editing
term being used at the end of an utterance.
\begin{example}{d93-12.4 utt96}
we'd be in Elmira at five a.m.\ip five p.m. I m- I mean
\end{example}
In this example, there is an intonational phrase ending on the word
``a.m.'', making it questionable whether this is a speech repair or a
repair at a deeper cognitive level.  If ``I mean'' is being used as an
editing term in this example, then it also illustrates how editing
terms can be the subject of speech repairs, a phenomena that we have
also not explored in this thesis.

Another problem in annotating editing terms is that discourse markers
are sometimes ambiguous as to whether they are part of the editing
term or part of the alteration.  Consider the following example.
\begin{example}{d92a-4.2 utt97}
\label{ex:d92a-4.2:utt97}
well we could go\ip  well we have time to spare right 
\end{example}
In this example, one could posit that the second instance of ``well''
is being used by the speaker as a comment about the relationship
between the reparandum and alteration and hence would be viewed as an
editing term. A second alternative is that the second occurrence of
``well'' is part of the alteration since it seems to be used as a word
correspondence with the first ``well''.

\section{POS Annotations}
\label{sec:corpus:pos}

We have also annotated the Trains corpus with part-of-speech (POS)
tags.  As our starting point, we used the tagset provided with the
Penn Treebank \cite {Marcus-etal93:cl,Santorini90:tr}.  We have
modified their tagset to add POS tags for discourse markers and turns.
We have also modified their tagset so that it provides more precise
syntactic information.  The list below gives the changes we have made.
\begin{enumerate}
\item
Removed all of the punctuation tags, since punctuation does not occur
in spoken dialog.  Instead, we add tags that are more appropriate for
spoken dialog.  We add the tag {\bf TURN} to indicate change in
speaker turn, which is marked with the pseudo-word {\bf $<$turn$>$}.
In Section~\ref {sec:detection:addtags}, we add extra tags for
marking boundary tones and speech repairs.
\item 
Divided the {\bf IN} class into prepositions {\bf PREP} and
subordinating conjunctions {\bf SC}.
\item 
Moved instances of ``to'' that are used as a preposition from the
class {\bf TO} to the class of prepositions {\bf PREP}.  The tag {\bf
TO} is now only used for the instances of ``to'' that are part of a
to-infinitive.
\item 
Separated conjugations of ``be'', ``have'', and ``do'' from the other
verbs.  For the base form, we use {\bf BE}, {\bf HAVE}, and {\bf DO},
respectively.  Note the present and past participles for ``have'' and
``do'' have not been separated.
\item 
Separated interjections into single word acknowledgments {\bf AC},
discourse interjections {\bf UH\_D}, and filled pauses {\bf UH\_FP}.
\item 
Added discourse marker versions for {\bf CC} and {\bf RB} by adding
the suffix `{\bf \_D}'.
\item 
Removed the pro-form of determiners from the class {\bf DT} and put them
into the new class of {\bf DP}.
\item 
Redefined the class {\bf WDT} to be strictly for `wh-determiners' by
moving the pro-form usages of ``which'' to {\bf WP}.
\item 
Added the class {\bf PPREP}, which is for the leading preposition of a 
phrasal preposition.
\end{enumerate}

Table~\ref{tab:pos} gives a complete listing of the resulting
tagset.
\begin{table}
\begin{minipage}[t]{2.8in}
\begin{list}{}{
\setlength{\baselineskip}{1.20em}
\setlength{\parsep}{0em}
\setlength{\itemsep}{0.20em}
\settowidth{\labelwidth}{\bf VBG\_TO}
\setlength{\leftmargin}{\labelwidth}
\addtolength{\leftmargin}{\labelsep}
\renewcommand{\makelabel}[1]{\makebox[\labelwidth][l]{#1}}
}
\item[\bf AC] Acknowledgement
\item[\bf BE] Base form of ``be''
\item[\bf BED] Past tense
\item[\bf BEG] Present participle
\item[\bf BEN] Past participle
\item[\bf BEP] Present
\item[\bf BEZ] 3rd person singular present
\item[CC] Co-ordinating conjunction
\item[\bf CC\_D] Discourse connective
\item[CD] Cardinal number
\item[\bf DO] Base form of ``do''
\item[\bf DOD] Past tense
\item[\bf DOP] Present
\item[\bf DOZ] 3rd person singular present
\item[\bf DP] Pro-form 
\item[DT] Determiner
\item[EX] Existential ``there''
\item[\bf HAVE] Base form of ``have''
\item[\bf HAVED] Past tense
\item[\bf HAVEP] Present
\item[\bf HAVEZ] 3rd person singular present
\item[JJ] Adjective
\item[JJR] Relative Adjective
\item[JJS] Superlative Adjective
\item[MD] Modal
\item[NN] Noun
\item[NNS] Plural noun
\item[NNP] Proper Noun
\end{list}
\end{minipage}
\hfill
\begin{minipage}[t]{2.8in}
\begin{list}{}{
\setlength{\baselineskip}{1.20em}
\setlength{\parsep}{0em}
\setlength{\itemsep}{0.20em}
\settowidth{\labelwidth}{\bf VBG\_TO}
\setlength{\leftmargin}{\labelwidth}
\addtolength{\leftmargin}{\labelsep}
\renewcommand{\makelabel}[1]{\makebox[\labelwidth][l]{#1}}}
\item[NNPS] Plural proper Noun
\item[PDT] Pre-determiner
\item[POS] Possessive
\item[\bf PPREP] Pre-preposition
\item[\bf PREP] Preposition
\item[PRP] Personal pronoun
\item[PRP\$] Possessive pronoun
\item[RB] Adverb
\item[RBR] Relative Adverb
\item[RBS] Superlative Adverb
\item[\bf RB\_D] Discourse adverbial
\item[RP] Reduced particle
\item[\bf SC] Subordinating conjunction
\item[TO] To-infinitive
\item[\bf TURN] Turn marker
\item[\bf UH\_D] Discourse interjection 
\item[\bf UH\_FP] Filled pause
\item[VB] Base form of verb (other than `do', `be', or `have') 
\item[VBD] Past tense
\item[VBG] Present participle
\item[VBN] Past participle
\item[VBP] Present tense
\item[VBZ] 3rd person singular present
\item[WDT] Wh-determiner
\item[WP]  Wh-pronoun
\item[WRB] Wh-adverb
\item[WP\$] Processive Wh-pronoun
\end{list}
\vspace{0.5em}
\end{minipage}
\caption{Part-of-Speech Tags Used in the Trains Corpus}
\label{tab:pos}
\end{table}%
The tags in bold font are those that differ
from the Penn Treebank tagset. The tags {\bf POS}, {\bf NNPS} and {\bf
WP\$} did not occur in the Trains corpus, but are included for
completeness. There are other tagsets that capture much more
information \cite {Greene-Rubin81:brown,Johansson-etal86:lob}.
However, because of the small size of the Trains corpus, there might
not be enough data to capture the additional distinctions.\footnote
{See page~\pageref {page:model:q:tags} for how finer grain syntactic
distinctions that are not captured by the tagset can be automatically
learned.}

Contractions, such as ``can't'' and ``gonna'', are composed of two
separate words, each having a separate syntactic role.  Rather than
create special tags for these words, we annotate them in a manner
analogous to how we annotate contractions with the speech repair word
correspondences.  We annotate such words with both POS tags and use
the symbol `$\wedge$' to separate them; for instance, ``can't'' is
annotated as `{\bf MD}$\wedge${\bf RB}'.  For language modeling,
contractions are split into two separate words, each with their
respective POS tag, as is described in Section~\ref {sec:model:setup}.

\section{Discourse Marker Annotations}
\label{sec:corpus:dm}

Our strategy for annotating discourse markers is to mark such usages
with special POS tags, as specified in the previous section.  Four
special POS tags are used.
\begin{description}
\item[AC] Single word acknowledgments, such as ``okay'', ``right'', ``mm-hm'',
``yeah'', ``yes'', ``alright'', ``no'', and ``yep''. 
\item[UH\_D] Interjections with discourse purpose, such as ``oh'''', ``well'', ``hm'', ``mm'', and ``like''.
\item[CC\_D] Co-ordinating conjuncts used as discourse markers, such as
``and'', ``so'', ``but'', ``oh'', and ``because''.
\item[RB\_D] Adverbials used as discourse markers, such as ``then'', ``now'',
``actually'', ``first'', and ``anyway''.
\end{description}
Verbs used as discourse markers, such as ``wait'', and ``see'', are
not given special markers, but are annotated as \mbox{\bf VB}.  Also,
no attempt has been made at analyzing multi-word discourse markers,
such as ``by the way'' and ``you know''.  However, phrases such as
``oh really'' and ``and then'' are treated as two individual discourse
markers.  Note, however, that when these phrases are used as editing
terms of speech repairs, such as ``let's see'', their
usage is captured by the editing term annotations given in
Section~\ref {sec:corpus:et}.  Lastly, although the filled pause words
``uh'', ``um'' and ``er'' are marked with {\bf UH\_FP}, we do not
consider them as discourse markers, but simply as filled pauses.

\cleardoublepage
\chapter{POS-Based Language Model}
\label{chapter:model}

The underlying model that we use to account for speakers' utterances
is a statistical language model.  Statistical language models that
predict the next word given the prior words, henceforth referred to
as {\em word-based} language models, have proven effective in helping
speech recognizers prune acoustic candidates.  Statistical language
models that predict the POS categories for a given sequence of
words---{\em POS taggers}---have proven effective in processing
written text and in providing the base probabilities for statistical
parsing.  In this chapter, we present a language model intended for
speech recognition that also performs POS tagging.  The goal of this
model is to find the best word and POS sequence, rather
than simply the best word sequence.  We refer to this model as a
POS-based language model. A concise overview of the work presented in
this chapter is given by Heeman and Allen \shortcite
{HeemanAllen97:eurospeech}.\footnote{The results given in this chapter
reflect a number of small improvements over the approach given by
Heeman and Allen \shortcite {HeemanAllen97:eurospeech}.}

Our original motivation for proposing a POS-based language model was
to make available shallow syntactic information in a speech
recognition language model, since such information is needed for
modeling the occurrence of speech repairs and boundary tones.
However, the POS tags are useful in their own right.  Recognizing the
words in a speaker's turn is only the first step towards understanding
a speaker's contribution to a dialog.  One also needs to determine the
syntactic structure of the words involved, their semantic meaning, and
the speaker's intention.  In fact, this higher level processing is
needed to help the speech recognizer constrain the alternative
hypotheses.  Hence, a tighter coupling is needed between speech
recognition and the rest of the interpretation process.  As a starting
point, we integrate shallow syntactic processing, as realized by POS
tags, into a speech recognition language model.

In the rest of this chapter, we first redefine the speech recognition
problem so that it incorporates POS tagging and discourse marker
identification.  Next, we introduce the decision tree algorithm, which
we use to estimate the probabilities that the POS-based language model
requires.  To allow the decision tree to ask meaningful questions
about the words and POS tags in the context, we use the clustering
algorithm of Brown~\etal~\shortcite {Brown-etal92:cl}, but adapted to
better deal with the combination of POS tags and word identities.  We
then derive the word perplexity measure for our POS-based language
model.  This is then followed by a section giving the results of our
model, in which we explore the various trade-offs that we have made.
Next, we contrast the POS-based model with a word-based model, a
class-based model, and a POS-based model that does not distinguish
discourse markers.  We also explore the effect of using a decision
tree algorithm to estimate the probability distributions.  In the
final section, we make some concluding remarks about both using POS
tags in a language model and the use of the decision tree algorithm in
estimating the probability distributions.

\section{Redefining the Speech Recognition Problem}
\label{sec:model:tagger}

As we mentioned in Section~\ref{sec:related:wordmodel}, the goal of a
speech recognition language model is to find the sequence of words
$\hat{W}$ that is most probable given the acoustic signal $A$.	
\begin{eqnarray}
\hat{W} = \arg\max_{W} \Pr(W|A) \label{eqn:model:model1}
\end{eqnarray}
To add POS tags into this language model, we refrain from simply
summing over all POS sequences as illustrated in Section~\ref
{sec:related:aclass}.  Instead, we redefine the speech recognition
problem as finding the best word and POS sequence.  Let $P$ be a
POS sequence for the word sequence $W$, where each POS tag is an
element of the tagset ${\cal P}$.  The goal of the speech recognition
process is to now solve the following.
\begin{eqnarray}
\label{eqn:model:2}
\hat{W}\hat{P} & = & \arg\max_{W,P} \Pr(WP|A) \label{eqn:model:model2}
\end{eqnarray}

Now that we have introduced the POS tags, we need to derive the
equations for the language model.  Using Bayes' rule, we rewrite
Equation~\ref{eqn:model:2} in the following manner.
\begin{eqnarray}
\label{eqn:model:3}
\hat{W}\hat{P} & = & \arg\max_{WP} \frac{\Pr(A|WP)\Pr(WP)}{\Pr(A)}
\end{eqnarray}
Since $\Pr(A)$ is independent of the choice of $W$ and $P$, we can
simplify Equation~\ref{eqn:model:3} as follows.
\begin{eqnarray}
\hat{W}\hat{P} & = & \arg\max_{WP} \Pr(A|WP)\Pr(WP)
\end{eqnarray}
The first term $\Pr(A|WP)$ is the probability due to the acoustic
model, which traditionally excludes the category assignment.  In fact,
the acoustic model can probably be reasonably approximated by
$\Pr(A|W)$.\footnote{But see Lea \shortcite{Lea80} for how POS can
affect acoustics.}

The second term $\Pr(WP)$ is the probability due to the POS-based
language model and this accounts for both the sequence of words and
the POS assignment for those words.  We rewrite the sequence $WP$
explicitly in terms of the $N$ words and their corresponding POS tags,
thus giving us the sequence $W_{1,N}P_{1,N}$.  As we showed in
Equation~\ref {eqn:related:tagger:3} of Section~\ref
{sec:related:tagger}, the probability $\Pr(W_{1,N}P_{1,N})$ forms the
basis for POS taggers, with the exception that POS taggers work from a
sequence of given words.  Hence the POS tagging equation can be used
as a basis for a speech recognition language model.

As in Equation~\ref{eqn:related:tagger:4}, we rewrite the probability
$\Pr(W_{1,N}P_{1,N})$ as follows using the definition of conditional
probability.
\begin{eqnarray}
\Pr(W_{1,N}P_{1,N})
&=& \prod_{i=1}^N \Pr(W_iP_i|W_{\rim}P_{\rim})  \\
&=& \prod_{i=1}^N \Pr(W_i|W_{\rim}P_{\ri})
                  \Pr(P_i|W_{\rim}P_{\rim}) \label{eqn:model:6}
\end{eqnarray}
Equation~\ref {eqn:model:6} involves two probability distributions
that need to be estimated.  As we discussed in Section~\ref
{sec:related:tagger} and Section~\ref {sec:related:aclass}, most POS
taggers and previous attempts at using POS tags in a language model
simplify these probability distributions, as given in Equations~\ref
{eqn:related:tagger:a1} and \ref {eqn:related:tagger:a2}.  However, to
successfully incorporate POS information, we need to account for the
full richness of the probability distributions.  Hence, we need to
learn the probability distributions while working under the following
assumptions.
\begin{eqnarray}
\Pr(W_i|W_\rim P_\ri)  & \not\approx & \Pr(W_i|P_i) \label{eqn:model:a1} \\
\Pr(P_i|W_\rim P_\rim) & \not\approx & \Pr(P_i|P_\rim) \label{eqn:model:a2}
\end{eqnarray}
Section~\ref {sec:model:r:richer} will give results contrasting
various simplification assumptions.

As we mentioned at the beginning of this section, our approach to
using POS tags as part of language modeling is novel in that we view
the POS tags as part of the output of the speech recognition process,
rather than as intermediate objects.  Hence, our approach does not sum
over all of the POS alternatives; rather, we search for the best word
and POS interpretation.  This approach can in fact lead to different
word sequences being found.  Consider the following contrived example
in which there are two possibilities for the $i$th word---$w$ and
$x$---and three possible POS tags $p$, $q$ and $r$.  Also assume that
there is only one choice for the POS tags and words for the prior
context $P_{\rim}W_{\rim}$, which we will refer to as $\mbox{\em
prior}_i$, and only a single choice for the words and POS tags that
follow the $i$th word.  Let the lexical and POS probabilities for the
$i$th word be as given in the first two columns of Table~\ref
{tab:model:findbest}, and let all other probabilities involving $w$
and $x$ and the three POS tags be the same.
\begin{table}
\begin{center}
\begin{tabular}{|ll|ll|ll|} \hline
$\Pr(w|p\ \mbox{\em prior}_i)$        &\hspace*{-0.5em}= 0.7 & 
$\Pr(p|\mbox{\em prior}_i)$           &\hspace*{-0.5em}= 0.5 &
$\Pr(w p|\mbox{\em prior}_i)$         &\hspace*{-0.5em}= 0.35 \\
$\Pr(w|q\ \mbox{\em prior}_i)$        &\hspace*{-0.5em}= 0.0 & 
$\Pr(q|\mbox{\em prior}_i)$           &\hspace*{-0.5em}= 0.3 &
$\Pr(w q|\mbox{\em prior}_i)$         &\hspace*{-0.5em}= 0.00 \\
$\Pr(w|r\ \mbox{\em prior}_i)$        &\hspace*{-0.5em}= 0.0 & 
$\Pr(r|\mbox{\em prior}_i)$           &\hspace*{-0.5em}= 0.2 &
$\Pr(w r|\mbox{\em prior}_i)$         &\hspace*{-0.5em}= 0.00 \\ \hline
$\Pr(x|p\ \mbox{\em prior}_i)$   &\hspace*{-0.5em}= 0.2 & 
$\Pr(p|\mbox{\em prior}_i)$           &\hspace*{-0.5em}= 0.5 &
$\Pr(x p|\mbox{\em prior}_i)$    &\hspace*{-0.5em}= 0.10 \\
$\Pr(x|q\ \mbox{\em prior}_i)$   &\hspace*{-0.5em}= 0.6 & 
$\Pr(q|\mbox{\em prior}_i)$           &\hspace*{-0.5em}= 0.3 &
$\Pr(x q|\mbox{\em prior}_i)$    &\hspace*{-0.5em}= 0.18 \\
$\Pr(x|r\ \mbox{\em prior}_i)$   &\hspace*{-0.5em}= 0.5 & 
$\Pr(r|\mbox{\em prior}_i)$           &\hspace*{-0.5em}= 0.2 &
$\Pr(x r|\mbox{\em prior}_i)$    &\hspace*{-0.5em}= 0.10 \\ \hline
\end{tabular}
\end{center}
\caption{Finding the Best Interpretation}
\label{tab:model:findbest}
\end{table}
From the third column of Table~\ref {tab:model:findbest}, we see that
using the traditional approach of deciding the word based on summing
over the POS alternatives gives a probability of 0.38 for word $x$ and
0.35 for word $w$.  Thus word $x$ is preferred word $w$.  However, our
approach, which chooses the best word and POS combination, prefers
word $w$ with POS $p$ with a probability of 0.35.  Hence, our approach
takes into account higher level syntactic information that the
traditional model just sums over.

For simple word classes, where each word belongs to only one class,
the use of the classes does not add any complexity to the task.
However, POS tags are ambiguous.  For each word assignment, there
might be many candidate POS interpretations.  Just as speech
recognizers keep multiple candidate word assignments, we need to
do the same for POS assignments as well.

\section{Learning the Probabilities}

In the previous section, we derived the probability distributions
needed for a POS-based language model.  To estimate these, we need to
take advantage of both the POS tags and word identities in the
context.  Traditional backoff approaches \cite{Katz87:assp} require a
hand-crafted strategy that specifies how to simplify the context if
there is not enough data.  Take for instance the context of $P_{i-1}
W_{i-1} P_{i-2}$.  Here, one would need to specify whether to back off
to $P_{i-1} W_{i-1}$ or $P_{i-1} P_{i-2}$.\footnote{The third choice
is $W_{i-1} P_{i-2}$, but is probably not worth considering.}  This
problem is compounded if one also wants to incorporate in a
class-based approach (introduced in Section~\ref {sec:related:class})
to allow generalizations between similar words and between similar POS
tags.  To illustrate, let $P^\prime$ represent the class that the POS
tag $P$ is in, and $W^\prime$ represent the class that word $W$ is in.
Now, when backing off from the context of say $P_{i-1} W^\prime_{i-1}
P^\prime_{i-2}$, we would have to choose between $P_{i-1}
W_{i-1}^\prime$, $P_{i-1} P_{i-2}^\prime$ and $P_{i-1}^\prime
P_{i-2}^\prime$.  If one wants to take advantage of hierarchical
classes, the problem is yet further compounded.

The problem of deciding the backoff strategy is not just limited to
backoff based approaches.  Even using interpolated
estimation, one would need to interpolate between all
possible strategies.  For instance, a trigram model with classes
for both the POS tags and words would have 64 interpolated
probabilities for the context of $P_{i-1} W_{i-1} P_{i-2}
W_{i-2}$.\footnote{Each term can be represented in entirety, by its
class, or removed from consideration.  Since there are four terms, each
having three possible backoff states, we get $4^3$ terms in the
interpolation formula.}

An alternative approach, as described in Section~\ref
{sec:related:dt}, is to use decision trees \cite
{Breiman-etal84:book,Bahl-etal89:tassp} to estimate the probability
distributions.  Decision trees have the advantage that they use
information theoretic measures to automatically choose how to
subdivide the context to provide more specific information and which
contexts are equivalent. Hence, there is less danger in adding extra
conditioning information, since we can rely on the decision tree
algorithm to decide what information is relevant. This allows us to
use extra information about the context that traditional approaches do
not use (e.g.~\cite{DeRose88:cl,Church88:anlp,Charniak-etal93:aaai}),
such as both word identities and POS tags, and even hierarchical
clusterings of them.  The approach of using decision trees will become
even more critical in the next two chapters where the probability
distributions will be conditioned on even richer context. This will
make it almost impossible to hand-craft a backoff strategy and
equivalence classes of contexts. Thus the decision tree algorithm will
be the only realistic option for estimating the probability
distributions involved in those models.  Of course, since the decision
tree algorithm is greedy and searches for locally optimal questions,
there is still some need for hand-crafting restrictions.

In Section~\ref{sec:related:dt}, we discussed the basis of decision
trees. In the rest of this section, we first discuss the type of
questions that we allow the decision tree to ask.  We then discuss how
we actually grow the trees, and then how we compute the probability
distribution given the hierarchy of equivalence classes that the
decision tree algorithm has found.

The decision tree algorithm uses training data from which the tree is
built. The training data is divided into two parts: growing data, and
heldout data. The heldout data is used by the algorithm to help reject
spurious correlations found in the growing data.  We use 30\% of the
training data as heldout data and the remaining 70\% for growing data.

\subsection{The Questions}

One of the most important aspects of using a decision tree algorithm
is the form of the questions that it is allowed to ask.  Our
implementation of the decision tree algorithm allows two basic types
of information to be used as part of the context: numeric and
categorical.  For a numeric variable $N$, the decision tree searches
for questions of the form `is $N >= n$', where $n$ is a numeric
constant. For a categorical variable $C$, it searches over questions
of the form `is $C \in S$' where $S$ is a subset of the possible
values of $C$.  However, neither of these two types are adequate for
representing word and POS information. (They will, however, be used
when we augment the POS-based model to account for occurrences of
speech repairs and boundary tones.)  Hence, we next address the
problem of dealing with word and POS information in the context.  In
the final part of this section, we address the issue of how we allow
more complicated questions.

\subsubsection{Word Identities and POS tags}
\label{sec:model:ctrees}

The context that we use for estimating the probabilities includes both
the word identities and their POS tags. As discussed in
Section~\ref{sec:related:dt:words}, there are two ways of allowing
decision trees to make use of this information.  We can let the
decision tree view them as categorical data and hence search for good
partitionings of the POS tags and word identities, or we can build
binary classification trees of the POS tags and word identities and
use these to restrict the partitionings that the decision tree can ask
about.  For the reasons discussed in Section~\ref{sec:related:dt:words},
we use the latter approach.

Previous approaches that use both POS tags and word identities as part
of the context for a decision tree algorithm (e.g.~\cite
{Black-etal92:darpa:hbg,Black-etal92:darpa:pos,Magerman94:thesis})
have viewed the POS tags and word identities as two separate sources
of information.  However, we take the approach of viewing the word
identities as a further refinement of the POS tags.  Hence, we build a
word classification tree for each POS tag, and only allow the decision
tree to ask word identity questions only when the POS tag for the
particular word is uniquely determined by the previous questions.
There are a number of advantages to the approach of building a word
classification tree for each POS tag (see Section~\ref
{sec:model:r:ctrees} for results).
\begin{enumerate}
\item If a word classification tree is built that does not take 
into account the POS tags, then the word information will often not be
consistent with the POS tags. This inconsistency will lead to
unnecessary data fragmentation.
\item By building a word classification tree for each POS tag, the 
word classification hierarchy will not be polluted by words that are
ambiguous as to their POS tag, as exemplified by the word ``may'',
which can be used as a modal or as the name of a month (cf.~\cite
{Charniak93:book}).  In the Trains corpus, this ambiguity does not
exist for the word ``may'', but it does exist for many other words.
For instance, the word ``that'' is used as a subordinating conjunct
{\bf SC}, a determiner {\bf DT}, a demonstrative pronoun {\bf DP} and
as a relative pronoun {\bf WP}; and the word ``loads'' is used as a
third-person present tense verb {\bf VBZ} and as a plural noun {\bf
NNS}.\footnote{Note that the POS tags, however, will not capture all
variations in meaning or syntactic role.  For instance, in the
examples ``We {\em need} a boxcar'' and ``We {\em need} to send a
boxcar to Avon'', both uses of the word ``need'' are tagged as a
present tense verb ({\bf VBP}).  Since each distinct word can appear
at most once in the classification tree, there is no facility in which
the classification tree will be able to cleanly separate verbs
according to their subcategorization.  However, ``need'' could be
categorized with other verbs that behave in the same way.}
\item Building a word classification tree for each POS tag simplifies
the problem because the hand annotations of the POS tags resolve a lot
of the difficulty that the classification algorithm would otherwise
have to handle.  This means that it is possible to build effective
classification trees even when only a small amount of data is
available, as is the case with the Trains corpus.
\item As a result of the large number of candidate merges that will
not be allowed, building a word classification tree for each POS tag 
significantly speeds up the clustering process.
\end{enumerate}

\begin{figure}
\begin{picture}(0,0)%
\includegraphics{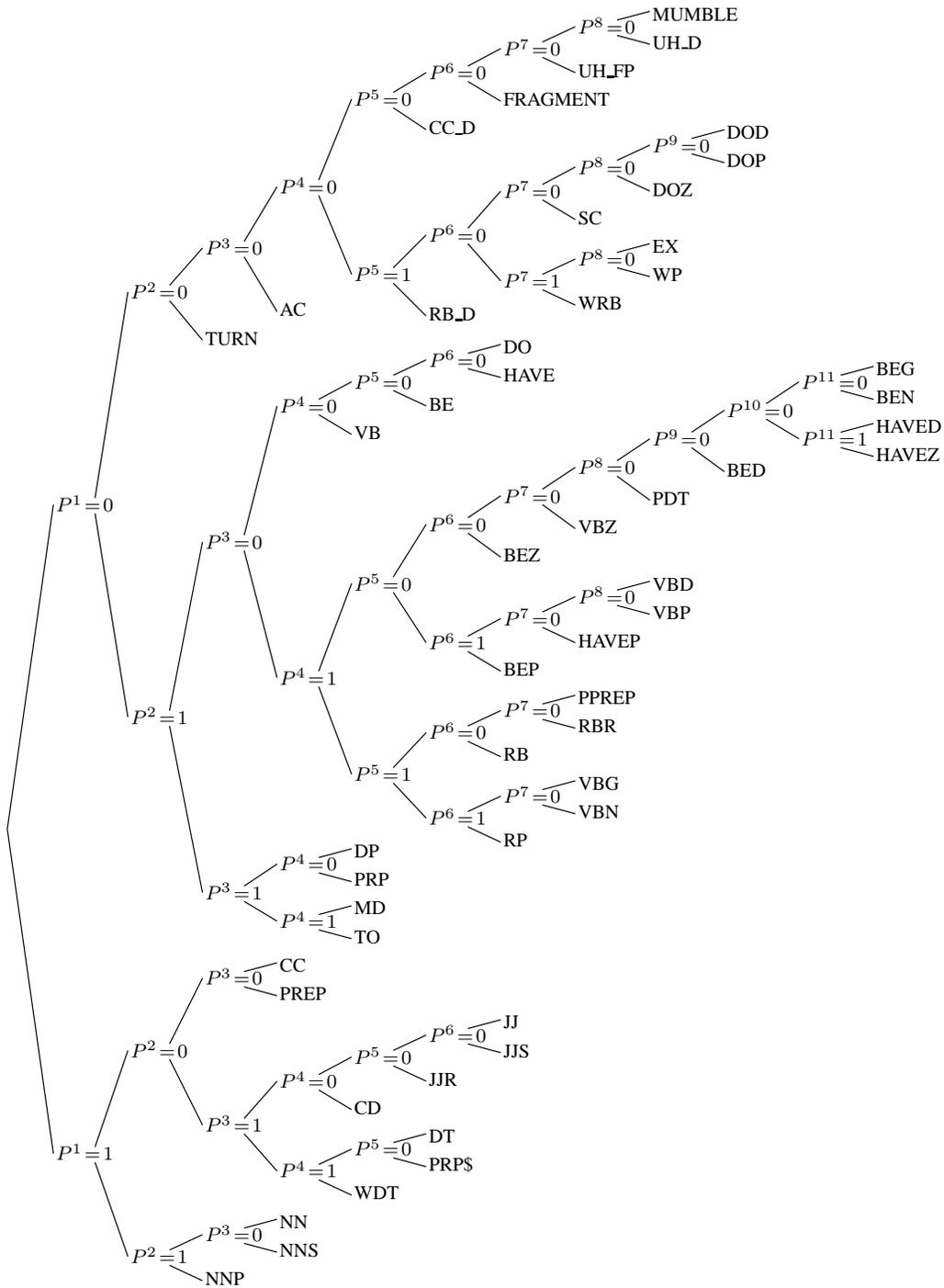}%
\end{picture}%
\setlength{\unitlength}{0.01250000in}%
\begingroup\makeatletter\ifx\SetFigFont\undefined
\def\x#1#2#3#4#5#6#7\relax{\def\x{#1#2#3#4#5#6}}%
\expandafter\x\fmtname xxxxxx\relax \def\y{splain}%
\ifx\x\y   
\gdef\SetFigFont#1#2#3{%
  \ifnum #1<17\tiny\else \ifnum #1<20\small\else
  \ifnum #1<24\normalsize\else \ifnum #1<29\large\else
  \ifnum #1<34\Large\else \ifnum #1<41\LARGE\else
     \huge\fi\fi\fi\fi\fi\fi
  \csname #3\endcsname}%
\else
\gdef\SetFigFont#1#2#3{\begingroup
  \count@#1\relax \ifnum 25<\count@\count@25\fi
  \def\x{\endgroup\@setsize\SetFigFont{#2pt}}%
  \expandafter\x
    \csname \romannumeral\the\count@ pt\expandafter\endcsname
    \csname @\romannumeral\the\count@ pt\endcsname
  \csname #3\endcsname}%
\fi
\fi\endgroup
\begin{picture}(399,593)(23,240)
\put(318,820){\makebox(0,0)[lb]{\smash{\SetFigFont{8}{9.6}{rm} MUMBLE}}}
\put(318,807){\makebox(0,0)[lb]{\smash{\SetFigFont{8}{9.6}{rm} UH\_D}}}
\put(284,814){\makebox(0,0)[lb]{\smash{\SetFigFont{8}{9.6}{rm} $P^{8}\!=\!0$}}}
\put(284,794){\makebox(0,0)[lb]{\smash{\SetFigFont{8}{9.6}{rm} UH\_FP}}}
\put(250,804){\makebox(0,0)[lb]{\smash{\SetFigFont{8}{9.6}{rm} $P^{7}\!=\!0$}}}
\put(250,781){\makebox(0,0)[lb]{\smash{\SetFigFont{8}{9.6}{rm} FRAGMENT}}}
\put(216,793){\makebox(0,0)[lb]{\smash{\SetFigFont{8}{9.6}{rm} $P^{6}\!=\!0$}}}
\put(216,768){\makebox(0,0)[lb]{\smash{\SetFigFont{8}{9.6}{rm} CC\_D}}}
\put(182,781){\makebox(0,0)[lb]{\smash{\SetFigFont{8}{9.6}{rm} $P^{5}\!=\!0$}}}
\put(352,766){\makebox(0,0)[lb]{\smash{\SetFigFont{8}{9.6}{rm} DOD}}}
\put(352,753){\makebox(0,0)[lb]{\smash{\SetFigFont{8}{9.6}{rm} DOP}}}
\put(318,760){\makebox(0,0)[lb]{\smash{\SetFigFont{8}{9.6}{rm} $P^{9}\!=\!0$}}}
\put(318,740){\makebox(0,0)[lb]{\smash{\SetFigFont{8}{9.6}{rm} DOZ}}}
\put(284,750){\makebox(0,0)[lb]{\smash{\SetFigFont{8}{9.6}{rm} $P^{8}\!=\!0$}}}
\put(284,727){\makebox(0,0)[lb]{\smash{\SetFigFont{8}{9.6}{rm} SC}}}
\put(250,739){\makebox(0,0)[lb]{\smash{\SetFigFont{8}{9.6}{rm} $P^{7}\!=\!0$}}}
\put(318,714){\makebox(0,0)[lb]{\smash{\SetFigFont{8}{9.6}{rm} EX}}}
\put(318,701){\makebox(0,0)[lb]{\smash{\SetFigFont{8}{9.6}{rm} WP}}}
\put(284,708){\makebox(0,0)[lb]{\smash{\SetFigFont{8}{9.6}{rm} $P^{8}\!=\!0$}}}
\put(284,688){\makebox(0,0)[lb]{\smash{\SetFigFont{8}{9.6}{rm} WRB}}}
\put(250,698){\makebox(0,0)[lb]{\smash{\SetFigFont{8}{9.6}{rm} $P^{7}\!=\!1$}}}
\put(216,719){\makebox(0,0)[lb]{\smash{\SetFigFont{8}{9.6}{rm} $P^{6}\!=\!0$}}}
\put(216,683){\makebox(0,0)[lb]{\smash{\SetFigFont{8}{9.6}{rm} RB\_D}}}
\put(182,701){\makebox(0,0)[lb]{\smash{\SetFigFont{8}{9.6}{rm} $P^{5}\!=\!1$}}}
\put(148,741){\makebox(0,0)[lb]{\smash{\SetFigFont{8}{9.6}{rm} $P^{4}\!=\!0$}}}
\put(148,685){\makebox(0,0)[lb]{\smash{\SetFigFont{8}{9.6}{rm} AC}}}
\put(114,713){\makebox(0,0)[lb]{\smash{\SetFigFont{8}{9.6}{rm} $P^{3}\!=\!0$}}}
\put(114,672){\makebox(0,0)[lb]{\smash{\SetFigFont{8}{9.6}{rm} TURN}}}
\put( 80,693){\makebox(0,0)[lb]{\smash{\SetFigFont{8}{9.6}{rm} $P^{2}\!=\!0$}}}
\put(250,668){\makebox(0,0)[lb]{\smash{\SetFigFont{8}{9.6}{rm} DO}}}
\put(250,655){\makebox(0,0)[lb]{\smash{\SetFigFont{8}{9.6}{rm} HAVE}}}
\put(216,662){\makebox(0,0)[lb]{\smash{\SetFigFont{8}{9.6}{rm} $P^{6}\!=\!0$}}}
\put(216,642){\makebox(0,0)[lb]{\smash{\SetFigFont{8}{9.6}{rm} BE}}}
\put(182,652){\makebox(0,0)[lb]{\smash{\SetFigFont{8}{9.6}{rm} $P^{5}\!=\!0$}}}
\put(182,629){\makebox(0,0)[lb]{\smash{\SetFigFont{8}{9.6}{rm} VB}}}
\put(148,641){\makebox(0,0)[lb]{\smash{\SetFigFont{8}{9.6}{rm} $P^{4}\!=\!0$}}}
\put(420,658){\makebox(0,0)[lb]{\smash{\SetFigFont{8}{9.6}{rm} BEG}}}
\put(420,645){\makebox(0,0)[lb]{\smash{\SetFigFont{8}{9.6}{rm} BEN}}}
\put(386,652){\makebox(0,0)[lb]{\smash{\SetFigFont{8}{9.6}{rm} $P^{11}\!=\!0$}}}
\put(420,632){\makebox(0,0)[lb]{\smash{\SetFigFont{8}{9.6}{rm} HAVED}}}
\put(420,619){\makebox(0,0)[lb]{\smash{\SetFigFont{8}{9.6}{rm} HAVEZ}}}
\put(386,626){\makebox(0,0)[lb]{\smash{\SetFigFont{8}{9.6}{rm} $P^{11}\!=\!1$}}}
\put(352,639){\makebox(0,0)[lb]{\smash{\SetFigFont{8}{9.6}{rm} $P^{10}\!=\!0$}}}
\put(352,612){\makebox(0,0)[lb]{\smash{\SetFigFont{8}{9.6}{rm} BED}}}
\put(318,626){\makebox(0,0)[lb]{\smash{\SetFigFont{8}{9.6}{rm} $P^{9}\!=\!0$}}}
\put(318,599){\makebox(0,0)[lb]{\smash{\SetFigFont{8}{9.6}{rm} PDT}}}
\put(284,613){\makebox(0,0)[lb]{\smash{\SetFigFont{8}{9.6}{rm} $P^{8}\!=\!0$}}}
\put(284,586){\makebox(0,0)[lb]{\smash{\SetFigFont{8}{9.6}{rm} VBZ}}}
\put(250,600){\makebox(0,0)[lb]{\smash{\SetFigFont{8}{9.6}{rm} $P^{7}\!=\!0$}}}
\put(250,573){\makebox(0,0)[lb]{\smash{\SetFigFont{8}{9.6}{rm} BEZ}}}
\put(216,587){\makebox(0,0)[lb]{\smash{\SetFigFont{8}{9.6}{rm} $P^{6}\!=\!0$}}}
\put(318,560){\makebox(0,0)[lb]{\smash{\SetFigFont{8}{9.6}{rm} VBD}}}
\put(318,547){\makebox(0,0)[lb]{\smash{\SetFigFont{8}{9.6}{rm} VBP}}}
\put(284,554){\makebox(0,0)[lb]{\smash{\SetFigFont{8}{9.6}{rm} $P^{8}\!=\!0$}}}
\put(284,534){\makebox(0,0)[lb]{\smash{\SetFigFont{8}{9.6}{rm} HAVEP}}}
\put(250,544){\makebox(0,0)[lb]{\smash{\SetFigFont{8}{9.6}{rm} $P^{7}\!=\!0$}}}
\put(250,521){\makebox(0,0)[lb]{\smash{\SetFigFont{8}{9.6}{rm} BEP}}}
\put(216,533){\makebox(0,0)[lb]{\smash{\SetFigFont{8}{9.6}{rm} $P^{6}\!=\!1$}}}
\put(182,560){\makebox(0,0)[lb]{\smash{\SetFigFont{8}{9.6}{rm} $P^{5}\!=\!0$}}}
\put(284,508){\makebox(0,0)[lb]{\smash{\SetFigFont{8}{9.6}{rm} PPREP}}}
\put(284,495){\makebox(0,0)[lb]{\smash{\SetFigFont{8}{9.6}{rm} RBR}}}
\put(250,502){\makebox(0,0)[lb]{\smash{\SetFigFont{8}{9.6}{rm} $P^{7}\!=\!0$}}}
\put(250,482){\makebox(0,0)[lb]{\smash{\SetFigFont{8}{9.6}{rm} RB}}}
\put(216,492){\makebox(0,0)[lb]{\smash{\SetFigFont{8}{9.6}{rm} $P^{6}\!=\!0$}}}
\put(284,469){\makebox(0,0)[lb]{\smash{\SetFigFont{8}{9.6}{rm} VBG}}}
\put(284,456){\makebox(0,0)[lb]{\smash{\SetFigFont{8}{9.6}{rm} VBN}}}
\put(250,463){\makebox(0,0)[lb]{\smash{\SetFigFont{8}{9.6}{rm} $P^{7}\!=\!0$}}}
\put(250,443){\makebox(0,0)[lb]{\smash{\SetFigFont{8}{9.6}{rm} RP}}}
\put(216,453){\makebox(0,0)[lb]{\smash{\SetFigFont{8}{9.6}{rm} $P^{6}\!=\!1$}}}
\put(182,473){\makebox(0,0)[lb]{\smash{\SetFigFont{8}{9.6}{rm} $P^{5}\!=\!1$}}}
\put(148,517){\makebox(0,0)[lb]{\smash{\SetFigFont{8}{9.6}{rm} $P^{4}\!=\!1$}}}
\put(114,579){\makebox(0,0)[lb]{\smash{\SetFigFont{8}{9.6}{rm} $P^{3}\!=\!0$}}}
\put(182,438){\makebox(0,0)[lb]{\smash{\SetFigFont{8}{9.6}{rm} DP}}}
\put(182,425){\makebox(0,0)[lb]{\smash{\SetFigFont{8}{9.6}{rm} PRP}}}
\put(148,432){\makebox(0,0)[lb]{\smash{\SetFigFont{8}{9.6}{rm} $P^{4}\!=\!0$}}}
\put(182,412){\makebox(0,0)[lb]{\smash{\SetFigFont{8}{9.6}{rm} MD}}}
\put(182,399){\makebox(0,0)[lb]{\smash{\SetFigFont{8}{9.6}{rm} TO}}}
\put(148,406){\makebox(0,0)[lb]{\smash{\SetFigFont{8}{9.6}{rm} $P^{4}\!=\!1$}}}
\put(114,419){\makebox(0,0)[lb]{\smash{\SetFigFont{8}{9.6}{rm} $P^{3}\!=\!1$}}}
\put( 80,499){\makebox(0,0)[lb]{\smash{\SetFigFont{8}{9.6}{rm} $P^{2}\!=\!1$}}}
\put( 46,596){\makebox(0,0)[lb]{\smash{\SetFigFont{8}{9.6}{rm} $P^{1}\!=\!0$}}}
\put(148,386){\makebox(0,0)[lb]{\smash{\SetFigFont{8}{9.6}{rm} CC}}}
\put(148,373){\makebox(0,0)[lb]{\smash{\SetFigFont{8}{9.6}{rm} PREP}}}
\put(114,380){\makebox(0,0)[lb]{\smash{\SetFigFont{8}{9.6}{rm} $P^{3}\!=\!0$}}}
\put(250,360){\makebox(0,0)[lb]{\smash{\SetFigFont{8}{9.6}{rm} JJ}}}
\put(250,347){\makebox(0,0)[lb]{\smash{\SetFigFont{8}{9.6}{rm} JJS}}}
\put(216,354){\makebox(0,0)[lb]{\smash{\SetFigFont{8}{9.6}{rm} $P^{6}\!=\!0$}}}
\put(216,334){\makebox(0,0)[lb]{\smash{\SetFigFont{8}{9.6}{rm} JJR}}}
\put(182,344){\makebox(0,0)[lb]{\smash{\SetFigFont{8}{9.6}{rm} $P^{5}\!=\!0$}}}
\put(182,321){\makebox(0,0)[lb]{\smash{\SetFigFont{8}{9.6}{rm} CD}}}
\put(148,333){\makebox(0,0)[lb]{\smash{\SetFigFont{8}{9.6}{rm} $P^{4}\!=\!0$}}}
\put(216,308){\makebox(0,0)[lb]{\smash{\SetFigFont{8}{9.6}{rm} DT}}}
\put(216,295){\makebox(0,0)[lb]{\smash{\SetFigFont{8}{9.6}{rm} PRP\$}}}
\put(182,302){\makebox(0,0)[lb]{\smash{\SetFigFont{8}{9.6}{rm} $P^{5}\!=\!0$}}}
\put(182,282){\makebox(0,0)[lb]{\smash{\SetFigFont{8}{9.6}{rm} WDT}}}
\put(148,292){\makebox(0,0)[lb]{\smash{\SetFigFont{8}{9.6}{rm} $P^{4}\!=\!1$}}}
\put(114,313){\makebox(0,0)[lb]{\smash{\SetFigFont{8}{9.6}{rm} $P^{3}\!=\!1$}}}
\put( 80,347){\makebox(0,0)[lb]{\smash{\SetFigFont{8}{9.6}{rm} $P^{2}\!=\!0$}}}
\put(148,269){\makebox(0,0)[lb]{\smash{\SetFigFont{8}{9.6}{rm} NN}}}
\put(148,256){\makebox(0,0)[lb]{\smash{\SetFigFont{8}{9.6}{rm} NNS}}}
\put(114,263){\makebox(0,0)[lb]{\smash{\SetFigFont{8}{9.6}{rm} $P^{3}\!=\!0$}}}
\put(114,243){\makebox(0,0)[lb]{\smash{\SetFigFont{8}{9.6}{rm} NNP}}}
\put( 80,253){\makebox(0,0)[lb]{\smash{\SetFigFont{8}{9.6}{rm} $P^{2}\!=\!1$}}}
\put( 46,300){\makebox(0,0)[lb]{\smash{\SetFigFont{8}{9.6}{rm} $P^{1}\!=\!1$}}}
\end{picture}
\caption{Binary Classification Tree for POS Tags}
\label{fig:model:postree}
\end{figure}

To build the POS tag encoding, we use an approach similar to the
classification tree algorithm of Brown \etal~\shortcite
{Brown-etal92:cl}, which was reviewed in Section~\ref
{sec:related:class}.  We start with a separate class for each POS tag,
and then successively merge classes that result in the smallest
decrease in mutual information between adjacent classes.
Figure~\ref{fig:model:postree} displays the binary classification tree
that we built for a training partition of the Trains corpus.\footnote
{As will be explained in Section~\ref {sec:model:setup}, we collected
our results using a six-fold cross-validation procedure. The
classification tree given in Figure~\ref {fig:model:postree} is the
one that was grown for the first partition of the training data.}
%
%
The binary encoding for a POS tag is computed by starting at the root
and following the path to the POS tag.  At each branching point, a
`$0$' is concatenated to the code if the top branch is taken and a
`$1$' if the bottom branch is taken.  In the figure, this is shown
by the labels on each node.  The label $P^i=j$ indicates that all POS
tags below this node have a $j$ as the $i^{\rm th}$ bit.  (Note that
the root node is not labeled.)  For example, the POS tag {\bf VBG} has
$P^1=0$, $P^2=1$, $P^3=0$, $P^4=1$, $P^5=1$, $P^6=1$, $P^7=0$, and
$P^8=0$, giving a binary encoding of $00111010$.

To build the binary encoding for the word identities, we start with a
separate class for each word and each POS tag that it takes on
(according to the training data).  We then successively merge classes
that result in the smallest decrease in mutual information between
adjacent classes.  However, we only consider two classes for merging
if the POS tags for the words in both classes are the same.  Hence, we
stop clustering when we have a single class for each POS tag.  Words
that occur rarely in the training data for a given POS tag will be
difficult to cluster.  Hence, we group such words together into the
group {\bf low}, which is distinct for each POS tag.\label
{page:model:q:low} For our work in POS-based language modeling, we
have found it best to include all of the {\em singletons}---words that
only occur once for a given POS tag.  This group is not only used for
initially grouping the low-occurring words, but is also used for
estimating the probabilities of {\em unknown} words, words that do not
occur in the training data, as is explained in Section~\ref
{sec:model:unknown}.

\begin{figure}
\begin{center}
\begin{picture}(0,0)%
\includegraphics{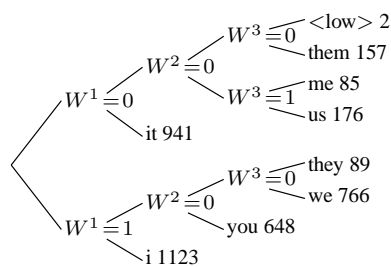}%
\end{picture}%
\setlength{\unitlength}{0.01250000in}%
\begingroup\makeatletter\ifx\SetFigFont\undefined
\def\x#1#2#3#4#5#6#7\relax{\def\x{#1#2#3#4#5#6}}%
\expandafter\x\fmtname xxxxxx\relax \def\y{splain}%
\ifx\x\y   
\gdef\SetFigFont#1#2#3{%
  \ifnum #1<17\tiny\else \ifnum #1<20\small\else
  \ifnum #1<24\normalsize\else \ifnum #1<29\large\else
  \ifnum #1<34\Large\else \ifnum #1<41\LARGE\else
     \huge\fi\fi\fi\fi\fi\fi
  \csname #3\endcsname}%
\else
\gdef\SetFigFont#1#2#3{\begingroup
  \count@#1\relax \ifnum 25<\count@\count@25\fi
  \def\x{\endgroup\@setsize\SetFigFont{#2pt}}%
  \expandafter\x
    \csname \romannumeral\the\count@ pt\expandafter\endcsname
    \csname @\romannumeral\the\count@ pt\endcsname
  \csname #3\endcsname}%
\fi
\fi\endgroup
\begin{picture}(127,115)(23,714)
\put(148,816){\makebox(0,0)[lb]{\smash{\SetFigFont{8}{9.6}{rm} $<$low$>$ 2}}}
\put(148,803){\makebox(0,0)[lb]{\smash{\SetFigFont{8}{9.6}{rm} them 157}}}
\put(114,810){\makebox(0,0)[lb]{\smash{\SetFigFont{8}{9.6}{rm} $W^{3}\!=\!0$}}}
\put(148,790){\makebox(0,0)[lb]{\smash{\SetFigFont{8}{9.6}{rm} me 85}}}
\put(148,777){\makebox(0,0)[lb]{\smash{\SetFigFont{8}{9.6}{rm} us 176}}}
\put(114,784){\makebox(0,0)[lb]{\smash{\SetFigFont{8}{9.6}{rm} $W^{3}\!=\!1$}}}
\put( 80,797){\makebox(0,0)[lb]{\smash{\SetFigFont{8}{9.6}{rm} $W^{2}\!=\!0$}}}
\put( 80,769){\makebox(0,0)[lb]{\smash{\SetFigFont{8}{9.6}{rm} it 941}}}
\put( 46,783){\makebox(0,0)[lb]{\smash{\SetFigFont{8}{9.6}{rm} $W^{1}\!=\!0$}}}
\put(148,756){\makebox(0,0)[lb]{\smash{\SetFigFont{8}{9.6}{rm} they 89}}}
\put(148,743){\makebox(0,0)[lb]{\smash{\SetFigFont{8}{9.6}{rm} we 766}}}
\put(114,750){\makebox(0,0)[lb]{\smash{\SetFigFont{8}{9.6}{rm} $W^{3}\!=\!0$}}}
\put(114,730){\makebox(0,0)[lb]{\smash{\SetFigFont{8}{9.6}{rm} you 648}}}
\put( 80,740){\makebox(0,0)[lb]{\smash{\SetFigFont{8}{9.6}{rm} $W^{2}\!=\!0$}}}
\put( 80,717){\makebox(0,0)[lb]{\smash{\SetFigFont{8}{9.6}{rm} i 1123}}}
\put( 46,729){\makebox(0,0)[lb]{\smash{\SetFigFont{8}{9.6}{rm} $W^{1}\!=\!1$}}}
\end{picture}
\end{center}
\caption{Binary Classification Tree for the Personal Pronouns}
\label{fig:model:prptree}
\end{figure}

Figure~\ref {fig:model:prptree} gives the binary classification tree
for the POS class of the personal pronouns ({\bf PRP}).
For reference, we also list the number of occurrences of each word for
the POS tag. In the figure, we see that the clustering algorithm
distinguished between the subjective pronouns `I', `we', and `they',
and the objective pronouns `me', `us' and `them'. The pronouns `you'
and `it' can take both the subjective and objective cases and the
clustering algorithm probably partitioned them according to their most
common usage in the training corpus, but immediately split `it' from
the other subjective pronouns.  Although we could have added extra POS
tags to distinguish between these two types of pronouns, it seems that
the clustering algorithm can make up for some of the shortcomings of
the POS tagset. As a side note, the words included in the {\bf low}
class are the reflexive pronouns ``themselves'', and ``itself''.\label
{page:model:q:tags} Since these just occurred once each, there was not
enough data to treat them individually, nor enough for the clustering
algorithm to learn that they are different from the subjective
pronouns.

\begin{figure}
\begin{center}
\begin{picture}(0,0)%
\includegraphics{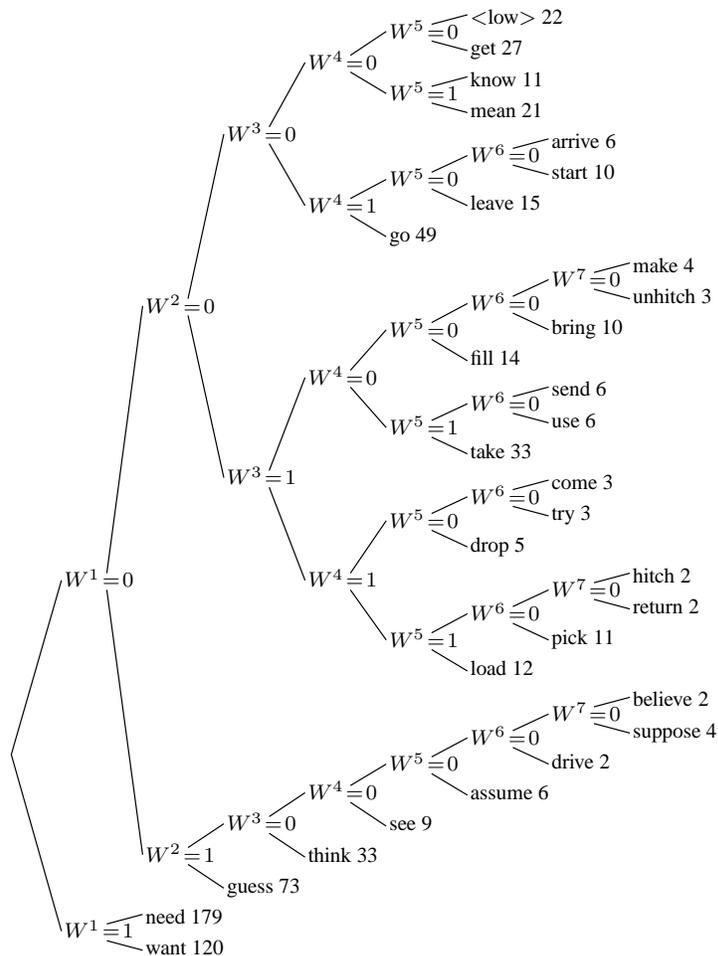}%
\end{picture}%
\setlength{\unitlength}{0.01250000in}%
\begingroup\makeatletter\ifx\SetFigFont\undefined
\def\x#1#2#3#4#5#6#7\relax{\def\x{#1#2#3#4#5#6}}%
\expandafter\x\fmtname xxxxxx\relax \def\y{splain}%
\ifx\x\y   
\gdef\SetFigFont#1#2#3{%
  \ifnum #1<17\tiny\else \ifnum #1<20\small\else
  \ifnum #1<24\normalsize\else \ifnum #1<29\large\else
  \ifnum #1<34\Large\else \ifnum #1<41\LARGE\else
     \huge\fi\fi\fi\fi\fi\fi
  \csname #3\endcsname}%
\else
\gdef\SetFigFont#1#2#3{\begingroup
  \count@#1\relax \ifnum 25<\count@\count@25\fi
  \def\x{\endgroup\@setsize\SetFigFont{#2pt}}%
  \expandafter\x
    \csname \romannumeral\the\count@ pt\expandafter\endcsname
    \csname @\romannumeral\the\count@ pt\endcsname
  \csname #3\endcsname}%
\fi
\fi\endgroup
\begin{picture}(263,406)(23,423)
\put(216,816){\makebox(0,0)[lb]{\smash{\SetFigFont{8}{9.6}{rm} $<$low$>$ 22}}}
\put(216,803){\makebox(0,0)[lb]{\smash{\SetFigFont{8}{9.6}{rm} get 27}}}
\put(182,810){\makebox(0,0)[lb]{\smash{\SetFigFont{8}{9.6}{rm} $W^{5}\!=\!0$}}}
\put(216,790){\makebox(0,0)[lb]{\smash{\SetFigFont{8}{9.6}{rm} know 11}}}
\put(216,777){\makebox(0,0)[lb]{\smash{\SetFigFont{8}{9.6}{rm} mean 21}}}
\put(182,784){\makebox(0,0)[lb]{\smash{\SetFigFont{8}{9.6}{rm} $W^{5}\!=\!1$}}}
\put(148,797){\makebox(0,0)[lb]{\smash{\SetFigFont{8}{9.6}{rm} $W^{4}\!=\!0$}}}
\put(250,764){\makebox(0,0)[lb]{\smash{\SetFigFont{8}{9.6}{rm} arrive 6}}}
\put(250,751){\makebox(0,0)[lb]{\smash{\SetFigFont{8}{9.6}{rm} start 10}}}
\put(216,758){\makebox(0,0)[lb]{\smash{\SetFigFont{8}{9.6}{rm} $W^{6}\!=\!0$}}}
\put(216,738){\makebox(0,0)[lb]{\smash{\SetFigFont{8}{9.6}{rm} leave 15}}}
\put(182,748){\makebox(0,0)[lb]{\smash{\SetFigFont{8}{9.6}{rm} $W^{5}\!=\!0$}}}
\put(182,725){\makebox(0,0)[lb]{\smash{\SetFigFont{8}{9.6}{rm} go 49}}}
\put(148,737){\makebox(0,0)[lb]{\smash{\SetFigFont{8}{9.6}{rm} $W^{4}\!=\!1$}}}
\put(114,767){\makebox(0,0)[lb]{\smash{\SetFigFont{8}{9.6}{rm} $W^{3}\!=\!0$}}}
\put(284,712){\makebox(0,0)[lb]{\smash{\SetFigFont{8}{9.6}{rm} make 4}}}
\put(284,699){\makebox(0,0)[lb]{\smash{\SetFigFont{8}{9.6}{rm} unhitch 3}}}
\put(250,706){\makebox(0,0)[lb]{\smash{\SetFigFont{8}{9.6}{rm} $W^{7}\!=\!0$}}}
\put(250,686){\makebox(0,0)[lb]{\smash{\SetFigFont{8}{9.6}{rm} bring 10}}}
\put(216,696){\makebox(0,0)[lb]{\smash{\SetFigFont{8}{9.6}{rm} $W^{6}\!=\!0$}}}
\put(216,673){\makebox(0,0)[lb]{\smash{\SetFigFont{8}{9.6}{rm} fill 14}}}
\put(182,685){\makebox(0,0)[lb]{\smash{\SetFigFont{8}{9.6}{rm} $W^{5}\!=\!0$}}}
\put(250,660){\makebox(0,0)[lb]{\smash{\SetFigFont{8}{9.6}{rm} send 6}}}
\put(250,647){\makebox(0,0)[lb]{\smash{\SetFigFont{8}{9.6}{rm} use 6}}}
\put(216,654){\makebox(0,0)[lb]{\smash{\SetFigFont{8}{9.6}{rm} $W^{6}\!=\!0$}}}
\put(216,634){\makebox(0,0)[lb]{\smash{\SetFigFont{8}{9.6}{rm} take 33}}}
\put(182,644){\makebox(0,0)[lb]{\smash{\SetFigFont{8}{9.6}{rm} $W^{5}\!=\!1$}}}
\put(148,665){\makebox(0,0)[lb]{\smash{\SetFigFont{8}{9.6}{rm} $W^{4}\!=\!0$}}}
\put(250,621){\makebox(0,0)[lb]{\smash{\SetFigFont{8}{9.6}{rm} come 3}}}
\put(250,608){\makebox(0,0)[lb]{\smash{\SetFigFont{8}{9.6}{rm} try 3}}}
\put(216,615){\makebox(0,0)[lb]{\smash{\SetFigFont{8}{9.6}{rm} $W^{6}\!=\!0$}}}
\put(216,595){\makebox(0,0)[lb]{\smash{\SetFigFont{8}{9.6}{rm} drop 5}}}
\put(182,605){\makebox(0,0)[lb]{\smash{\SetFigFont{8}{9.6}{rm} $W^{5}\!=\!0$}}}
\put(284,582){\makebox(0,0)[lb]{\smash{\SetFigFont{8}{9.6}{rm} hitch 2}}}
\put(284,569){\makebox(0,0)[lb]{\smash{\SetFigFont{8}{9.6}{rm} return 2}}}
\put(250,576){\makebox(0,0)[lb]{\smash{\SetFigFont{8}{9.6}{rm} $W^{7}\!=\!0$}}}
\put(250,556){\makebox(0,0)[lb]{\smash{\SetFigFont{8}{9.6}{rm} pick 11}}}
\put(216,566){\makebox(0,0)[lb]{\smash{\SetFigFont{8}{9.6}{rm} $W^{6}\!=\!0$}}}
\put(216,543){\makebox(0,0)[lb]{\smash{\SetFigFont{8}{9.6}{rm} load 12}}}
\put(182,555){\makebox(0,0)[lb]{\smash{\SetFigFont{8}{9.6}{rm} $W^{5}\!=\!1$}}}
\put(148,580){\makebox(0,0)[lb]{\smash{\SetFigFont{8}{9.6}{rm} $W^{4}\!=\!1$}}}
\put(114,623){\makebox(0,0)[lb]{\smash{\SetFigFont{8}{9.6}{rm} $W^{3}\!=\!1$}}}
\put( 80,695){\makebox(0,0)[lb]{\smash{\SetFigFont{8}{9.6}{rm} $W^{2}\!=\!0$}}}
\put(284,530){\makebox(0,0)[lb]{\smash{\SetFigFont{8}{9.6}{rm} believe 2}}}
\put(284,517){\makebox(0,0)[lb]{\smash{\SetFigFont{8}{9.6}{rm} suppose 4}}}
\put(250,524){\makebox(0,0)[lb]{\smash{\SetFigFont{8}{9.6}{rm} $W^{7}\!=\!0$}}}
\put(250,504){\makebox(0,0)[lb]{\smash{\SetFigFont{8}{9.6}{rm} drive 2}}}
\put(216,514){\makebox(0,0)[lb]{\smash{\SetFigFont{8}{9.6}{rm} $W^{6}\!=\!0$}}}
\put(216,491){\makebox(0,0)[lb]{\smash{\SetFigFont{8}{9.6}{rm} assume 6}}}
\put(182,503){\makebox(0,0)[lb]{\smash{\SetFigFont{8}{9.6}{rm} $W^{5}\!=\!0$}}}
\put(182,478){\makebox(0,0)[lb]{\smash{\SetFigFont{8}{9.6}{rm} see 9}}}
\put(148,491){\makebox(0,0)[lb]{\smash{\SetFigFont{8}{9.6}{rm} $W^{4}\!=\!0$}}}
\put(148,465){\makebox(0,0)[lb]{\smash{\SetFigFont{8}{9.6}{rm} think 33}}}
\put(114,478){\makebox(0,0)[lb]{\smash{\SetFigFont{8}{9.6}{rm} $W^{3}\!=\!0$}}}
\put(114,452){\makebox(0,0)[lb]{\smash{\SetFigFont{8}{9.6}{rm} guess 73}}}
\put( 80,465){\makebox(0,0)[lb]{\smash{\SetFigFont{8}{9.6}{rm} $W^{2}\!=\!1$}}}
\put( 46,580){\makebox(0,0)[lb]{\smash{\SetFigFont{8}{9.6}{rm} $W^{1}\!=\!0$}}}
\put( 80,439){\makebox(0,0)[lb]{\smash{\SetFigFont{8}{9.6}{rm} need 179}}}
\put( 80,426){\makebox(0,0)[lb]{\smash{\SetFigFont{8}{9.6}{rm} want 120}}}
\put( 46,433){\makebox(0,0)[lb]{\smash{\SetFigFont{8}{9.6}{rm} $W^{1}\!=\!1$}}}
\end{picture}
\end{center}
\caption{Binary Classification Tree for the Present Tense Verbs}
\label{fig:model:vbptree}
\end{figure}

In Figure~\ref{fig:model:vbptree}, 
we give the binary classification tree for the present-tense verbs
{\bf VBP} (this class does not include the third-person present-tense
verbs, which are tagged as {\bf VBZ}).  From the figure, we see that
the clustering algorithm made a number of relevant distinctions, such
as grouping together ``want'' and ``need'', and grouping together
``arrive'', ``start'', ``leave'' and ``go'', which deal with sending
trains from one city to another.  One unfortunate mistake was grouping
together ``drive'' with ``believe'', ``suppose'', and ``assume'', but
this is probably a result of ``drive'' having only two occurrences as
a present tense verb in the training corpus.  As with the POS trees,
the binary encoding is derived by following the path from the root of
the tree.

\subsubsection{Composite Questions}

So far in this section, we have focused on the elementary types of
questions that the decision tree can ask. However, there might be a
relevant partitioning of the data that can not be expressed as a
simple question with respect to how the context was encoded as
variables.  For instance, a good partitioning of a node might involve
asking whether questions $q_1$ and $q_2$ are both true.  Using
elementary questions, the decision tree would need to first ask
question $q_1$ and then ask $q_2$ in the true subnode created by
$q_1$.  However, the false case is now split into two separate
nodes: the false subnode created by $q_1$ and the false subnode
created by $q_2$.  This means that there will be less data from which
to further partition the case of $q_1$ or $q_2$ being false since the
data is now split into two nodes, thus causing unnecessary data
fragmentation.

Unnecessary data fragmentation can be avoided by allowing composite
questions.  With composite questions, the nodes of the decision tree
are allowed to go beyond asking elementary questions about the
context, and instead can ask boolean combinations of elementary
questions.\footnote{For numerical data, there are many other types of
combinations that one might want to support.} In fact, allowing
composite questions can alleviate shortcomings in using the
predetermined POS and word partitions, since questions can be
formulated that go beyond the strict partitioning of the
classification trees.

Bahl \etal~\shortcite{Bahl-etal89:tassp} introduced a simple but
effective approach for constructing composite questions.  Rather than
allowing any boolean combination of elementary questions, they
restrict the typology of the combinations to {\em pylons}, which have
the following form ({\em true} maps all data into the true subset).
\begin{eqnarray*}
\mbox{\em pylon} & \Rightarrow & \mbox{\em true} \\
\mbox{\em pylon} & \Rightarrow & (\mbox{\em pylon} \wedge \mbox{\em elementary}) \\
\mbox{\em pylon} & \Rightarrow & (\mbox{\em pylon} \vee \mbox{\em elementary})
\end{eqnarray*}
The effect of any binary question is to divide the data into true and
false subsets.  The advantage of pylons is that each successive
elementary question has the effect of swapping data from the true
subnode into the false or vice versa.  Hence, one can compute the
change in node impurity that results from each successive elementary
question.  This allows one to use a greedy algorithm, which picks the
successive elementary question that results in the largest decrease in
node impurity.  One can also do a beam search and explore the best $n$
alternatives at each level of the pylon.  In Figure~\ref
{code:model:pylon}, we give the algorithm for finding the best pylon
for a node.
\begin{figure}
\noindent
              {\bf function} {\em Find-Best-Pylon}$(${\em Node}$)$ \vspace*{0.5em} \\
              {\em Agenda} $\leftarrow$ \{ pylon consisting of the true question \} \\
              {\em Contenders} $\leftarrow \emptyset$ \\
              {\bf while (} {\em Agenda} $\not = \emptyset$ {\bf )} \\
\hspace*{1em}   {\bf foreach} {\em p $\in$ Agenda}  \\
\hspace*{2em}     {\em p.done} $\leftarrow$ {\bf True} \\   
\hspace*{2em}     {\bf foreach} {\em q $\in$ ElementaryQuestions} \\
\hspace*{3em}        {\em p}$^\prime \leftarrow$ {add question {\em q} to pylon {\em p}} \\
\hspace*{3em}        {\bf if (}
                            {\em impurity(p$^\prime$,Node,GD) $-$
			         impurity(p,Node,GD) $>$ $\varepsilon$} {\bf )}\\
\hspace*{4em}		{\em p.heldoutChange $\leftarrow$
			   impurity(p$^\prime$,Node,HD) $-$
			   impurity(p,Node,HD)} \\
\hspace*{4em}          {\em add(p$^\prime$,Contenders)} \\
\hspace*{4em}          {\em p.done} $\leftarrow$ {\bf False} \\
\hspace*{3em}        {\bf end-if} \\
\hspace*{2em}     {\bf end-foreach} \\
\hspace*{2em}     {\bf if (} {\em p.done $=$ {\bf False}} {\bf )} \\
\hspace*{3em}        {\em add(p,Contenders)} \\
\hspace*{1em}   {\bf end-foreach} \\
\hspace*{1em}   {\em Agenda} $\leftarrow \emptyset$ \\
\hspace*{1em}   {\bf foreach} {\em p $\in$ Contender} \\
\hspace*{2em}      {\bf if (} {\em p.done} $=$ {\bf False} \ {\bf and} \  
			        {\em p.heldoutChange} $\ge 0$ {\bf )} \\
\hspace*{3em}         move {\em p} from {\em Contenders} to {\em Agenda} \\
              {\bf end-while} \\
\newlength{\argmaxx}
\settowidth{\argmaxx}{\small $q \in$ {\em Contenders}}
              {\em p} $\leftarrow$ 
		\parbox[t]{\argmaxx}{\makebox[\argmaxx][c]{\bf arg\,max}
				     \small \vspace*{-2.7em} \\
                                     $q \in ${\em Contenders}}\
		{\em impurity(q,Node,GD)} \vspace*{0.8em} \\
              {\bf if (} {\em p.change} $< 0$ {\bf )} \\
\hspace*{1em}   remove last question from {\em p} \\
              {\bf return} {\em p} 
\caption{Algorithm for Finding the Best Pylon}
\label{code:model:pylon}
\end{figure}%
In the algorithm, {\em GD} refers to the growing data, {\em HD} the
heldout data, and $\varepsilon$ is a constant used to guard against
round-off errors (we use $\varepsilon$ $=1^{-7}$).  The function {\em
impurity(p,Node,Data)} returns the decrease in tree impurity that
would result from splitting {\em Node} with pylon {\em p} with respect
to either the growing data or heldout data as indicated by {\em Data}.

Given that we have heldout data, it would be worthwhile to make use of
this in finding the best pylon. We must be careful not to make too
much use of the heldout data for otherwise it will become as biased as
the training data. For instance, we have found that adding the
restriction of only adding elementary questions that lead to an
improvement with respect to the heldout data (in addition to the
training data) does not lead to a good improvement in finding the best
pylon. Instead, we use the heldout data only to decide when to stop
growing a particular pylon. If the last question added to a candidate
pylon results in an increase in node impurity with respect to the
heldout data, we stop growing that alternative. When there are no
further candidates that can be grown, we choose the winning pylon as
the one with the best decrease in node impurity with respect to the
training data. If the last elementary question of the chosen pylon
results in an increase in node impurity with respect to the heldout
data, then we remove this question from the pylon. In Section~\ref
{sec:model:r:pylons}, we report on the significance that using pylons
has in estimating the probability distributions.

\subsection{Growing the Decision Trees}
\label{sec:model:grow}

In this section, we discuss how we grow the decision trees given the
set of questions that the decision tree can ask.  We first discuss
how we take advantage of the POS tags to simplify the task
of estimating the word probability distribution.  We then give
the actual algorithm for finding the decision tree, and we conclude
this section with an example of the decision tree that was found
for the POS probability distribution.

\subsubsection{The Events for the Word Decision Tree}

For the POS probability distribution, there are approximately 60 POS
tags, which form the events that the decision tree algorithm is trying
to predict given the context.  For the word probability distribution,
there are approximately 820 distinct words in the training data,
meaning that there are a lot more events that the decision tree
algorithm needs to predict.  Furthermore, unlike the POS tagset, the
word vocabulary size is bound to grow larger for more complex
application domains and as the size of the training corpus increases.
In this section, we show that the number of events that the decision
tree algorithm must consider can be significantly decreased.

The approach of viewing a word as a further specification of the POS
tag can also be used to advantage in estimating the word probability
distribution. Rather than using the decision tree algorithm to
completely determine the equivalence classes, one can start the
decision tree with an equivalence class for each POS tag that the
words take on according to the training data (or equivalently build a
word decision tree for each POS tag). This means that the decision
tree starts with the probability of $\Pr(W_i|P_i)$ as estimated by
relative frequency. This is the same value with which non-decision
tree approaches start (and end). This approach has the following
advantages.
\begin{enumerate}
\item Starting with a single root node containing all of the training data is 
adversely affected by the smoothing algorithm. Even if the decision
tree first subdivides training data by the POS tags, the probabilities
will be diluted by the smoothing algorithm. For instance, consider the
superlative adjectives ({\bf JJS}) and relative adjectives ({\bf
JJR}), and assume that the clustering algorithm has grouped them
together.  Whether a word is a superlative or relative adjective is
completely determined by the word identity.  Now assume that the
decision tree algorithm splits the context on the basis of whether the
current word is a superlative versus relative adjective.  The
smoothing algorithm, however, will undoubtedly smooth the word
probabilities for these two contexts with the word probabilities of
their common parent node, even though there is no need to do
this. This will make the word probabilities less accurate.

\item Starting with an equivalence class for each POS tag means 
that most word probabilities for a context will be zero. In fact, for
the Trains corpus, with approximately 820 different words for the
training data, each POS tag takes on at most 140 different
words. Rather than training the decision tree with the events being
the word identities, we let the events be word indices relative to the
POS tag. For the Trains corpus, this means that there will only be 140
different events that the decision tree needs to predict rather than
820, which significantly speeds up the algorithm and requires about
one-sixth as much space.
\end{enumerate}

Above, we explained how we can take advantage of the POS tags to both
improve the probability estimates of the words, and improve the
efficiency of the algorithm since we now have fewer events that the
decision tree needs to consider.  In fact, we can further decrease the
number of events.  A significant number of words in the training
corpus have a small number of occurrences.  Such words will prove
problematic for the decision tree algorithm to predict.  Just as we
initially clustered all of the low-occurring words before invoking the
clustering algorithm to build the word classification tree (see
page~\pageref {page:model:g:low}), we can cluster such words into a
single event for the decision tree to predict.\label
{page:model:g:low} This not only leads to better probability
estimates, but also leads to a reduction in the number of events that
the decision tree must estimate.  Grouping together all singleton
words leads to a reduction in the number of word events from 140 to
approximately 90.  With this ability to dramatically reduce the number
of word events we need to deal with, we are in a good position to deal
with corpora with larger vocabulary sizes.

\subsubsection{The Algorithm}

For growing the decision trees, we follow the approach of growing the
tree as far as possible.  For each leaf node of the tree, as long as
there is some minimum number of data points ({\bf MinSize}) in the
leaf with respect to the growing data (we set {\bf MinSize}$=10$), we
look for the best composite question.  If no question was found, or
the question that was found results in a change in tree impurity of
less than $\varepsilon$ with respect to either the growing data or the
heldout data, then we do not expand the leaf \cite
{Bahl-etal89:tassp}. The algorithm for finding the decision tree is
given in Figure~\ref{code:model:tree}, which is invoked by passing it
the root node.
\begin{figure}[bht]
\noindent
              {\bf function} {\em Find-Best-Tree(Leaf)} \vspace*{0.5em} \\
              {\bf if (} {\em Leaf} has less than {\bf MinSize} data in {\em GD} {\bf )} \\
\hspace*{1em}   {\bf return} \\
              {\em p} $\leftarrow$ {\em Find-Best-Pylon(Leaf)} \\
              {\bf if (}\ {\em impurity(p,Leaf,GD)} $<$ $\varepsilon$ 
			 \ {\bf or} \
			 {\em impurity(p,Leaf,HD)} $<$ $\varepsilon$ {\bf )} \\
\hspace*{1em}      {\bf return} \\
              split {\em Leaf} with {\em q} creating {\em TrueLeaf} and {\em FalseLeaf} \\
              {\em Find-Best-Tree(TrueLeaf)} \\
              {\em Find-Best-Tree(FalseLeaf)} \\
	      {\bf return}
\caption{Algorithm for Finding the Best Decision Tree}
\label{code:model:tree}
\end{figure}

\subsubsection{A Sample Decision Tree}

\begin{figure}
\begin{picture}(0,0)%
\includegraphics{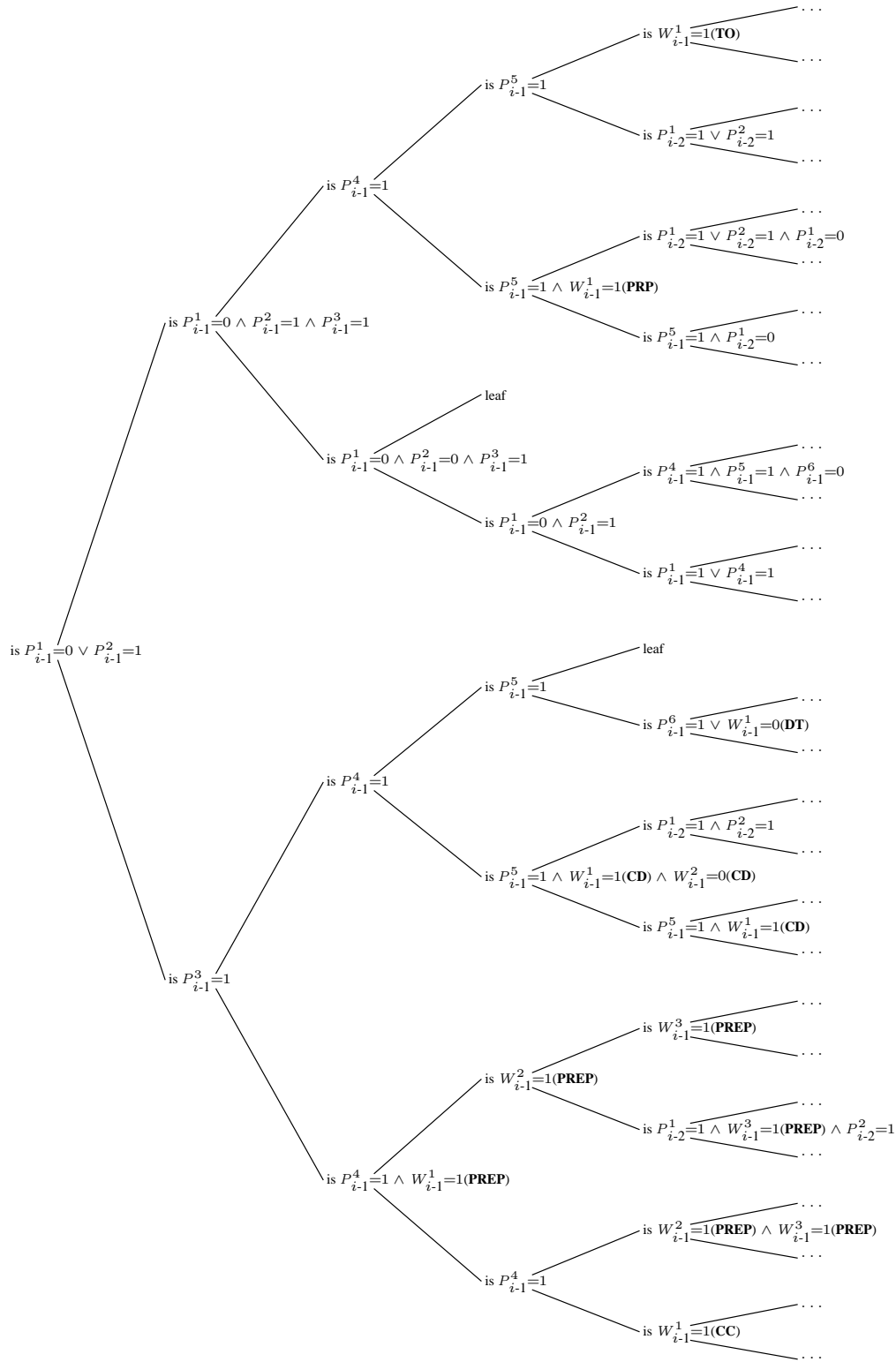}%
\end{picture}%
\setlength{\unitlength}{0.01250000in}%
\begingroup\makeatletter\ifx\SetFigFont\undefined
\def\x#1#2#3#4#5#6#7\relax{\def\x{#1#2#3#4#5#6}}%
\expandafter\x\fmtname xxxxxx\relax \def\y{splain}%
\ifx\x\y   
\gdef\SetFigFont#1#2#3{%
  \ifnum #1<17\tiny\else \ifnum #1<20\small\else
  \ifnum #1<24\normalsize\else \ifnum #1<29\large\else
  \ifnum #1<34\Large\else \ifnum #1<41\LARGE\else
     \huge\fi\fi\fi\fi\fi\fi
  \csname #3\endcsname}%
\else
\gdef\SetFigFont#1#2#3{\begingroup
  \count@#1\relax \ifnum 25<\count@\count@25\fi
  \def\x{\endgroup\@setsize\SetFigFont{#2pt}}%
  \expandafter\x
    \csname \romannumeral\the\count@ pt\expandafter\endcsname
    \csname @\romannumeral\the\count@ pt\endcsname
  \csname #3\endcsname}%
\fi
\fi\endgroup
\begin{picture}(377,656)(21,163)
\put(396,806){\makebox(0,0)[lb]{\smash{\SetFigFont{6}{7.2}{rm} $\cdots$}}}
\put(396,782){\makebox(0,0)[lb]{\smash{\SetFigFont{6}{7.2}{rm} $\cdots$}}}
\put(321,794){\makebox(0,0)[lb]{\smash{\SetFigFont{6}{7.2}{rm} is ${\em W}_{\mbox{$i$-1}}^1\!\!=\!\!1 \mbox{({\bf TO})}$}}}
\put(396,758){\makebox(0,0)[lb]{\smash{\SetFigFont{6}{7.2}{rm} $\cdots$}}}
\put(396,734){\makebox(0,0)[lb]{\smash{\SetFigFont{6}{7.2}{rm} $\cdots$}}}
\put(321,746){\makebox(0,0)[lb]{\smash{\SetFigFont{6}{7.2}{rm} is ${\em P}_{\mbox{$i$-2}}^1\!\!=\!\!1 \vee {\em P}_{\mbox{$i$-2}}^2\!\!=\!\!1$}}}
\put(246,770){\makebox(0,0)[lb]{\smash{\SetFigFont{6}{7.2}{rm} is ${\em P}_{\mbox{$i$-1}}^5\!\!=\!\!1$}}}
\put(396,710){\makebox(0,0)[lb]{\smash{\SetFigFont{6}{7.2}{rm} $\cdots$}}}
\put(396,686){\makebox(0,0)[lb]{\smash{\SetFigFont{6}{7.2}{rm} $\cdots$}}}
\put(321,698){\makebox(0,0)[lb]{\smash{\SetFigFont{6}{7.2}{rm} is ${\em P}_{\mbox{$i$-2}}^1\!\!=\!\!1 \vee {\em P}_{\mbox{$i$-2}}^2\!\!=\!\!1 \wedge {\em P}_{\mbox{$i$-2}}^1\!\!=\!\!0$}}}
\put(396,662){\makebox(0,0)[lb]{\smash{\SetFigFont{6}{7.2}{rm} $\cdots$}}}
\put(396,638){\makebox(0,0)[lb]{\smash{\SetFigFont{6}{7.2}{rm} $\cdots$}}}
\put(321,650){\makebox(0,0)[lb]{\smash{\SetFigFont{6}{7.2}{rm} is ${\em P}_{\mbox{$i$-1}}^5\!\!=\!\!1 \wedge {\em P}_{\mbox{$i$-2}}^1\!\!=\!\!0$}}}
\put(246,674){\makebox(0,0)[lb]{\smash{\SetFigFont{6}{7.2}{rm} is ${\em P}_{\mbox{$i$-1}}^5\!\!=\!\!1 \wedge {\em W}_{\mbox{$i$-1}}^1\!\!=\!\!1 \mbox{({\bf PRP})}$}}}
\put(171,722){\makebox(0,0)[lb]{\smash{\SetFigFont{6}{7.2}{rm} is ${\em P}_{\mbox{$i$-1}}^4\!\!=\!\!1$}}}
\put(246,622){\makebox(0,0)[lb]{\smash{\SetFigFont{6}{7.2}{rm} leaf}}}
\put(396,598){\makebox(0,0)[lb]{\smash{\SetFigFont{6}{7.2}{rm} $\cdots$}}}
\put(396,574){\makebox(0,0)[lb]{\smash{\SetFigFont{6}{7.2}{rm} $\cdots$}}}
\put(321,586){\makebox(0,0)[lb]{\smash{\SetFigFont{6}{7.2}{rm} is ${\em P}_{\mbox{$i$-1}}^4\!\!=\!\!1 \wedge {\em P}_{\mbox{$i$-1}}^5\!\!=\!\!1 \wedge {\em P}_{\mbox{$i$-1}}^6\!\!=\!\!0$}}}
\put(396,550){\makebox(0,0)[lb]{\smash{\SetFigFont{6}{7.2}{rm} $\cdots$}}}
\put(396,526){\makebox(0,0)[lb]{\smash{\SetFigFont{6}{7.2}{rm} $\cdots$}}}
\put(321,538){\makebox(0,0)[lb]{\smash{\SetFigFont{6}{7.2}{rm} is ${\em P}_{\mbox{$i$-1}}^1\!\!=\!\!1 \vee {\em P}_{\mbox{$i$-1}}^4\!\!=\!\!1$}}}
\put(246,562){\makebox(0,0)[lb]{\smash{\SetFigFont{6}{7.2}{rm} is ${\em P}_{\mbox{$i$-1}}^1\!\!=\!\!0 \wedge {\em P}_{\mbox{$i$-1}}^2\!\!=\!\!1$}}}
\put(171,592){\makebox(0,0)[lb]{\smash{\SetFigFont{6}{7.2}{rm} is ${\em P}_{\mbox{$i$-1}}^1\!\!=\!\!0 \wedge {\em P}_{\mbox{$i$-1}}^2\!\!=\!\!0 \wedge {\em P}_{\mbox{$i$-1}}^3\!\!=\!\!1$}}}
\put( 96,657){\makebox(0,0)[lb]{\smash{\SetFigFont{6}{7.2}{rm} is ${\em P}_{\mbox{$i$-1}}^1\!\!=\!\!0 \wedge {\em P}_{\mbox{$i$-1}}^2\!\!=\!\!1 \wedge {\em P}_{\mbox{$i$-1}}^3\!\!=\!\!1$}}}
\put(321,502){\makebox(0,0)[lb]{\smash{\SetFigFont{6}{7.2}{rm} leaf}}}
\put(396,478){\makebox(0,0)[lb]{\smash{\SetFigFont{6}{7.2}{rm} $\cdots$}}}
\put(396,454){\makebox(0,0)[lb]{\smash{\SetFigFont{6}{7.2}{rm} $\cdots$}}}
\put(321,466){\makebox(0,0)[lb]{\smash{\SetFigFont{6}{7.2}{rm} is ${\em P}_{\mbox{$i$-1}}^6\!\!=\!\!1 \vee {\em W}_{\mbox{$i$-1}}^1\!\!=\!\!0 \mbox{({\bf DT})}$}}}
\put(246,484){\makebox(0,0)[lb]{\smash{\SetFigFont{6}{7.2}{rm} is ${\em P}_{\mbox{$i$-1}}^5\!\!=\!\!1$}}}
\put(396,430){\makebox(0,0)[lb]{\smash{\SetFigFont{6}{7.2}{rm} $\cdots$}}}
\put(396,406){\makebox(0,0)[lb]{\smash{\SetFigFont{6}{7.2}{rm} $\cdots$}}}
\put(321,418){\makebox(0,0)[lb]{\smash{\SetFigFont{6}{7.2}{rm} is ${\em P}_{\mbox{$i$-2}}^1\!\!=\!\!1 \wedge {\em P}_{\mbox{$i$-2}}^2\!\!=\!\!1$}}}
\put(396,382){\makebox(0,0)[lb]{\smash{\SetFigFont{6}{7.2}{rm} $\cdots$}}}
\put(396,358){\makebox(0,0)[lb]{\smash{\SetFigFont{6}{7.2}{rm} $\cdots$}}}
\put(321,370){\makebox(0,0)[lb]{\smash{\SetFigFont{6}{7.2}{rm} is ${\em P}_{\mbox{$i$-1}}^5\!\!=\!\!1 \wedge {\em W}_{\mbox{$i$-1}}^1\!\!=\!\!1 \mbox{({\bf CD})}$}}}
\put(246,394){\makebox(0,0)[lb]{\smash{\SetFigFont{6}{7.2}{rm} is ${\em P}_{\mbox{$i$-1}}^5\!\!=\!\!1 \wedge {\em W}_{\mbox{$i$-1}}^1\!\!=\!\!1 \mbox{({\bf CD})} \wedge {\em W}_{\mbox{$i$-1}}^2\!\!=\!\!0 \mbox{({\bf CD})}$}}}
\put(171,439){\makebox(0,0)[lb]{\smash{\SetFigFont{6}{7.2}{rm} is ${\em P}_{\mbox{$i$-1}}^4\!\!=\!\!1$}}}
\put(396,334){\makebox(0,0)[lb]{\smash{\SetFigFont{6}{7.2}{rm} $\cdots$}}}
\put(396,310){\makebox(0,0)[lb]{\smash{\SetFigFont{6}{7.2}{rm} $\cdots$}}}
\put(321,322){\makebox(0,0)[lb]{\smash{\SetFigFont{6}{7.2}{rm} is ${\em W}_{\mbox{$i$-1}}^3\!\!=\!\!1 \mbox{({\bf PREP})}$}}}
\put(396,286){\makebox(0,0)[lb]{\smash{\SetFigFont{6}{7.2}{rm} $\cdots$}}}
\put(396,262){\makebox(0,0)[lb]{\smash{\SetFigFont{6}{7.2}{rm} $\cdots$}}}
\put(321,274){\makebox(0,0)[lb]{\smash{\SetFigFont{6}{7.2}{rm} is ${\em P}_{\mbox{$i$-2}}^1\!\!=\!\!1 \wedge {\em W}_{\mbox{$i$-1}}^3\!\!=\!\!1 \mbox{({\bf PREP})} \wedge {\em P}_{\mbox{$i$-2}}^2\!\!=\!\!1$}}}
\put(246,298){\makebox(0,0)[lb]{\smash{\SetFigFont{6}{7.2}{rm} is ${\em W}_{\mbox{$i$-1}}^2\!\!=\!\!1 \mbox{({\bf PREP})}$}}}
\put(396,238){\makebox(0,0)[lb]{\smash{\SetFigFont{6}{7.2}{rm} $\cdots$}}}
\put(396,214){\makebox(0,0)[lb]{\smash{\SetFigFont{6}{7.2}{rm} $\cdots$}}}
\put(321,226){\makebox(0,0)[lb]{\smash{\SetFigFont{6}{7.2}{rm} is ${\em W}_{\mbox{$i$-1}}^2\!\!=\!\!1 \mbox{({\bf PREP})} \wedge {\em W}_{\mbox{$i$-1}}^3\!\!=\!\!1 \mbox{({\bf PREP})}$}}}
\put(396,190){\makebox(0,0)[lb]{\smash{\SetFigFont{6}{7.2}{rm} $\cdots$}}}
\put(396,166){\makebox(0,0)[lb]{\smash{\SetFigFont{6}{7.2}{rm} $\cdots$}}}
\put(321,178){\makebox(0,0)[lb]{\smash{\SetFigFont{6}{7.2}{rm} is ${\em W}_{\mbox{$i$-1}}^1\!\!=\!\!1 \mbox{({\bf CC})}$}}}
\put(246,202){\makebox(0,0)[lb]{\smash{\SetFigFont{6}{7.2}{rm} is ${\em P}_{\mbox{$i$-1}}^4\!\!=\!\!1$}}}
\put(171,250){\makebox(0,0)[lb]{\smash{\SetFigFont{6}{7.2}{rm} is ${\em P}_{\mbox{$i$-1}}^4\!\!=\!\!1 \wedge {\em W}_{\mbox{$i$-1}}^1\!\!=\!\!1 \mbox{({\bf PREP})}$}}}
\put( 96,345){\makebox(0,0)[lb]{\smash{\SetFigFont{6}{7.2}{rm} is ${\em P}_{\mbox{$i$-1}}^3\!\!=\!\!1$}}}
\put( 21,501){\makebox(0,0)[lb]{\smash{\SetFigFont{6}{7.2}{rm} is ${\em P}_{\mbox{$i$-1}}^1\!\!=\!\!0 \vee {\em P}_{\mbox{$i$-1}}^2\!\!=\!\!1$}}}
\end{picture}
\caption{Decision Tree for POS Tags}
\label{fig:model:Ptree}
\end{figure}

In Figure~\ref {fig:model:Ptree}, we illustrate part of the tree
that was grown for estimating the POS tag of the current
word.
At each node, we indicate the question that was asked.  Questions are
about the binary encoding of the POS tags and the word identities of
the words in the context of the probability distribution.  Questions
have the form of ``is $X_{i-j}^b\!=\!1$'', where $X$ is either a POS
tag $P$ or a word identify $W$, $j$ indicates the word before the
current word that is being queried, and $b$ indicates which bit in the
encoding is being queried.  If an elementary question is about a word
identity, we show the corresponding POS in brackets following the
question, e.g.~``is $W_{\im}^1\!=\!0 (\mbox{\bf DT})$''.

We now discuss the decision tree that was grown.  The question on the
root node ``is $P_{\im}^1\!=\!0 \vee P_{\im}^2\!=\!1$'' is asking
whether the POS tag of the previous word has a ``0'' as the first bit
or a ``1'' as the second bit of its binary encoding.  If the answer is
``yes'', the top branch is followed, otherwise the bottom branch is
followed.  Referring to the POS classification tree given in
Figure~\ref{fig:model:postree}, we see that the partition created by
the bottom branch is as follows.
\[ P_{\im} \in \{ \mbox{\bf CC}, \mbox{\bf PREP}, \mbox{\bf JJ}, 
\mbox{\bf JJS}, \mbox{\bf JJR}, \mbox{\bf CD}, \mbox{\bf DT}, 
\mbox{\bf PRP\$}, \mbox{\bf WDT} \} \]
Following the bottom branch of the decision tree, we see that the next
question is ``is $P_{\im}^3\!=\!1$''.  The true partition of this
question is as follows.
\[ P_{\im} \in \{ \mbox{\bf JJ}, \mbox{\bf JJS},
\mbox{\bf JJR}, \mbox{\bf CD}, \mbox{\bf DT}, \mbox{\bf PRP\$}, 
\mbox{\bf WDT} \} \]
Following the top branch, we see that the next question is ``is
$P_{\im}^4\!=\!1$'', whose true partition is $P_{\im} \in \{
\mbox{\bf DT}, \mbox{\bf PRP\$}, \mbox{\bf WDT}\}$.  The next question
along the top branch is ``is $P_{\im}^5\!=\!1$''.  The true partition
is $P_{\im}\!=\!\mbox{\bf WDT}$.  As indicated in the figure, this is
a leaf node, and so no suitable question was found to ask of this
context, nor of the prior POS tags, nor of the word identities, which
for this POS tag consists of ``which'', ``what'', and ``whatever''.
Here we see that the syntactic category captures the relevant features
as far as predicting the next POS tag, at least as far as the training
data is able to supply.

The false partition of the question ``is $P_{\im}^5\!=\!1$'' is
$P_{\im} \in \{ \mbox{\bf DT},
\mbox{\bf PRP\$}\}$.  The question that is then asked of this context is 
the following.
\[ \mbox{``is } P_{\im}^6\!=\!1 \vee W_{\im}^1\!=\!0\mbox{''} \]
The first part splits the context so that the true partition
corresponds to $P_{\im}\!=\!\mbox{\bf PRP\$}$ and the false is
$P_{\im}\!=\!\mbox{\bf DT}$.  The second part of the pylon moves the
determiners ({\bf DT}) that have a zero as the first bit of their
binary encoding to the true context leaving just the determiners ``a''
and ``an'' in the false partition.  The reason why the decision tree
probabably did not split solely on the distinction between possessive
pronouns and determiners is that the determiners occur 27 times more
frequently in the Trains corpus than do the possessive pronouns.
Hence, the distinction between the possessive pronouns and determiners
might not lead to as great of a decrease in node impurity than a major
distinction of the determiners.  The second explanation is that the
POS tags might not be capturing the optimal features for predicting
the next word.  Composite questions, in this case, allow the decision
tree to go beyond the strict partitioning imposed by the POS tags.

\subsection{Computing the Probabilities}

After a decision tree is grown, probabilities for each event in each
leaf node of the decision tree can be computed based on relative
frequencies.  In this section, we first briefly mention how we use
interpolated estimation to smooth these probability distributions.  We
then discuss a specific case where the probability estimates are not
reliable, and how we overcome this deficiency.  We end the section
with a discussion on how we compute probability estimates for low
occurring and unknown words.

\subsubsection{Smoothing the Probilities}

As with some previous decision tree approaches
(e.g.~\cite{Black-etal92:darpa:hbg,Black-etal92:darpa:pos,Magerman94:thesis}),
we divide the training data into two parts, one used for growing the
tree, and a second, the heldout data.  We have already discussed the
use of the heldout data to decide when to stop growing a pylon and
when to stop growing a leaf node.  We also use the heldout data to
smooth the probability distribution of each node with that of its
parent, using interpolated estimation, which was described in
Section~\ref {sec:related:sparseness}.  We follow the approach
described in Magerman \shortcite {Magerman94:thesis} for using
interpolated estimation with decision trees.  Here, each node of the
decision tree has its own lambda, unless the number of events in the
node is less than 100, in which case the node is grouped with other
nodes to achieve the minimum number.

\subsubsection{Unreliable Estimates}

When growing a decision tree, the question that leads to the biggest
decrease in impurity is used to split a node.  However, especially
when there might be a large number of events, as with estimating word
or POS probabilities, some events might not have enough data to give
reliable probability estimates for the child nodes.  The situation
might arise where the data is split such that one of the child nodes
has occurrences of an event in either its growing data or its heldout
data, but not in both.  We take this as a sign that there is not
enough data for this event to allow the counts in the subnode to
reliably estimate the probability of the event, nor be reliably used
in smoothing the node's probability of the event with that of its
parent.  If this occurs, then we propagate the parent's probability
for the event to all of the descendant nodes.  We then normalize all
of the probabilities to ensure that each node adds up to one.  More
work is needed to further expore using decision trees for the
estimation of large number of events.\footnote{See for example
Section~\ref {sec:model:c:dt:class}, which shows that predicting a
word by first predicting a class does better than directly predicting
the word.}
%

\subsubsection{Low-Occurring Words}

On page~\pageref {page:model:g:low}, we discussed how we cluster all
of the low-occurring words (namely the singletons) into a single
event, called {\bf low}, that the decision tree algorithm tries to
predict.  Hence, for a given context, we have the probability of the
low-ocurring event.  We estimate the probability of a given
low-occurring word $W_i$ as the probility of the low-occurring event
given the context, as computed by the decision tree probabilities,
multiplied by the ratio of the number of occurrences of the low
occurring word over the number of low-occurring words in the training
corpus for the POS tag.
\begin{eqnarray}
{\Pr}_L(W_i|\mbox{\em Context}) &=& 
  \frac{\mbox{c($W_iP_i$)}}
       {\mbox{c({\bf low }$P_i$)}}
  {\Pr}_T(\mbox{\bf low}|\mbox{\em Context})
\end{eqnarray}

\subsubsection{Unknown Words}
\label{sec:model:unknown}

Words that are in the test corpus but not in the training corpus are
referred to as {\em unknown} words.  Unknown words pose a major
difficulty for both POS tagging and speech recognition.  Traditional
POS taggers, which find the best POS tags for a given sequence of
words, can employ morphological or orthographic information to
determine the most likely POS tag for the unknown word (e.g.~\cite
{Weischedel-etal93:cl,Brill95:cl,Mikheev96:acl}). For instance, if a
word ends in ``-ing'', then it is very likely to be a present tense
verb ({\bf VBG}), and if the word begins with a capital, then it is
likely to be a proper noun ({\bf NNP}). They can also look up the
word in a dictionary, which could list all possible POS tags that the
word can take on.  They can also take into account the distribution of
rarely seen words in the training corpus to estimate the likely POS
tag for an unknown word; for instance one can use words that occur
only once in the training corpus to determine the probability of the
unknown word given the POS tag \cite{DermatasKokkinakis95:cl}.

For speech recognition, modeling unknown words is much more difficult.
Unless one expands the dictionary used by the acoustic model, the
acoustic model will not be able to find the word.  Enlarging the
acoustic dictionary, however, will increase the difficulty of the
problem that the acoustic model is faced with, since there are now
more candidate words from which to choose.  Because of this problem,
we do not fully exploit techniques that have proven effective for POS
tagging of written text, since they are not adequate for speech
recognition.  We do, however, go a step beyond the strategy of
Dermatas and Kokkinakis \shortcite {DermatasKokkinakis95:cl} and
estimate the probability of an unknown word based not on its POS tag,
but on the probability estimate of low-occurring words given the
context, where the context includes the POS tag of the word, as well
as the preceding words and their POS tags.\footnote{In the case where
the low-occurring words are not the same as the singletons, we use the
ratio $\frac{\mbox{c({\bf singletons }$P_i$)}} {\mbox{c({\bf low
}$P_i$)}}$ to adjust the probability so that it uses the occurrence
rate of the singletons to predict the unknown words.}
\begin{eqnarray}
{\Pr}_U(\mbox{\bf unknown}|\mbox{\em Context}) 
&=& \frac{\mbox{c({\bf singletons }$P_i$)}}
         {\mbox{c({\bf low }$P_i$)}}
    {\Pr}_T(\mbox{\bf low}|\mbox{\em Context})
\end{eqnarray}
To ensure that all of the probabilities add up to one, we normalize
the probabilities to take into account the probability mass we assign
to the unknowns.

If an unknown word has been predicted, it will consequently be used
as part of the context for later decisions.  As explained on 
page~\pageref {page:model:q:low}, unknown words are treated as
being in the same group as the low-occurring words in the word
classification tree.

\section{Perplexity}
\label{sec:model:perplexity}

In Section~\ref{sec:related:perplexity}, we introduced perplexity as a
way to estimate how well the language model is able to predict the
next word in terms of the number of alternatives that need to be
considered at each point.  For traditional language models, with
estimated probability distribution of $\hat{\Pr}(w_i|w_{\rim})$, the
perplexity of a test set of $N$ words $w_{1,N}$ is calculated as
$2^H$, where $H$ is the entropy and is defined as follows.
\begin{eqnarray}
H & = & -\frac 1 N \sum_{i=1}^N \log_2 \hat{\Pr}(w_i|w_{\rim})
\end{eqnarray}

\subsection{Branching Perplexity}
\label{sec:model:p:branching}

The question now arises as to how to calculate the perplexity of a
POS-based language model.  Here the language model is not only
predicting the next word, but its POS category as well.  To determine
the branching factor, and thus estimate the size of the search space,
we need to look at the estimated probability
\[ \hat{\Pr}(w_ip_i|w_{\rim}p_{\rim}) 
= \hat{\Pr}(w_i|w_{\rim}p_{\ri}) \hat{\Pr}(p_i|w_{\rim}p_{\rim}) \]
where $p_i$ is the POS tag for word $w_i$.  The corresponding
perplexity measure will be as follows.
\begin{eqnarray}
H&=&-\frac{1}{N} \sum_{i=1}^N \log_2 \hat{\Pr}(w_ip_i|w_{\rim}p_{\rim})
\end{eqnarray}
If the perplexity, as given by the above formula, is higher than that
for a traditional word-based language model, it tells us that the
recognizer will have more alternatives that it must explore since it
now must consider alternative POS tags as well.  It also tells us that
the accuracy in correctly predicting both the word and POS assignment
will probably not be as high as a traditional language model does on
words alone.

\subsection{Word Perplexity}

In order to compare a POS-based language model against a traditional
language model, we should not penalize the POS-based language model
for incorrect POS tags, and hence we should ignore them when defining
the perplexity.  Just as with a traditional model, we base the
perplexity measure on $\Pr(w_i|w_{\rim})$.  The problem is that for a
POS-based language model, this probability is not estimated.  Hence,
this probability must be rewritten in terms of the probability of the
partial sequence that we derived in Equation~\ref{eqn:model:6}.  To do
this, our only recourse is to sum over all possible POS sequences, in
a similar way as is shown in Equation~\ref {eqn:related:jel} and
\ref {eqn:related:jel2} of Section~\ref {sec:related:aclass}.
\begin{eqnarray}
\Pr(w_i|w_{\rim})
&=& \sum_{P_{\ri}} \Pr(w_i P_{\ri} | w_{\rim}) \nonumber \\ 
&=& \sum_{P_{\ri}} \frac {\Pr(w_{\ri} P_{\ri})} {\Pr(w_{\rim})} \nonumber \\
&=& \frac{\sum_{P_{\ri}} \Pr(w_{\ri} P_{\ri})} {\Pr(w_{\rim})}
\nonumber \\ 
&=& \frac{\sum_{P_\ri} \Pr(w_{\ri} P_{\ri})}
{\sum_{P_\rim} \Pr(w_{\rim}P_{\rim})} \nonumber \\ 
&=& \frac{\sum_{P_\ri} \Pr(w_iP_i|w_{\rim} P_{\rim})\Pr(w_{\rim}
P_{\rim})} {\sum_{P_\rim} \Pr(w_{\rim}P_{\rim})}
\label{eqn:model:wp-w}
\end{eqnarray}
The result is that we sum over all POS sequences and normalize this by
the sum of all the POS sequences up to the previous word.

Equation~\ref{eqn:model:wp-w} is defined in terms of
$\Pr(w_{\ri}P_{\ri})$.  However, this can be easily written in terms
of the estimated probability distributions, as is shown in the
following.
\[ \Pr(w_{\ri}P_{\ri}) = 
   \prod_{j=1}^i\Pr(w_jP_j|w_{1,j{\dash}1}P_{1,j{\dash}1}) \]
Hence, we define $\hat{\Pr}(w_{\ri}P_{\ri})$ as follows.
\[ \hat{\Pr}(w_{\ri}P_{\ri}) = 
   \prod_{j=1}^i\hat{\Pr}(w_jP_j|w_{1,j{\dash}1}P_{1,j{\dash}1}) \]

Given this, we can now compute the entropy that the language model assigns to the test corpus as follows.
\begin{eqnarray}
H &=& -\frac1N \sum_{i=1}^N \log_2 
    \frac{\sum_{P_{\ri}} \hat{\Pr}(w_iP_i|w_\rim P_\rim) 
			 \hat{\Pr}(w_\rim P_\rim)}
	 {\sum_{P_{\rim}}\hat{\Pr}(w_\rim P_\rim)}  
\label{eqn:model:wp-h}
\end{eqnarray}
We reiterate that we are only summing over all of the POS tags in
order to give a perplexity measure that is comparable to the
measure used by traditional language modeling approaches,
which do not include POS tags as part of the definition of the
speech recognition problem.

\subsection{Word Perplexity with Pruning}
\label{sec:model:p:word}

As Equation~\ref{eqn:model:wp-h} shows, the calculation of word
perplexity relies on using all possible POS sequences.  For an
$n$-gram POS model, a Markov model would have $|{\cal P}|^{n-1}$
states with $|{\cal P}|$ ways of extending each state, resulting in a
Viterbi search of complexity $|{\cal P}|^n$.  To overcome this, we use
a {\em path}-based approach, as explained in Section~\ref
{sec:related:paths}, where only the most likely POS sequences are kept
\cite {ChowSchwartz89:darpa}.  In terms of estimating the perplexity,
we only have some of the POS sequences that we need to sum over.  So,
we define the probability of word $w_i$ in terms of the set of POS
sequences $\Pi_i$ for the words $w_{\rim}$ that survive being pruned.
Hence, we rewrite Equation~\ref {eqn:model:wp-w} as shown below, where
$P_\ri \in \Pi_i{\cal P}$ means that $P_\rim \in
\Pi_i$ and $P_i$ is any POS tag in the tagset ${\cal P}$.
\begin{eqnarray}
\Pr(w_i|w_{\rim})
&=& \frac{\sum_{P_\ri} \Pr(w_iP_i|w_{\rim} P_{\rim})\Pr(w_{\rim} P_{\rim})}
	 {\sum_{P_\rim} \Pr(w_{\rim}P_{\rim})} \nonumber \\
&\approx& \frac{\sum_{P_\ri \in \Pi_i{\cal P}} 
		\Pr(w_iP_i|w_{\rim} P_{\rim})\Pr(w_{\rim} P_{\rim})}
	       {\sum_{P_\rim \in \Pi_i} \Pr(w_{\rim}P_{\rim})}
\end{eqnarray}
This leads to the following estimate for the perplexity.
\begin{eqnarray}
H & \approx & - \frac1N \sum_{i=1}^N \log_2 
	  \frac{\sum_{P_\ri \in \Pi_i{\cal P}} 
		\hat{\Pr}(w_iP_i|w_{\rim} P_{\rim})
		\hat{\Pr}(w_{\rim} P_{\rim})}
	       {\sum_{P_\rim \in \Pi_i} 
		\hat{\Pr}(w_{\rim}P_{\rim})}
\label{eqn:model:wpp-h}
\end{eqnarray}
Doing this will make the word perplexity sensitive to how well the
pruning algorithm is able to include the appropriate POS assignments.
Given that the pruning algorithm removes POS assignments that have
very low probability, the effect of pruning should not significantly
alter the perplexity score.

\section{Results}
\label{sec:model:results}

To test our POS-based language model, we ran a number of
experiments.  
The first set of experiments varies the amount of previous context
that the decision tree can query in building the equivalence classes,
ranging between just using the previous word (a bigram model), up to
using the previous four words (a 5-gram model).
The second set of experiments examines the effect that using richer
contexts for estimating the word and POS probability distributions has
on the perplexity and POS tagging results.  Here we show that the
context used by traditional language models that incorporate POS
information is not rich enough, and that better models can be built by
not simplifying the context.
The third set of experiments examines how well the decision tree's
greedy algorithm for choosing questions translates into globally
optimal decision trees.
The fourth set of experiments shows the benefit of using composite
questions in building the decision tree.
The fifth set of experiments examines the question of how the word
and POS trees should be grown.  Here, we show that it is best to take
into account the POS tags when growing the word classification trees.
Before we give the results, however, we first explain the methodology
that we use throughout all experiments.

\subsection{Experimental Setup}
\label{sec:model:setup}

In order to make the best use of our limited data, we tested our
model using a six-fold cross-validation procedure.  We divided the
dialogs into six partitions, and each partition was tested using a
model built from the data of the other five partitions.\footnote{For
expository reasons, we will refer to the results on the test corpus.
This should be interpreted as the cumulative results over all six
partitions of the corpus.}  The dialogs involve 34 different speakers;
and so the results are for multiple speakers.  However, in dividing up
the dialogs into the six partitions, we distributed the speaker pairs
into the six partitions as evenly as possible.  As a result, for each
dialog tested (except one), the model used to test it was trained on at 
least one dialog from the same speaker pair.  So, the results are not
fully speaker-independent.

Since current speech recognition rates for spontaneous speech are
quite low, these experiments use the ideal output from a
speech recognizer, namely the hand-collected transcripts.

Information about changes in speaker is incorporated into the model by
using a special token {\bf $<$turn$>$} and POS tag {\bf TURN} to mark
them, in the same way that sentence indicators are used by POS taggers
for written text.  The POS results do not include our ability to tag
the end of turn marker {\bf $<$turn$>$} as {\bf TURN}, but we do
include our ability to predict the end of turn marker {\bf $<$turn$>$}
in our perplexity results.

We treat contractions, such as ``that'll'' and ``gonna'', as separate
words, treating them as ``that'' and ``'ll'' for the first example,
and ``going'' and ``ta'' in the second.\footnote {See Heeman and
Damnati \shortcite {HeemanDamnati97:asru} for implications of treating
contractions as separate words in an actual speech recognizer.}  We
also changed all word fragments into a common token {\bf
$<$fragment$>$}.

\subsection{Using Larger Histories}
\label{sec:model:r:ngram}

The first set of experiments show the effect of how much of the
previous context we allow the decision tree algorithm to query.  We
vary the amount of context (or history) from just the previous word (a
bigram model), to the previous four words (a 5-gram model).  The
results given in Table~\ref{tab:model:ngram} reflect the cumulative
results of the model over all six partitions of the Trains corpus.
\begin{table}
\begin{center}
\begin{tabular}{|l|r|r|r|r|} \hline
                          & Bigram&Trigram& 4-gram& 5-gram\\ \hline \hline
{\em POS Tags}            &       &       &       &       \\
\ Errors                  &  1838 &  1711 &  1751 &  1739 \\
\ Error Rate              &  3.15 &  2.93 &  3.00 &  2.98 \\ \hline
{\em Discourse Markers}   &       &       &       &       \\
\ Errors                  &   606 &   630 &   655 &   643 \\
\ Error Rate              &  7.32 &  7.61 &  7.91 &  7.76 \\
\ Recall                  & 96.66 & 96.75 & 96.41 & 96.52 \\
\ Precision               & 96.03 & 95.68 & 95.70 & 95.74 \\ \hline
{\em Perplexity}          &       &       &       &       \\
\ Word                    & 27.24 & 24.04 & 23.99 & 24.00 \\ 
\ Branching               & 29.95 & 26.35 & 26.30 & 26.31 \\ \hline
\end{tabular}
\end{center}
\caption{Using Larger Histories to Estimate Probabilities}
\label{tab:model:ngram}
\end{table}%
We give the performance of the POS tagging, detection of discourse
markers, and the perplexity.  As discussed in the Section~\ref
{sec:model:p:word}, the word perplexity measures how well the model
can predict the next word, regardless of whether it guesses the right
POS tag.  The branching perplexity, as discussed in Section~\ref
{sec:model:p:branching}, measures how well the model can predict both
the next word and its POS tag, and hence gives an indication of number
of alternatives that must be pursued.  The second column of the table
gives the results for a bigram model, where we restrict the decision
tree to only asking about the prior word and its POS tag.  The third
column gives the results for a trigram model, the fourth gives the
results of a 4-gram model, and the fifth column gives the results for
a 5-gram model.

The bigram model incorrectly tagged 1838 of the 58298 words, giving a
POS error rate of 3.15\%.  For the discourse markers, it made 606
errors in comparison to the 8278 discourse markers, giving an error
rate of 7.32\%.  Of these errors, 276 resulted from not tagging a
discourse marker with one of the discourse marker POS tags,
translating into a recall rate of 96.66\%.  The 330 other errors
resulted from tagging a word that is not a discourse marker with one
of the discourse marker POS tags, which gives a precision rate of
96.03\%.\footnote {See Section~\ref {sec:model:c:dm} for experiments
that show the impact of modeling discourse markers on word
perplexity.}  In terms of perplexity, the model achieved a word
perplexity of 24.04 and a branching perplexity of 26.35.

From the results, we see that moving from a bigram to a trigram
results in a 7.1\% reduction in the POS error rate and an 11.3\%
reduction in perplexity, while actually giving a slight degradation in
performance in identifying discourse markers: 630 mistakes versus 606
mistakes for the bigram model.  The difference between the trigram
language model and the 4-gram, however, is much less pronounced.
Although there is a slight improvement in perplexity, there is a
degradation in the POS error rate, and actually a slight decline for
the identification of discourse markers.  We also include the results
for a 5-gram language model.  Here, we see that there is a very small
improvement in perplexity in comparison to the 4-gram model, but a
noticeable decline in POS tagging.  Because of the small size of the
Trains corpus, there is not enough data to warrant a 5-gram language
model, and perhaps not even a 4-gram language model.  The decision
tree approach, however, seems capable of deciding how much of the
available context to use, and hence it is not necessary to restrict
how much of the context it can look at.  In the subsequent
experiments, we use the 3-gram version as the basis of comparison.

\subsection{Using Richer Histories}
\label{sec:model:r:richer}

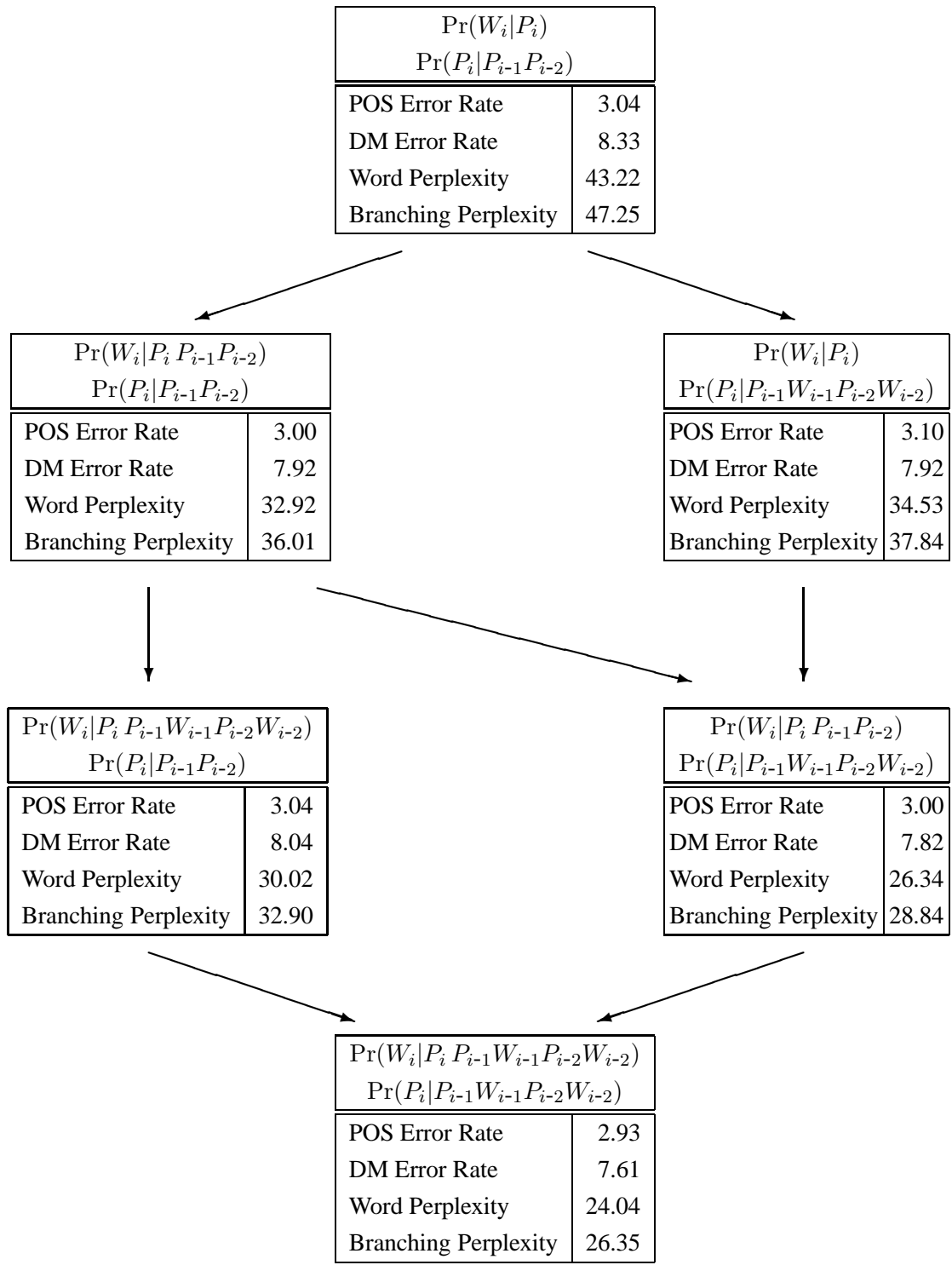
\begin{figure}
\small
\setlength{\unitlength}{\textwidth}
\begin{picture}(1,1.4)
\put(0.52,1.25){\makebox(0,0){\begin{tabular}{|l|r|} \hline
\multicolumn{2}{|c|}{$\Pr(W_i|P_i)$} \\    
\multicolumn{2}{|c|}{$\Pr(P_i|P_{\im}P_{\imm})$} \\ \hline \hline
POS Error Rate            &  3.04 \\
DM Error Rate             &  8.33 \\
Word Perplexity           & 43.22 \\ 
Branching Perplexity      & 47.25 \\ \hline
\end{tabular}}}

\put(0.17,0.90){\makebox(0,0){
\begin{tabular}{|l|r|} \hline
\multicolumn{2}{|c|}{$\Pr(W_i|P_i\,P_{\im}P_{\imm})$} \\
\multicolumn{2}{|c|}{$\Pr(P_i|P_{\im}P_{\imm})$} \\ \hline \hline
POS Error Rate            &  3.00 \\
DM Error Rate             &  7.92 \\
Word Perplexity           & 32.92 \\ 
Branching Perplexity      & 36.01 \\ \hline
\end{tabular}}}

\put(0.17,0.50){\makebox(0,0){\begin{tabular}{|l|r|} \hline
\multicolumn{2}{|c|}{$\Pr(W_i|P_i\,P_{\im}W_{\im}P_{\imm}W_{\imm})$} \\
\multicolumn{2}{|c|}{$\Pr(P_i|P_{\im}P_{\imm})$} \\ \hline \hline
POS Error Rate            &  3.04 \\
DM Error Rate             &  8.04 \\
Word Perplexity           & 30.02 \\ 
Branching Perplexity      & 32.90 \\ \hline
\end{tabular}}}

\put(0.85,0.90){\makebox(0,0){\setlength{\tabcolsep}{0.2em}
\begin{tabular}{|l|r|} \hline
\multicolumn{2}{|c|}{$\Pr(W_i|P_i)$} \\
\multicolumn{2}{|c|}{$\Pr(P_i|P_{\im}W_{\im}P_{\imm}W_{\imm})$}\\ \hline \hline
POS Error Rate            &  3.10 \\
DM Error Rate             &  7.92 \\
Word Perplexity           & 34.53 \\ 
Branching Perplexity      & 37.84 \\ \hline
\end{tabular}}}

\put(0.85,0.50){\makebox(0,0){\setlength{\tabcolsep}{0.2em}
\begin{tabular}{|l|r|} \hline
\multicolumn{2}{|c|}{$\Pr(W_i|P_i\,P_{\im}P_{\imm})$} \\
\multicolumn{2}{|c|}{$\Pr(P_i|P_{\im}W_{\im}P_{\imm}W_{\imm})$}\\ \hline \hline
POS Error Rate            &  3.00 \\
DM Error Rate             &  7.82 \\
Word Perplexity           & 26.34 \\ 
Branching Perplexity      & 28.84 \\ \hline
\end{tabular}}}

\put(0.52,0.15){\makebox(0,0){\begin{tabular}{|l|r|} \hline
\multicolumn{2}{|c|}{$\Pr(W_i|P_i\,P_{\im}W_{\im}P_{\imm}W_{\imm})$} \\
\multicolumn{2}{|c|}{$\Pr(P_i|P_{\im}W_{\im}P_{\imm}W_{\imm})$}\\ \hline \hline
POS Error Rate            &  2.93 \\
DM Error Rate             &  7.61 \\
Word Perplexity           & 24.04 \\ 
Branching Perplexity      & 26.35 \\ \hline
\end{tabular}}}

\thicklines
\put(0.42,1.11){\vector(-3,-1){0.22}}
\put(0.62,1.11){\vector(3,-1){0.22}}
\put(0.15,0.75){\vector(0,-1){0.10}}
\put(0.33,0.75){\vector(4,-1){0.40}}
\put(0.85,0.75){\vector(0,-1){0.10}}
\put(0.15,0.36){\vector(3,-1){0.22}}
\put(0.85,0.36){\vector(-3,-1){0.22}}

\end{picture}
\caption{Using Richer Histories to Estimate Probabilities}
\label{fig:model:context}
\end{figure}

As we mentioned earlier, typical language models that use POS
tags---both POS taggers and speech recognition language
models---estimate the probability of a word based simply on its POS
tag and so do not take into account the previous POS tags or their
word identities.  They also estimate the probability of the POS tag of
the current word based only on the POS tags of the previous words and
thus ignore the word identities.  Figure~\ref{fig:model:context}
shows the effect of varying the richness of the information that the
decision tree algorithm is allowed to use in estimating the POS and
word probabilities.  We consider three choices for the context for the
word probability.
\begin{enumerate} 
\item the POS tag 
\item the POS tag and the POS tags of the two previous words
\item the POS tag and the POS tags and word identities of the two 
previous words
\end{enumerate}
We consider two choices for the context for the POS probability.
\begin{enumerate} 
\item the POS tags of the two previous words
\item the POS tags and word identities of the two previous words
\end{enumerate}
In all, there are six combinations that we consider, and there is a
table showing the results for each.  The top table gives the results
of the combination typically used by POS taggers and by previous
attempts at using POS tags in a speech recognition language model,
namely estimating the word probabilities solely on the basis of their
POS tags and estimating the POS tags on the basis of the previous POS
tags.  The bottom table gives the results for the combination that
uses the full richness of the information in the context.  The rest of
the tables give the results for intermediate amounts of information.
The arrows between the tables indicate that more context is being
added, either in estimating the probability of the POS tag or of the
word.

The results show that adding in the extra context has the biggest
effect on the perplexity measures, decreasing the word perplexity from
43.22 to 24.04, a reduction of 44.4\%, with the perplexity decreasing
as each additional piece of context is added.  The effect on the POS
tagging rate is less pronounced, but still with a decrease in the
error rate from 3.04\% to 2.93\%, giving an error rate reduction of
3.8\%.  We also see a 8.7\% reduction in the error rate for
identifying discourse markers.  Hence, in order to use POS tags in a
speech recognition language model, we need to use a richer context for
estimating the probabilities than what is typically used.

\subsection{Constraining the Decision Tree}
\label{sec:model:r:constraints}

In this section, we examine the decision tree algorithm's ability to
find the best overall decision tree.  Since searching for the best
overall decision tree will computationally explode, decision tree
algorithms use a greedy approach in constructing the decision tree.
The algorithm successively picks the composite question that gives the
best decrease in node impurity given the previous questions that were
asked.  This greedy algorithm gives no guarantee that the best overall
set of composite questions will be found.  In fact, we have found it
worthwhile to restrict which questions the decision tree algorithm can ask, in
order to help it find the best overall tree.

As we mentioned earlier, the word identity information $W_{i{\dash}j}$
is viewed as further refining the POS tag of the word $P_{i{\dash}j}$.
Hence, questions about the word encoding are only allowed if the POS
tag is uniquely defined.  Furthermore, for both POS and word
questions, we restrict the algorithm so that it only asks about more
specific bits of the POS tag and word encodings if it has already
uniquely identified the less specific bits.  However, this still
leaves a lot of alternatives for the decision tree to consider.  For
instance, in estimating the POS tag $P_i$, the decision tree is free
to query the POS tags of any of the previous words, and if the POS tag
of one of the previous words is uniquely identified, it can ask about
the word encoding.  In Table~\ref{tab:model:constraints},
\begin{table}
\begin{center}
\begin{tabular}{|l|r|r|r|} \hline
                        &\mc{Minimal} &\mc{POS} &\mc{Full} \\ \hline \hline
{\em POS Tags}          &          &         &          \\
\ Errors                &     1756 &    1711 &     1722 \\
\ Error Rate            &     3.01 &    2.93 &     2.95 \\ \hline
{\em Discourse Markers} &          &         &          \\
\ Errors                &      660 &     630 &      614 \\
\ Error Rate            &     7.97 &    7.61 &     7.41 \\
\ Recall                &    96.61 &   96.75 &    96.85 \\
\ Precision             &    95.46 &   95.68 &    95.77 \\ \hline
{\em Perplexity}        &          &         &          \\
\ Word                  &    24.43 &   24.04 &    24.24 \\ 
\ Branching             &    26.81 &   26.35 &    26.59 \\ \hline
\end{tabular}\end{center}
\caption{Adding Additional Constraints}
\label{tab:model:constraints}
\end{table}%
we contrast the effectiveness of adding additional constraints.  The
second column gives the results with the minimal constraints, which we
outlined above.  The third column adds the constraint that the POS tag
$P_{i{\dash}j}$ can only be queried if the POS tag $P_{i{\dash}j+1}$
has been fully explored.  The fourth column adds the constraint that
the $P_{i{\dash}j}$ can only be queried if the word $W_{i{\dash}j+1}$
has been fully explored.

From comparing the results of the second and third columns, we see
that the decision tree, unless constrained, will ask questions about
the POS tags of earlier occurring words before it has fully explored
the POS information in the later ones.  Constraining the decision
tree results in an improved language model, with a perplexity
reduction of 1.6\%.  The effect is amplified as we increase the amount
of context that the decision tree can consider: adding the constraint
to a 4-gram model gives a reduction of 3.1\%, and adding it to a
5-gram model gives a reduction of 5.5\%.\footnote{See Section~\ref
{sec:model:c:dt:word} for a comparison of this effect with a
word-based model.}  Hence, we see that the greedy algorithm that the
decision tree algorithm uses for deciding what questions to ask is not
optimal, and can be improved by imposing some constraints.

In comparing the results of the third and fourth columns, we see that
constraining the decision tree to fully explore the word identities
(in addition to the POS tags) before looking at the POS tags of
earlier words is not worthwhile.  Hence, the decision tree algorithm
is able to make use of word generalizations, and it is worthwhile to
not always fully explore the word information.

\subsection{Allowing Composite Questions}
\label{sec:model:r:pylons}

The next area that we explore is the use of composite questions in
estimating the probability distributions.  In Table~\ref
{tab:model:composites}, we give the results of varying whether
composite questions are use, and, if they are, the number of
alternatives that are kept around.
\begin{table}
\begin{center}
\begin{tabular}{|l|r|r|r|r|r|} \hline
Alternate Composites 	  & None  & Single&\mc{5} &\mc{10} &\mc{20} \\ 
\hline \hline
{\em POS Tags}            &       &       &       &        &        \\
\ Error Rate              &  3.02 &  2.95 &  2.95 &  2.93  &   2.93 \\ \hline
{\em Discourse Markers}   &       &       &       &        &        \\
\ Recall                  & 96.93 & 96.87 & 96.77 & 96.75  &  96.76 \\
\ Precision               & 95.72 & 95.68 & 95.71 & 95.68  &  95.75 \\
\ Error Rate              &  7.39 &  7.50 &  7.56 &  7.61  &   7.52 \\ \hline
{\em Perplexity}          &       &       &       &        &        \\
\ Word                    & 25.34 & 24.28 & 24.08 & 24.04  &  24.05 \\ 
\ Branching               & 27.83 & 26.61 & 26.39 & 26.35  &  26.36 \\ \hline
\end{tabular}
\end{center}
\caption{Using Composite Questions in Estimating Probabilities}
\label{tab:model:composites}
\end{table}%
The second column gives the results when composite questions are not
used by the decision tree algorithm.  The third column gives the
results when the number of alternatives that the decision tree
explores is limited to the single best one at each level of the pylon.
The fourth, fifth and sixth columns give the results when the decision
tree algorithm keeps the best five, ten, and twenty alternatives,
respectively.

As can be seen in the table, allowing composite questions improves the
performance of the language model, especially in terms of perplexity.
Even using a very greedy algorithm, as illustrated by the results in
the third column, using composites results in a 4.2\% reduction in
word perplexity.  Using a less greedy algorithm that explores the five
best alternatives at each level of the pylon results in a further
improvement in perplexity, bringing the overall reduction to 5.0\%.
Increasing the number beyond this does not have much of an effect, but
will prove useful for the richer contexts that are needed in the next
chapter for modeling speech repairs and boundary tones.

\subsection{Building the Classification Trees}
\label{sec:model:r:ctrees}

The word and POS classification trees determine what questions the
decision tree can ask about the words and POS tags in the context.
Hence the quality of these classification trees has a significant
impact on how well the decision tree algorithm can estimate the
probability distributions.  In Section~\ref {sec:model:ctrees}, we
advocated that the word classification tree should be viewed as a
further specification of the POS tags.  In this section, we contrast
this approach with two other approaches for building the word trees.
The competing approaches view the word and POS information of the
words in the context as two separate sources of information.  Below we
describe the three approaches.
\begin{description}
\item[{\em Force POS}:] 
This is the method advocated in Section~\ref {sec:model:ctrees}.  The
word classification tree is grown by starting with a separate class
for each word and POS combination that occurs in the training corpus.
Combinations that occur only once are initially clustered into the
group {\bf low}, which is distinct for each POS tag.  Only words that
have the same POS tag are allowed to be merged, thus giving a separate
tree for each POS tag.

\item[{\em Mix POS}:]
This method allows the classification algorithm to merge classes that
have words of different POS tags.  The resulting word classification
tree mixes together words of different POS tags, and hence the words
are no longer viewed as a further refinement of the POS tags. 

\item[{\em Ignore POS}:]
This method completely ignores the POS tags when building the
classification tree.  We start with a class for each word, which might
combine several POS senses, such as ``loads'' as a plural noun and as
a present-tense third-person verb.  The resulting tree is hence
entirely insensitive to the POS information.
\end{description}

As was shown in Section~\ref {sec:model:r:constraints}, the
restrictions that are imposed on the decision tree algorithm in
choosing the questions can have a significant impact on the
performance of the resulting language model.  Hence we compare the
three methods under two different sets of constraints.\footnote {For
all three methods, we start the probability of $W_i$ with an
equivalence class for each POS tag $P_i$.}
\begin{description}
\item [{\em Minimal}:] 
This version imposes the minimal constraints.
For the {\em Force POS} approach, the information about $W_{i-j}$ can only
be queried if $P_{i-j}$ is unique. For the other two methods, there is
no need to force the decision tree to uniquely determine the POS tag
before looking at the word identity information, as given by the word
classification tree.
\item [{\em POS}:]
In Section~\ref {sec:model:r:constraints}, we found it best to allow
the decision tree algorithm to examine the POS tag of earlier words
only when it has fully examined the POS tag of later words.  Because
of the emphasis on the POS tags, we add the constraint to the {\em
Ignore POS} and {\em Mix POS} that questions about $W_{i-j}$ can only
be asked if $P_{i-j}$ is unique.  Hence, we force all methods to fully
explore the POS tag of a word before asking about word information.
\end{description}

The second consideration in comparing the three approaches is how the
clustering algorithm initially builds the group of low-occurring
words.  For the {\em Force POS} approach, we find that the best
results occur by just including the singletons.  However, this is not
the case for the other two methods.  Hence, we vary the cutoff for
including words in this group and contrast the {\em Force POS}
approach with the cutoff that gives the best results.  The reason why
the cutoff is important is because without the POS constraints on
growing the word classification tree, words that seldomly occur will
be more difficult to properly cluster.  However, including words in
the class of low-occurring words prevents the decision tree from
asking questions about them.

Table~\ref{tab:model:r:ctrees} 
\begin{table}
\begin{center}
\begin{tabular}{|l|r|r||r|r||r|r|} \hline
     & \multicolumn{2}{c||}{Ignore POS}
     & \multicolumn{2}{c||}{Mix POS}
     & \multicolumn{2}{c|}{Force POS} \\ \cline{2-7}
                        &Minimal&  POS&Minimal& POS &Minimal&POS \\\hline\hline
Low-Occurring           &   5  &   1  &   4  &   3  &   1  &   1   \\ \hline
{\em POS Tags}          &      &      &      &      &      &       \\
\ Error Rate            &  3.05&  2.96&  3.00&  2.98&  3.01&  2.93 \\ \hline
{\em Discourse Markers} &      &      &      &      &      &       \\
\ Recall                & 96.66& 96.67& 96.47& 96.48& 96.61& 96.75 \\
\ Precision             & 96.52& 95.71& 95.60& 95.45& 95.46& 95.68 \\
\ Error Rate            &  7.86&  7.64&  7.96&  8.10&  7.97&  7.61 \\ \hline
{\em Perplexity}        &      &      &      &      &      &       \\
\ Word                  & 24.37& 24.31& 24.22& 24.28& 24.43& 24.04 \\
\ Branching             & 26.74& 26.69& 26.59& 26.61& 26.81& 26.35 \\ \hline
\end{tabular}
\end{center}
\caption{Building the Classification Trees}
\label{tab:model:r:ctrees}
\end{table}%
gives the results of the language model using the three different
approaches for building the word classification trees and the two
types of constraints.  The first row of the table gives the value for
the cutoff that was optimal for the test data.  For instance, for {\em
Mix POS} using no constraints, we find that it is best to include all
words that occur four times or fewer in the group of low-occurring
words.

From the table, we see that the best results are obtained using the
{\em Force POS} approach, in which the word classification trees are
built by not allowing classes to be merged that have words of
different POS tags.  In other words, when asking questions about the
context, it is best to treat the words as a further refinement of the
POS tags.  The best results were obtained using the {\em POS}
constraint, in which we force the POS tag of later words to be fully
explored before asking questions of the earlier POS tags.  However,
the {\em POS} constraint did not have much of an effect on the {\em
Ignore POS} and {\em Mix POS} approaches.  In fact, when using no
constraints, both of these methods did better than the {\em Force POS}
approach.

\section{Comparison}
\label{sec:model:comparison}

In this section, we contrast the results of our POS-based language
model with other language models that have been proposed for speech
recognition.  We remind the reader that our language model does more
than just identify the most probable word interpretation; it also
gives the most probable POS interpretation as well as identifies
discourse markers.  Hence our model is accomplishing part of the
linguistic analysis that would otherwise have to be done as a later
process.  This leads to the following questions.
\begin{enumerate}
\item Does modeling syntactic information, in the form of POS tags,
lead to better language modeling?
\item Does modeling discourse marker usage, by way of additional POS tags, 
lead to better language modeling?
\end{enumerate}

Our model also relies heavily on a decision tree learning algorithm to
estimate the probability distributions.  Although they have been used
before for speech recognition language models \cite
{Bahl-etal89:tassp}, they have not been used with hierarchical
partitionings of the words, which more recent approaches in
statistical language learning (e.g.~\cite
{Black-etal92:darpa:hbg,Black-etal92:darpa:pos,Magerman94:thesis})
have used.  This leads to the following question.
\begin{enumerate}
\setcounter{enumi}{2}
\item Does the decision tree learning algorithm along with hierarchical
partitionings of the words (and POS tags) lead to better probability
estimates?
\end{enumerate}

In this comparison, we do not consider POS taggers, which are text
based.  Section~\ref{sec:model:r:richer} already explored the issue of
using richer contexts than what POS taggers typically use, and we
showed that this results in a reduction in the POS error rate of
3.8\%.  For text, the POS tagging rate can be further improved by
better modeling unknown words.  But as explained in Section~\ref
{sec:model:unknown}, such attempts are not always appropriate for
speech recognition.  We also do not consider the effect of corpus
size, and other types of dialog scenarios, such as human-machine
dialogs, and information querying dialogs.

\subsection{Modeling Discourse Markers}
\label{sec:model:c:dm}

Our first set of experiments show the effect of explicitly modeling
discourse markers with the special POS tags, as described in
Section~\ref {sec:corpus:dm}.\footnote {See Section~\ref
{sec:results:c:dm} for a comparison of the results of modeling
discourse markers with the work of Litman \shortcite {Litman96:jair}.}
We contrast the model that explicitly models discourse markers by way
of special POS tags with a model in which discourse markers are not
identified.  This model, which we refer to as {\em No DM}, collapses
the discourse usages with the sentential usages.  Thus, the discourse
conjunct {\bf CC\_D} is collapsed into {\bf CC}, the discourse
adverbial {\bf RB\_D} is collapsed into {\bf RB}, and the
acknowledgment {\bf AC} and discourse interjection {\bf UH\_D} are
collapsed into {\bf UH\_FP}.  Thus, the tagset of {\em No DM} removes
the extra discourse marker tags that we added to the Penn Treebank
tagset \cite {Marcus-etal93:cl,Santorini90:tr}.

Table~\ref {tab:model:c:dm} gives the results of the experiment.  
\begin{table}
\begin{center}
\begin{tabular}{|l|r|r|r|} \hline
                     &       &\multicolumn{2}{c|}{DM} \\ \cline{3-4}
                     &       & \mc{Report} & \mc{Ignore}  \\ 
                     & No DM & DM Errors   & DM Errors    \\ \hline \hline
POS Errors           &  1219 &  1711 &  1189 \\
POS Error Rate       &  2.09 &  2.93 &  2.04 \\
DM Errors            &  n.a. &   630 &  same \\
Word Perplexity      & 24.20 & 24.04 &  same \\ 
Branching Perplexity & 26.08 & 26.35 &  same \\ \hline
\end{tabular}
\end{center}
\caption{Effect of including Discourse Markers Identification}
\label{tab:model:c:dm}
\end{table}%
The second column gives the results of the {\em No DM} model, which
does not distinguish discourse marker usage.  The third column gives
the results of the model that does distinguish discourse marker usage.
We see that word perplexity improves, from 24.20 to 24.04, by modeling
the discourse markers as distinct POS tags, which shows that we are
better able to predict the next word by explicitly modeling discourse
markers.  However, we see that the branching perplexity is worse for
modeling the discourse markers, which results from the increase in the
number of possible POS tags.  We also see that the POS error rate is
worse for modeling the discourse markers; in fact, modeling discourse
markers results in 492 more POS errors.  However, the discourse marker
version made 630 errors in identifying the discourse markers, which
each correspond to a POS error, and so the increase in POS errors
might be solely attributable to errors in identifying discourse
markers.

In the fourth column we give the results of ignoring POS errors that
are attributable to misidentifying discourse marker usage.  Here, any
errors that are attributable to confusion between {\bf CC\_D} and {\bf
CC}, between {\bf RB\_D} and {\bf RB}, or between {\bf AC}, {\bf
UH\_D} and {\bf UH\_FP} are ignored.  This reduces the number of POS
errors down to 1189 for the discourse marker version, versus 1219 for
the version that does not distinguish discourse markers.  Hence, we
see that modeling discourse markers actually results in a small
improvement to POS tagging, decreasing the number of errors not
associated with discourse markers from 1219 to 1189, a reduction of
2.5\%.  Although the improvements in perplexity and POS tagging are
small, the result indicates that there are interactions, and hence
discourse markers should be resolved at the same time as POS tagging
and speech recognition word prediction.

\subsection{Word-Based Decision-Tree Model}
\label{sec:model:c:dt:word}

Our second comparison explores the effectiveness of modeling syntactic
information in the language model.  We contrast the POS-based model
with a word-based model, which does not employ POS tags.  To make the
comparison as focused as possible, we use the decision tree algorithm
and hierarchical clustering of the words to estimate the probability
distributions.  Since the word-based model does not employ POS tags,
the word classification tree that is built is the same one that is
used in the {\em Ignore POS} method of Section~\ref
{sec:model:r:ctrees}.  Here we found it best to define the
low-occurring words as the singletons.

Since the word model uses a word classification tree, the decision
tree algorithm can still make generalizations about similar words when
finding equivalence classes of the context.  In fact, it can even ask
questions about the classifications of a word $W_{i{\dash}j{\dash}1}$
before fully exploring $W_{i{\dash}j}$, and it can even later return
to asking questions about $W_{i{\dash}j}$.  In Section~\ref
{sec:model:r:constraints}, we found it best to partially constrain the
decision tree algorithm for the POS-based model.  Hence, in this
section, we build two versions of the word-based decision tree model:
one with minimal constraints (i.e.~the decision tree algorithm can
only ask about more specific bits of the word encodings if it has
already asked about the less specific bits) and a second constrained
to fully explore a word before exploring prior words.  Table~\ref
{tab:model:c:dt:word} gives the word perplexity results for the
word-based model and contrasts the results with the POS model.  We
give the results varying the size of the context from a bigram model
through to a 5-gram model.\footnote{For the bigram model, the
constraints have no impact since there is only one word of context
that is being used.}
\begin{table}
\begin{center}
\begin{tabular}{|l|r|r||r|r|r|} \hline
        & \multicolumn{2}{c||}{Word-Based} 
	& \multicolumn{3}{c|}{POS-Based} \\ \cline{2-6}
        &\mc{Minimal} & \multicolumn{1}{c||}{Full}
	&\mc{Minimal} & \mc{POS}   & \mc{Full} \\ \hline \hline
Bigram  & 29.07 & 29.07 & 27.24 & 27.24 & 27.24 \\
Trigram & 25.53 & 25.67 & 24.43 & 24.04 & 24.24 \\
4-gram  & 25.64 & 25.45 & 24.81 & 23.99 & 24.15 \\
5-gram  & 25.98 & 25.40 & 25.43 & 24.00 & 24.15 \\ \hline
\end{tabular}
\end{center}
\caption{Comparison between Word and POS-Based Decision Tree Models}
\label{tab:model:c:dt:word}
\end{table}%
Interestingly enough, with no constraints, the 4-gram and 5-gram
word-based models do worse than the trigram model, while the models
with full constraints show improved performance as the amount of
context is increased.  This could be the same phenomena as we observed
with the POS-based model in Section~\ref {sec:model:r:constraints}: as
more alternatives become available, the decision tree algorithm's
greedy approach is not able to properly distinguish between them.

The main contrast that we want to draw between the word-based and
POS-based versions is that the POS-based model does significantly
better: it achieved a perplexity of 24.00 for the 5-gram model while
the word-based model achieved a perplexity of 25.40.  This 5.5\%
improvement is not simply because of the better constraints that are
imposed for the POS model, since the POS-based model still does better
when both models are fully constrained: 25.40 for the word-based model
and 24.15 for the POS-based model.

The reason for the improved performance of the POS-based model over
the word-based model is that the POS tags can help distinguish
contexts based on syntactic information, which is probably more
informative than just the words in the context.  In addition, taking
advantage of the POS tags in growing the word classification trees
results in better classification trees for the POS-based model.  The
trees are not polluted by words that take on more than one POS sense,
and they are easier to grow since we constrain the growing process to
respect the POS tags.

\subsection{Class-Based Decision-Tree Model}
\label{sec:model:c:dt:class}

We next examine whether POS tags can be replaced by automatically
created unambiguous word classes.  The word classes are obtained from
the word clustering algorithm, but stopping once the number of classes
reaches a certain number.  Again, in order to focus the comparison on
using classes versus POS tags, we estimate the probabilities for the
class-based model using the decision tree algorithm with hierarchical
classification trees of the word classes and the words within each
word.  This clustering effectively just gives the word hierarchy of
the previous section, but with a cut through the tree marking the
classes.  Hence, the only difference between the word-based model of
the previous section and the class-based approach explored in this
section is that the probability estimate of a word given the previous
words is split into two parts: we first estimate the class given the
previous words and classes, and then estimate the word given the class
and the previous words and classes.

In using a class based approach, the question arises as to how many
classes one should use.  Unfortunately, the clustering algorithm given
by Brown~\etal~does not have a mechanism to decide when to stop
clustering words.  Hence, to give an optimal evaluation of the
class-based approach, we choose the number of classes that gives the
best results, in increments of 25.  We found that the best results
were obtained at 100 classes.

Table~\ref{tab:model:c:dt:class} gives the results of running three
sets of experiments: one with minimal restrictions, one forcing the
classes to be uniquely identified before identifying the next word,
and a third forcing the words to be fully identified.
\begin{table}
\begin{center}
\begin{tabular}{|l|r|r||r|r|r||r|r|r|} \hline
                      & \multicolumn{2}{c||}{Word-Based} 
                      & \multicolumn{3}{c||}{Class-Based} 
		      & \multicolumn{3}{c|}{POS-Based} \\ \cline{2-9}
        &\mc{Minimal} & \multicolumn{1}{c||}{Full}
	&\mc{Minimal} & \mc{Class} & \multicolumn{1}{c||}{Full}
	&\mc{Minimal} & \mc{POS}   & \mc{Full} \\ \hline \hline
Bigram  & 29.07& 29.07& 28.56& 28.56 & 28.56& 27.24& 27.24& 27.24 \\ 
Trigram & 25.53& 25.67& 25.26& 25.24 & 25.26& 24.43& 24.04& 24.24 \\ 
4-gram  & 25.64& 25.45& 25.58& 25.05 & 25.08& 24.81& 23.99& 24.15 \\ 
5-gram  & 25.98& 25.40& 25.99& 25.04 & 25.11& 25.43& 24.00& 24.15 \\ 
\hline
\end{tabular}
\end{center}
\caption{Comparison between Word, Class and POS-Based Decision Tree Models}
\label{tab:model:c:dt:class}
\end{table}%
Just as with the POS and word-based models, we find that the
unconstrained version cannot effectively deal with the increase in
context.  We also see that the version that does not enforce fully
exploring the word information does better than the version that just
enforces exploring the classes.

We also find that the class-based model does better than the
word-based model.  Hence, splitting the problem of predicting a word
into the two parts---first predict the class and then predict the
word---results in better estimates of the probability distributions,
as evidence by the 1.4\% reduction in perplexity for the 5-gram
versions.  However, the class-based model does not match the
performance of the POS-based model as evidenced by the POS-based
model's 4.2\% reduction in perplexity over the class-based model for
the 5-gram versions.  Hence, the linguistic information, as captured
by the POS tags, results in a better model than automatically created
classes.

\subsection{Word-Based Backoff Model}
\label{sec:model:c:bo:word}

Using a decision tree algorithm to estimate the probability
distribution is not the only option.  In this section, we contrast the
decision tree models with a word-based model where the probabilities
are estimated using a backoff approach \cite {Katz87:assp}.  We used
the CMU statistical language modeling toolkit \cite {Rosenfeld95:arpa}
to build the word-based backoff models.\footnote{This toolkit is
available by anonymous FTP from {\tt ftp.cs.cmu.edu} in the directory
{\tt project/fgdata} under the name {\tt
CMU\_SLM\_Toolkit\_V1.0\_release.tar.Z}.  A newer version of the
toolkit is now available from Cambridge University.} We trained the
model using the exact same information (with the exception of the POS
tags) and we obtained the results in the same manner, namely using a
six-fold cross-validation procedure. A comparison of the results
achieved using the word-based backoff model, word-based decision-tree
model, and POS-based decision tree model is given in Table~\ref
{tab:model:c:bo:word}.\footnote {As described by Katz
\shortcite{Katz87:assp}, one can choose to exclude some of the low
occurring bigrams and trigrams when estimating the bigram and trigram
probabilities, and instead distribute this probability amongst the
unseen bigrams and trigrams, respectively. Doing this results in a
smaller model since fewer bigrams and trigrams need to be explicitly
kept; however, this is at the expense of a small degradation in
perplexity.  Hence the results reported here do not make use of
this option.}
\begin{table}
\begin{center}
\begin{tabular}{|l|r|r|r|} \hline
           & \multicolumn{1}{c|}{Backoff}
           & \multicolumn{2}{c|}{Decision Tree} \\ \cline{2-4}
           & Word-Based & Word-Based & POS-Based \\ \hline \hline
Bigram     &      29.30 &      29.07 &     27.24 \\
Trigram    &      26.13 &      25.53 &     24.04 \\ \hline
\end{tabular}
\end{center}
\caption{Comparison between Backoff and Decision Trees}
\label{tab:model:c:bo:word}
\end{table}%
The word-based backoff bigram model achieved a perplexity of 29.30 and
the trigram model a perplexity of 26.13.  These results are in
contrast to the POS-based bigram perplexity of 27.24 and trigram
perplexity of 24.04.  Thus, our bigram model gives a perplexity
reduction of 7.0\% and our trigram model a reduction of 8.0\% over the
word-based backoff models.  Hence we see that our model, based on
using decision trees and incorporating POS tags, is better able to
predict the next word.  In comparison to the word-based decision tree
model, we also see an improvement over the backoff method; however,
as we discuss at the end of this section, this is the result of better
handling of unknown words.

\begin{figure}
\centerline{\psfig{figure=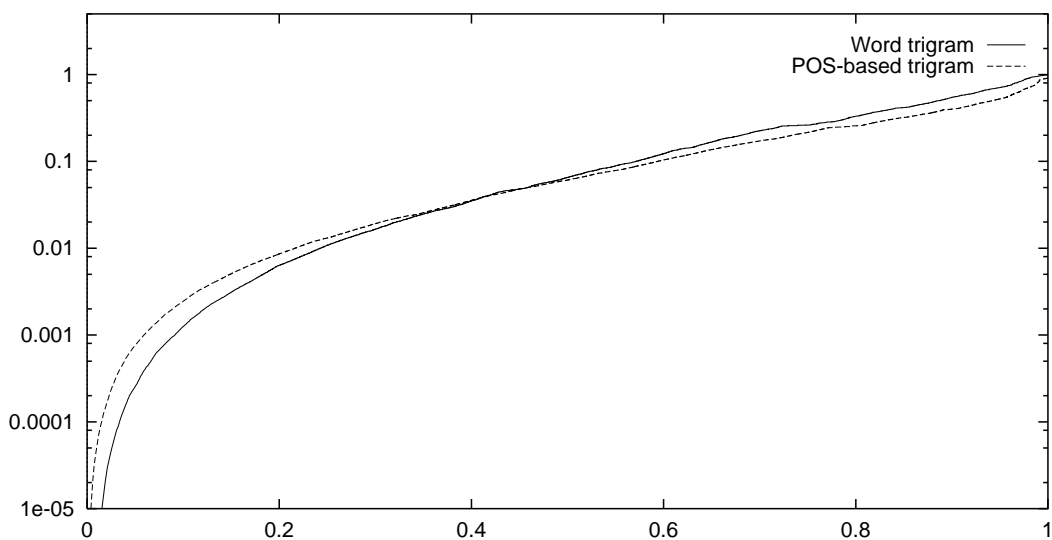,width=\textwidth}}
\caption{Cumulative Distribution of Word Probabilities}
\label{fig:model:perplexity}
\end{figure}

We next look at where the difference in perplexity is realized between
the word-based backoff model and the POS-based model.  In Figure~\ref
{fig:model:perplexity},
we give the distributions of probabilities assigned to each word of
the test corpus.  The y-axis shows the probability assigned to a word,
and the x-axis shows the percentage of words that have at least that
probability.  From the figure, we see that the POS-based model better
estimates lower probability words, while the word-based model better
estimates higher probability words.\footnote {Bahl~\etal~\shortcite
{Bahl-etal89:tassp} found that their word-based decision tree approach
also better predicts lower probability words than a word-based model using
interpolated estimation.}  The cross-over point occurs at the 43\%
mark.  Our model does well enough on the lower 43\% of the words, in terms of
perplexity, to more than compensate for the better performance of the
word-based model on the higher 57\%.  One of the implications of this,
however, is that if the speech recognition word error rate is greater
than 43\%, the POS-based model might not result in a decrease in word
error rate because the speech recognizer might just be recognizing the
57\% of the words that the language model assigns the high
probability.  Recent speech recognition error results for spontaneous
speech are now starting to fall below this rate (e.g.~\cite
{Zeppenfeld-etal97:icassp}).

In comparing the backoff and decision tree approaches, we need to
discuss the effect of unknown words.  We have already mentioned that
the POS-based decision tree model better predicts lower probability
words than the backoff approach.  This is because it first predicts
the POS tag based both on the POS tags and word identities of the
previous words.  Having this extra step allows the model to generalize
over syntactic categories and hence is not as affected by sparseness
of data.  Unknown words are definitely affected by sparseness of data.
In the test corpus, there are 356 unknown words.  The perplexity
improvement of the POS-based model is partially attributable to better
predicting the occurrence of unknown words.  Note that in our model,
prediction of unknown words is not as difficult since we only need to
predict them with respect to the POS tag.  Closed word categories,
such as determiners ({\bf DT}), are much less likely to have an
unknown word than open word categories, such as nouns and verbs.  Our
model also assigns more probability weight to the unknown words.  In
fact, the amount of weight assigned is in accordance with the
occurrence of singleton words: words that only occur once for a POS
tag in the training corpus.

To gauge the extent to which the improvement of our model is a result
of better handling of unknown words, we computed the perplexity
ignoring the probabilities assigned to unknown words; for the decision
tree model, we gave the unknown words a probability mass similar to
that used by the backoff model: assume a single
occurrence.\footnote{We are actually still giving the unknown words
too much weight, which adversely affects our results for this
comparison, but the difference is not significant.}  The results are
given in Table~\ref {tab:model:c:bo:word:known}.
\begin{table}
\begin{center}
\begin{tabular}{|l|r|r|r|} \hline
        & \multicolumn{1}{c|}{Backoff}
	& \multicolumn{2}{c|}{Decision Tree}  \\ \cline{3-4}
        & Word-Based & Word-Based & POS-Based \\ \hline \hline
Bigram  &      27.85 &      28.64 &     26.83 \\
Trigram &      24.78 &      25.14 &     23.78 \\ \hline
\end{tabular}
\end{center}
\caption{Comparison between Backoff and Decision Trees for Known Words}
\label{tab:model:c:bo:word:known}
\end{table}%
For the trigram model, this results in a perplexity of 24.78 for the
word-based model, and a perplexity of 23.78 for the POS-based model.
Thus the difference between the two models drops to an improvement of
4.0\%.  Which perplexity figure (with or without unknown words) better
predicts the speech recognition error rate is difficult to say, and
depends to a large extent on the acoustic modeling.  Acoustic models
that incorporate a {\em garbage} category, which is used when an
acoustic signal does not match any of the phonetic entries in the
dictionary, will undoubtedly benefit from our better modeling of
unknown words.  So far, such techniques have just been used in
key-word spotting (e.g.~\cite{Junkawitsch-etal96:icslp}).

We now compare the word-based backoff model to the word-based decision
tree model.  As can be seen in Table~\ref {tab:model:c:bo:word:known},
excluding the unknown words results in the backoff model doing better
than the decision tree word-based model, even though the decision tree
approach can generalize over words as a result of its use of a word
classification tree. For the bigram version, the backoff approach
achieves a perplexity reduction of 2.8\% in comparison to the decision
tree approach and 1.5\% for the trigram versions.  These results are
contrary to the improvement reported by
Bahl~\etal~\shortcite{Bahl-etal89:tassp} (reviewed in Section~\ref
{sec:related:dt:results}).  However, our comparison involved matching
the amount of context that both approaches have access to and involves
a much smaller corpus size.  Hence, there might not be enough data to
adequately grow a word classification tree (without using POS
information) that can compete with the simpler backoff approach.

As the last point in our comparison, we address the size of the
language models. The decision trees for the POS tags and word
probabilities (for the trigram model) have in total about 4300 leaf
nodes, and the word-based trigram backoff model has approximately 9100
distinct contexts for the trigrams.  Of course, each of the contexts
of the word-based trigram backoff has many zero entries (which are
thus predicted based on bigram counts).  In fact, there are on average
less than three non-zero trigrams for each distinct context.  This is
not the case for the decision trees, in which every possible value for
a leaf is given a value.  Hence, in terms of overall size, the backoff
model is more concise; however, this is an area that has not been
explored for decision trees.

\subsection{Class-Based Backoff Model}
\label{sec:model:c:bo:class}

We now compare our model with a class-based approach.  Class-based
approaches offer the advantage of being able to generalize over
similar words.  This generalization happens in two ways.  First, the
equivalence classes of the context are in terms of the classes that
were found.  Second, the probability of a word is assumed to be simply
the probability that that word occurs as the class.  Hence, the class
for the word we are predicting completely captures the effect of the
preceding context.  For instance, in the Trains corpus, the names of
the towns---Avon, Bath, Corning, Dansville, and Elmira---could be
grouped into a class, without much loss in information, but with an
increase in the amount of generalization.

The equations used for a class-based model are the following.
\begin{eqnarray}
\Pr(W_i|W_\rim) &\approx& \Pr(W_i|g(W_i))\Pr(g(W_i)|g(W_{i-1}) g(W_{i-2}))
\end{eqnarray}
We use the CMU toolkit to build a trigram backoff model on the classes,
and then multiply in the probability of the word given the class.

We use the clustering algorithm of Brown~\etal~\shortcite
{Brown-etal92:cl} to build the set of classes for the words (just as
we do for {\em Ignore POS} in Section~\ref {sec:model:r:ctrees}).
There are two factors that need to be considered.  First, we need to
determine the number of words to initially group in the class of
low-occurring words.  We have found that the best results were arrived
at automatically grouping all singletons in this class.  Furthermore,
the class of singletons can be used for predicting the occurrence of
unknown words.  The second factor is the number of classes.  We choose
the number of classes that results in the optimal performance, just as
we did in Section~\ref {sec:model:c:dt:class}.
The best results occur at 550 classes, which gives a perplexity rate
of 25.85 for the trigram version.  However, if we remove the influence
of unknown words, the perplexity is 24.79, which is not significantly
different than the corresponding results for the word-based model
given in the previous section.  Hence, the only improvement is in
modeling unknown words, which resulted from using the singletons to
predict the unknown words.  The lack of improvement in modeling known
words is somewhat surprising.  This probably points to how difficult
it is to construct worthwhile classes, especially from small corpora,
a fact borne out by our experiments with the {\em Ignore POS} method
in Section~\ref {sec:model:r:ctrees}.  These results might also
explain the poor performance of the word-based decision tree model,
which relies on the {\em Ignore POS} method to build the word
hierarchy.

\section{Conclusion}

In this chapter, we have presented a POS-based language model.  Unlike
previous approaches that use POS tags in language modeling, we
redefine the speech recognition problem so that it includes finding
the best word sequence and best POS tag interpretation for those
words.  In order to make use of the POS tags, we use a decision tree
algorithm to learn the probability distributions, and a clustering
algorithm to build hierarchical partitionings of the POS tags and the
word identities.  Furthermore, we take advantage of the POS tags in
building the word classification trees and in estimating the word
probabilities, which both results in better performance, and
significantly speeds up the training procedure.  We find that using
the rich context that using a decision tree allows results in a
perplexity reduction of 44.4\%.  We also find that the POS-based model
gives a 4.2\% reduction in perplexity over a class-based model, also
built with the decision tree and clustering algorithms.  In comparison
to using a backoff approach, it gives a 7.0\% reduction in perplexity
over a class-based model using a backoff approach, and a 4.1\%
reduction when unknown words are not considered.  Hence, using a
POS-based model results in an improved language model as well as
accomplishes the first part of the task in linguistic understanding.

The decision tree algorithm still needs work.  As shown in
Section~\ref {sec:model:r:constraints}, we find we must add
constraints to help it find better probability estimates that it
cannot learn on its own.  We also find that in comparing the
word-based decision tree version with a backoff version, the decision
tree version only does better when we include in its better handling
of unknown words.  This might be a result of inadequate data from
which to build the word classification trees.  Luckily, with the
POS-based approach, we can make use of the hand-labeled POS tags to
ease the problem of constructing the word classification tree, as
explained in Section~\ref {sec:model:ctrees}.

There are a number of issues that we have not explored in this
chapter.  First, we have not compared the decision tree approach with
directly using interpolated estimation.  We have also not explored
issues relating to size of training data.  The Trains corpus has less
than 60,000 words of data, whereas many other corpora used for speech
recognition have been much larger, typically containing at least a
million words of data.  However, these larger corpora have sometimes
been text corpora, or read-speech corpora.  One of the few exceptions
is the Switchboard corpus; however, as we noted in the beginning of
Chapter~\ref {chapter:corpus}, this corpus is not task-oriented, nor
is the domain limited.  Thus, it is of less interest to those
interested in building a spoken dialog system.

\cleardoublepage
\chapter{Detecting Speech Repairs and Boundary Tones}
\label{chapter:detection}

In order to understand spontaneous speech, one needs to eventually
segment the speech into utterance units, or intonational units, and
resolve the speech repairs.  As we argued in Section~\ref
{sec:intro:interactions} and \ref {sec:intro:tagging}, this can only
be done by a model that accounts for the interactions between these
two tasks, as well as the tasks of identifying discourse markers, POS
tagging, and the speech recognition task of predicting the next word.

In the previous chapter, we presented a POS-based language model that
uses special tags to denote discourse markers.  However, this model
does not take into account the occurrence of speech repairs and
intonational phrase boundaries.  Ignoring these events when building a
statistical language will lead to probabilistic estimates for the
words and POS tags that are less precise, since they mix contexts that
cross boundary tones and interruption points of speech repairs with
{\em fluent} stretches of speech.  Consider the following example of a
speech repair.
\begin{example}{d93-13.1 utt90} 
\label{ex:d93-13.1:utt90} I
can run trains \reparandum{on the}\ip\alteration{in the} opposite
direction 
\end{example} 
Here we have a preposition following a determiner, an event that only
happens across the interruption point of a speech repair. Now consider
the following example of a mid-turn boundary tone.
\begin{example}{d93-18.1 utt58} 
\label{ex:d93-18.1:utt58}
so let's see {\bf \%} \\
what time do we have to get done by {\bf \%} \\
by two p.m. {\bf \%}
\end{example} 
After asking the question ``what time do we have to get done by'', the
speaker refines the question to be whether they have to be done by a
certain time---``by two p.m.''  The result, however, is that there is
a repetition of the words ``by'', but separated by a boundary tone.
If such examples are included in the training data for a statistical
language model, then the model might incorrectly use this pattern to
label the POS tags in ambiguous fluent speech, or in predicting the
next word of a fluent stretch of speech.  However, if we exclude
examples with speech repairs and boundary tones from the training
data, then we will be unable to properly assign the POS tags across
the interruption point of speech repairs and across boundary tones,
nor will be able to recognize the words involved.  Thus, a language
model for spontaneous speech must take into account the occurrence of
speech repairs and boundary tones.

Given that a language model must account for the occurrence of speech
repairs and boundary tones, we are faced with the problem that there
is not a reliable signal for detecting the interruption point of
speech repairs \cite {Bear-etal92:acl} nor the occurrence of boundary
tones.  Rather, there are a number of different sources of information
that give evidence as to the occurrence of these events.  These
sources include the presence of pauses, filled pauses, cue phrases,
discourse markers, word fragments, word correspondences, syntactic
anomalies, as well as other acoustic cues in addition to pauses.
Table~\ref {tab:detection:cues} gives the number of occurrences for
some of these features in the Trains corpus.
\begin{table}
\begin{center}
\newlength{\xxa}
\settowidth{\xxa}{Boundary}
\begin{tabular} {|l|r|r|r|r|r|} \hline
        & \makebox[\xxa][c]{Fluent}&\makebox[\xxa][c]{Abridged}&\mc{Modification}&\makebox[\xxa][c]{Fresh} &\makebox[\xxa][c]{Boundary}\\
Feature & \mc{Speech}&\mc{Repairs} &\mc{Repairs}& \mc{Starts}&\mc{Tones}\\ \hline
all & 43439 & 423 & 1301 & 671 & 5211 \\
fragments & 7 & 0 & 481 & 150 & 0 \\
filled pauses & 97 & 374 & 114 & 71 & 358 \\
short pauses & 4415 & 146 & 711 & 313 & 1710 \\
long pauses & 1537 & 121 & 176 & 186 & 1622 \\
cue phrases & 0 & 49 & 72 & 166 & 0 \\
matching (2) & 2629 & 27 & 869 & 197 & 373 \\
matching (5) & 11479 & 94 & 1517 & 575 & 1375 \\
\hline
\end{tabular}
\end{center}
\caption{Occurrence of Features that Signal Speech Repairs and
Boundary Tones}
\label{tab:detection:cues}
\end{table}
For each word in the corpus that is not turn-final nor part of the
editing term of a speech repair, we report the occurrence of these
features in terms of whether the word has a boundary tone marked on
it, or it is marked with the interruption point of an abridged repair,
a modification repair, or a fresh start.  All other words are
categorized as {\em fluent}.\footnote {Where an interruption point
also has a boundary tone, we count it as an interruption point.  Also,
one of the modification repairs reported in Table~\ref
{tab:corpus:occurrences} was marked on the last word of a turn, and
hence is not included in Table~\ref{tab:detection:cues}.  The seven
word fragments that were not marked as the interruption point of a
speech repair in Table~\ref{tab:detection:cues} are problematic cases
and need to be reviewed.}

The first row, labeled ``all'', gives the number of occurrences of
these speech repairs, (turn-internal) boundary tones, and fluent
speech.  The second row gives the number of occurrences in which the
word is a fragment.  From the table, we see that 481 of the 1301
modification repairs has a reparandum that ends in a word fragment,
whereas 150 of the 671 reparanda of fresh starts end with a fragment.
The row labeled ``filled pauses'' reports on the number of words that
are followed by a filled pause.  As explained in Section~\ref
{sec:intro:classification}, not all filled pauses are viewed as part
of the editing term of a speech repair.  In fact, we see that 455 of
the 1014 filled pauses (which follow a non-editing term word) are not
part of an editing term of a speech repair, and hence are viewed as
part of the sentential content.  The next row, ``short pauses'',
reports on the words that are followed by a {\em short pause}, a pause
that is less than 0.5 seconds in duration.  The next row reports on
the number of pauses that are at least 0.5 seconds.  Pause durations
were computed automatically with a speech recognizer constrained to
the word transcription \cite {Entropic94:aligner}.  The next row,
``cue phrases'', reports on the number of words that are followed by
an editing term starting with a cue phrase (rather than a filled
pause).  Note that this row only applies to words that are marked with
the interruption point of a speech repair.  The row labeled ``matching
(2)'' reports on the number of times that there is a word matching
that crosses the word under consideration with at most 2 intervening
words between the match.  The first word of the match is allowed to be
on the current word.  For speech repairs, this shows how often the
interruption point has word matches that cross it.  The next row,
``matching (5)'', is for those with at most 5 intervening words.

From the table, it is clear that none of the cues on their own are a
reliable indicator of speech repairs or boundary tones.  For instance,
14.5\% of all pauses occur after the interruption point of a speech
repair and 32.9\% occur after a boundary tone.  Conversely, 69.0\% of
all repairs are followed by a pause while 71.9\% of all boundary tones
are followed by a pause.\footnote {Blackmer and Mitton \shortcite
{BlackmerMitton91:cog} also found that pauses are not obligatory for
speech repairs, even when the alteration does not simply repeat the
reparandum.} Hence, pauses alone do not give a complete picture as to
whether a speech repair or a boundary tone occurred.  The same holds
for filled pauses, which can occur both after the interruption point
of a speech repair and in non-speech repair contexts, namely between
utterances or after utterance-initial discourse markers.  Even word
matchings can be spurious, as evidenced by the 27 occurrences of
matching with at most two intervening words within the vicinity of an
abridged repair, as well as the matchings across boundary tones and
fluent speech.  As for cue phrases, although we see that they only
follow the interruption point of speech repairs, this is solely due to
our definition of cue phrases.  One still faces the problem of
deciding if a word, or a sequence of words, is being used as a cue
phrase, as explained in Section~\ref {sec:intro:repairs_dm} with
Example~\ref{ex:d92a-4.2:utt62}, repeated below.
\begin{example}{d92a-4.2 utt62}
\label{ex:d92a-4.2:utt62:a}
\reparandum{I don't know if the}\ip \et{okay} that'll be three hours right
\end{example}

Even syntactic ill-formedness at the interruption point is not always
guaranteed, as the following example illustrates.
\begin{example}{d93-13.2 utt53} 
\label{ex:d93-13.2:utt53}
load two \reparandum{boxes of}\ip boxcars with oranges 
\end{example} 
In the example above, it is only after domain knowledge is applied
that ``boxes of boxcars'' would be found to be ill-formed.  Hence
using parser failures to find speech repairs, as is done by Dowding
\etal~\shortcite {Dowding-etal93:acl} and Sagawa \etal~\shortcite
{Sagawa-etal94:coling}, will not be robust.  Syntactic irregularities
are just one source of information that is available for detecting
speech repairs and identifying boundary tones.

In this chapter, we augment our POS-based language model so that it
can also detect intonational phrase boundary tones and speech repairs,
along with their editing terms.\footnote {The material presented in
this chapter, Chapter~\ref {chapter:correction} and Chapter~\ref
{chapter:acoustic} is a slight improvement over the results given in
Heeman and Allen \shortcite {HeemanAllen97:acl}.}  Although not all
speech repairs have obvious syntactic anomalies, as displayed in
Example~\ref {ex:d93-13.1:utt90}, the probability distributions for
words and POS tags are going to be different depending on whether they
follow the interruption point of a speech repair, a boundary tone, or
fluent speech.  So, it makes sense to take the speech repairs and
boundary tones into account by directly modeling them when building
the language model, which automatically gives us a means of detecting
these events and better predicting the speech that follows.

\begin{figure}[bth]
\begin{center}
\setlength{\unitlength}{0.00083300in}%
\begingroup\makeatletter\ifx\SetFigFont\undefined
\def\x#1#2#3#4#5#6#7\relax{\def\x{#1#2#3#4#5#6}}%
\expandafter\x\fmtname xxxxxx\relax \def\y{splain}%
\ifx\x\y   
\gdef\SetFigFont#1#2#3{%
  \ifnum #1<17\tiny\else \ifnum #1<20\small\else
  \ifnum #1<24\normalsize\else \ifnum #1<29\large\else
  \ifnum #1<34\Large\else \ifnum #1<41\LARGE\else
     \huge\fi\fi\fi\fi\fi\fi
  \csname #3\endcsname}%
\else
\gdef\SetFigFont#1#2#3{\begingroup
  \count@#1\relax \ifnum 25<\count@\count@25\fi
  \def\x{\endgroup\@setsize\SetFigFont{#2pt}}%
  \expandafter\x
    \csname \romannumeral\the\count@ pt\expandafter\endcsname
    \csname @\romannumeral\the\count@ pt\endcsname
  \csname #3\endcsname}%
\fi
\fi\endgroup
\begin{picture}(4524,848)(139,-3678)
\thicklines
\put(901,-3361){\circle{600}}
\put(3901,-3361){\circle{618}}
\put(2401,-3361){\circle{526}}
\put(151,-3361){\vector( 1, 0){375}}
\put(1276,-3361){\vector( 1, 0){780}}
\put(2738,-3361){\vector( 1, 0){788}}
\put(4276,-3361){\vector( 1, 0){375}}
\put(901,-2986){\makebox(0,0)[b]{\smash{\SetFigFont{12}{14.4}{rm}$W_{i-1}$}}}
\put(3901,-2986){\makebox(0,0)[b]{\smash{\SetFigFont{12}{14.4}{rm}$W_i$}}}
\put(2401,-2986){\makebox(0,0)[b]{\smash{\SetFigFont{11}{13.2}{rm}null}}}
\put(901,-3421){\makebox(0,0)[b]{\smash{\SetFigFont{12}{14.4}{rm}$P_{i-1}$}}}
\put(2401,-3421){\makebox(0,0)[b]{\smash{\SetFigFont{12}{14.4}{rm}$U_i$}}}
\put(3901,-3421){\makebox(0,0)[b]{\smash{\SetFigFont{12}{14.4}{rm}$P_i$}}}
\end{picture}
\end{center}
\caption{Tagging Null Tokens with an Utterance Tag}
\label{fig:simple}
\end{figure}
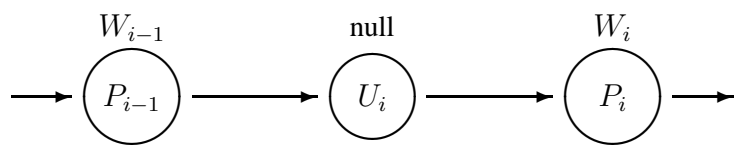

To model the occurrence of boundary tones and speech repairs, we
introduce extra variables into the language model, which will be
tagged as to the occurrence of boundary tones and speech repairs.
Another way of viewing this is that we are introducing a null token
between each pair of consecutive words, which will be assigned a POS
tag, just like any other word \cite{HeemanAllen94:arpa}.  For these
null tokens, we restrict the types of tags that it can assign to a
small set of {\em utterance} tags.  These tags will indicate whether
an intonational boundary tone, an interruption point of a speech
repair, or an editing term sequence has occurred.  Hence these tags
capture the discontinuities in the speaker's turn, and use these
discontinuities to better model the speech that follows.  If we let
variable $U_i$ denote the utterance-level tag between word $W_\im$ and
$W_i$, we can picture the language modeling task as illustrated in
Figure~\ref{fig:simple}.

\section{Splitting the Utterance Tag}

The utterance tag needs to capture the occurrence of three different
types of events: intonational boundary tones, the interruption point
of speech repairs, and the presence of editing terms, which sometimes
accompany speech repairs.  Hence, we split the utterance tag into
three separate ones, one for each type of these events.  The {\em
repair\/} tag $R_i$ models the occurrence of speech repairs; the {\em
editing term\/} tag $E_i$ models the occurrence of editing terms; and
the {\em tone\/} tag $T_i$ models the occurrence of boundary tones.

\subsection{Speech Repairs}
\label{sec:detection:R}

The speech repair tag indicates the occurrence of speech repairs.
However, we not only want to know whether a repair occurred, but also
the type of repair: whether it is a modification repair, fresh start,
or an abridged repair.  The type of repair is important since the
strategy that a hearer uses to correct the repair depends on the type
of repair.  For fresh starts, the hearer must determine the beginning
of the current utterance, and use this in deciding the onset of the
reparandum. For modification repairs, the hearer can make use of the
repair structure, the parallel structure that often exists between the
reparandum and alteration, to determine the extent of the
reparandum. For abridged repairs, there is no reparandum, and so
simply knowing that it is abridged automatically gives the correction.

We now address how the occurrence of a repair is modeled by the repair
tag.  For speech repairs that do not have an editing term, the
interruption point is where the local context is disrupted, and hence
is the logical spot to tag such repairs.  Below, we repeat
Example~\ref {ex:d93-13.2:utt53}, with the repair tag marked between
the appropriate words.
\begin{example}{d93-13.2 utt53}
load two boxes of {\bf Mod} boxcars with oranges
\end{example}
Hence the repair is marked on the transition between $W_{i-1}$ and
$W_i$ where $W_\im$ is the end of the reparandum and $W_i$ is the
onset of the alteration.\footnote{As mentioned in Footnote~\ref
{ft:alteration} in Section~\ref {sec:ft:alteration}, for repairs
without an obvious alteration, we define the alteration as the first
non-editing term word after the interruption point.}

For speech repairs that have an editing term, there are two choices
for marking the speech repair: either directly following the end of
the reparandum, or directly preceding the onset of the alteration.
The following example illustrates these two choices, marking them with
{\bf Mod?}.
\begin{example}{d92a-5.2 utt34}
\label{ex:d92a-5.2:utt34}
so we'll pick up a tank of {\bf Mod?} uh {\bf Mod?} the
tanker of oranges
\end{example}
Several researchers \cite{Levelt83:cog,Hindle83:acl} have noted that
the editing term gives evidence as to the type of repair.  Hence,
using the latter alternative means that we will have the editing term
available as part of the context in determining the repair tag.
Furthermore, the latter alternative might be a better psychological
model, since it does not claim that the speaker decides which type of
repair (abridged, fresh start or modification repair) she is going to
make before she utters the editing term.

The above leads to the following definition of the repair variable
$R_i$ for the transition between word $W_\im$ and $W_i$.
\[ R_i = \left\{ \begin{array}{ll}
{\bf Mod}  & \parbox[t]{4in}{\setlength{\baselineskip}{1.3em}
		if $W_i$ is the alteration onset of a 
		modification repair} \\ 
{\bf Can}  & \parbox[t]{4in}{\setlength{\baselineskip}{1.3em}
		if $W_i$ is the alteration onset of a 
		fresh start (or {\em cancel})} \\
{\bf Abr}  & \parbox[t]{4in}{\setlength{\baselineskip}{1.3em}
		if $W_i$ is the alteration onset of an abridged repair} \\
{\bf null} & \mbox{otherwise}
\end{array} \right. \]

\subsection{Editing Terms}
\label{sec:et_tag}

Editing terms are problematic for tagging speech repairs since they
separate the end of the reparandum from the onset of the alteration,
thus separating the discontinuity that gives evidence that a fresh
start or modification repair occurred.  For abridged repairs, they
separate the word that follows the editing term from the context that
is needed to determine the identity of the word and its POS tag.

If editing terms could be identified without having to consider their
context, we could simply skip over them, but still use them as part of
the context for deciding the repair tag.  In fact this was a
simplifying assumption that we made in earlier work \cite
{HeemanAllen94:acl}.  However, this assumption is not valid for words
that are ambiguous as to whether they are an editing term, such as
``let me see''.  Even filled pauses are problematic since they are not
necessarily part of the editing term of a speech repair.  So we need
to model the occurrence of editing terms, and to use the editing term
tags in deciding the repair tags.

To model editing terms, we use the variable $E_i$ to indicate the type
of editing term transition between word $W_\im$ and $W_i$.
\[ E_i = \left\{ \begin{array}{ll}
{\bf Push}  & \parbox[t]{4in}{if $W_\im$ is not part of an editing term but $W_i$ is} \\
{\bf ET}    & \parbox[t]{4in}{if $W_\im$ and $W_i$ are both part of
an editing term} \\
{\bf Pop}   & \parbox[t]{4in}{if $W_\im$ is part of an editing term but $W_i$ is not} \\
{\bf null}  & \parbox[t]{4in}{if neither $W_\im$ nor $W_i$ are part
of an editing term}
\end{array}
\right. \]

The editing term tags are by no means independent of the repair
tags.  Below, we list the dependencies between these two tags.
\begin{enumerate}
\item if $E_i = {\bf Pop}$ then $R_i \in \{{\bf Mod},{\bf Can},{\bf Abr}\}$
\item if $E_i \in \{{\bf ET},{\bf Push}\}$ then $R_i = {\bf null}$
\item if $E_i = {\bf null}$ then $R_i \in \{{\bf null},{\bf Mod},{\bf Can}\}$
\end{enumerate}
Although the repair and editing term variables are dependent on each
other, using a single variable to model both is not advantageous.  In
the case where the previous word is part of an editing term, we need
to determine whether to end the editing term, and we need to determine
the type of repair.  However, the relevant context for these two
decisions is different.  Deciding if the next word continues an
editing term depends to a large extent on what the previous editing
term words are.  For instance, if the previous two words have been tagged
as editing terms and are the words ``let'' and ``us'', then the next
word will probably continue the editing term.  In the case of deciding
the repair tag, this will not only depend on the editing term that has
been seen, but also on the words before the editing term.  Speakers
for instance might be more likely to make an abridged repair following
a determiner than following a noun.  Because of the different
information that is relevant, it should require less training data to
train a model that has separate variables for editing terms and speech
repairs than one that just uses a single variable.

Below, we give an example of a speech repair with a multi-word editing term, and we show all non-null editing term and repair
tags.
\begin{example}{d93-10.4 utt30}
\label{ex:d93-10.4:utt30}
that'll get there at four a.m. {\bf Push} oh {\bf ET} sorry {\bf Pop}
{\bf Mod} at eleven a.m.
\end{example}

\subsection{Boundary Tones}

The last variable we introduce is for modeling intonational phrase
boundaries, or boundary tones.  Detecting boundary tones not
only gives us better probability estimates for the language model, but
will also be used in Chapter~\ref {chapter:correction} to help
determine the extent of the reparandum of a fresh start, which often
starts after the previous boundary tone.  We use the variable $T_i$ to
mark the occurrence of a boundary tone on the previous word.
\[ T_i = \left\{ \begin{array}{ll}
{\bf Tone} & \parbox[t]{4in}{if $W_\im$ has a boundary tone} \\
{\bf null} & \parbox[t]{4in}{if $W_\im$ does not have a boundary tone}
\end{array} \right. \]

The tone variable is separate from the editing term and repair
variables since this variable is not restricted by the value of the
other two.  For instance, an editing term could have a boundary tone,
especially on the end of a cue phrase such as ``let's see''.  In
addition, the end of the reparandum (the interruption point) could
also be marked as an intonational phrase ending, as
Example~\ref {ex:d92a-2.1:utt29:a} below demonstrates.

Below we give an example (Example~\ref{ex:d92a-2.1:utt29} given
earlier) showing all non-null tone, editing term and repair tags
marked. In this example, the end of the intonational phrase also
corresponds to the interruption point of a modification repair with an
editing term.
\begin{example}{d92a-2.1 utt29}
\label{ex:d92a-2.1:utt29:a}
that's the one with the bananas {\bf Tone} {\bf Push} I {\bf ET} mean
{\bf Pop} {\bf Mod} that's taking the bananas
\end{example}

\section{Redefining the Speech Recognition Problem}
\label{sec:detection:model}

Now that we have introduced the tone, editing term, and repair
variables, we redefine the speech recognition problem.  Similar to
Section~\ref {sec:model:tagger}, we redefine the problem so that the
goal is to find the sequence of words and the corresponding POS, tone,
editing term and repair tags that is most probable given the acoustic
signal.
\newlength{\argmaxd}
\settowidth{\argmaxd}{$\arg\max$}
\begin{eqnarray}
\hat{W}\hat{P}\hat{R}\hat{E}\hat{T} 
&=& \parbox[t]{\argmaxd}{$\arg\max$ 
		\footnotesize \vspace*{-2.4em} \\ $WPRET$ \vspace*{-0.8em}}\
    \Pr(WPRET|A) \\
&=& \parbox[t]{\argmaxd}{$\arg\max$ 
		\footnotesize \vspace*{-2.4em} \\ $WPRET$ \vspace*{-0.8em}}\
    \frac{\Pr(A|WPRET) \Pr(WPRET)}{\Pr(A)} \\
&=& \parbox[t]{\argmaxd}{$\arg\max$ \footnotesize \vspace*{-2.4em} \\ $WPRET$}
    \Pr(A|WPRET) \Pr(WPRET) \label{eqn:detection:model}
\end{eqnarray} 
Again, the first term of Equation~\ref {eqn:detection:model} is the
acoustic model, and the second term is the language model.  We can
rewrite the language model term as
\[\Pr(W_{1,N}P_{1,N}R_{1,N}E_{1,N}T_{1,N})\]
where $N$ is the number of words in the sequence.  We now rewrite this
term as the following.
\begin{eqnarray}
\lefteqn{\Pr(W_{1,N}P_{1,N}R_{1,N}E_{1,N}T_{1,N})} \nonumber \\
&=&\prod_{i=1}^N\Pr(W_iP_iR_iE_iT_i|W_{\rim}P_{\rim}R_{\rim}E_{\rim}T_{\rim})\\
&=&\prod_{i=1}^N\Pr(T_i|W_{\rim}P_{\rim}R_{\rim}E_{\rim}T_{\rim}) \nonumber\\
  &&\hspace{1.5em}\Pr(E_i|W_{\rim}P_{\rim}R_{\rim}E_{\rim}T_{\ri}) \nonumber\\
  &&\hspace{1.5em}\Pr(R_i|W_{\rim}P_{\rim}R_{\rim}E_{\ri}T_{\ri}) \nonumber\\
  &&\hspace{1.5em}\Pr(P_i|W_{\rim}P_{\rim}R_{\ri}E_{\ri}T_{\ri}) \nonumber\\
  &&\hspace{1.5em}\Pr(W_i|W_{\rim}P_{\ri}R_{\ri}E_{\ri}T_{\ri}) \label{eqn:detection:model2}
\end{eqnarray}

As can be seen in the last line of the derivation, we have chosen the
order of separating the utterance tags so that the following hold.
\begin{enumerate}
\item $T_i$ depends only on the previous context 
\item $E_i$ depends on the previous context and $T_i$.
\item $R_i$ depends on the previous context and $T_i$ and $E_i$.
\item $P_i$ depends on the previous context and $T_i$, $E_i$ and $R_i$.
\item $W_i$ depends on the previous context and $T_i$, $E_i$, $R_i$ and $P_i$.
\end{enumerate}
Although any choice would be correct in terms of probability theory,
we are constrained by a sparsity of data in estimating the
distributions.  Hence, we choose the ordering that seems as
psycholingistically appealing as possible.  Speakers probably choose
whether to end a word with an intonational boundary before deciding
they need to revise what they just said.  Editing terms are often
viewed as a stalling technique, perhaps even to stall while deciding
the type of repair.  Furthermore, since we separate the editing terms
from repairs by using two separate tags, it makes sense to decide
whether to end the editing term and then decide on the type of repair,
since otherwise deciding the repair tag would automatically give the
editing term tag.

\begin{figure}[tbh]
\begin{center}
\setlength{\unitlength}{0.00083300in}%
\begingroup\makeatletter\ifx\SetFigFont\undefined
\def\x#1#2#3#4#5#6#7\relax{\def\x{#1#2#3#4#5#6}}%
\expandafter\x\fmtname xxxxxx\relax \def\y{splain}%
\ifx\x\y   
\gdef\SetFigFont#1#2#3{%
  \ifnum #1<17\tiny\else \ifnum #1<20\small\else
  \ifnum #1<24\normalsize\else \ifnum #1<29\large\else
  \ifnum #1<34\Large\else \ifnum #1<41\LARGE\else
     \huge\fi\fi\fi\fi\fi\fi
  \csname #3\endcsname}%
\else
\gdef\SetFigFont#1#2#3{\begingroup
  \count@#1\relax \ifnum 25<\count@\count@25\fi
  \def\x{\endgroup\@setsize\SetFigFont{#2pt}}%
  \expandafter\x
    \csname \romannumeral\the\count@ pt\expandafter\endcsname
    \csname @\romannumeral\the\count@ pt\endcsname
  \csname #3\endcsname}%
\fi
\fi\endgroup
\begin{picture}(5428,863)(139,-3692)
\thicklines
\put(1954,-3366){\circle{530}}
\put(1954,-3418){\makebox(0,0)[b]{\smash{\SetFigFont{12}{14.4}{rm}$T_i$}}}
\put(2846,-3366){\circle{530}}
\put(2843,-3424){\makebox(0,0)[b]{\smash{\SetFigFont{12}{14.4}{rm}$E_i$}}}
\put(3753,-3359){\circle{530}}
\put(3750,-3424){\makebox(0,0)[b]{\smash{\SetFigFont{12}{14.4}{rm}$R_i$}}}
\put(4810,-3375){\circle{618}}
\put(5180,-3370){\vector( 1, 0){375}}
\put(4795,-2985){\makebox(0,0)[b]{\smash{\SetFigFont{12}{14.4}{rm}$W_i$}}}
\put(4800,-3425){\makebox(0,0)[b]{\smash{\SetFigFont{12}{14.4}{rm}$P_i$}}}
\put(901,-3361){\circle{600}}
\put(151,-3361){\vector( 1, 0){375}}
\put(1276,-3361){\vector( 1, 0){375}}
\put(2259,-3361){\vector( 1, 0){292}}
\put(3159,-3361){\vector( 1, 0){292}}
\put(4059,-3361){\vector( 1, 0){367}}
\put(901,-2986){\makebox(0,0)[b]{\smash{\SetFigFont{12}{14.4}{rm}$W_{i-1}$}}}
\put(901,-3414){\makebox(0,0)[b]{\smash{\SetFigFont{12}{14.4}{rm}$P_{i-1}$}}}
\end{picture}
\end{center}
\caption{Tagging Null Tokens with Tone, Editing Term, and Repair Tags}
\label{fig:RET}
\end{figure}
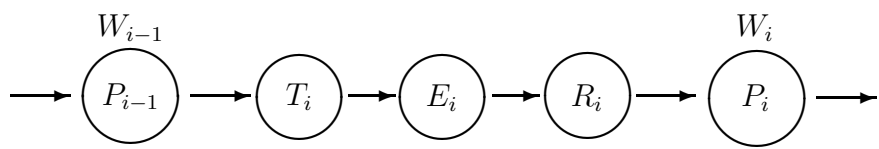

The order in which the probabilities were expanded lets us view the
speech recognition problem as illustrated in Figure~\ref{fig:RET}.
Here the language model, in addition to recognizing the words and
assigning them a POS tag, must also assign tags to the three null
words, a tone tag, an editing term tag, and a repair tag.
Note that even though the tone, editing term, and repair tags for word
$W_i$ do not directly depend on the word $W_i$ or POS tag $P_i$, the
probabilities $W_i$ and $P_i$ do depend on the tone, editing term and
repair tags for the current word as well as on the previous context.
So, the probability of these utterance tags will (indirectly) depend both on
the following word and its POS tag.

\section{Discontinuities in the Context}

Equation \ref{eqn:detection:model2} involves five probability
distributions that need to be estimated.  The context for each
includes all of the previous context, as well as the variables of the
current word that have already been predicted.  As is typically done
with language modeling, questions are asked relative to the current
word.  In other words, the decision tree algorithm can ask about the
value that has been assigned to a variable for the current word, or
the previous word, etc., but it cannot ask what value has been
assigned to the first word in the turn.

In principal, we could give all of the context to the decision tree
algorithm and let it decide what information is relevant in
constructing equivalence classes of the contexts.  However, editing
terms, tones, and repairs introduce discontinuities into the context,
which current techniques for estimating probability distributions are
not sophisticated enough to handle.  This will prevent them from
making relevant generalizations, leading to unnecessary data
fragmentation.  But for these tags, we do not have the data to spare,
since repairs, editing terms, and even tones do not occur in the same
abundance as fluent speech and are not as constrained.  In the
following, we illustrate the problems that can lead to unnecessary
data fragmentation.

\subsection{After Abridged Repairs}
\label{sec:detection:context:abr}

For abridged repairs, editing terms can interfere with predicting the
word, and its POS tag, that follows the editing term.  Consider the
following two examples.
\begin{example}{d93-11.1 utt46}
so we need to get the three tankers
\end{example}
\begin{example}{d92a-2.2 utt6}
so we need to {\bf Push} um {\bf Pop} {\bf Abr} get a tanker of OJ to Avon
\end{example}
Here, both examples have the verb ``get'' following the words ``so we
need to'', with the only difference being that the second example has
an editing term in between.  For this example, once we know that the
repair is abridged, the editing term merely gets in the way of
predicting the word ``get'' (and its POS tag) for it prohibits the
decision tree algorithm from generalizing with non-abridged examples.
This would force it to estimate the probability of the verb based
solely on the abridged examples in the training corpus.  Of course,
there might be instances where it is best to differentiate based on
the presence of an editing term, but this should not be forced from
the onset.

\subsection{After Repairs with Editing Terms}
\label{sec:detection:context:et}

The prediction of the word, and its POS tag, after an abridged repair
are not the only examples that suffer from the discontinuity that
editing terms introduce.  Consider the next two examples of
modification repairs, differing by the presence of an editing term in
the second example.
\begin{example}{d93-23.1 utt25}
so it should get there at {\bf Mod} to Bath a little bit after five
\end{example}
\begin{example}{d92a-3.2 utt45}
\label{ex:d92a-3.2:utt45}
engine E three will be there at {\bf Push} uh {\bf Pop} {\bf Mod} in
three hours
\end{example}
Here, both examples have a preposition as the last word of the
reparandum, and the repair replaces this by another preposition.  For
the task of predicting the POS tag and the word identity of the onset
of the alteration, the presence of the editing term in the second
example should not prevent generalizations over these two examples.

Although we have focused on predicting the word (and its POS tag) that
follows the repair, the same argument also holds for even predicting
the repair.  The presence of an editing term and its identity are
certainly an important source in deciding if a repair occurred.  But
also of importance are the words that precede the editing term.  So,
we should be able to generalize over the words that precede the
interruption point, without regard to whether the repair has an
editing term.

\subsection{After Repairs and Boundary Tones}
\label{sec:detection:context:tones}

Speech repairs, even those without editing terms, and boundary tones
also introduce discontinuities in the context.  For instance, in the
following example, in predicting the word ``takes'' or its POS tag, it
is probably inappropriate to ask about the word ``picks'' if we
haven't yet asked whether there is a modification repair in between.
\begin{example}{d92-1 utt53}
engine E two picks {\bf Mod} takes the two boxcars
\end{example}
The same also holds for boundary tones.  In the example below, if the
word ``is'' is going to be used to provide context for later words, it
should only be in the realization that it ends an intonational phrase.
\begin{example}{d92a-1.2 utt3}
you'll have to tell me what the problem is {\bf Tone} 
I don't have their labels
\end{example}

Although the repair and tone tags are part of the context and so can
be used in partitioning it, the question is whether this will happen.
The problem is that null-tones and null-repairs dominate the training
examples.  So, we are bound to run into contexts in which there are
not enough tones and repairs for the decision tree algorithm to learn
the importance of using this information, and instead might blindly
subdivide the context based on some subdivision of the POS tags.  The
solution we propose is analogous to what is done in tagging written
text: view the repair and tone tags as words, rather than as extra
tags.  This way, it will be more difficult for the learning algorithm
to ignore these tags, and much easier for it to group these tags with
POS tags and words that behave in a similar way, such as change in
speaker turn, and discourse markers.

\section{Representing the Context}

As we discussed in the previous section, we need to be careful about
how we represent the context so as to allow relevant generalizations about
contexts that contain editing terms, repairs, and boundary tones.
Rather than supplying the full context to the decision tree algorithm
and letting it decide what information is relevant in constructing
equivalence classes of the contexts, we instead will be using the full
context to construct a more relevant set of variables for it to query.

\subsection{Utterance-Sensitive Word and POS Tags}
\label{sec:detection:cleanup}

We refer to the first set of variables that we use as the {\em
utterance-sensitive} Word and POS variables. These correspond to the
POS and word variables, but take into account the utterance
tags. First, as motivated in Section~\ref
{sec:detection:context:tones}, we insert the non-null tone and
modification and fresh start tags into the POS and word variables so
as to allow generalizations over tone and repair contexts and lexical
contexts that behave in a similar way, such as change in speaker turn,
and discourse markers. Second, as we argued in Section~\ref
{sec:detection:context:abr} and Section~\ref
{sec:detection:context:et}, in order to allow generalizations over
different editing term contexts, we need to make available a context
that cleans up completed editing terms.  Hence, when an editing term
is completed, as signaled by an editing term {\bf Pop}, we remove the
words involved in the editing term as well as the {\bf Push} tag.
Thus the utterance-sensitive word and POS tags give us a view of the
previous words and POS tags that accounts for the utterance tags that
have been hypothesized. This approach is similar to how Kompe
\etal~\shortcite {Kompe-etal94:icassp} insert boundary tones into the
context used by their language model and how Stolcke and Shriberg
\shortcite {StolckeShriberg96:icassp} clean up mid-utterance filled
pauses.

\begin{figure}
\begin{center}
\begin{picture}(0,0)%
\includegraphics{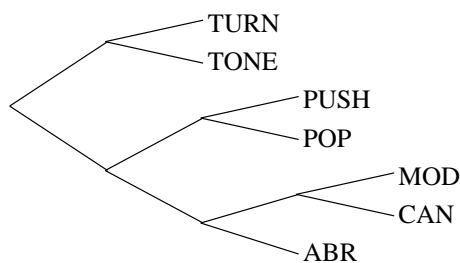}%
\end{picture}%
\setlength{\unitlength}{0.01250000in}%
\begingroup\makeatletter\ifx\SetFigFont\undefined
\def\x#1#2#3#4#5#6#7\relax{\def\x{#1#2#3#4#5#6}}%
\expandafter\x\fmtname xxxxxx\relax \def\y{splain}%
\ifx\x\y   
\gdef\SetFigFont#1#2#3{%
  \ifnum #1<17\tiny\else \ifnum #1<20\small\else
  \ifnum #1<24\normalsize\else \ifnum #1<29\large\else
  \ifnum #1<34\Large\else \ifnum #1<41\LARGE\else
     \huge\fi\fi\fi\fi\fi\fi
  \csname #3\endcsname}%
\else
\gdef\SetFigFont#1#2#3{\begingroup
  \count@#1\relax \ifnum 25<\count@\count@25\fi
  \def\x{\endgroup\@setsize\SetFigFont{#2pt}}%
  \expandafter\x
    \csname \romannumeral\the\count@ pt\expandafter\endcsname
    \csname @\romannumeral\the\count@ pt\endcsname
  \csname #3\endcsname}%
\fi
\fi\endgroup
\begin{picture}(165,112)(23,701)
\put(106,800){\makebox(0,0)[lb]{\smash{\SetFigFont{10}{12.0}{rm} TURN}}}
\put(106,784){\makebox(0,0)[lb]{\smash{\SetFigFont{10}{12.0}{rm} TONE}}}
\put(146,768){\makebox(0,0)[lb]{\smash{\SetFigFont{10}{12.0}{rm} PUSH}}}
\put(146,752){\makebox(0,0)[lb]{\smash{\SetFigFont{10}{12.0}{rm} POP}}}
\put(186,736){\makebox(0,0)[lb]{\smash{\SetFigFont{10}{12.0}{rm} MOD}}}
\put(186,720){\makebox(0,0)[lb]{\smash{\SetFigFont{10}{12.0}{rm} CAN}}}
\put(146,704){\makebox(0,0)[lb]{\smash{\SetFigFont{10}{12.0}{rm} ABR}}}
\end{picture}
\end{center}
\caption{Adding Extra Tags to the POS Classification Tree}
\label{fig:detection:turn}
\end{figure}

The above approach means that the utterance-sensitive word and POS
tags will have {\bf Tone}, {\bf Mod}, {\bf Can} and {\bf Push} tags
interspersed in them.  Hence, we treat these tags just as if they were
lexical items,\label{sec:detection:addtags} and associate a POS tag
with these tokens, which will simply be themselves. We have manually
added these new POS tags into the POS classification tree, grouping
them with the {\bf TURN} tag. Figure~\ref{fig:detection:turn} shows
the subtree that replaces the {\bf TURN} tag in the POS classification
tree that was given in Figure~\ref{fig:model:postree}.
%
%

To illustrate how the values of the utterance-sensitive word and
POS tags are determined, consider the following example.
\begin{example}{d93-18.1 utt47}
\label{ex:d93-18.1:utt47} 
it takes one {\bf Push} you {\bf ET} know {\bf Pop} {\bf Mod} two
hours {\bf Tone}
\end{example}
In predicting the POS tag for the word ``you'' given the correct
interpretation of the previous context, these variables will be set as
follows, where the utterance-sensitive word and POS tags are denoted
by {\em pW} and {\em pP}, and the top row indicates the indices.
\begin{center}
\begin{tabular}{|l|c|c|c|c|} \hline
           & i-4     & i-3     & i-2    & i-1      \\ \hline
{\em pP}   & \bf PRP & \bf VBP & \bf CD & \bf Push \\
{\em pW}   & it      & takes   & one    & \bf Push \\ \hline
\end{tabular}
\end{center}%
For predicting the word ``you'' given the correct interpretation of
the previous context, we also have access to its hypothesized POS tag,
as shown below.
\begin{center}
\begin{tabular}{|l|c|c|c|c|c|} \hline
           & i-4     & i-3     & i-2    & i-1      & i       \\ \hline
{\em pP}   & \bf PRP & \bf VBP & \bf CD & \bf Push & \bf PRP \\
{\em pW}   & it      & takes   & one    & \bf Push &         \\ \hline
\end{tabular}
\end{center}%

After we have finished hypothesizing the editing term, we will have
hypothesized a {\bf Pop} editing term tag, and then have hypothesized
a {\bf Mod} repair tag.  Since the {\bf Pop} causes the editing term
of ``you know'' to be cleaned up, as well as the {\bf Push}, the
resulting context for predicting the POS tag of the current word,
which is ``two'', will be as follows.\footnote{If a modification
repair or fresh start is proposed on the same word that a boundary
tone has been proposed on, only the speech repair is marked in the
utterance-sensitive words and POS tags.}
\begin{center}
\begin{tabular}{|l|c|c|c|c|} \hline
         & i-4   & i-3   & i-2  & i-1    \\ \hline
{\em pP} &\bf PRP&\bf VBP&\bf CD&\bf Mod \\
{\em pW} & it    & takes & one  &\bf Mod \\ \hline
\end{tabular}
\end{center}

The reader should note that the {\em pP} and {\em pW} variables are
actually only applicable for predicting the word and POS tag.\footnote
{The {\em p} prefix was chosen because these variables are used as the
context for the POS tags (as well as the words).}  We actually need
variations of these for predicting the tone, editing term, and repair
tag.  We define two additional sets.  The first, which we refer to as
{\em tP} and {\em tW}, capture the context before the {\em t\/}one,
editing term and repair tags of the current word are predicted.  The
second set, which we refer to as {\em rP} and {\em rW}, also capture
the context before the tone, editing term and repair tags, but also
before any editing term that we might be processing. Hence, the {\em
rP} and {\em rW} variables capture the words in the {\em
r\/}eparandum.

To show how the {\em tP}, {\em tW}, {\em rP}, and {\em rW} variables
are determined, we return to the example above.  Below we give the values of 
these variables that are used to predict the tags after the word
``one'', which happens to be right before the editing term starts.
\begin{center}
\begin{tabular}{|l|c|c|c|} \hline
         & i-3   & i-2   & i-1   \\ \hline
{\em tP} &\bf PRP&\bf VBP&\bf CD \\
{\em tW} & it    & takes & one   \\ \hline
{\em rP} &\bf PRP&\bf VBP&\bf CD \\
{\em rW} & it    & takes & one   \\ \hline
\end{tabular}
\end{center}
Here we see that since the previous word is not an editing term,
the two sets of variables are the same.

Below we give the values of these variables that are used to predict
the tags after the word ``know'', which happens to be the last word
of the editing term.
\begin{center}
\begin{tabular}{|l|c|c|c|c|c|c|} \hline
         & i-6   & i-5   & i-4  & i-3    & i-2   & i-1   \\ \hline
{\em tP} &\bf PRP&\bf VBP&\bf CD&\bf Push&\bf PRP&\bf VBP\\
{\em tW} & it    & takes & one  &\bf Push& you   & know  \\ \hline
{\em rP} &       &       &      &\bf PRP &\bf VBP&\bf CD \\
{\em rW} &       &       &      & it     & takes & one   \\ \hline
\end{tabular}
\end{center}
Here we see that {\em rP} and {\em rW} capture the context of the
reparandum.  In fact, this set of variables will be mainly used
in predicting the repair tag.

\subsection{Other Variables}

We also include other variables that the decision tree algorithm can
use. We include a variable to indicate if we are currently processing
an editing term, and whether a non-filled pause editing term was seen.
We also include a variable that indicates the number of words in the
editing term so far. This lets the decision tree easily determine this
information without forcing it to look for a previous {\bf Push} in
the utterance-sensitive POS tags.
\begin{description}
\item{\em ET-state:} Indicates if we are in the middle of processing an editing term, and also if the editing term includes any non filled pauses.
\item{\em ET-prev:} Indicates the number of words in the editing term so far.
\end{description}
We actually have two sets of these variables, one set describes the state
before the editing term tag is predicted, and a second set, used in predicting
the word and POS tags, takes into account the editing term tag.

For the context for the editing term tag, we include how the tone was
just tagged, and for the context for the repair tag, we include the
tone tag and the editing term tag.

\subsection{The Decision Trees}

\begin{figure}
\begin{picture}(0,0)%
\includegraphics{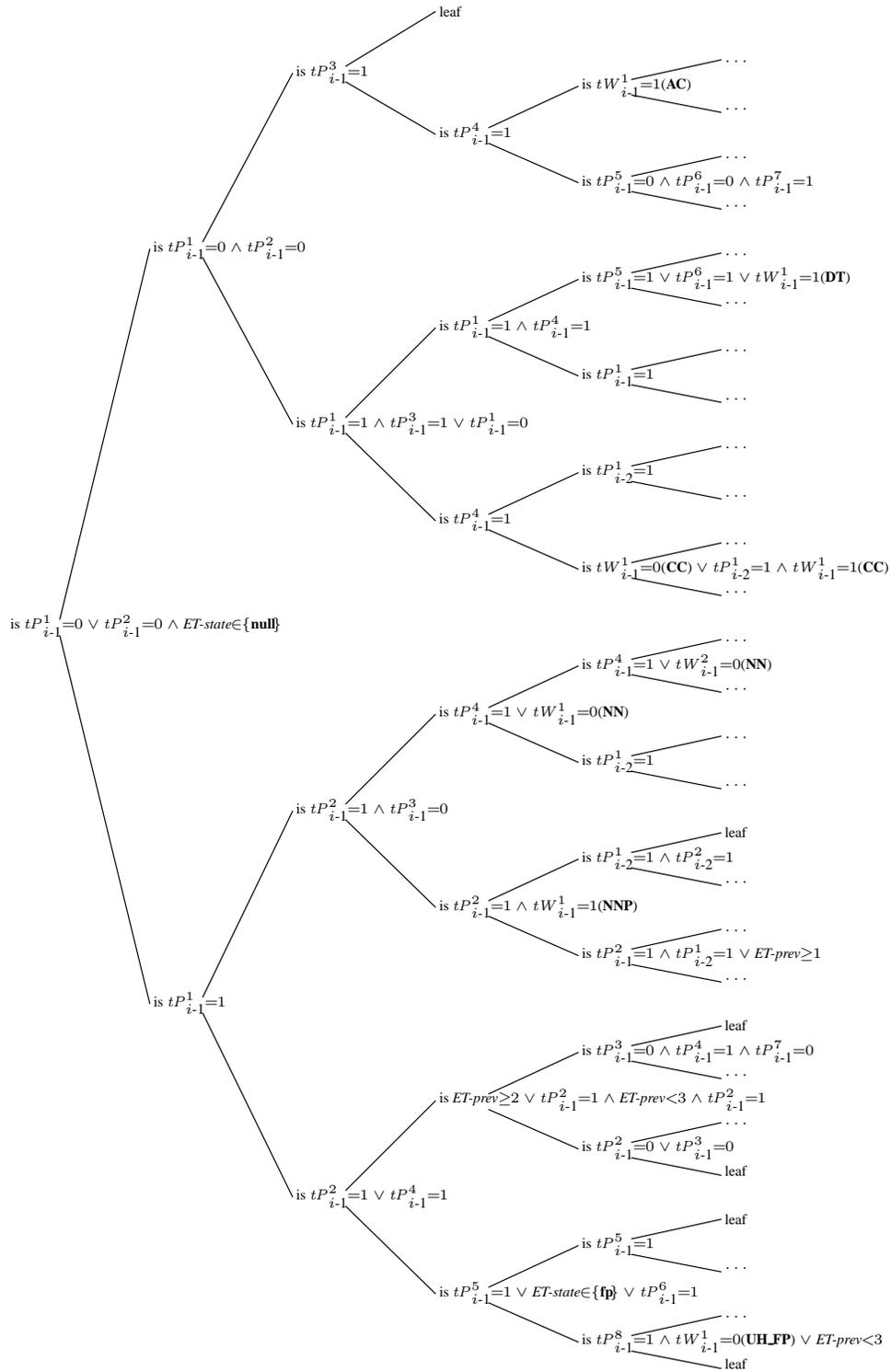}%
\end{picture}%
\setlength{\unitlength}{0.01250000in}%
\begingroup\makeatletter\ifx\SetFigFont\undefined
\def\x#1#2#3#4#5#6#7\relax{\def\x{#1#2#3#4#5#6}}%
\expandafter\x\fmtname xxxxxx\relax \def\y{splain}%
\ifx\x\y   
\gdef\SetFigFont#1#2#3{%
  \ifnum #1<17\tiny\else \ifnum #1<20\small\else
  \ifnum #1<24\normalsize\else \ifnum #1<29\large\else
  \ifnum #1<34\Large\else \ifnum #1<41\LARGE\else
     \huge\fi\fi\fi\fi\fi\fi
  \csname #3\endcsname}%
\else
\gdef\SetFigFont#1#2#3{\begingroup
  \count@#1\relax \ifnum 25<\count@\count@25\fi
  \def\x{\endgroup\@setsize\SetFigFont{#2pt}}%
  \expandafter\x
    \csname \romannumeral\the\count@ pt\expandafter\endcsname
    \csname @\romannumeral\the\count@ pt\endcsname
  \csname #3\endcsname}%
\fi
\fi\endgroup
\begin{picture}(327,632)(21,189)
\put(216,808){\makebox(0,0)[lb]{\smash{\SetFigFont{6}{7.2}{rm} leaf}}}
\put(346,786){\makebox(0,0)[lb]{\smash{\SetFigFont{6}{7.2}{rm} $\cdots$}}}
\put(346,764){\makebox(0,0)[lb]{\smash{\SetFigFont{6}{7.2}{rm} $\cdots$}}}
\put(281,775){\makebox(0,0)[lb]{\smash{\SetFigFont{6}{7.2}{rm} is ${\em tW}_{\mbox{$i$-1}}^1\!\!=\!\!1 \mbox{({\bf AC})}$}}}
\put(346,742){\makebox(0,0)[lb]{\smash{\SetFigFont{6}{7.2}{rm} $\cdots$}}}
\put(346,720){\makebox(0,0)[lb]{\smash{\SetFigFont{6}{7.2}{rm} $\cdots$}}}
\put(281,731){\makebox(0,0)[lb]{\smash{\SetFigFont{6}{7.2}{rm} is ${\em tP}_{\mbox{$i$-1}}^5\!\!=\!\!0 \wedge {\em tP}_{\mbox{$i$-1}}^6\!\!=\!\!0 \wedge {\em tP}_{\mbox{$i$-1}}^7\!\!=\!\!1$}}}
\put(216,753){\makebox(0,0)[lb]{\smash{\SetFigFont{6}{7.2}{rm} is ${\em tP}_{\mbox{$i$-1}}^4\!\!=\!\!1$}}}
\put(151,781){\makebox(0,0)[lb]{\smash{\SetFigFont{6}{7.2}{rm} is ${\em tP}_{\mbox{$i$-1}}^3\!\!=\!\!1$}}}
\put(346,698){\makebox(0,0)[lb]{\smash{\SetFigFont{6}{7.2}{rm} $\cdots$}}}
\put(346,676){\makebox(0,0)[lb]{\smash{\SetFigFont{6}{7.2}{rm} $\cdots$}}}
\put(281,687){\makebox(0,0)[lb]{\smash{\SetFigFont{6}{7.2}{rm} is ${\em tP}_{\mbox{$i$-1}}^5\!\!=\!\!1 \vee {\em tP}_{\mbox{$i$-1}}^6\!\!=\!\!1 \vee {\em tW}_{\mbox{$i$-1}}^1\!\!=\!\!1 \mbox{({\bf DT})}$}}}
\put(346,654){\makebox(0,0)[lb]{\smash{\SetFigFont{6}{7.2}{rm} $\cdots$}}}
\put(346,632){\makebox(0,0)[lb]{\smash{\SetFigFont{6}{7.2}{rm} $\cdots$}}}
\put(281,643){\makebox(0,0)[lb]{\smash{\SetFigFont{6}{7.2}{rm} is ${\em tP}_{\mbox{$i$-1}}^1\!\!=\!\!1$}}}
\put(216,665){\makebox(0,0)[lb]{\smash{\SetFigFont{6}{7.2}{rm} is ${\em tP}_{\mbox{$i$-1}}^1\!\!=\!\!1 \wedge {\em tP}_{\mbox{$i$-1}}^4\!\!=\!\!1$}}}
\put(346,610){\makebox(0,0)[lb]{\smash{\SetFigFont{6}{7.2}{rm} $\cdots$}}}
\put(346,588){\makebox(0,0)[lb]{\smash{\SetFigFont{6}{7.2}{rm} $\cdots$}}}
\put(281,599){\makebox(0,0)[lb]{\smash{\SetFigFont{6}{7.2}{rm} is ${\em tP}_{\mbox{$i$-2}}^1\!\!=\!\!1$}}}
\put(346,566){\makebox(0,0)[lb]{\smash{\SetFigFont{6}{7.2}{rm} $\cdots$}}}
\put(346,544){\makebox(0,0)[lb]{\smash{\SetFigFont{6}{7.2}{rm} $\cdots$}}}
\put(281,555){\makebox(0,0)[lb]{\smash{\SetFigFont{6}{7.2}{rm} is ${\em tW}_{\mbox{$i$-1}}^1\!\!=\!\!0 \mbox{({\bf CC})} \vee {\em tP}_{\mbox{$i$-2}}^1\!\!=\!\!1 \wedge {\em tW}_{\mbox{$i$-1}}^1\!\!=\!\!1 \mbox{({\bf CC})}$}}}
\put(216,577){\makebox(0,0)[lb]{\smash{\SetFigFont{6}{7.2}{rm} is ${\em tP}_{\mbox{$i$-1}}^4\!\!=\!\!1$}}}
\put(151,621){\makebox(0,0)[lb]{\smash{\SetFigFont{6}{7.2}{rm} is ${\em tP}_{\mbox{$i$-1}}^1\!\!=\!\!1 \wedge {\em tP}_{\mbox{$i$-1}}^3\!\!=\!\!1 \vee {\em tP}_{\mbox{$i$-1}}^1\!\!=\!\!0$}}}
\put( 86,701){\makebox(0,0)[lb]{\smash{\SetFigFont{6}{7.2}{rm} is ${\em tP}_{\mbox{$i$-1}}^1\!\!=\!\!0 \wedge {\em tP}_{\mbox{$i$-1}}^2\!\!=\!\!0$}}}
\put(346,522){\makebox(0,0)[lb]{\smash{\SetFigFont{6}{7.2}{rm} $\cdots$}}}
\put(346,500){\makebox(0,0)[lb]{\smash{\SetFigFont{6}{7.2}{rm} $\cdots$}}}
\put(281,511){\makebox(0,0)[lb]{\smash{\SetFigFont{6}{7.2}{rm} is ${\em tP}_{\mbox{$i$-1}}^4\!\!=\!\!1 \vee {\em tW}_{\mbox{$i$-1}}^2\!\!=\!\!0 \mbox{({\bf NN})}$}}}
\put(346,478){\makebox(0,0)[lb]{\smash{\SetFigFont{6}{7.2}{rm} $\cdots$}}}
\put(346,456){\makebox(0,0)[lb]{\smash{\SetFigFont{6}{7.2}{rm} $\cdots$}}}
\put(281,467){\makebox(0,0)[lb]{\smash{\SetFigFont{6}{7.2}{rm} is ${\em tP}_{\mbox{$i$-2}}^1\!\!=\!\!1$}}}
\put(216,489){\makebox(0,0)[lb]{\smash{\SetFigFont{6}{7.2}{rm} is ${\em tP}_{\mbox{$i$-1}}^4\!\!=\!\!1 \vee {\em tW}_{\mbox{$i$-1}}^1\!\!=\!\!0 \mbox{({\bf NN})}$}}}
\put(346,434){\makebox(0,0)[lb]{\smash{\SetFigFont{6}{7.2}{rm} leaf}}}
\put(346,412){\makebox(0,0)[lb]{\smash{\SetFigFont{6}{7.2}{rm} $\cdots$}}}
\put(281,423){\makebox(0,0)[lb]{\smash{\SetFigFont{6}{7.2}{rm} is ${\em tP}_{\mbox{$i$-2}}^1\!\!=\!\!1 \wedge {\em tP}_{\mbox{$i$-2}}^2\!\!=\!\!1$}}}
\put(346,390){\makebox(0,0)[lb]{\smash{\SetFigFont{6}{7.2}{rm} $\cdots$}}}
\put(346,368){\makebox(0,0)[lb]{\smash{\SetFigFont{6}{7.2}{rm} $\cdots$}}}
\put(281,379){\makebox(0,0)[lb]{\smash{\SetFigFont{6}{7.2}{rm} is ${\em tP}_{\mbox{$i$-1}}^2\!\!=\!\!1 \wedge {\em tP}_{\mbox{$i$-2}}^1\!\!=\!\!1 \vee \mbox{\em ET-prev}\!\!\geq\!\!1$}}}
\put(216,401){\makebox(0,0)[lb]{\smash{\SetFigFont{6}{7.2}{rm} is ${\em tP}_{\mbox{$i$-1}}^2\!\!=\!\!1 \wedge {\em tW}_{\mbox{$i$-1}}^1\!\!=\!\!1 \mbox{({\bf NNP})}$}}}
\put(151,445){\makebox(0,0)[lb]{\smash{\SetFigFont{6}{7.2}{rm} is ${\em tP}_{\mbox{$i$-1}}^2\!\!=\!\!1 \wedge {\em tP}_{\mbox{$i$-1}}^3\!\!=\!\!0$}}}
\put(346,346){\makebox(0,0)[lb]{\smash{\SetFigFont{6}{7.2}{rm} leaf}}}
\put(346,324){\makebox(0,0)[lb]{\smash{\SetFigFont{6}{7.2}{rm} $\cdots$}}}
\put(281,335){\makebox(0,0)[lb]{\smash{\SetFigFont{6}{7.2}{rm} is ${\em tP}_{\mbox{$i$-1}}^3\!\!=\!\!0 \wedge {\em tP}_{\mbox{$i$-1}}^4\!\!=\!\!1 \wedge {\em tP}_{\mbox{$i$-1}}^7\!\!=\!\!0$}}}
\put(346,302){\makebox(0,0)[lb]{\smash{\SetFigFont{6}{7.2}{rm} $\cdots$}}}
\put(346,280){\makebox(0,0)[lb]{\smash{\SetFigFont{6}{7.2}{rm} leaf}}}
\put(281,291){\makebox(0,0)[lb]{\smash{\SetFigFont{6}{7.2}{rm} is ${\em tP}_{\mbox{$i$-1}}^2\!\!=\!\!0 \vee {\em tP}_{\mbox{$i$-1}}^3\!\!=\!\!0$}}}
\put(216,313){\makebox(0,0)[lb]{\smash{\SetFigFont{6}{7.2}{rm} is $\mbox{\em ET-prev}\!\!\geq\!\!2 \vee {\em tP}_{\mbox{$i$-1}}^2\!\!=\!\!1 \wedge \mbox{\em ET-prev}\!\!<\!\!3 \wedge {\em tP}_{\mbox{$i$-1}}^2\!\!=\!\!1$}}}
\put(346,258){\makebox(0,0)[lb]{\smash{\SetFigFont{6}{7.2}{rm} leaf}}}
\put(346,236){\makebox(0,0)[lb]{\smash{\SetFigFont{6}{7.2}{rm} $\cdots$}}}
\put(281,247){\makebox(0,0)[lb]{\smash{\SetFigFont{6}{7.2}{rm} is ${\em tP}_{\mbox{$i$-1}}^5\!\!=\!\!1$}}}
\put(346,214){\makebox(0,0)[lb]{\smash{\SetFigFont{6}{7.2}{rm} $\cdots$}}}
\put(346,192){\makebox(0,0)[lb]{\smash{\SetFigFont{6}{7.2}{rm} leaf}}}
\put(281,203){\makebox(0,0)[lb]{\smash{\SetFigFont{6}{7.2}{rm} is ${\em tP}_{\mbox{$i$-1}}^8\!\!=\!\!1 \wedge {\em tW}_{\mbox{$i$-1}}^1\!\!=\!\!0 \mbox{({\bf UH\_FP})} \vee \mbox{\em ET-prev}\!\!<\!\!3$}}}
\put(216,225){\makebox(0,0)[lb]{\smash{\SetFigFont{6}{7.2}{rm} is ${\em tP}_{\mbox{$i$-1}}^5\!\!=\!\!1 \vee \mbox{\em ET-state}\!\!\in\!\!\{\mbox{\bf fp$\!$}\} \vee {\em tP}_{\mbox{$i$-1}}^6\!\!=\!\!1$}}}
\put(151,269){\makebox(0,0)[lb]{\smash{\SetFigFont{6}{7.2}{rm} is ${\em tP}_{\mbox{$i$-1}}^2\!\!=\!\!1 \vee {\em tP}_{\mbox{$i$-1}}^4\!\!=\!\!1$}}}
\put( 86,357){\makebox(0,0)[lb]{\smash{\SetFigFont{6}{7.2}{rm} is ${\em tP}_{\mbox{$i$-1}}^1\!\!=\!\!1$}}}
\put( 21,529){\makebox(0,0)[lb]{\smash{\SetFigFont{6}{7.2}{rm} is ${\em tP}_{\mbox{$i$-1}}^1\!\!=\!\!0 \vee {\em tP}_{\mbox{$i$-1}}^2\!\!=\!\!0 \wedge \mbox{\em ET-state}\!\!\in\!\!\{\mbox{\bf null$\!$}\}$}}}
\end{picture}
\caption{Decision Tree for Tone Tags}
\label{fig:detection:Ttree}
\end{figure}

In Figure~\ref{fig:detection:Ttree}, we give the top part of the
decision tree that was grown for the tone tags (for the first
partition of the training data).
%
%
In learning the probability distribution for the tones, the null case
corresponds to a number of different events.  It could be the
beginning of an editing term ({\bf Push}), the end of an editing term
({\bf Pop}), a modification repair or a fresh start (without an
editing term).  We find that we get a better probability estimate for
the null tone event if we train the decision tree to predict each type
of these null events, rather then treat them as a single class.  The
probability of the null tone is simply the sum of probabilities of the
non-tone classes.

\begin{figure}
\begin{picture}(0,0)%
\includegraphics{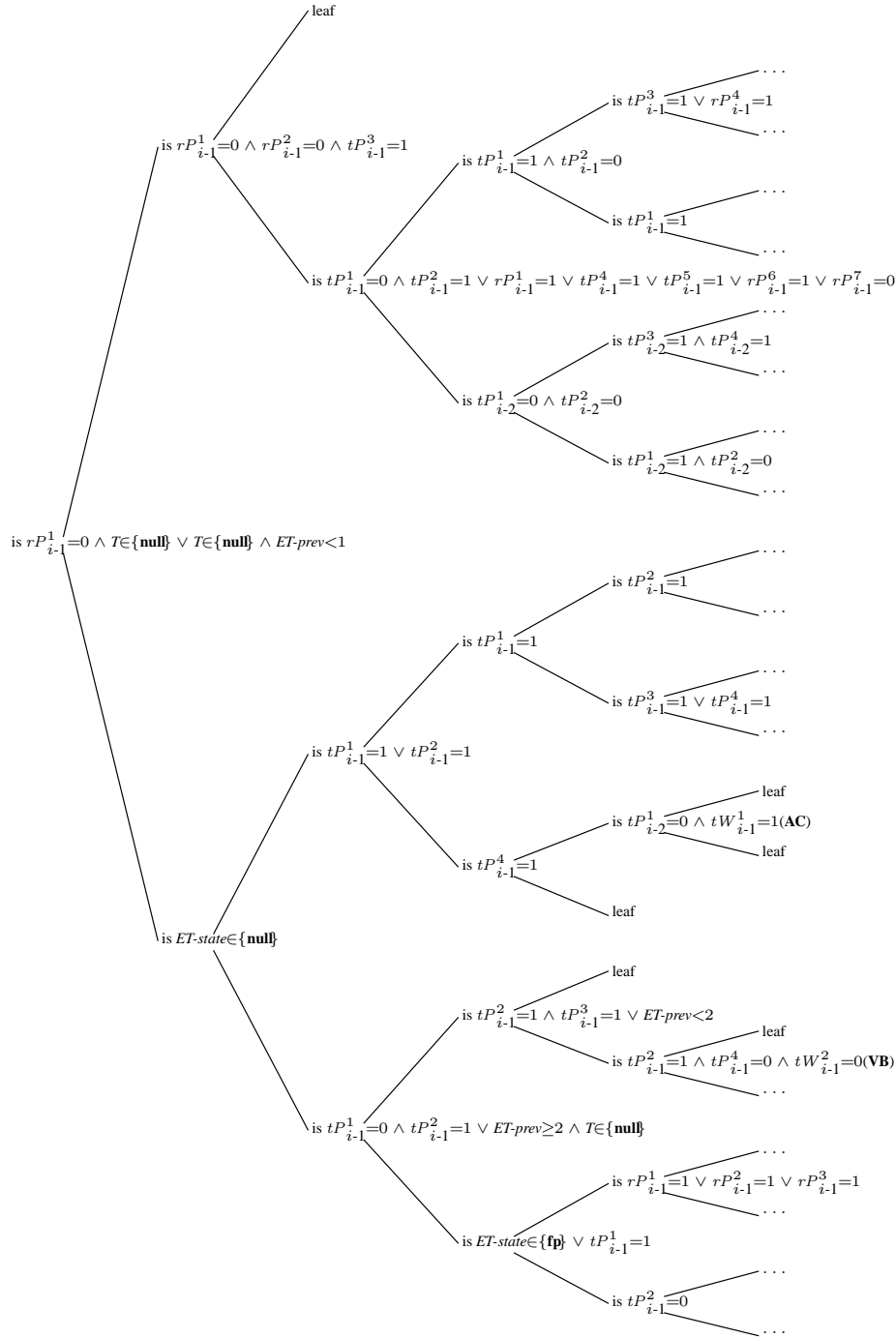}%
\end{picture}%
\setlength{\unitlength}{0.01250000in}%
\begingroup\makeatletter\ifx\SetFigFont\undefined
\def\x#1#2#3#4#5#6#7\relax{\def\x{#1#2#3#4#5#6}}%
\expandafter\x\fmtname xxxxxx\relax \def\y{splain}%
\ifx\x\y   
\gdef\SetFigFont#1#2#3{%
  \ifnum #1<17\tiny\else \ifnum #1<20\small\else
  \ifnum #1<24\normalsize\else \ifnum #1<29\large\else
  \ifnum #1<34\Large\else \ifnum #1<41\LARGE\else
     \huge\fi\fi\fi\fi\fi\fi
  \csname #3\endcsname}%
\else
\gdef\SetFigFont#1#2#3{\begingroup
  \count@#1\relax \ifnum 25<\count@\count@25\fi
  \def\x{\endgroup\@setsize\SetFigFont{#2pt}}%
  \expandafter\x
    \csname \romannumeral\the\count@ pt\expandafter\endcsname
    \csname @\romannumeral\the\count@ pt\endcsname
  \csname #3\endcsname}%
\fi
\fi\endgroup
\begin{picture}(327,588)(21,229)
\put(151,804){\makebox(0,0)[lb]{\smash{\SetFigFont{6}{7.2}{rm} leaf}}}
\put(346,778){\makebox(0,0)[lb]{\smash{\SetFigFont{6}{7.2}{rm} $\cdots$}}}
\put(346,752){\makebox(0,0)[lb]{\smash{\SetFigFont{6}{7.2}{rm} $\cdots$}}}
\put(281,765){\makebox(0,0)[lb]{\smash{\SetFigFont{6}{7.2}{rm} is ${\em tP}_{\mbox{$i$-1}}^3\!\!=\!\!1 \vee {\em rP}_{\mbox{$i$-1}}^4\!\!=\!\!1$}}}
\put(346,726){\makebox(0,0)[lb]{\smash{\SetFigFont{6}{7.2}{rm} $\cdots$}}}
\put(346,700){\makebox(0,0)[lb]{\smash{\SetFigFont{6}{7.2}{rm} $\cdots$}}}
\put(281,713){\makebox(0,0)[lb]{\smash{\SetFigFont{6}{7.2}{rm} is ${\em tP}_{\mbox{$i$-1}}^1\!\!=\!\!1$}}}
\put(216,739){\makebox(0,0)[lb]{\smash{\SetFigFont{6}{7.2}{rm} is ${\em tP}_{\mbox{$i$-1}}^1\!\!=\!\!1 \wedge {\em tP}_{\mbox{$i$-1}}^2\!\!=\!\!0$}}}
\put(346,674){\makebox(0,0)[lb]{\smash{\SetFigFont{6}{7.2}{rm} $\cdots$}}}
\put(346,648){\makebox(0,0)[lb]{\smash{\SetFigFont{6}{7.2}{rm} $\cdots$}}}
\put(281,661){\makebox(0,0)[lb]{\smash{\SetFigFont{6}{7.2}{rm} is ${\em tP}_{\mbox{$i$-2}}^3\!\!=\!\!1 \wedge {\em tP}_{\mbox{$i$-2}}^4\!\!=\!\!1$}}}
\put(346,622){\makebox(0,0)[lb]{\smash{\SetFigFont{6}{7.2}{rm} $\cdots$}}}
\put(346,596){\makebox(0,0)[lb]{\smash{\SetFigFont{6}{7.2}{rm} $\cdots$}}}
\put(281,609){\makebox(0,0)[lb]{\smash{\SetFigFont{6}{7.2}{rm} is ${\em tP}_{\mbox{$i$-2}}^1\!\!=\!\!1 \wedge {\em tP}_{\mbox{$i$-2}}^2\!\!=\!\!0$}}}
\put(216,635){\makebox(0,0)[lb]{\smash{\SetFigFont{6}{7.2}{rm} is ${\em tP}_{\mbox{$i$-2}}^1\!\!=\!\!0 \wedge {\em tP}_{\mbox{$i$-2}}^2\!\!=\!\!0$}}}
\put(151,687){\makebox(0,0)[lb]{\smash{\SetFigFont{6}{7.2}{rm} is ${\em tP}_{\mbox{$i$-1}}^1\!\!=\!\!0 \wedge {\em tP}_{\mbox{$i$-1}}^2\!\!=\!\!1 \vee {\em rP}_{\mbox{$i$-1}}^1\!\!=\!\!1 \vee {\em tP}_{\mbox{$i$-1}}^4\!\!=\!\!1 \vee {\em tP}_{\mbox{$i$-1}}^5\!\!=\!\!1 \vee {\em rP}_{\mbox{$i$-1}}^6\!\!=\!\!1 \vee {\em rP}_{\mbox{$i$-1}}^7\!\!=\!\!0$}}}
\put( 86,746){\makebox(0,0)[lb]{\smash{\SetFigFont{6}{7.2}{rm} is ${\em rP}_{\mbox{$i$-1}}^1\!\!=\!\!0 \wedge {\em rP}_{\mbox{$i$-1}}^2\!\!=\!\!0 \wedge {\em tP}_{\mbox{$i$-1}}^3\!\!=\!\!1$}}}
\put(346,570){\makebox(0,0)[lb]{\smash{\SetFigFont{6}{7.2}{rm} $\cdots$}}}
\put(346,544){\makebox(0,0)[lb]{\smash{\SetFigFont{6}{7.2}{rm} $\cdots$}}}
\put(281,557){\makebox(0,0)[lb]{\smash{\SetFigFont{6}{7.2}{rm} is ${\em tP}_{\mbox{$i$-1}}^2\!\!=\!\!1$}}}
\put(346,518){\makebox(0,0)[lb]{\smash{\SetFigFont{6}{7.2}{rm} $\cdots$}}}
\put(346,492){\makebox(0,0)[lb]{\smash{\SetFigFont{6}{7.2}{rm} $\cdots$}}}
\put(281,505){\makebox(0,0)[lb]{\smash{\SetFigFont{6}{7.2}{rm} is ${\em tP}_{\mbox{$i$-1}}^3\!\!=\!\!1 \vee {\em tP}_{\mbox{$i$-1}}^4\!\!=\!\!1$}}}
\put(216,531){\makebox(0,0)[lb]{\smash{\SetFigFont{6}{7.2}{rm} is ${\em tP}_{\mbox{$i$-1}}^1\!\!=\!\!1$}}}
\put(346,466){\makebox(0,0)[lb]{\smash{\SetFigFont{6}{7.2}{rm} leaf}}}
\put(346,440){\makebox(0,0)[lb]{\smash{\SetFigFont{6}{7.2}{rm} leaf}}}
\put(281,453){\makebox(0,0)[lb]{\smash{\SetFigFont{6}{7.2}{rm} is ${\em tP}_{\mbox{$i$-2}}^1\!\!=\!\!0 \wedge {\em tW}_{\mbox{$i$-1}}^1\!\!=\!\!1 \mbox{({\bf AC})}$}}}
\put(281,414){\makebox(0,0)[lb]{\smash{\SetFigFont{6}{7.2}{rm} leaf}}}
\put(216,434){\makebox(0,0)[lb]{\smash{\SetFigFont{6}{7.2}{rm} is ${\em tP}_{\mbox{$i$-1}}^4\!\!=\!\!1$}}}
\put(151,483){\makebox(0,0)[lb]{\smash{\SetFigFont{6}{7.2}{rm} is ${\em tP}_{\mbox{$i$-1}}^1\!\!=\!\!1 \vee {\em tP}_{\mbox{$i$-1}}^2\!\!=\!\!1$}}}
\put(281,388){\makebox(0,0)[lb]{\smash{\SetFigFont{6}{7.2}{rm} leaf}}}
\put(346,362){\makebox(0,0)[lb]{\smash{\SetFigFont{6}{7.2}{rm} leaf}}}
\put(346,336){\makebox(0,0)[lb]{\smash{\SetFigFont{6}{7.2}{rm} $\cdots$}}}
\put(281,349){\makebox(0,0)[lb]{\smash{\SetFigFont{6}{7.2}{rm} is ${\em tP}_{\mbox{$i$-1}}^2\!\!=\!\!1 \wedge {\em tP}_{\mbox{$i$-1}}^4\!\!=\!\!0 \wedge {\em tW}_{\mbox{$i$-1}}^2\!\!=\!\!0 \mbox{({\bf VB})}$}}}
\put(216,369){\makebox(0,0)[lb]{\smash{\SetFigFont{6}{7.2}{rm} is ${\em tP}_{\mbox{$i$-1}}^2\!\!=\!\!1 \wedge {\em tP}_{\mbox{$i$-1}}^3\!\!=\!\!1 \vee \mbox{\em ET-prev}\!\!<\!\!2$}}}
\put(346,310){\makebox(0,0)[lb]{\smash{\SetFigFont{6}{7.2}{rm} $\cdots$}}}
\put(346,284){\makebox(0,0)[lb]{\smash{\SetFigFont{6}{7.2}{rm} $\cdots$}}}
\put(281,297){\makebox(0,0)[lb]{\smash{\SetFigFont{6}{7.2}{rm} is ${\em rP}_{\mbox{$i$-1}}^1\!\!=\!\!1 \vee {\em rP}_{\mbox{$i$-1}}^2\!\!=\!\!1 \vee {\em rP}_{\mbox{$i$-1}}^3\!\!=\!\!1$}}}
\put(346,258){\makebox(0,0)[lb]{\smash{\SetFigFont{6}{7.2}{rm} $\cdots$}}}
\put(346,232){\makebox(0,0)[lb]{\smash{\SetFigFont{6}{7.2}{rm} $\cdots$}}}
\put(281,245){\makebox(0,0)[lb]{\smash{\SetFigFont{6}{7.2}{rm} is ${\em tP}_{\mbox{$i$-1}}^2\!\!=\!\!0$}}}
\put(216,271){\makebox(0,0)[lb]{\smash{\SetFigFont{6}{7.2}{rm} is $\mbox{\em ET-state}\!\!\in\!\!\{\mbox{\bf fp$\!$}\} \vee {\em tP}_{\mbox{$i$-1}}^1\!\!=\!\!1$}}}
\put(151,320){\makebox(0,0)[lb]{\smash{\SetFigFont{6}{7.2}{rm} is ${\em tP}_{\mbox{$i$-1}}^1\!\!=\!\!0 \wedge {\em tP}_{\mbox{$i$-1}}^2\!\!=\!\!1 \vee \mbox{\em ET-prev}\!\!\geq\!\!2 \wedge \mbox{\em T}\!\!\in\!\!\{\mbox{\bf null$\!$}\}$}}}
\put( 86,402){\makebox(0,0)[lb]{\smash{\SetFigFont{6}{7.2}{rm} is $\mbox{\em ET-state}\!\!\in\!\!\{\mbox{\bf null$\!$}\}$}}}
\put( 21,574){\makebox(0,0)[lb]{\smash{\SetFigFont{6}{7.2}{rm} is ${\em rP}_{\mbox{$i$-1}}^1\!\!=\!\!0 \wedge \mbox{\em T}\!\!\in\!\!\{\mbox{\bf null$\!$}\} \vee \mbox{\em T}\!\!\in\!\!\{\mbox{\bf null$\!$}\} \wedge \mbox{\em ET-prev}\!\!<\!\!1$}}}
\end{picture}
\caption{Decision Tree for Editing Term Tags}
\label{fig:detection:Etree}
\end{figure}

In Figure~\ref{fig:detection:Etree}, we give the top part of the
decision tree for the editing term tags.
Just as with learning the probability distribution of the tone tag, we
subdivide the null editing term case into whether there is a
modification repair or a fresh start (without an editing term).
Again, this lets us better predict the null editing term tag.

\begin{figure}
\begin{picture}(0,0)%
\includegraphics{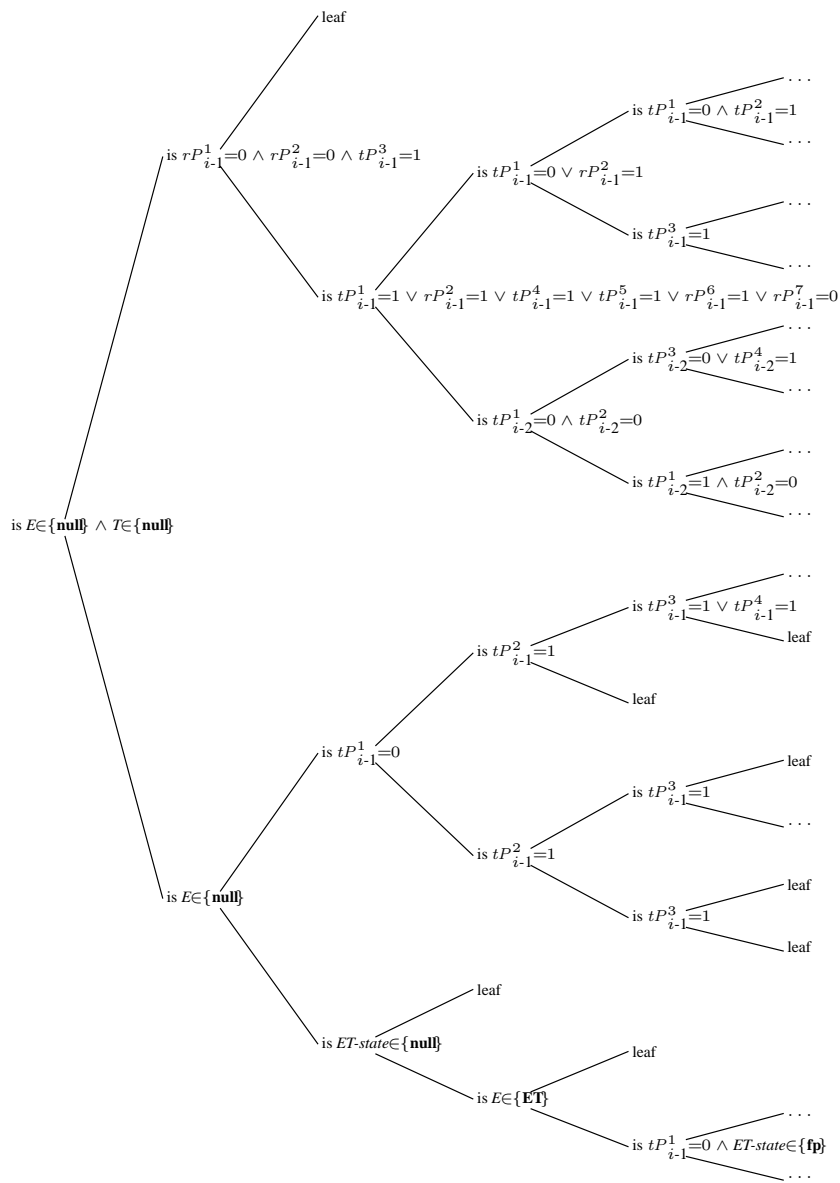}%
\end{picture}%
\setlength{\unitlength}{0.01250000in}%
\begingroup\makeatletter\ifx\SetFigFont\undefined
\def\x#1#2#3#4#5#6#7\relax{\def\x{#1#2#3#4#5#6}}%
\expandafter\x\fmtname xxxxxx\relax \def\y{splain}%
\ifx\x\y   
\gdef\SetFigFont#1#2#3{%
  \ifnum #1<17\tiny\else \ifnum #1<20\small\else
  \ifnum #1<24\normalsize\else \ifnum #1<29\large\else
  \ifnum #1<34\Large\else \ifnum #1<41\LARGE\else
     \huge\fi\fi\fi\fi\fi\fi
  \csname #3\endcsname}%
\else
\gdef\SetFigFont#1#2#3{\begingroup
  \count@#1\relax \ifnum 25<\count@\count@25\fi
  \def\x{\endgroup\@setsize\SetFigFont{#2pt}}%
  \expandafter\x
    \csname \romannumeral\the\count@ pt\expandafter\endcsname
    \csname @\romannumeral\the\count@ pt\endcsname
  \csname #3\endcsname}%
\fi
\fi\endgroup
\begin{picture}(327,502)(21,315)
\put(151,804){\makebox(0,0)[lb]{\smash{\SetFigFont{6}{7.2}{rm} leaf}}}
\put(346,778){\makebox(0,0)[lb]{\smash{\SetFigFont{6}{7.2}{rm} $\cdots$}}}
\put(346,752){\makebox(0,0)[lb]{\smash{\SetFigFont{6}{7.2}{rm} $\cdots$}}}
\put(281,765){\makebox(0,0)[lb]{\smash{\SetFigFont{6}{7.2}{rm} is ${\em tP}_{\mbox{$i$-1}}^1\!\!=\!\!0 \wedge {\em tP}_{\mbox{$i$-1}}^2\!\!=\!\!1$}}}
\put(346,726){\makebox(0,0)[lb]{\smash{\SetFigFont{6}{7.2}{rm} $\cdots$}}}
\put(346,700){\makebox(0,0)[lb]{\smash{\SetFigFont{6}{7.2}{rm} $\cdots$}}}
\put(281,713){\makebox(0,0)[lb]{\smash{\SetFigFont{6}{7.2}{rm} is ${\em tP}_{\mbox{$i$-1}}^3\!\!=\!\!1$}}}
\put(216,739){\makebox(0,0)[lb]{\smash{\SetFigFont{6}{7.2}{rm} is ${\em tP}_{\mbox{$i$-1}}^1\!\!=\!\!0 \vee {\em rP}_{\mbox{$i$-1}}^2\!\!=\!\!1$}}}
\put(346,674){\makebox(0,0)[lb]{\smash{\SetFigFont{6}{7.2}{rm} $\cdots$}}}
\put(346,648){\makebox(0,0)[lb]{\smash{\SetFigFont{6}{7.2}{rm} $\cdots$}}}
\put(281,661){\makebox(0,0)[lb]{\smash{\SetFigFont{6}{7.2}{rm} is ${\em tP}_{\mbox{$i$-2}}^3\!\!=\!\!0 \vee {\em tP}_{\mbox{$i$-2}}^4\!\!=\!\!1$}}}
\put(346,622){\makebox(0,0)[lb]{\smash{\SetFigFont{6}{7.2}{rm} $\cdots$}}}
\put(346,596){\makebox(0,0)[lb]{\smash{\SetFigFont{6}{7.2}{rm} $\cdots$}}}
\put(281,609){\makebox(0,0)[lb]{\smash{\SetFigFont{6}{7.2}{rm} is ${\em tP}_{\mbox{$i$-2}}^1\!\!=\!\!1 \wedge {\em tP}_{\mbox{$i$-2}}^2\!\!=\!\!0$}}}
\put(216,635){\makebox(0,0)[lb]{\smash{\SetFigFont{6}{7.2}{rm} is ${\em tP}_{\mbox{$i$-2}}^1\!\!=\!\!0 \wedge {\em tP}_{\mbox{$i$-2}}^2\!\!=\!\!0$}}}
\put(151,687){\makebox(0,0)[lb]{\smash{\SetFigFont{6}{7.2}{rm} is ${\em tP}_{\mbox{$i$-1}}^1\!\!=\!\!1 \vee {\em rP}_{\mbox{$i$-1}}^2\!\!=\!\!1 \vee {\em tP}_{\mbox{$i$-1}}^4\!\!=\!\!1 \vee {\em tP}_{\mbox{$i$-1}}^5\!\!=\!\!1 \vee {\em rP}_{\mbox{$i$-1}}^6\!\!=\!\!1 \vee {\em rP}_{\mbox{$i$-1}}^7\!\!=\!\!0$}}}
\put( 86,746){\makebox(0,0)[lb]{\smash{\SetFigFont{6}{7.2}{rm} is ${\em rP}_{\mbox{$i$-1}}^1\!\!=\!\!0 \wedge {\em rP}_{\mbox{$i$-1}}^2\!\!=\!\!0 \wedge {\em tP}_{\mbox{$i$-1}}^3\!\!=\!\!1$}}}
\put(346,570){\makebox(0,0)[lb]{\smash{\SetFigFont{6}{7.2}{rm} $\cdots$}}}
\put(346,544){\makebox(0,0)[lb]{\smash{\SetFigFont{6}{7.2}{rm} leaf}}}
\put(281,557){\makebox(0,0)[lb]{\smash{\SetFigFont{6}{7.2}{rm} is ${\em tP}_{\mbox{$i$-1}}^3\!\!=\!\!1 \vee {\em tP}_{\mbox{$i$-1}}^4\!\!=\!\!1$}}}
\put(281,518){\makebox(0,0)[lb]{\smash{\SetFigFont{6}{7.2}{rm} leaf}}}
\put(216,538){\makebox(0,0)[lb]{\smash{\SetFigFont{6}{7.2}{rm} is ${\em tP}_{\mbox{$i$-1}}^2\!\!=\!\!1$}}}
\put(346,492){\makebox(0,0)[lb]{\smash{\SetFigFont{6}{7.2}{rm} leaf}}}
\put(346,466){\makebox(0,0)[lb]{\smash{\SetFigFont{6}{7.2}{rm} $\cdots$}}}
\put(281,479){\makebox(0,0)[lb]{\smash{\SetFigFont{6}{7.2}{rm} is ${\em tP}_{\mbox{$i$-1}}^3\!\!=\!\!1$}}}
\put(346,440){\makebox(0,0)[lb]{\smash{\SetFigFont{6}{7.2}{rm} leaf}}}
\put(346,414){\makebox(0,0)[lb]{\smash{\SetFigFont{6}{7.2}{rm} leaf}}}
\put(281,427){\makebox(0,0)[lb]{\smash{\SetFigFont{6}{7.2}{rm} is ${\em tP}_{\mbox{$i$-1}}^3\!\!=\!\!1$}}}
\put(216,453){\makebox(0,0)[lb]{\smash{\SetFigFont{6}{7.2}{rm} is ${\em tP}_{\mbox{$i$-1}}^2\!\!=\!\!1$}}}
\put(151,496){\makebox(0,0)[lb]{\smash{\SetFigFont{6}{7.2}{rm} is ${\em tP}_{\mbox{$i$-1}}^1\!\!=\!\!0$}}}
\put(216,396){\makebox(0,0)[lb]{\smash{\SetFigFont{6}{7.2}{rm} leaf}}}
\put(281,370){\makebox(0,0)[lb]{\smash{\SetFigFont{6}{7.2}{rm} leaf}}}
\put(346,344){\makebox(0,0)[lb]{\smash{\SetFigFont{6}{7.2}{rm} $\cdots$}}}
\put(346,318){\makebox(0,0)[lb]{\smash{\SetFigFont{6}{7.2}{rm} $\cdots$}}}
\put(281,331){\makebox(0,0)[lb]{\smash{\SetFigFont{6}{7.2}{rm} is ${\em tP}_{\mbox{$i$-1}}^1\!\!=\!\!0 \wedge \mbox{\em ET-state}\!\!\in\!\!\{\mbox{\bf fp$\!$}\}$}}}
\put(216,351){\makebox(0,0)[lb]{\smash{\SetFigFont{6}{7.2}{rm} is $\mbox{\em E}\!\!\in\!\!\{\mbox{\bf ET$\!$}\}$}}}
\put(151,374){\makebox(0,0)[lb]{\smash{\SetFigFont{6}{7.2}{rm} is $\mbox{\em ET-state}\!\!\in\!\!\{\mbox{\bf null$\!$}\}$}}}
\put( 86,435){\makebox(0,0)[lb]{\smash{\SetFigFont{6}{7.2}{rm} is $\mbox{\em E}\!\!\in\!\!\{\mbox{\bf null$\!$}\}$}}}
\put( 21,591){\makebox(0,0)[lb]{\smash{\SetFigFont{6}{7.2}{rm} is $\mbox{\em E}\!\!\in\!\!\{\mbox{\bf null$\!$}\} \wedge \mbox{\em T}\!\!\in\!\!\{\mbox{\bf null$\!$}\}$}}}
\end{picture}
\caption{Decision Tree for Repair Tags}
\label{fig:detection:Rtree}
\end{figure}

In Figure~\ref{fig:detection:Rtree}, we give the decision tree for the
repair tags.
Note that new versions of the word tree and POS tree are also grown,
which take into account the utterance-sensitive words and POS tags
afforded by modeling the occurrence of boundary tones and speech
repairs.

\cleardoublepage
\chapter{Correcting Speech Repairs}
\label{chapter:correction}

In the previous chapter, we showed how a statistical language model
can be augmented to detect the occurrence of speech repairs, editing
terms and intonational boundaries.  But for speech repairs, we have
only addressed half of the problem; the other half is determining the
extent of the reparandum, which we refer to as correcting the speech
repair.  As we discussed in Section~\ref {sec:related:repairs}, many
different approaches have been employed in correcting speech repairs.
Hindle \shortcite {Hindle83:acl} and Kikui and Morimoto \shortcite
{KikuiMorimoto94:icslp} both separate the task of correcting a repair
from detecting it by assuming that there is an acoustic editing signal
that marks the interruption point of speech repairs.  As discussed in
the introduction of Chapter~\ref {chapter:detection}, a reliable
signal has not yet been found.  Although the previous chapter presents
a model that detects the occurrence of speech repairs, this model is
not effective enough.  In fact, we feel that one of its crucial
shortcomings is that it does not take into consideration the task of
correcting speech repairs \cite {HeemanLokenkimAllen96:icslp}.  Since
hearers are often unaware of speech repairs \cite {MartinStrange68},
they must be able to correct them as the utterance is unfolding and as
an indistinguishable event from detecting them and recognizing the
words involved.

Bear \etal~\shortcite{Bear-etal92:acl} proposed that multiple
information sources need to be combined in order to detect and correct
speech repairs.  One of these sources includes a pattern matching
routine that looks for simple cases of word correspondences that could
indicate a speech repair.  However, pattern matching is too limited
too capture the variety of word correspondence patterns that speech
repairs exhibit \cite{HeemanAllen94:acl}.  In the Trains corpus, there
are 160 different repair structures, not including variations of
fragments and editing terms, for the 1302 modification repairs. Of
these 160, only 47 occurred more than one time, and these are listed
in Table~\ref{tab:patterns}.  Each word in the reparandum and
alteration is represented by its label type: `{\bf m}' for word match,
`{\bf r}' for replacement, `{\bf p}' for multi-word replacements, and
`{\bf x}' for deletions from the reparandum or insertions in the
alteration.  A period `{\bf .}' marks the interruption point.  For
example, the structure for the repair given below (given earlier as
Example~\ref {ex:d93-15.2:utt42}) would be `{\bf mrm.mrm}'.
\begin{example}{d93-5.2 utt42}
\reparandum{engine two from Elmi(ra)-}\ip 
\et{or} 	
\alteration{engine three from Elmira}
\end{example}	
\begin{table}[hbp]
\begin{center}
\begin{tabular}{|l|r||l|r|} \hline
x. & 357 & mmmr.mmmr & 4 \\ 
m.m & 249 & mm.mxm & 4 \\ 
r.r & 136 & xmmm.mmm & 3 \\ 
mm.mm & 85 & mrx.mr & 3 \\ 
mx.m & 76 & mrr.mrr & 3 \\ 
mmx.mm & 35 & mrmx.mrm & 3 \\ 
mr.mr & 29 & mmmmm.mmmmm & 3 \\ 
mmm.mmm & 22 & mm.xmm & 3 \\ 
rx.r & 20 & xr.r & 2 \\ 
rm.rm & 20 & xmx.m & 2 \\ 
xx. & 12 & xmmx.mm & 2 \\ 
mmmm.mmmm & 12 & rr.rr & 2 \\ 
mmr.mmr & 10 & rm.rxm & 2 \\ 
m.xm & 10 & r.xr & 2 \\ 
mxx.m & 8 & mxmx.mm & 2 \\ 
mmmx.mmm & 8 & mrmm.mrmm & 2 \\ 
m.xxm & 8 & mmmxx.mmm & 2 \\ 
mrm.mrm & 7 & mmmmx.mmmm & 2 \\ 
mx.xm & 6 & mmm.xxxmmm & 2 \\ 
xm.m & 5 & mmm.mxmm & 2 \\ 
p.pp & 5 & mmm.mmxm & 2 \\ 
mmmmr.mmmmr & 5 & mm.xxmm & 2 \\ 
rmm.rmm & 4 & mm.mxxm & 2 \\ 
mmxx.mm & 4 &  &  \\ 
\hline
\end{tabular}
\end{center}
\caption{Occurrences of Common Repair Structures}
\label{tab:patterns}
\end{table}

To remedy the limitation of Bear \etal, we proposed that the structure
of the word correspondences between the reparandum and alteration
could be accounted for by a set of well-formedness rules \cite
{HeemanAllen94:acl}.  Potential repair structures found by the rules
were passed to a statistical language model (an early predecessor of
the model presented in the Chapter~\ref {chapter:detection}), which
was used to prune out false positives.  The statistical language model
took into account the word matches found by the repair structure.  We
then cleaned up this approach \cite {HeemanLokenkimAllen96:icslp} by
using the potential repair structures as part of the context used by
the statistical model, rather than just the word matches.  However,
even this approach is still lacking in how it incorporates speech
repair correction into the language model.  The alteration of a
repair, which makes up half of the repair structure, occurs after the
interruption point and hence should not be used to predict the
occurrence of a repair.  Hence these models are of limited use in
helping a speech recognizer predict the next word given the previous
context.

Recently, Stolcke and Shriberg \shortcite{StolckeShriberg96:icassp}
presented a word-based model for speech recognition that models simple
word deletion and word repetition patterns.  They used the prediction
of the repair to clean up the context and help predict what word will
occur next.  Although their model is limited to simple types of
repairs, it provides a starting point for incorporating speech repair
correction into a statistical language model.

\section{Sources of Information}
\label{sec:correction:sources}

Before we lay out our model of incorporating speech repair correction
into a statistical language model, we first review the information
that gives evidence of the extent of the reparandum.  Probably the
most widely used is the presence of word correspondences between the
reparandum and alteration, both at the word level, and at the level of
syntactic constituents \cite
{Levelt83:cog,Hindle83:acl,Bear-etal92:acl,HeemanAllen94:acl,KikuiMorimoto94:icslp}.

The second source is to simply look for a fluent transition from the
speech that precedes the onset of the reparandum to alteration
\cite{KikuiMorimoto94:icslp}.  Although closely related to the first
source, it is different, especially for speech repairs that do not
have initial retracing.  This source of information is a mainstay of
the ``parser-first'' approach (e.g.~\cite{Dowding-etal93:acl})---keep
trying alternative corrections until one of them parses.

A third source of information is that speakers tend to restart at the
beginning of constituent boundaries \cite{Nooteboom80}.  Levelt
\shortcite {Levelt83:cog} refined this observation by noting that
reparandum onsets tend to occur where a co-ordinated constituent can
be placed.  Hence, reparandum onsets can be partially predicted based
on a syntactic analysis of the speech that precedes the interruption
point.

\section{Our Proposal}

Most previous approaches to correcting speech repairs have taken the
standpoint of finding the best reparandum given the neighboring words.
Instead, we view the problem as finding the reparandum that best
predicts the following words.  Since speech repairs are often
accompanied by word correspondences, the actual reparandum will better
predict the words involved in the alteration of the repair.  Consider
the following speech repair involving repeated words.
\begin{example}{d93-3.2 utt45}
which engine \reparandum{are we}\ip are we taking
\end{example}
In this example, if we predicted that a modification repair occurred
and that the reparandum consists of ``are we'', then the probability of
``are'' being the first word of the alteration would be very high,
since it matches the first word of the reparandum.  Conversely, if we
are not predicting a modification repair whose reparandum is ``are
we'', then the probability of seeing this word would be much lower.
The same reasoning holds for predicting the next word, ``we'': it is
much more likely under the repair interpretation.  So, as we process
the words involved in the alteration, the repair interpretation will
better account for the words that follow it, strengthening the
interpretation.

When predicting the words in the alteration, it is not just the words
in the proposed reparandum that can be taken into account.  When
predicting the first word of the alteration, we can also take into
account the context provided by the words that precede the reparandum.
Consider the following repair in which the first two words of the
alteration are inserted words.
\begin{example}{d93-16.2 utt66}
and two tankers \reparandum{to}\ip of OJ to Dansville
\end{example}
Here, if we know the reparandum is ``to'', then we know that the first
word of the reparandum must be a fluent continuation of the speech
before the onset of the reparandum.  In fact, we see that the repair
interpretation (with the correct reparandum onset) provides better
context for predicting the first word of the alteration than a
hypothesis that predicts either the wrong reparandum onset or predicts
no speech repair at all.  Hence, by predicting the reparandum of a
speech repair, we no longer need to predict the onset of the
alteration on the basis of the ending of the reparandum, as we did in
Section~\ref {sec:detection:cleanup} in the previous chapter.  Such
predictions are based on limited amounts of training data since just
examples of speech repairs can be used.  Rather, by first predicting
the reparandum, we can use examples of fluent transitions to help
predict the first word of the alteration.

We can also make use of the third source of correction information
identified in the previous section.  When we initially hypothesize the
reparandum onset, we can take into account the a priori probability
that it will occur at that point.  Consider the following example.
\begin{example}{d92a-2.1 utt77}
that way \reparandum{the other one can be free}\ip the orange juice
one can travel \mbox{back and forth}
\end{example}
According to Levelt, some of the possible reparandum onsets are not
well-formed.  For this example, reparandum onsets of ``one'',
``other'', and ``way'' would be ill-formed, and so should have a lower
probability assigned to them.

\section{Adding in Correction Tags}

In order to incorporate correction processing into our language model,
we need to add some extra variables.  After we predict a repair, we need to
predict the reparandum onset.  Knowing the reparandum onset then
allows us to predict the word correspondences between the reparandum
and alteration, thus allowing us to use the repair to better predict
the words and their POS tags that make up the alteration.  Just as in
Chapter~\ref{chapter:detection}, we can view this as adding extra null
tokens that will be labeled with the correction tags.  In the rest of
this section, we introduce the variables that we tag.

\subsection{Reparandum Onset}
\label{sec:correction:onset}

If we have just predicted a modification repair or a fresh start, we
need to predict the reparandum onset.  We define the reparandum onset
tag $O_i$ as follows.\footnote{We are actually predicting the length
of the removed speech, which for overlapping repairs might not be the
same as the reparandum, as explained in Section~\ref{sec:corpus:overlap}.}
\[ O_i = \left\{
\begin{array}{ll}
{\bf null} & \parbox[t]{4in}{if $R_i \in \{{\bf null},{\bf Abr}\}$} \\
j          & \parbox[t]{4in}{if $W_j$ is the reparandum onset corresponding to  $R_i$}
\end{array}
\right.\]
This definition is not very useful for actually learning the
distribution because $O_i$ can take on so many different values.  The
longest speaker turn (in terms of the number of words) in the Trains
corpus involves 211 words (d93-26.5 utt48); hence $O_i$ can take on
potentially 210 different values.  Hence, there will not be enough
data to learn this distribution.

As an alternative, one could equivalently define the tag in terms of
the length of the reparandum.  The longest speech repair has a
reparandum length of 16 words (d93-13.1 utt56), and so a probability
distribution based on using the reparandum length will be much easier
to estimate.  However, even this probability distribution will be
difficult to estimate due to unnecessary data fragmentation.  Consider
the following two examples of modification repairs.
\begin{example}{d93-16.3 utt9}
to fill \reparandum{the engine}\ip the boxcars with bananas 
\end{example}
\begin{example}{d93-25.6 utt31}
drop off \reparandum{the one tanker}\ip the two tankers 
\end{example}
Although the examples differ in the length of the reparandum, their
reparanda both start at the beginning of a noun phrase.  This same
phenomena also exists for fresh starts where reparandum onsets are
likely to follow a boundary tone, the beginning of the turn, or a
discourse marker, rather than be of a particular reparandum length.

In order to allow generalizations across different reparandum lengths,
it is best not to define $O_i$ in terms of the reparandum length. A
better alternative is to query each potential onset individually to
see how likely it is as the onset, thus reducing the problem to a
binary classification problem. 
For $R_i \in\{{\bf Mod},{\bf Can}\}$ and $j < i$, we
define $O_{ij}$ as follows.
\[ O_{ij} = \left\{
\begin{array}{ll}
{\bf Onset} & \parbox[t]{4in}{if $W_j$ is the reparandum onset of repair $R_i$}  \\
{\bf null}  & \parbox[t]{4in}{otherwise}
\end{array}
\right.\]
The probability distribution for $O_{ij}$ is simply a reformulation of
that of $O_i$ as the two can be related by the following,
\begin{eqnarray}
\Pr(O_i = j|{\em Context})
  & = & \Pr(O_{iX} = {\bf Onset}|X\!=\!j\,{\em Context}) 
\end{eqnarray}
where $X$ is the variable that denotes the proposed onset of the reparandum.

\subsection{The Active Repair}
\label{sec:correction:active}

After we predict a speech repair and its reparandum, we need to
predict the words of the alteration.  As we discussed in
Section~\ref{sec:corpus:overlap}, 36\% of non-abridged repairs overlap
in the Trains corpus, and sometimes the correspondences are not just
with the removed speech of the last repair.  So, we need to decide
with which repair's removed speech to make the correspondence.  To
illustrate, consider Example~\ref {ex:d93-16.3:utt4}, repeated below.
\begin{example}{d93-16.3:utt4}
what's the shortest route from engine\ip from\ip for engine two at Elmira
\end{example}
For the second repair, it is ambiguous as to whether ``for'' should
correspond to the second instance of ``from'' (in the removed
speech of the second repair), or to the first instance (in the removed
speech of the first repair).  In either case, the second instance of
``engine'' corresponds to the word ``engine'', which is part of the
removed speech of the first repair.

The approach we take is to always choose the most recent repair that
has words in its removed speech that have not yet licensed a
correspondence (other than a word fragment).  Hence, the active repair
for predicting the word ``for'' is the second repair, while the active
repair for predicting the second instance of ``engine'' is the first
repair.  For predicting the word ``two'', neither the first nor the
second repair has any unlicensed words in their removed speech, and
hence it will not have an active repair.  In future work, we are
planning to allow it to choose between the removed speech of
alternative speech repairs, as is allowed by the annotation scheme.

\subsection{Licensing a Correspondence}
\label{sec:correction:license}

\begin{figure}
\centerline{\psfig{file=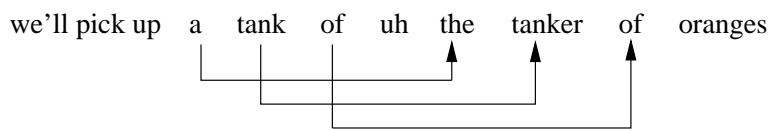}}
\caption{Cross Serial Correspondences between Reparandum and Alteration}
\label{fig:correction:cross}
\end{figure}

If we are in the midst of processing a repair, we can use the
reparandum to help predict the current word $W_i$ and its POS tag
$P_i$.  In order to do this, we need to determine which word in the
reparandum of the active repair will {\em license} the current word,
which we indicate by the tag $L_i$.  However, there are definite
restrictions as to which words of the reparandum can license a
correspondence.  As illustrated in Figure~\ref{fig:correction:cross},
word correspondences for speech repairs tend to exhibit a cross serial
dependency \cite{HeemanAllen94:acl}; in other words, if we have a
correspondence between $w_j$ in the reparandum and $w_k$ in the
alteration, any correspondence with a word in the alteration after
$w_k$ will be to a word that is after $w_j$, as illustrated in
Figure~\ref{fig:correction:cross}.  This regularity does have
exceptions, as the following example illustrates.
\begin{example}{d93-19.4 utt37}
\reparandum{can we have}\ip we can have three engines 
in Corning at the same time
\end{example}
However, such exceptions are rare, and so we currently do not support
such correspondences to predict the next word.  This means that if
there is already a correspondence from the reparandum to the
alteration then we can restrict $L_i$ to the words following the last
such correspondence in the reparandum.

To illustrate the use of $L_i$, consider the following example,
in which there is a deleted word in the reparandum (marked by
{\bf x} in the annotation scheme).
\begin{example}{d92a-1.2 utt40}
you can \reparandum{carry them both on}\ip tow both on the same engine 
\end{example}
In this example, when processing the word ``tow'', $L_i$ needs to
indicate that the word correspondence will be from the word ``carry''.
When processing the next word of the alteration, $L_i$ will need to
indicate that the next word of the reparandum, namely ``them'', should
be skipped over, and that the correspondence should be with the next
word in the reparandum, namely ``both''.

The above example illustrates how deleted words in the reparandum can
be handled.  There is also the problem of inserted words in the
alteration: words in the alteration for which there is no word in the
reparandum that licenses it. Consider the following example, in which
the word ``two'' is inserted in the alteration and hence has no
correspondence with any word in the reparandum.
\begin{example}{d93-15.4 utt45}
and fill \reparandum{my boxcars full of oranges}\ip my two boxcars full of oranges 
\end{example}
For such examples, we simply let $L_i$ point to the next word in the
alternation that could have a correspondence, in this case
``boxcars''.  We leave it to the next variable, the correspondence variable, to
encode that there is no correspondence.

Given the above, we define $L_i$ as follows.
\[ L_i = \left\{
\begin{array}{ll}
{\bf null} & \parbox[t]{4in}{There is no active repair} \\
0	   & \parbox[t]{4in}{$w_i$ is an inserted word for the active
repair} \\
i          & \parbox[t]{4in}{\setlength{\baselineskip}{1.3em}
			     $i$ words are deleted in the reparandum
                             of an active repair since the last correspondence}
\end{array}
\right. \]
Just as with estimating the probability of the reparandum onset, we
rephrase this probability so that it is a series of binary
classifications, querying each potential alternative.  We have found
from previous work \cite{HeemanAllen94:acl}, that if there is going to
be a word correspondence to the reparandum then it will be within 4 words.
Hence, we restrict this to just asking about the last four words in
order not to unduly increase the number of alternatives that need to
be considered.

\subsection{The Word Correspondence}

Now that we have decided which word in the reparandum will potentially
license the current word, we need to predict the type of
correspondence to this word.  We focus on correspondences involving
exact word match (identical POS tag and word), word
replacements (same POS tag), or no such correspondence.\footnote
{Predicting that there is no correspondence only applies if $L_i$ is
zero.}
\[ C_i = \left\{
\begin{array}{ll}
{\bf null} & \parbox[t]{4in}{There is no active repair} \\
{\bf x}    & \parbox[t]{4in}{$W_i$ is an inserted word for an active repair} \\
{\bf r}    & \parbox[t]{4in}{\setlength{\baselineskip}{1.3em}
			     $W_i$ is a word replacement of the word
			     indicated in the reparandum by $L_i$} \\
{\bf m}    & \parbox[t]{4in}{$W_i$ is a word match of the word
indicated in the reparandum by $L_i$}
\end{array}
\right. \]

\section{Redefining the Speech Recognition Problem}

Now that we have introduced the correction tags, we again need to
redefine the speech recognition problem.  Similar to
Section~\ref{sec:model:tagger} and Section~\ref{sec:detection:model},
we redefine the problem so that the goal is to find the sequence of
words, and the corresponding POS, tone, editing term, repair, and
correction tags that is most probable given the acoustic signal.
\newlength{\argmaxc}
\settowidth{\argmaxc}{\footnotesize $WPCLORET$}
\begin{eqnarray}
\lefteqn{\hat{W}\hat{P}\hat{C}\hat{L}\hat{O}\hat{R}\hat{E}\hat{T}} \nonumber \\
&=& \parbox[t]{\argmaxc}{\makebox[\argmaxc][c]{$\arg\max$}
		\footnotesize \vspace{-2.4em} \\ $WPCLORET$ \vspace*{-0.8em}}\
	 \Pr(WPCLORET|A) \nonumber \\
&=& \parbox[t]{\argmaxc}{\makebox[\argmaxc][c]{$\arg\max$}
		\footnotesize \vspace{-2.4em} \\ $WPCLORET$ \vspace*{-0.8em}}\
	 \frac{\Pr(A|WPCLORET) \Pr(WPCLORET)}{\Pr(A)} \nonumber \vspace*{1.5em} \\
&=& \parbox[t]{\argmaxc}{\makebox[\argmaxc][c]{$\arg\max$}
		\footnotesize \vspace{-2.4em} \\ $WPCLORET$}\
	 \Pr(A|WPCLORET) \Pr(WPCLORET) \label{eqn:correction:model} 
\end{eqnarray}\vspace*{-1em}\\
Again, the first term of Equation~\ref {eqn:correction:model} is the
acoustic model, and the second term is the language model.  We can
rewrite the language model term as
\[ \Pr(W_{1,N}P_{1,N}C_{1,N}L_{1,N}O_{1,N}R_{1,N}E_{1,N}T_{1,N}) \]
where $N$ is the number of words in the sequence.  We now rewrite
this term as the following.
\begin{eqnarray}
\lefteqn{\Pr(W_{1,N}P_{1,N}C_{1,N}L_{1,N}O_{1,N}R_{1,N}E_{1,N}T_{1,N})} \nonumber \\
&=&\prod_{i=1}^N\Pr(W_iP_iC_iL_iO_iR_iE_iT_i|
W_{\rim}P_{\rim}C_{\rim}L_{\rim}O_{\rim}R_{\rim}E_{\rim}T_{\rim}) \nonumber\\
&=&\prod_{i=1}^N\Pr(T_i|W_{\rim}P_{\rim}C_{\rim}L_{\rim}O_{\rim}R_{\rim}E_{\rim}T_{\rim}) \nonumber\\
  &&\hspace{1.5em}\Pr(E_i|W_{\rim}P_{\rim}C_{\rim}L_{\rim}O_{\rim}R_{\rim}E_{\rim}T_{\ri}) \nonumber\\
  &&\hspace{1.5em}\Pr(R_i|W_{\rim}P_{\rim}C_{\rim}L_{\rim}O_{\rim}R_{\rim}E_{\ri}T_{\ri}) \nonumber\\
  &&\hspace{1.5em}\Pr(O_i|W_{\rim}P_{\rim}C_{\rim}L_{\rim}O_{\rim}R_{\ri}E_{\ri}T_{\ri}) \nonumber\\
  &&\hspace{1.5em}\Pr(L_i|W_{\rim}P_{\rim}C_{\rim}L_{\rim}O_{\ri}R_{\ri}E_{\ri}T_{\ri}) \nonumber\\
  &&\hspace{1.5em}\Pr(C_i|W_{\rim}P_{\rim}C_{\rim}L_{\ri}O_{\ri}R_{\ri}E_{\ri}T_{\ri}) \nonumber\\
  &&\hspace{1.5em}\Pr(P_i|W_{\rim}P_{\rim}C_{\ri}L_{\ri}O_{\ri}R_{\ri}E_{\ri}T_{\ri}) \nonumber\\
  &&\hspace{1.5em}\Pr(W_i|W_{\rim}P_{\ri}C_{\ri}L_{\ri}O_{\ri}R_{\ri}E_{\ri}T_{\ri}) \label{eqn:correction:model2}
\end{eqnarray}

\section{Representing the Context}

From Equation~\ref {eqn:correction:model2} of the previous section, we
see that we now have three additional probability distributions that
we need to estimate, as well as a richer context for the other
distributions.  In this section, we first discuss how we alter the
utterance-sensitive word and POS tags, introduced in Section~\ref
{sec:detection:cleanup}, to take advantage of the additional context
provided by the correction variables.  We then describe the set of
variables that we use for estimating the probability distribution
for the reparandum onset, for the correspondence licensor, and for the
correspondence type.

\subsection{Utterance-Sensitive Word and POS Tags}

In Section~\ref{sec:detection:cleanup}, we presented the
utterance-sensitive word and POS tags, which are used to encode the
effect that the tone, editing term and repair tags have on the lexical
context.  One of the results of doing this was that after an editing
term is completed (as signaled by the editing term tag {\bf Pop}), we
can clean up the editing term and thus allow generalizations across
different editing term contexts.  Since this chapter added speech
repair correction, we can now do the same for the reparandum of speech
repairs.  This is in fact what Stolcke and Shriberg \shortcite
{StolckeShriberg96:icassp} do for the reparandum of simple repair
patterns. 

Consider the modification repair in the following example.
\begin{example}{d93-13.1 utt64}
pick up and load \reparandum{two}\ip um the two boxcars on engine two 
\end{example}
For the task of predicting the word ``the'' and its POS tag, if we
have predicted that it follows a modification repair with editing term
``um'' and reparandum ``two'', then we should be able to generalize
with fluent examples as the following.
\begin{example}{d93-12.4 utt97}
and  to make the orange juice and load the tankers 
\end{example}
Hence, for modification repairs, once the reparandum onset has been
predicted, we can clean up the words in the reparandum along with the
modification repair marker {\bf Mod}.  This will allow us to make use
of the second knowledge source that was identified in
Section~\ref{sec:correction:sources}.  

Cleaning up the reparandum of modification repairs also helps in
understanding overlapping repairs.  Consider the following example,
given earlier as Example~\ref {ex:d92a-2.1:utt95}.
\begin{example}{d92a-2.1 utt95}
\label{ex:d92a-2.1:utt95:a}
and that will take a $\underbrace{\makebox{total of}}_{\makebox[3em][r]{\em reparandum} \hfill}$\ip \et{um let's see} total 
$\underbrace{\makebox{of s-}}_{\makebox[2em][r]{\em reparandum \hfill}}$\ip of seven hours
\end{example}
In predicting the POS tag for the third instance of ``of'', the
utterance-sensitive words and POS tags will be as follows.
\begin{center}
\begin{tabular}{|l|c|c|c|c|c|c|} \hline
         & i-6     & i-5   & i-4  & i-3   & i-2   & i-1   \\ \hline
{\em pP} &\bf CC\_D&\bf DP &\bf MD&\bf VBP&\bf DT &\bf NN \\
{\em pW} & and     & that  & will & take  & a     & total \\ \hline
\end{tabular}
\end{center}
Note here that the reparandum of the current repair, involving the
words ``of s-'', has been cleaned up, as well as the reparandum of the
previous repair, ``total of''.  Hence, for predicting ``of'' and its
POS tag, we can make use of the cleaned-up linguistic context.  Of
course, this is only part of the context that is used for deciding the
current word and its POS tag.  Since we predict correspondences with
words of the reparandum, we also take this information into account
when predicting the POS tag and the word identity.

The above illustrates how modification repairs are handled.  Fresh
starts require a slightly different treatment.  Fresh starts are used
by speakers to abandon the current utterance, and hence the alteration
of a fresh start should be starting a new utterance.  But this new
utterance will start differently then most utterances in that it will
not begin with initial filled pauses, or phrases such as ``let's
see'', since these would have been counted as part of the editing term
of the fresh start.  Hence, we need to leave a indicator that a fresh
start occurred, which serves a similar purpose to marking that a
boundary tone occurred.  Hence, although we clean up the reparandum of
fresh starts, we do not clean up the fresh start marker {\bf Can},
just as we do not clean up the boundary tone marker {\bf Tone}.
Consider the following fresh start, which occurred at the beginning of
the speaker's turn.
\begin{example}{d92a-1.3 utt67}
\reparandum{is there}\ip there's no other quicker way right
\end{example}
After we hypothesize that a fresh start occurred after the first
instance of ``there'', and hypothesized its reparandum to be ``is
there'', the context for predicting the POS tag and word will include
the {\bf Can} tag, as indicated below.\footnote{Note that if the fresh
start marker follows a boundary tone marker, the boundary tone marker
is removed.}
\begin{center}
\begin{tabular}{|l|c|c|c|c|c|c|} \hline
         & i-1     \\ \hline
{\em pP} &\bf Can  \\
{\em pW} &\bf Can  \\ \hline
\end{tabular}
\end{center}

Just as the utterance-sensitive words and POS tags for predicting the
POS tag and tone take into account the reparandum of speech repairs,
so do the versions that capture the context for the tone, editing term
and repair tags. Below we give their values when predicting the tone,
editing term and repair tags associated with the third instance of
``of'' for Example~\ref {ex:d92a-2.1:utt95:a}.
\begin{center}
\begin{tabular}{|l|c|c|c|c|c|c|c|c|} \hline
         & i-8     & i-7   & i-6  & i-5   & i-4  & i-3  & i-2    & i-1\\ \hline
{\em tP} &\bf CC\_D&\bf DP &\bf MD&\bf VBP&\bf DT&\bf NN&\bf PREP&\bf FRAG \\
{\em tW} & and     & that  & will & take  & a    & total& of     & s- \\ \hline
{\em rP} &\bf CC\_D&\bf DP &\bf MD&\bf VBP&\bf DT&\bf NN&\bf PREP&\bf FRAG \\
{\em rW} & and     & that  & will & take  & a    & total& of     & s- \\ \hline
\end{tabular}
\end{center}
Note that since the second repair has not yet been predicted (nor its
reparandum onset), the context includes the reparandum of this repair,
but not that of the first repair.  Also note that since there is no
editing term in progress the two sets of variables have the same
value.

\subsection{Context for the Reparandum Onset}

Like the probability distributions for the tone, editing term and
repair tags, we need to define a set of relevant variables.
Table~\ref{tab:correction:Otree} gives a complete list of the
variables that are used, and Figure~\ref{fig:correction:Otree} gives
the top part of the decision tree that was grown for the first
partition of the training data.
\begin{table}
\begin{description}
\item [$\mbox{\em R}$:] The repair tag $R_i$, which will either be a 
modification repair or a fresh start.
\item [$\mbox{\em Len}$:] Length of the proposed reparandum, not 
including words that are part of the reparandum or editing term of an
already resolved repair.
\item [$\mbox{\em OP}$:] POS tag of the proposed onset word.
\item [$\mbox{\em oP}$:] Utterance sensitive POS tags of context prior to the 
proposed onset.  Note that after predicting a fresh start and its
reparandum onset, {\em pP} will always end with the fresh start marker
{\bf Can}.  The variables {\em oP} differ from {\em pP} in that this
final {\bf Can} marker is not appended.	
\item [$\mbox{\em oW}$:] Utterance sensitive word identities of context prior 
to the proposed onset.  See {\em oP} for how these differ from {\em pW}.
\item [$\mbox{\em Tones}$:] Number of boundary tones that are embedded in the 
proposed reparandum. If the last word of the reparandum has a boundary
tone, it is not included in the count.
\item [$\mbox{\em DM}$:] Number of discourse markers that are embedded 
in the reparandum. If more than one discourse marker appears in a
role, the stretch of discourse markers is only counted once. Discourse
markers at the beginning of the reparandum are not counted.
\item [$\mbox{\em FP}$:] Number of filled-pause words that the 
reparandum contains.
\item [$\mbox{\em Prev}$:] Indicates whether there is a previous repair, 
and if so whether the proposed reparandum onset is earlier, coincides
with, or is after the alteration onset of that repair.
\end{description}
\caption{Variables used for Predicting Reparandum Onset}
\label{tab:correction:Otree}
\end{table}
\begin{figure}
\begin{picture}(0,0)%
\includegraphics{Otree.pstex}%
\end{picture}%
\setlength{\unitlength}{0.01250000in}%
\begingroup\makeatletter\ifx\SetFigFont\undefined
\def\x#1#2#3#4#5#6#7\relax{\def\x{#1#2#3#4#5#6}}%
\expandafter\x\fmtname xxxxxx\relax \def\y{splain}%
\ifx\x\y   
\gdef\SetFigFont#1#2#3{%
  \ifnum #1<17\tiny\else \ifnum #1<20\small\else
  \ifnum #1<24\normalsize\else \ifnum #1<29\large\else
  \ifnum #1<34\Large\else \ifnum #1<41\LARGE\else
     \huge\fi\fi\fi\fi\fi\fi
  \csname #3\endcsname}%
\else
\gdef\SetFigFont#1#2#3{\begingroup
  \count@#1\relax \ifnum 25<\count@\count@25\fi
  \def\x{\endgroup\@setsize\SetFigFont{#2pt}}%
  \expandafter\x
    \csname \romannumeral\the\count@ pt\expandafter\endcsname
    \csname @\romannumeral\the\count@ pt\endcsname
  \csname #3\endcsname}%
\fi
\fi\endgroup
\begin{picture}(287,616)(21,203)
\put(192,806){\makebox(0,0)[lb]{\smash{\SetFigFont{6}{7.2}{rm} leaf}}}
\put(249,782){\makebox(0,0)[lb]{\smash{\SetFigFont{6}{7.2}{rm} leaf}}}
\put(306,758){\makebox(0,0)[lb]{\smash{\SetFigFont{6}{7.2}{rm} leaf}}}
\put(306,734){\makebox(0,0)[lb]{\smash{\SetFigFont{6}{7.2}{rm} $\cdots$}}}
\put(249,746){\makebox(0,0)[lb]{\smash{\SetFigFont{6}{7.2}{rm} is $\mbox{\em Prev}\!\!\in\!\!\{\mbox{\bf null,=$\!$}\}$}}}
\put(192,764){\makebox(0,0)[lb]{\smash{\SetFigFont{6}{7.2}{rm} is $\mbox{\em Tones}\!\!\geq\!\!1 \wedge \mbox{\em Len}\!\!\geq\!\!7 \vee \mbox{\em Len}\!\!<\!\!5 \wedge \mbox{\em Len}\!\!\geq\!\!3$}}}
\put(135,785){\makebox(0,0)[lb]{\smash{\SetFigFont{6}{7.2}{rm} is $\mbox{\em Len}\!\!\geq\!\!10 \vee \mbox{\em Tones}\!\!\geq\!\!2$}}}
\put(306,710){\makebox(0,0)[lb]{\smash{\SetFigFont{6}{7.2}{rm} leaf}}}
\put(306,686){\makebox(0,0)[lb]{\smash{\SetFigFont{6}{7.2}{rm} leaf}}}
\put(249,698){\makebox(0,0)[lb]{\smash{\SetFigFont{6}{7.2}{rm} is $\mbox{\em OP}^1\!\!=\!\!1 \vee \mbox{\em OP}^2\!\!=\!\!0 \wedge \mbox{\em Prev}\!\!\in\!\!\{\mbox{\bf null,$<$,$>$$\!$}\}$}}}
\put(249,662){\makebox(0,0)[lb]{\smash{\SetFigFont{6}{7.2}{rm} leaf}}}
\put(192,680){\makebox(0,0)[lb]{\smash{\SetFigFont{6}{7.2}{rm} is ${\em oP}_{\mbox{$i$-1}}^5\!\!=\!\!1 \vee \mbox{\em DM}\!\!\geq\!\!1$}}}
\put(306,638){\makebox(0,0)[lb]{\smash{\SetFigFont{6}{7.2}{rm} $\cdots$}}}
\put(306,614){\makebox(0,0)[lb]{\smash{\SetFigFont{6}{7.2}{rm} $\cdots$}}}
\put(249,626){\makebox(0,0)[lb]{\smash{\SetFigFont{6}{7.2}{rm} is $\mbox{\em Len}\!\!<\!\!7 \vee \mbox{\em Len}\!\!\geq\!\!10 \vee {\em oP}_{\mbox{$i$-1}}^3\!\!=\!\!1$}}}
\put(249,590){\makebox(0,0)[lb]{\smash{\SetFigFont{6}{7.2}{rm} leaf}}}
\put(192,608){\makebox(0,0)[lb]{\smash{\SetFigFont{6}{7.2}{rm} is $\mbox{\em Prev}\!\!\in\!\!\{\mbox{\bf null,=,$>$$\!$}\}$}}}
\put(135,644){\makebox(0,0)[lb]{\smash{\SetFigFont{6}{7.2}{rm} is ${\em oP}_{\mbox{$i$-1}}^3\!\!=\!\!1 \wedge \mbox{\em Tones}\!\!<\!\!1$}}}
\put( 78,715){\makebox(0,0)[lb]{\smash{\SetFigFont{6}{7.2}{rm} is $\mbox{\em R}\!\!\in\!\!\{\mbox{\bf Mod$\!$}\} \vee {\em oP}_{\mbox{$i$-1}}^1\!\!=\!\!1 \vee {\em oP}_{\mbox{$i$-1}}^2\!\!=\!\!1 \vee \mbox{\em Tones}\!\!\geq\!\!2 \vee \mbox{\em Len}\!\!<\!\!3 \vee \mbox{\em FP}\!\!\geq\!\!1$}}}
\put(306,566){\makebox(0,0)[lb]{\smash{\SetFigFont{6}{7.2}{rm} leaf}}}
\put(306,542){\makebox(0,0)[lb]{\smash{\SetFigFont{6}{7.2}{rm} leaf}}}
\put(249,554){\makebox(0,0)[lb]{\smash{\SetFigFont{6}{7.2}{rm} is $\mbox{\em Prev}\!\!\in\!\!\{\mbox{\bf null,$>$$\!$}\} \wedge \mbox{\em Len}\!\!<\!\!3$}}}
\put(306,518){\makebox(0,0)[lb]{\smash{\SetFigFont{6}{7.2}{rm} $\cdots$}}}
\put(306,494){\makebox(0,0)[lb]{\smash{\SetFigFont{6}{7.2}{rm} $\cdots$}}}
\put(249,506){\makebox(0,0)[lb]{\smash{\SetFigFont{6}{7.2}{rm} is $\mbox{\em Prev}\!\!\in\!\!\{\mbox{\bf null,$>$$\!$}\} \wedge \mbox{\em Len}\!\!<\!\!3 \vee \mbox{\em Len}\!\!<\!\!3$}}}
\put(192,530){\makebox(0,0)[lb]{\smash{\SetFigFont{6}{7.2}{rm} is ${\em oP}_{\mbox{$i$-1}}^1\!\!=\!\!1 \vee \mbox{\em Len}\!\!\geq\!\!3 \wedge \mbox{\em OP}^1\!\!=\!\!1 \wedge \mbox{\em OP}^2\!\!=\!\!1$}}}
\put(306,470){\makebox(0,0)[lb]{\smash{\SetFigFont{6}{7.2}{rm} $\cdots$}}}
\put(306,446){\makebox(0,0)[lb]{\smash{\SetFigFont{6}{7.2}{rm} $\cdots$}}}
\put(249,458){\makebox(0,0)[lb]{\smash{\SetFigFont{6}{7.2}{rm} is $\mbox{\em OP}^3\!\!=\!\!0$}}}
\put(306,422){\makebox(0,0)[lb]{\smash{\SetFigFont{6}{7.2}{rm} $\cdots$}}}
\put(306,398){\makebox(0,0)[lb]{\smash{\SetFigFont{6}{7.2}{rm} leaf}}}
\put(249,410){\makebox(0,0)[lb]{\smash{\SetFigFont{6}{7.2}{rm} is $\mbox{\em OP}^1\!\!=\!\!1 \vee {\em oP}_{\mbox{$i$-1}}^1\!\!=\!\!0$}}}
\put(192,434){\makebox(0,0)[lb]{\smash{\SetFigFont{6}{7.2}{rm} is $\mbox{\em Len}\!\!<\!\!3 \wedge \mbox{\em R}\!\!\in\!\!\{\mbox{\bf Mod$\!$}\} \wedge \mbox{\em OP}^2\!\!=\!\!1$}}}
\put(135,482){\makebox(0,0)[lb]{\smash{\SetFigFont{6}{7.2}{rm} is $\mbox{\em OP}^1\!\!=\!\!1 \vee {\em oP}_{\mbox{$i$-1}}^1\!\!=\!\!1 \wedge \mbox{\em Len}\!\!<\!\!5 \wedge \mbox{\em R}\!\!\in\!\!\{\mbox{\bf Mod$\!$}\}$}}}
\put(306,374){\makebox(0,0)[lb]{\smash{\SetFigFont{6}{7.2}{rm} $\cdots$}}}
\put(306,350){\makebox(0,0)[lb]{\smash{\SetFigFont{6}{7.2}{rm} $\cdots$}}}
\put(249,362){\makebox(0,0)[lb]{\smash{\SetFigFont{6}{7.2}{rm} is $\mbox{\em Prev}\!\!\in\!\!\{\mbox{\bf null,=,$>$$\!$}\} \wedge \mbox{\em R}\!\!\in\!\!\{\mbox{\bf Mod$\!$}\}$}}}
\put(306,326){\makebox(0,0)[lb]{\smash{\SetFigFont{6}{7.2}{rm} $\cdots$}}}
\put(306,302){\makebox(0,0)[lb]{\smash{\SetFigFont{6}{7.2}{rm} leaf}}}
\put(249,314){\makebox(0,0)[lb]{\smash{\SetFigFont{6}{7.2}{rm} is ${\em oP}_{\mbox{$i$-1}}^5\!\!=\!\!1$}}}
\put(192,338){\makebox(0,0)[lb]{\smash{\SetFigFont{6}{7.2}{rm} is ${\em oP}_{\mbox{$i$-1}}^1\!\!=\!\!1 \vee \mbox{\em Len}\!\!\geq\!\!2 \vee {\em oP}_{\mbox{$i$-1}}^2\!\!=\!\!1 \vee {\em oP}_{\mbox{$i$-1}}^3\!\!=\!\!0$}}}
\put(306,278){\makebox(0,0)[lb]{\smash{\SetFigFont{6}{7.2}{rm} $\cdots$}}}
\put(306,254){\makebox(0,0)[lb]{\smash{\SetFigFont{6}{7.2}{rm} $\cdots$}}}
\put(249,266){\makebox(0,0)[lb]{\smash{\SetFigFont{6}{7.2}{rm} is ${\em oP}_{\mbox{$i$-1}}^1\!\!=\!\!1 \vee {\em oP}_{\mbox{$i$-1}}^2\!\!=\!\!1$}}}
\put(306,230){\makebox(0,0)[lb]{\smash{\SetFigFont{6}{7.2}{rm} $\cdots$}}}
\put(306,206){\makebox(0,0)[lb]{\smash{\SetFigFont{6}{7.2}{rm} leaf}}}
\put(249,218){\makebox(0,0)[lb]{\smash{\SetFigFont{6}{7.2}{rm} is $\mbox{\em Len}\!\!\geq\!\!5 \vee {\em oP}_{\mbox{$i$-1}}^5\!\!=\!\!0 \wedge \mbox{\em Len}\!\!\geq\!\!4$}}}
\put(192,242){\makebox(0,0)[lb]{\smash{\SetFigFont{6}{7.2}{rm} is ${\em oP}_{\mbox{$i$-1}}^1\!\!=\!\!1 \vee \mbox{\em R}\!\!\in\!\!\{\mbox{\bf Mod$\!$}\} \vee {\em oP}_{\mbox{$i$-1}}^2\!\!=\!\!1 \vee {\em oP}_{\mbox{$i$-1}}^3\!\!=\!\!0$}}}
\put(135,290){\makebox(0,0)[lb]{\smash{\SetFigFont{6}{7.2}{rm} is $\mbox{\em R}\!\!\in\!\!\{\mbox{\bf Mod$\!$}\} \vee \mbox{\em Len}\!\!\geq\!\!2 \wedge \mbox{\em Len}\!\!<\!\!4$}}}
\put( 78,386){\makebox(0,0)[lb]{\smash{\SetFigFont{6}{7.2}{rm} is ${\em oP}_{\mbox{$i$-1}}^1\!\!=\!\!1 \vee {\em oP}_{\mbox{$i$-1}}^2\!\!=\!\!1 \wedge \mbox{\em Len}\!\!\geq\!\!2$}}}
\put( 21,551){\makebox(0,0)[lb]{\smash{\SetFigFont{6}{7.2}{rm} is $\mbox{\em FP}\!\!\geq\!\!1 \vee \mbox{\em Tones}\!\!\geq\!\!1 \vee \mbox{\em DM}\!\!\geq\!\!1 \vee \mbox{\em Len}\!\!\geq\!\!8$}}}
\end{picture}
\caption{Decision Tree for Reparandum Onset}
\label{fig:correction:Otree}
\end{figure}%
The reparandum onset, as explained in Section~\ref
{sec:correction:onset}, is estimating by querying each potential onset
$X$ to see how likely it is to be the onset. 
\[ \Pr(O_{iX} = \mbox{\bf Onset}|XW_{\rim}P_{\rim}C_{\rim}L_{\rim}O_{\rim}R_{\ri}E_{\ri}T_{\ri}) \]
Hence, the context used in estimating this distribution must indicate
which word we are querying.  The variable ${\em OP}$ is the POS tag of
the proposed reparandum onset X. The variables {\em oP} and {\em oW}
are the utterance sensitive POS tags and words.  These are computed
under the assumption that $X$ is in fact the reparandum onset, and
hence they provide the context prior to the onset.  Thus these
variables together with {\em OP} allow the decision tree to check if
the onset is at a suitable constituent boundary.  The variables {\em
oP} and {\em oW} are slight variations of {\em pP} and {\em pW}: for
fresh starts, {\em oP} and {\em oW} do not end with the fresh start
marker {\bf Can}.

Reparanda of speech repairs rarely extend over two utterance units;
rather, they tend to act only in the current utterance.  Hence,
we include three variables that help indicate whether the proposed
reparanda is crossing an utterance boundary.  The variable {\em Tones}
indicates the number of boundary tones that the proposed reparandum
includes.  Note that the interruption point of a speech repair might
occur on a word that has been marked with a boundary tone, as was
illustrated in Example~\ref {ex:d92a-2.1:utt29}.  Such boundary tones
are not included in {\em Tones}.  The second variable is {\em DM},
which counts the number of discourse markers in the reparandum.
Discourse markers at the beginning of the reparandum are not included,
and if discourse markers appear consecutively, the band is only
counted once.  The third variable is {\em FP}, which counts the number
of filled pause markers in the reparandum.  As with boundary tones and
discourse markers, filled pauses that are part of the editing term or
reparandum of a prior repair are not counted.

Another source of information is the presence of other repairs in the
turn. As indicated by Table~\ref{tab:corpus:overlap}, if a repair
overlaps a previous one then its reparandum onset is likely to
co-occur with the alteration onset of the previous repair. The
variable {\em Prev} indicates whether there is a previous repair, and
if there is, whether the proposed onset coincides with, is earlier
than, or is later than the alteration onset of the preceding repair.

\subsection{Context for the Correspondence Licensor}

\begin{figure}
\begin{picture}(0,0)%
\includegraphics{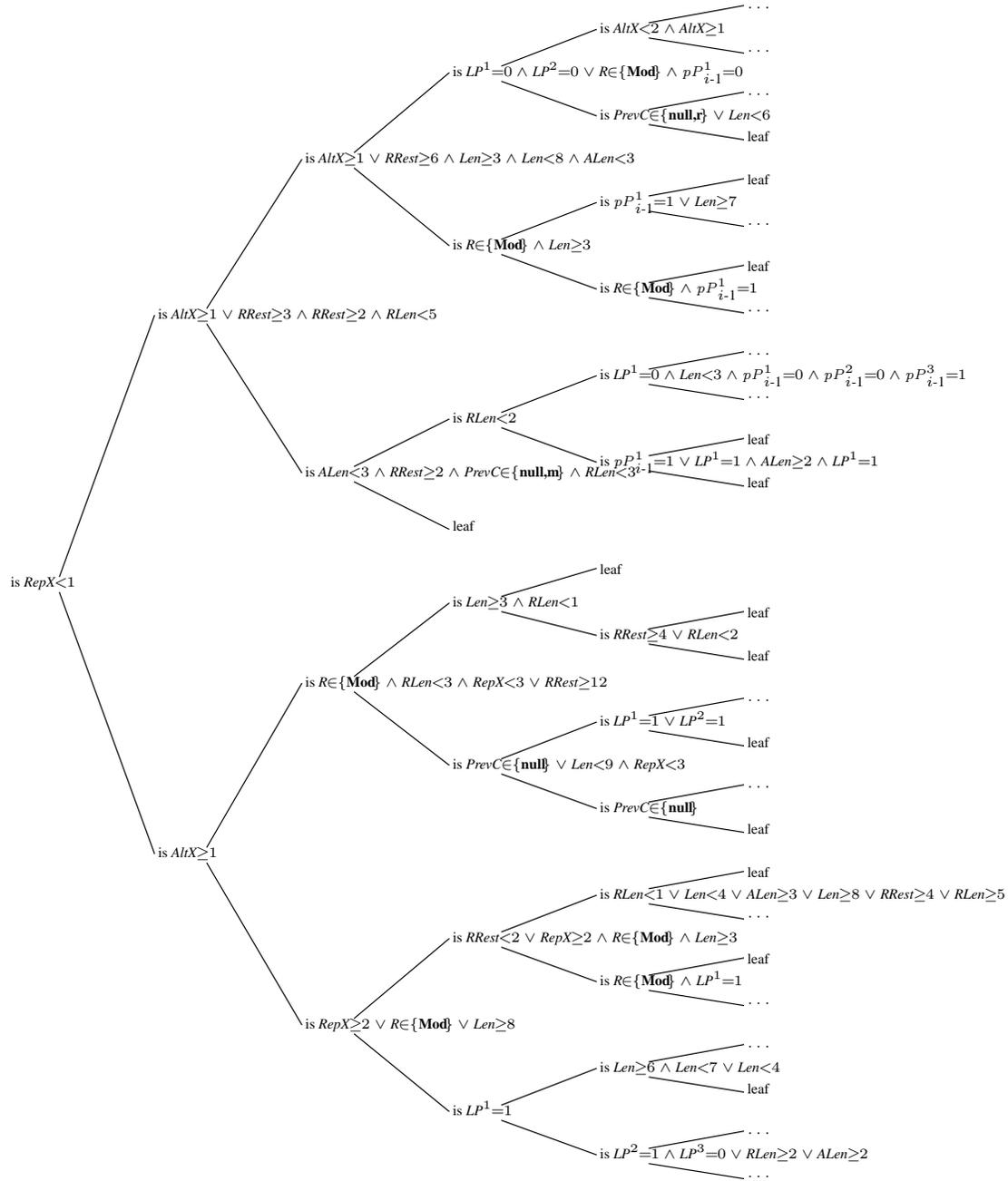}%
\end{picture}%
\setlength{\unitlength}{0.01250000in}%
\begingroup\makeatletter\ifx\SetFigFont\undefined
\def\x#1#2#3#4#5#6#7\relax{\def\x{#1#2#3#4#5#6}}%
\expandafter\x\fmtname xxxxxx\relax \def\y{splain}%
\ifx\x\y   
\gdef\SetFigFont#1#2#3{%
  \ifnum #1<17\tiny\else \ifnum #1<20\small\else
  \ifnum #1<24\normalsize\else \ifnum #1<29\large\else
  \ifnum #1<34\Large\else \ifnum #1<41\LARGE\else
     \huge\fi\fi\fi\fi\fi\fi
  \csname #3\endcsname}%
\else
\gdef\SetFigFont#1#2#3{\begingroup
  \count@#1\relax \ifnum 25<\count@\count@25\fi
  \def\x{\endgroup\@setsize\SetFigFont{#2pt}}%
  \expandafter\x
    \csname \romannumeral\the\count@ pt\expandafter\endcsname
    \csname @\romannumeral\the\count@ pt\endcsname
  \csname #3\endcsname}%
\fi
\fi\endgroup
\begin{picture}(342,556)(21,255)
\put(361,798){\makebox(0,0)[lb]{\smash{\SetFigFont{6}{7.2}{rm} $\cdots$}}}
\put(361,778){\makebox(0,0)[lb]{\smash{\SetFigFont{6}{7.2}{rm} $\cdots$}}}
\put(293,788){\makebox(0,0)[lb]{\smash{\SetFigFont{6}{7.2}{rm} is $\mbox{\em AltX}\!\!<\!\!2 \wedge \mbox{\em AltX}\!\!\geq\!\!1$}}}
\put(361,758){\makebox(0,0)[lb]{\smash{\SetFigFont{6}{7.2}{rm} $\cdots$}}}
\put(361,738){\makebox(0,0)[lb]{\smash{\SetFigFont{6}{7.2}{rm} leaf}}}
\put(293,748){\makebox(0,0)[lb]{\smash{\SetFigFont{6}{7.2}{rm} is $\mbox{\em PrevC}\!\!\in\!\!\{\mbox{\bf null,r$\!$}\} \vee \mbox{\em Len}\!\!<\!\!6$}}}
\put(225,768){\makebox(0,0)[lb]{\smash{\SetFigFont{6}{7.2}{rm} is $\mbox{\em LP}^1\!\!=\!\!0 \wedge \mbox{\em LP}^2\!\!=\!\!0 \vee \mbox{\em R}\!\!\in\!\!\{\mbox{\bf Mod$\!$}\} \wedge {\em pP}_{\mbox{$i$-1}}^1\!\!=\!\!0$}}}
\put(361,718){\makebox(0,0)[lb]{\smash{\SetFigFont{6}{7.2}{rm} leaf}}}
\put(361,698){\makebox(0,0)[lb]{\smash{\SetFigFont{6}{7.2}{rm} $\cdots$}}}
\put(293,708){\makebox(0,0)[lb]{\smash{\SetFigFont{6}{7.2}{rm} is ${\em pP}_{\mbox{$i$-1}}^1\!\!=\!\!1 \vee \mbox{\em Len}\!\!\geq\!\!7$}}}
\put(361,678){\makebox(0,0)[lb]{\smash{\SetFigFont{6}{7.2}{rm} leaf}}}
\put(361,658){\makebox(0,0)[lb]{\smash{\SetFigFont{6}{7.2}{rm} $\cdots$}}}
\put(293,668){\makebox(0,0)[lb]{\smash{\SetFigFont{6}{7.2}{rm} is $\mbox{\em R}\!\!\in\!\!\{\mbox{\bf Mod$\!$}\} \wedge {\em pP}_{\mbox{$i$-1}}^1\!\!=\!\!1$}}}
\put(225,688){\makebox(0,0)[lb]{\smash{\SetFigFont{6}{7.2}{rm} is $\mbox{\em R}\!\!\in\!\!\{\mbox{\bf Mod$\!$}\} \wedge \mbox{\em Len}\!\!\geq\!\!3$}}}
\put(157,728){\makebox(0,0)[lb]{\smash{\SetFigFont{6}{7.2}{rm} is $\mbox{\em AltX}\!\!\geq\!\!1 \vee \mbox{\em RRest}\!\!\geq\!\!6 \wedge \mbox{\em Len}\!\!\geq\!\!3 \wedge \mbox{\em Len}\!\!<\!\!8 \wedge \mbox{\em ALen}\!\!<\!\!3$}}}
\put(361,638){\makebox(0,0)[lb]{\smash{\SetFigFont{6}{7.2}{rm} $\cdots$}}}
\put(361,618){\makebox(0,0)[lb]{\smash{\SetFigFont{6}{7.2}{rm} $\cdots$}}}
\put(293,628){\makebox(0,0)[lb]{\smash{\SetFigFont{6}{7.2}{rm} is $\mbox{\em LP}^1\!\!=\!\!0 \wedge \mbox{\em Len}\!\!<\!\!3 \wedge {\em pP}_{\mbox{$i$-1}}^1\!\!=\!\!0 \wedge {\em pP}_{\mbox{$i$-1}}^2\!\!=\!\!0 \wedge {\em pP}_{\mbox{$i$-1}}^3\!\!=\!\!1$}}}
\put(361,598){\makebox(0,0)[lb]{\smash{\SetFigFont{6}{7.2}{rm} leaf}}}
\put(361,578){\makebox(0,0)[lb]{\smash{\SetFigFont{6}{7.2}{rm} leaf}}}
\put(293,588){\makebox(0,0)[lb]{\smash{\SetFigFont{6}{7.2}{rm} is ${\em pP}_{\mbox{$i$-1}}^1\!\!=\!\!1 \vee \mbox{\em LP}^1\!\!=\!\!1 \wedge \mbox{\em ALen}\!\!\geq\!\!2 \wedge \mbox{\em LP}^1\!\!=\!\!1$}}}
\put(225,608){\makebox(0,0)[lb]{\smash{\SetFigFont{6}{7.2}{rm} is $\mbox{\em RLen}\!\!<\!\!2$}}}
\put(225,558){\makebox(0,0)[lb]{\smash{\SetFigFont{6}{7.2}{rm} leaf}}}
\put(157,583){\makebox(0,0)[lb]{\smash{\SetFigFont{6}{7.2}{rm} is $\mbox{\em ALen}\!\!<\!\!3 \wedge \mbox{\em RRest}\!\!\geq\!\!2 \wedge \mbox{\em PrevC}\!\!\in\!\!\{\mbox{\bf null,m$\!$}\} \wedge \mbox{\em RLen}\!\!<\!\!3$}}}
\put( 89,656){\makebox(0,0)[lb]{\smash{\SetFigFont{6}{7.2}{rm} is $\mbox{\em AltX}\!\!\geq\!\!1 \vee \mbox{\em RRest}\!\!\geq\!\!3 \wedge \mbox{\em RRest}\!\!\geq\!\!2 \wedge \mbox{\em RLen}\!\!<\!\!5$}}}
\put(293,538){\makebox(0,0)[lb]{\smash{\SetFigFont{6}{7.2}{rm} leaf}}}
\put(361,518){\makebox(0,0)[lb]{\smash{\SetFigFont{6}{7.2}{rm} leaf}}}
\put(361,498){\makebox(0,0)[lb]{\smash{\SetFigFont{6}{7.2}{rm} leaf}}}
\put(293,508){\makebox(0,0)[lb]{\smash{\SetFigFont{6}{7.2}{rm} is $\mbox{\em RRest}\!\!\geq\!\!4 \vee \mbox{\em RLen}\!\!<\!\!2$}}}
\put(225,523){\makebox(0,0)[lb]{\smash{\SetFigFont{6}{7.2}{rm} is $\mbox{\em Len}\!\!\geq\!\!3 \wedge \mbox{\em RLen}\!\!<\!\!1$}}}
\put(361,478){\makebox(0,0)[lb]{\smash{\SetFigFont{6}{7.2}{rm} $\cdots$}}}
\put(361,458){\makebox(0,0)[lb]{\smash{\SetFigFont{6}{7.2}{rm} leaf}}}
\put(293,468){\makebox(0,0)[lb]{\smash{\SetFigFont{6}{7.2}{rm} is $\mbox{\em LP}^1\!\!=\!\!1 \vee \mbox{\em LP}^2\!\!=\!\!1$}}}
\put(361,438){\makebox(0,0)[lb]{\smash{\SetFigFont{6}{7.2}{rm} $\cdots$}}}
\put(361,418){\makebox(0,0)[lb]{\smash{\SetFigFont{6}{7.2}{rm} leaf}}}
\put(293,428){\makebox(0,0)[lb]{\smash{\SetFigFont{6}{7.2}{rm} is $\mbox{\em PrevC}\!\!\in\!\!\{\mbox{\bf null$\!$}\}$}}}
\put(225,448){\makebox(0,0)[lb]{\smash{\SetFigFont{6}{7.2}{rm} is $\mbox{\em PrevC}\!\!\in\!\!\{\mbox{\bf null$\!$}\} \vee \mbox{\em Len}\!\!<\!\!9 \wedge \mbox{\em RepX}\!\!<\!\!3$}}}
\put(157,486){\makebox(0,0)[lb]{\smash{\SetFigFont{6}{7.2}{rm} is $\mbox{\em R}\!\!\in\!\!\{\mbox{\bf Mod$\!$}\} \wedge \mbox{\em RLen}\!\!<\!\!3 \wedge \mbox{\em RepX}\!\!<\!\!3 \vee \mbox{\em RRest}\!\!\geq\!\!12$}}}
\put(361,398){\makebox(0,0)[lb]{\smash{\SetFigFont{6}{7.2}{rm} leaf}}}
\put(361,378){\makebox(0,0)[lb]{\smash{\SetFigFont{6}{7.2}{rm} $\cdots$}}}
\put(293,388){\makebox(0,0)[lb]{\smash{\SetFigFont{6}{7.2}{rm} is $\mbox{\em RLen}\!\!<\!\!1 \vee \mbox{\em Len}\!\!<\!\!4 \vee \mbox{\em ALen}\!\!\geq\!\!3 \vee \mbox{\em Len}\!\!\geq\!\!8 \vee \mbox{\em RRest}\!\!\geq\!\!4 \vee \mbox{\em RLen}\!\!\geq\!\!5$}}}
\put(361,358){\makebox(0,0)[lb]{\smash{\SetFigFont{6}{7.2}{rm} leaf}}}
\put(361,338){\makebox(0,0)[lb]{\smash{\SetFigFont{6}{7.2}{rm} $\cdots$}}}
\put(293,348){\makebox(0,0)[lb]{\smash{\SetFigFont{6}{7.2}{rm} is $\mbox{\em R}\!\!\in\!\!\{\mbox{\bf Mod$\!$}\} \wedge \mbox{\em LP}^1\!\!=\!\!1$}}}
\put(225,368){\makebox(0,0)[lb]{\smash{\SetFigFont{6}{7.2}{rm} is $\mbox{\em RRest}\!\!<\!\!2 \vee \mbox{\em RepX}\!\!\geq\!\!2 \wedge \mbox{\em R}\!\!\in\!\!\{\mbox{\bf Mod$\!$}\} \wedge \mbox{\em Len}\!\!\geq\!\!3$}}}
\put(361,318){\makebox(0,0)[lb]{\smash{\SetFigFont{6}{7.2}{rm} $\cdots$}}}
\put(361,298){\makebox(0,0)[lb]{\smash{\SetFigFont{6}{7.2}{rm} leaf}}}
\put(293,308){\makebox(0,0)[lb]{\smash{\SetFigFont{6}{7.2}{rm} is $\mbox{\em Len}\!\!\geq\!\!6 \wedge \mbox{\em Len}\!\!<\!\!7 \vee \mbox{\em Len}\!\!<\!\!4$}}}
\put(361,278){\makebox(0,0)[lb]{\smash{\SetFigFont{6}{7.2}{rm} $\cdots$}}}
\put(361,258){\makebox(0,0)[lb]{\smash{\SetFigFont{6}{7.2}{rm} $\cdots$}}}
\put(293,268){\makebox(0,0)[lb]{\smash{\SetFigFont{6}{7.2}{rm} is $\mbox{\em LP}^2\!\!=\!\!1 \wedge \mbox{\em LP}^3\!\!=\!\!0 \vee \mbox{\em RLen}\!\!\geq\!\!2 \vee \mbox{\em ALen}\!\!\geq\!\!2$}}}
\put(225,288){\makebox(0,0)[lb]{\smash{\SetFigFont{6}{7.2}{rm} is $\mbox{\em LP}^1\!\!=\!\!1$}}}
\put(157,328){\makebox(0,0)[lb]{\smash{\SetFigFont{6}{7.2}{rm} is $\mbox{\em RepX}\!\!\geq\!\!2 \vee \mbox{\em R}\!\!\in\!\!\{\mbox{\bf Mod$\!$}\} \vee \mbox{\em Len}\!\!\geq\!\!8$}}}
\put( 89,407){\makebox(0,0)[lb]{\smash{\SetFigFont{6}{7.2}{rm} is $\mbox{\em AltX}\!\!\geq\!\!1$}}}
\put( 21,532){\makebox(0,0)[lb]{\smash{\SetFigFont{6}{7.2}{rm} is $\mbox{\em RepX}\!\!<\!\!1$}}}
\end{picture}
\caption{Decision Tree for Correspondence Licensor}
\label{fig:correction:Ltree}
\end{figure}

Like the previous probability distributions, we define a set of
relevant variables that encode the context, which are given in
Table~\ref{tab:correction:Ltree}.  Figure~\ref{fig:correction:Ltree}
gives the top part of the decision tree for the correspondence licensor.
\begin{table}
\begin{description}
\item [$\mbox{\em LP}$:] POS tag of licensing word.
\item [$\mbox{\em pP}$:] Utterance sensitive POS tags of words preceding $W_i$.
\item [$\mbox{\em pW}$:] Utterance sensitive word identities of words 
preceding $W_i$.
\item [$\mbox{\em R}$:] Type of repair ({\bf Mod} or {\bf Can}).
\item [$\mbox{\em Len}$:] Length of reparandum of repair.
\item [$\mbox{\em RLen}$:] Length of partial reparandum accounted for so far.
the partial reparandum.
\item [$\mbox{\em RRest}$:] Length of reparandum not accounted for yet.
\item [$\mbox{\em ALen}$:] Length of partial alteration seen so far.
\item [$\mbox{\em PrevC}$:] Type of previous replacement or matching 
correspondence ({\bf m} or {\bf r}).
\item [$\mbox{\em RepX}$:] Number of {\bf x} correspondences at the end of
\item [$\mbox{\em AltX}$:] Number of {\bf x} correspondences at the end of
the partition alteration.
\end{description}
\caption{Variables used for Predicting Correspondence Licensor}
\label{tab:correction:Ltree}
\end{table}
Just as with the reparandum onset, we estimate the probability of
which word in the alteration licenses the current word in the
reparandum by querying each eligible word $X$. The context that we use
to condition this query must obviously include information about the
word in the reparandum that it is being proposed.
\[ \Pr(L_{iX}=\mbox{\bf Licensor}|XW_{\rim}P_{\rim}C_{\rim}L_{\rim}O_{\ri}R_{\ri}E_{\ri}T_{\ri}) \]
The variable {\em LP} indicates the POS tag of $X$.  We also take into
account the utterance-sensitive POS and word context prior
to the current word, the type of repair, and the reparandum length.

We also take into account information about the repair structure that
has been found so far.  If the previous word was a word match, there
is a good chance that the current word will involve a word match to
the next word. The features that we use are the number of words that
are skipped in the reparandum ({\em RepX}) and alteration ({\em AltX})
since the last correspondence of this repair, and the type of the last
correspondence ({\em PrevC}): whether it was a word match or word
replacement.  We also take into account the number of words since the
onset of the reparandum ({\em RLen}) and alteration ({\em ALen}), and
the number of words to the end of the reparandum ({\em RRest}),

\subsection{Context for the Type of Correspondence}

\begin{figure}
\begin{picture}(0,0)%
\includegraphics{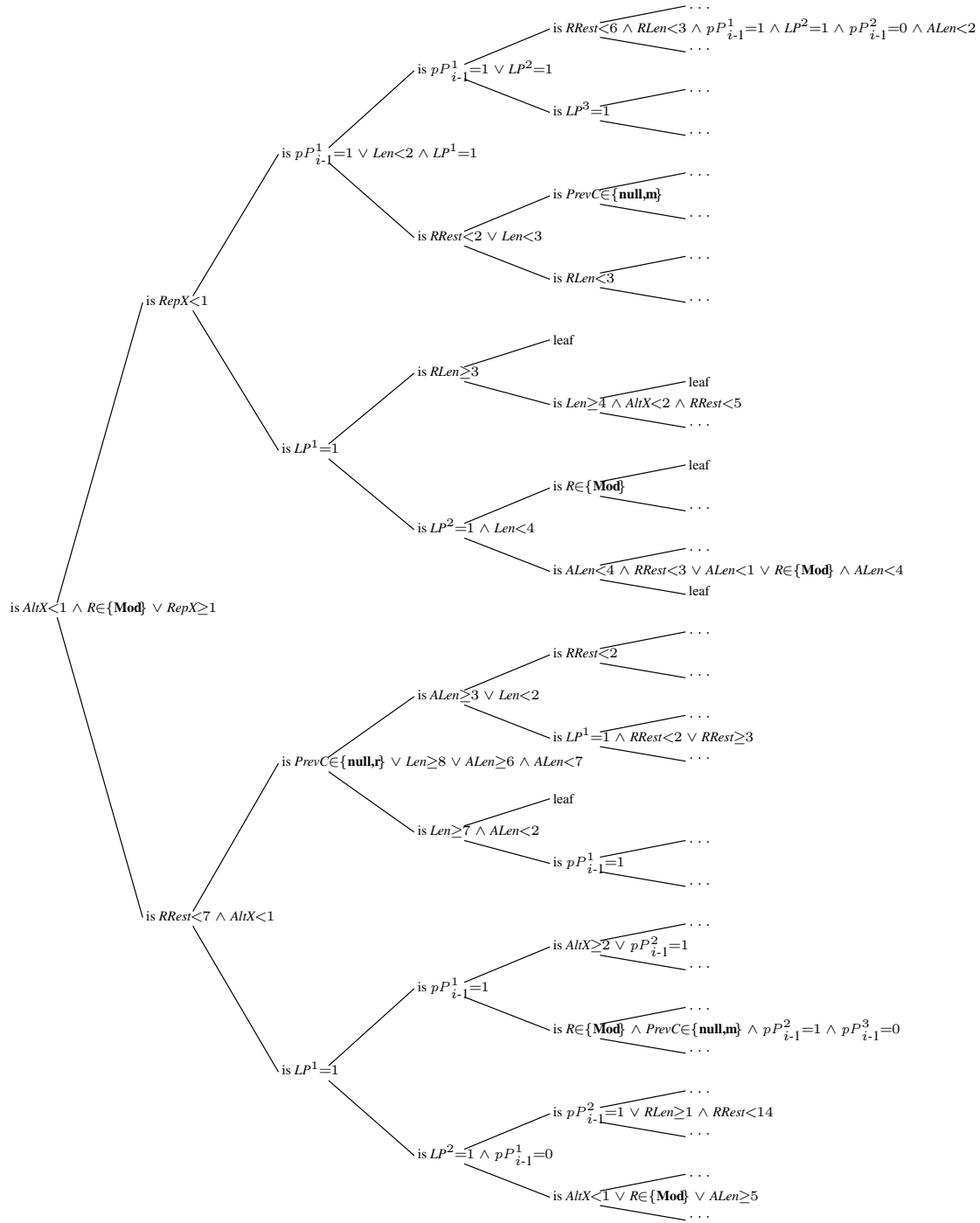}%
\end{picture}%
\setlength{\unitlength}{0.01250000in}%
\begingroup\makeatletter\ifx\SetFigFont\undefined
\def\x#1#2#3#4#5#6#7\relax{\def\x{#1#2#3#4#5#6}}%
\expandafter\x\fmtname xxxxxx\relax \def\y{splain}%
\ifx\x\y   
\gdef\SetFigFont#1#2#3{%
  \ifnum #1<17\tiny\else \ifnum #1<20\small\else
  \ifnum #1<24\normalsize\else \ifnum #1<29\large\else
  \ifnum #1<34\Large\else \ifnum #1<41\LARGE\else
     \huge\fi\fi\fi\fi\fi\fi
  \csname #3\endcsname}%
\else
\gdef\SetFigFont#1#2#3{\begingroup
  \count@#1\relax \ifnum 25<\count@\count@25\fi
  \def\x{\endgroup\@setsize\SetFigFont{#2pt}}%
  \expandafter\x
    \csname \romannumeral\the\count@ pt\expandafter\endcsname
    \csname @\romannumeral\the\count@ pt\endcsname
  \csname #3\endcsname}%
\fi
\fi\endgroup
\begin{picture}(327,596)(21,227)
\put(346,810){\makebox(0,0)[lb]{\smash{\SetFigFont{6}{7.2}{rm} $\cdots$}}}
\put(346,790){\makebox(0,0)[lb]{\smash{\SetFigFont{6}{7.2}{rm} $\cdots$}}}
\put(281,800){\makebox(0,0)[lb]{\smash{\SetFigFont{6}{7.2}{rm} is $\mbox{\em RRest}\!\!<\!\!6 \wedge \mbox{\em RLen}\!\!<\!\!3 \wedge {\em pP}_{\mbox{$i$-1}}^1\!\!=\!\!1 \wedge \mbox{\em LP}^2\!\!=\!\!1 \wedge {\em pP}_{\mbox{$i$-1}}^2\!\!=\!\!0 \wedge \mbox{\em ALen}\!\!<\!\!2$}}}
\put(346,770){\makebox(0,0)[lb]{\smash{\SetFigFont{6}{7.2}{rm} $\cdots$}}}
\put(346,750){\makebox(0,0)[lb]{\smash{\SetFigFont{6}{7.2}{rm} $\cdots$}}}
\put(281,760){\makebox(0,0)[lb]{\smash{\SetFigFont{6}{7.2}{rm} is $\mbox{\em LP}^3\!\!=\!\!1$}}}
\put(216,780){\makebox(0,0)[lb]{\smash{\SetFigFont{6}{7.2}{rm} is ${\em pP}_{\mbox{$i$-1}}^1\!\!=\!\!1 \vee \mbox{\em LP}^2\!\!=\!\!1$}}}
\put(346,730){\makebox(0,0)[lb]{\smash{\SetFigFont{6}{7.2}{rm} $\cdots$}}}
\put(346,710){\makebox(0,0)[lb]{\smash{\SetFigFont{6}{7.2}{rm} $\cdots$}}}
\put(281,720){\makebox(0,0)[lb]{\smash{\SetFigFont{6}{7.2}{rm} is $\mbox{\em PrevC}\!\!\in\!\!\{\mbox{\bf null,m$\!$}\}$}}}
\put(346,690){\makebox(0,0)[lb]{\smash{\SetFigFont{6}{7.2}{rm} $\cdots$}}}
\put(346,670){\makebox(0,0)[lb]{\smash{\SetFigFont{6}{7.2}{rm} $\cdots$}}}
\put(281,680){\makebox(0,0)[lb]{\smash{\SetFigFont{6}{7.2}{rm} is $\mbox{\em RLen}\!\!<\!\!3$}}}
\put(216,700){\makebox(0,0)[lb]{\smash{\SetFigFont{6}{7.2}{rm} is $\mbox{\em RRest}\!\!<\!\!2 \vee \mbox{\em Len}\!\!<\!\!3$}}}
\put(151,740){\makebox(0,0)[lb]{\smash{\SetFigFont{6}{7.2}{rm} is ${\em pP}_{\mbox{$i$-1}}^1\!\!=\!\!1 \vee \mbox{\em Len}\!\!<\!\!2 \wedge \mbox{\em LP}^1\!\!=\!\!1$}}}
\put(281,650){\makebox(0,0)[lb]{\smash{\SetFigFont{6}{7.2}{rm} leaf}}}
\put(346,630){\makebox(0,0)[lb]{\smash{\SetFigFont{6}{7.2}{rm} leaf}}}
\put(346,610){\makebox(0,0)[lb]{\smash{\SetFigFont{6}{7.2}{rm} $\cdots$}}}
\put(281,620){\makebox(0,0)[lb]{\smash{\SetFigFont{6}{7.2}{rm} is $\mbox{\em Len}\!\!\geq\!\!4 \wedge \mbox{\em AltX}\!\!<\!\!2 \wedge \mbox{\em RRest}\!\!<\!\!5$}}}
\put(216,635){\makebox(0,0)[lb]{\smash{\SetFigFont{6}{7.2}{rm} is $\mbox{\em RLen}\!\!\geq\!\!3$}}}
\put(346,590){\makebox(0,0)[lb]{\smash{\SetFigFont{6}{7.2}{rm} leaf}}}
\put(346,570){\makebox(0,0)[lb]{\smash{\SetFigFont{6}{7.2}{rm} $\cdots$}}}
\put(281,580){\makebox(0,0)[lb]{\smash{\SetFigFont{6}{7.2}{rm} is $\mbox{\em R}\!\!\in\!\!\{\mbox{\bf Mod$\!$}\}$}}}
\put(346,550){\makebox(0,0)[lb]{\smash{\SetFigFont{6}{7.2}{rm} $\cdots$}}}
\put(346,530){\makebox(0,0)[lb]{\smash{\SetFigFont{6}{7.2}{rm} leaf}}}
\put(281,540){\makebox(0,0)[lb]{\smash{\SetFigFont{6}{7.2}{rm} is $\mbox{\em ALen}\!\!<\!\!4 \wedge \mbox{\em RRest}\!\!<\!\!3 \vee \mbox{\em ALen}\!\!<\!\!1 \vee \mbox{\em R}\!\!\in\!\!\{\mbox{\bf Mod$\!$}\} \wedge \mbox{\em ALen}\!\!<\!\!4$}}}
\put(216,560){\makebox(0,0)[lb]{\smash{\SetFigFont{6}{7.2}{rm} is $\mbox{\em LP}^2\!\!=\!\!1 \wedge \mbox{\em Len}\!\!<\!\!4$}}}
\put(151,598){\makebox(0,0)[lb]{\smash{\SetFigFont{6}{7.2}{rm} is $\mbox{\em LP}^1\!\!=\!\!1$}}}
\put( 86,669){\makebox(0,0)[lb]{\smash{\SetFigFont{6}{7.2}{rm} is $\mbox{\em RepX}\!\!<\!\!1$}}}
\put(346,510){\makebox(0,0)[lb]{\smash{\SetFigFont{6}{7.2}{rm} $\cdots$}}}
\put(346,490){\makebox(0,0)[lb]{\smash{\SetFigFont{6}{7.2}{rm} $\cdots$}}}
\put(281,500){\makebox(0,0)[lb]{\smash{\SetFigFont{6}{7.2}{rm} is $\mbox{\em RRest}\!\!<\!\!2$}}}
\put(346,470){\makebox(0,0)[lb]{\smash{\SetFigFont{6}{7.2}{rm} $\cdots$}}}
\put(346,450){\makebox(0,0)[lb]{\smash{\SetFigFont{6}{7.2}{rm} $\cdots$}}}
\put(281,460){\makebox(0,0)[lb]{\smash{\SetFigFont{6}{7.2}{rm} is $\mbox{\em LP}^1\!\!=\!\!1 \wedge \mbox{\em RRest}\!\!<\!\!2 \vee \mbox{\em RRest}\!\!\geq\!\!3$}}}
\put(216,480){\makebox(0,0)[lb]{\smash{\SetFigFont{6}{7.2}{rm} is $\mbox{\em ALen}\!\!\geq\!\!3 \vee \mbox{\em Len}\!\!<\!\!2$}}}
\put(281,430){\makebox(0,0)[lb]{\smash{\SetFigFont{6}{7.2}{rm} leaf}}}
\put(346,410){\makebox(0,0)[lb]{\smash{\SetFigFont{6}{7.2}{rm} $\cdots$}}}
\put(346,390){\makebox(0,0)[lb]{\smash{\SetFigFont{6}{7.2}{rm} $\cdots$}}}
\put(281,400){\makebox(0,0)[lb]{\smash{\SetFigFont{6}{7.2}{rm} is ${\em pP}_{\mbox{$i$-1}}^1\!\!=\!\!1$}}}
\put(216,415){\makebox(0,0)[lb]{\smash{\SetFigFont{6}{7.2}{rm} is $\mbox{\em Len}\!\!\geq\!\!7 \wedge \mbox{\em ALen}\!\!<\!\!2$}}}
\put(151,448){\makebox(0,0)[lb]{\smash{\SetFigFont{6}{7.2}{rm} is $\mbox{\em PrevC}\!\!\in\!\!\{\mbox{\bf null,r$\!$}\} \vee \mbox{\em Len}\!\!\geq\!\!8 \vee \mbox{\em ALen}\!\!\geq\!\!6 \wedge \mbox{\em ALen}\!\!<\!\!7$}}}
\put(346,370){\makebox(0,0)[lb]{\smash{\SetFigFont{6}{7.2}{rm} $\cdots$}}}
\put(346,350){\makebox(0,0)[lb]{\smash{\SetFigFont{6}{7.2}{rm} $\cdots$}}}
\put(281,360){\makebox(0,0)[lb]{\smash{\SetFigFont{6}{7.2}{rm} is $\mbox{\em AltX}\!\!\geq\!\!2 \vee {\em pP}_{\mbox{$i$-1}}^2\!\!=\!\!1$}}}
\put(346,330){\makebox(0,0)[lb]{\smash{\SetFigFont{6}{7.2}{rm} $\cdots$}}}
\put(346,310){\makebox(0,0)[lb]{\smash{\SetFigFont{6}{7.2}{rm} $\cdots$}}}
\put(281,320){\makebox(0,0)[lb]{\smash{\SetFigFont{6}{7.2}{rm} is $\mbox{\em R}\!\!\in\!\!\{\mbox{\bf Mod$\!$}\} \wedge \mbox{\em PrevC}\!\!\in\!\!\{\mbox{\bf null,m$\!$}\} \wedge {\em pP}_{\mbox{$i$-1}}^2\!\!=\!\!1 \wedge {\em pP}_{\mbox{$i$-1}}^3\!\!=\!\!0$}}}
\put(216,340){\makebox(0,0)[lb]{\smash{\SetFigFont{6}{7.2}{rm} is ${\em pP}_{\mbox{$i$-1}}^1\!\!=\!\!1$}}}
\put(346,290){\makebox(0,0)[lb]{\smash{\SetFigFont{6}{7.2}{rm} $\cdots$}}}
\put(346,270){\makebox(0,0)[lb]{\smash{\SetFigFont{6}{7.2}{rm} $\cdots$}}}
\put(281,280){\makebox(0,0)[lb]{\smash{\SetFigFont{6}{7.2}{rm} is ${\em pP}_{\mbox{$i$-1}}^2\!\!=\!\!1 \vee \mbox{\em RLen}\!\!\geq\!\!1 \wedge \mbox{\em RRest}\!\!<\!\!14$}}}
\put(346,250){\makebox(0,0)[lb]{\smash{\SetFigFont{6}{7.2}{rm} $\cdots$}}}
\put(346,230){\makebox(0,0)[lb]{\smash{\SetFigFont{6}{7.2}{rm} $\cdots$}}}
\put(281,240){\makebox(0,0)[lb]{\smash{\SetFigFont{6}{7.2}{rm} is $\mbox{\em AltX}\!\!<\!\!1 \vee \mbox{\em R}\!\!\in\!\!\{\mbox{\bf Mod$\!$}\} \vee \mbox{\em ALen}\!\!\geq\!\!5$}}}
\put(216,260){\makebox(0,0)[lb]{\smash{\SetFigFont{6}{7.2}{rm} is $\mbox{\em LP}^2\!\!=\!\!1 \wedge {\em pP}_{\mbox{$i$-1}}^1\!\!=\!\!0$}}}
\put(151,300){\makebox(0,0)[lb]{\smash{\SetFigFont{6}{7.2}{rm} is $\mbox{\em LP}^1\!\!=\!\!1$}}}
\put( 86,374){\makebox(0,0)[lb]{\smash{\SetFigFont{6}{7.2}{rm} is $\mbox{\em RRest}\!\!<\!\!7 \wedge \mbox{\em AltX}\!\!<\!\!1$}}}
\put( 21,522){\makebox(0,0)[lb]{\smash{\SetFigFont{6}{7.2}{rm} is $\mbox{\em AltX}\!\!<\!\!1 \wedge \mbox{\em R}\!\!\in\!\!\{\mbox{\bf Mod$\!$}\} \vee \mbox{\em RepX}\!\!\geq\!\!1$}}}
\end{picture}
\caption{Decision Tree for Correspondence Type}
\label{fig:correction:Ctree}
\end{figure}

The context used for estimating the correspondence type $C_i$ is
exactly the same as the context used for estimating the proposed
licensor. Figure~\ref{fig:correction:Ctree} gives the top part of the
decision tree for the correspondence tree.

\subsection{The Other Trees}

Just as new trees were grown for the word and POS tags in
Chapter~\ref{chapter:detection}, which took advantage of the
utterance-sensitive words and POS tags afforded by modeling the
occurrence of boundary tones and speech repairs, new versions of the
word, POS tag, tone, editing term, and repair trees are grown.  These
trees take advantage of the better utterance-sensitive words and POS
tags that are afforded by modeling the correction of speech repairs.

For predicting the word and POS tags, we have an additional source of
information, namely the values of the correspondence licensor and the
correspondence type.  Rather then use these two variables as part of
the context that we give the decision tree algorithm, we use the
values of the tags to override the decision tree probability.  If the
correspondence type indicates a word replacement or a word match, we
assign all of the probability for the POS tag to the POS tag of the
word indicated by the correspondence licensor.  If the correspondence
type is a word match, we assign all of the probability for the word
variable to the word identity indicated by the correspondence
licensor.

\section{The Complexity}

The above approach might seem to explode the search space, for we need
to predict both the length of a repair and the possible
correspondences.  However the correspondences will be immediately
confirmed when we predict the category and the word, so the
correspondences only temporarily increase the search space size.  As
for guessing the length of the reparandum, most modification repairs
start with word retracing.  In fact, Levelt \shortcite{Levelt83:cog}
found that the onset of the reparandum is either the same syntactic
category or is a word match of the onset of the alteration for 48\% of
all repairs.  As we will observe in Section~\ref
{sec:results:tagging}, including the correction does not increase the
branching perplexity.

\cleardoublepage
\chapter{Acoustic Cues}
\label{chapter:acoustic}

Silence, as well as other acoustic information, can also give evidence
as to whether an intonational phrase, speech repair, or editing term
occurred.  Most speech recognizers can hypothesize the occurrence of a
pause, which they treat as a lexical-like item.  In the same way that
hypothesized words have an associated duration, so do the pauses.  In
Table~\ref {tab:detection:cues}, we saw that pauses have strong
correlations with the occurrence of boundary tones and speech
repairs.  Hence, the goal of this chapter is to revise the language
model so that these pauses can give evidence as to the occurrences of
these events.

There are several ways that pauses can be incorporated into a
language model.  First, one could treat pauses as language model
tokens, which would hence be predicted as any other word would be
\cite {Zeppenfeld-etal97:icassp}.\footnote{They used the silence
tokens as part of their speech recognition language model and found
that it resulted in a ``[word error rate] improvement of approximately
1\% absolute''.  This language model does not incorporate the
identification of boundary tones nor speech repairs.}  The second
alternative, explored in Section~\ref {sec:acoustic:redefine}, is to
predict the silence duration between each pair of consecutive words.
In Section~\ref {sec:acoustic:evidence}, we show how the prediction of
the silence durations can be used to give evidence for the occurrence
of boundary tones and speech repairs.

\section{Redefining the Speech Recognition Problem}
\label{sec:acoustic:redefine}

One possible way of incorporating silence information is to once again
redefine the speech recognition problem so that it also includes the
recognition of silence durations.  We define $S_i$ as the
silence duration between the words $W_\im$ and $W_i$.  The new speech
recognition equation is now the following.
\newlength{\argmaxa}
\settowidth{\argmaxa}{\footnotesize $WPCLORETS$}
\begin{eqnarray}
\lefteqn{\hat{W}\hat{P}\hat{C}\hat{L}\hat{O}\hat{R}\hat{E}\hat{T}\hat{S}} \nonumber \\
&=& \parbox[t]{\argmaxa}{\makebox[\argmaxa][c]{$\arg\max$}
		\footnotesize \vspace{-2.4em} \\ $WPCLORETS$}\
	 \Pr(A|WPCLORETS) \Pr(WPCLORETS) \label{eqn:acoustic:model}
\end{eqnarray} 
Again, the first term is the probability that results from the
acoustic model.  Note that one can make independence assumptions to
rewrite the acoustic model as $\Pr(A|WS)$, and thus reduce it to the
traditional language model.

The second term of Equation \ref{eqn:acoustic:model} is the
probability that results from the new language model, which we can
rewrite as the following, where $N$ is the number of words in the
sequence.
\[ \Pr(W_{1,N}P_{1,N}C_{1,N}L_{1,N}O_{1,N}R_{1,N}E_{1,N}T_{1,N}S_{1,N}) \]
We now rewrite this term as the following.
\begin{eqnarray}
\lefteqn{\Pr(W_{1,N}P_{1,N}C_{1,N}L_{1,N}O_{1,N}R_{1,N}E_{1,N}T_{1,N}S_{1,N})} \nonumber \\
&=&\prod_{i=1}^N\Pr(W_iP_iC_iL_iO_iR_iE_iT_iS_i|
W_{\rim}P_{\rim}C_{\rim}L_{\rim}O_{\rim}R_{\rim}E_{\rim}T_{\rim}S_{\rim})
\nonumber
\end{eqnarray}
We now need to decide when $S_i$ should be expanded.  From a
theoretical viewpoint, the choice is completely arbitrary.  However,
the choice impacts how easily the resulting probability distributions
can be estimated.  We choose to expand the silence variable first (as
indicated by our choice in writing the $S$ as the last variable in
$WPCLORETS$).  Hence, we first predict the silence following a word,
and then use the silence as part of the context in estimating the tags
for the remaining variables, namely $R$, $E$, and $T$.\footnote{A
second choice is to predict the silence duration after the repair,
editing term and tone variables have been predicted, thus giving a
model that could be referred to as $WPCLOSRET$.  Here, the prediction
of the silence can make use of the predictions that were already made
for the boundary tone, editing term, and speech repair variables.}
Below we give the expansion.
\begin{eqnarray} 
\lefteqn{Pr(W_iP_iC_iL_iO_iR_iE_iT_iS_i|
W_{\rim}P_{\rim}C_{\rim}L_{\rim}O_{\rim}R_{\rim}E_{\rim}T_{\rim}S_{\rim})} 
\nonumber \\
&=&\Pr(S_i|W_{\rim}P_{\rim}C_{\rim}L_{\rim}O_{\rim}R_{\rim}E_{\rim}T_{\rim}S_{\rim}) \nonumber \hspace{4em} \\
 &&\Pr(T_i|W_{\rim}P_{\rim}C_{\rim}L_{\rim}O_{\rim}R_{\rim}E_{\rim}T_{\rim}S_{\ri}) \nonumber\\
 &&\Pr(E_i|W_{\rim}P_{\rim}C_{\rim}L_{\rim}O_{\rim}R_{\rim}E_{\rim}T_{\ri}S_{\ri}) \nonumber\\
 &&\Pr(R_i|W_{\rim}P_{\rim}C_{\rim}L_{\rim}O_{\rim}R_{\rim}E_{\ri}T_{\ri}S_{\ri}) \nonumber\\
 &&\Pr(O_i|W_{\rim}P_{\rim}C_{\rim}L_{\rim}O_{\rim}R_{\ri}E_{\ri}T_{\ri}S_{\ri}) \nonumber\\
 &&\Pr(L_i|W_{\rim}P_{\rim}C_{\rim}L_{\rim}O_{\ri}R_{\ri}E_{\ri}T_{\ri}S_{\ri}) \nonumber\\
 &&\Pr(C_i|W_{\rim}P_{\rim}C_{\rim}L_{\ri}O_{\ri}R_{\ri}E_{\ri}T_{\ri}S_{\ri}) \nonumber\\
 &&\Pr(P_i|W_{\rim}P_{\rim}C_{\ri}L_{\ri}O_{\ri}R_{\ri}E_{\ri}T_{\ri}S_{\ri}) \nonumber\\
 &&\Pr(W_i|W_{\rim}P_{\ri}C_{\ri}L_{\ri}O_{\ri}R_{\ri}E_{\ri}T_{\ri}S_{\ri}) \label{eqn:acoustic:model2a}
\end{eqnarray}
The first consequence of doing this is that we have the silence
durations as part of the context for the other probabilities, and
hence we can make use of this information in determining the
probability of boundary tones, editing terms, and speech repairs.

The second consequence is that we have an extra probability in our
model, namely the probability of $S_i$ given the previous context.
The variable $S_i$ is the silence duration and hence will take on
values in accordance to the minimum time samples that the speech
recognizer uses.  To deal with limited amounts of training data, one
could collapse these silence durations into larger intervals.
Including this probability also means that we need to make sure that
the acoustic model does not have a bias against silences, since this
bias is now handled by the language model.  A final consideration is
that including this probability impacts the perplexity computation.
Usually, prediction of pauses or silence durations are not included in
the perplexity calculation.  In order to allow comparisons between the
perplexity rates of the model that includes silence durations and the
ones that do not, we will be excluding the probability of $S_i$ given
the previous context.  However, we will need to explore the use of
this probability when we actually use the model with a speech
recognizer to report word error rates.

\section{Using Silence as Part of the Context}
\label{sec:acoustic:evidence}

As indicated in the previous section, we can include the silence
durations as part of the context for predicting the values of the
other variables.  However, it is just the boundary tone, editing term,
and repair variables that this information is most appropriate.  The
simplest way of using this information would be to let the decision
tree algorithm use the silence duration as part of the context in
estimating the probability distributions for the boundary tone,
editing term and repair variables.  However, our attempts at doing
this have not met with success, perhaps because of sparseness of data
and limitations with the decision tree learning algorithm.  Asking
questions about the silence duration fragments the training data and
hence makes it difficult to model the influence of the other aspects
of the context. The alternative that we pursue is to assume that the
silence information is independent from the previous context.  Below
we give the derivation for the boundary tone variable, starting with
the probability for $T_i$ given in
Equation~\ref{eqn:acoustic:model2a}.  For expository ease, we define
${\em Prior}_i$ to be the prior context for deciding the probabilities
for word $S_i$.
\[ {\em Prior}_i = W_{\rim}P_{\rim}C_{\rim}L_{\rim}O_{\rim}R_{\rim}E_{\rim}T_{\rim}S_{\rim} \]
The derivation is as follows.
\begin{eqnarray}
\Pr(T_i|S_i{\em Prior}_i) 
&=& \frac{\Pr({\em Prior}_iS_i|T_i)\Pr(T_i)}
         {\Pr(S_i{\em Prior}_i)} \nonumber \\
&\approx& \frac{\Pr({\em Prior}_i|T_i)\Pr(S_i|T_i)\Pr(T_i)}
         {\Pr(S_i{\em Prior}_i)} \nonumber \\
&\approx& \frac{\Pr({\em Prior}_i|T_i)\Pr(S_i|T_i)\Pr(T_i)}
         {\Pr(S_i)\Pr({\em Prior}_i)} \nonumber \\
&=& \frac{\Pr(T_i|{\em Prior}_i)\Pr(S_i|T_i)}
         {\Pr(S_i)} \nonumber \\
&=& \Pr(T_i|{\em Prior}_i) \frac{\Pr(T_i|S_i)}
                                {\Pr(T_i)} \label{eqn:detection:silence}
\end{eqnarray}
In the first line, we applied Bayes' Rule.  The derivation of the next
two lines involves the following independence assumptions.
\begin{description}
\item[(1)] The prior context ${\em Prior}_i$ and the silence duration to 
the next word $S_i$ are independent.  Thus we simplify
$\Pr({\em Prior}_iS_i)$ to $\Pr({\em Prior}_i)\Pr(S_i)$.
\item[(2)] The prior context
${\em Prior}_i$ and the silence duration to the next word $S_i$ are
independent given the boundary tone variable $T_i$.  Thus we can simplify the
probability of $\Pr({\em Prior}_iS_i|T_i)$ to $\Pr({\em
Prior}_i|T_i)\Pr(S_i|T_i)$.
\end{description}
The first assumption is obviously too strong.  If the previous word is
a noun it is more likely that there will be a silence after it than if
the previous word was an article.  However, making these two
assumptions allows us to model the silence information independently
from the rest of the context, which gives us more data to estimate
their effect.  The result is we can use the factor
\[ \frac{\Pr(T_i|S_i)}{\Pr(T_i)} \]
to adjust the probabilities computed by the decision tree algorithm,
which does not take the silence duration into account.

\begin{figure}[tbh]
\centerline{\psfig{figure=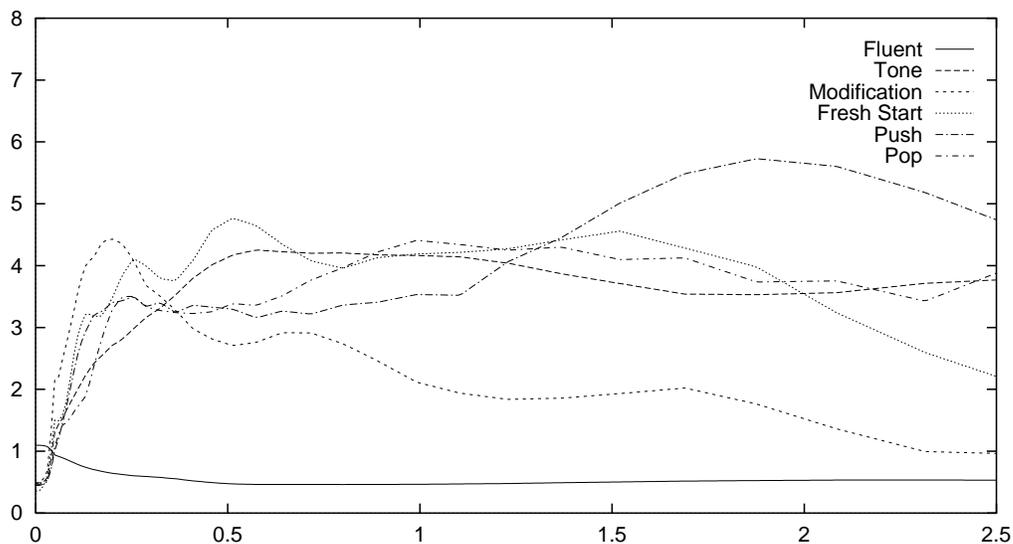,width=\textwidth}}
\caption{Preference for Utterance Tags given the Length of Silence}
\label{fig:silences}
\end{figure}

Rather than model the silence duration for all possible utterance tag
combinations, we define six equivalence classes of utterance tag
combinations.
\begin{itemize}
\item {\em Tone:} $T_i = {\bf Tone}$
\item {\em Push:} $E_i = {\bf Push}$ and $T_i = {\bf null}$
\item {\em Pop:} $E_i = {\bf Pop}$ and $T_i = {\bf null}$
\item {\em Cancel:} $R_i = {\bf Can}$ and $E_i = {\bf null}$ and $T_i
= {\bf null}$
\item {\em Modification:} $R_i = {\bf Mod}$ and $E_i = {\bf null}$ and
$T_i = {\bf null}$
\item {\em Fluent:} $T_i = {\bf null}$ and $E_i \in \{ {\bf null},{\bf
ET} \}$ and $R_i = {\bf null}$
\end{itemize}
We then compute the preference factors for these six equivalence classes
given the silence duration.  We do this by bucketing the silence
durations into 30 intervals and counting the number of occurrences of
each class for that bucket for the training data.  We
then smooth the counts using a gaussian filter and arrive at the
factors given in Figure~\ref {fig:silences}.
The ratio between the curves gives the preference for one type of
transition over another, for a given silence duration.  The silence
data was obtained from a word aligner \cite {Entropic94:aligner} that
annotates silences, and so is automatically derived. 

These factors are used to adjust the probabilities based on the
context.  Since the derivation of the silence factors is based on
several assumptions, we guard against any shortcomings of these
assumptions by normalizing the adjusted probabilities to insure that
they sum to one over all of the utterance tags.

Silences between speaker turns are not used in computing the
preference factor, nor is the preference factor used at such points.
As explained in Section~\ref {sec:related:tones}, the end of the
speaker's turn is determined jointly by both the speaker and the
hearer.  So when building a system that is designed to participate in
a conversation, these silence durations will be partially determined
by the system's turn-taking strategy.\label{sec:acoustic:endofturn} We
also do not include the silence durations after word fragments.  These
silence durations were hand-computed, rather than being automatically
derived.

\cleardoublepage
\chapter{Examples}
\label{chapter:examples}

In this chapter, we illustrate the workings of the algorithm and how
it makes use of the probability distributions to find the best
interpretation.  Speech repair detection and correction involves
five different tags that must be predicted; hence, it is the most
complicated part of our model.  Thus, the examples in this chapter
focus mainly upon this aspect of our model.

As in Section~\ref {sec:model:setup}, we illustrate the workings of
the algorithm constraining it to the word transcriptions.  The
algorithm incrementally considers all possible interpretations (at
least those that do not get pruned), proceeding one word at a time.
After fully processing a word, it will have a set of alternative
interpretations with a probability assigned to each.  In processing
the next word, it will iteratively expand each of the interpretations
for each possible value of the variables involved.  The probability of
each tag is multiplied into the probability of the interpretation.  A
beam search is used to prevent combinatoric explosion.

\section{First Example}

We first consider the following example, which was given earlier as
Example~\ref {ex:d92a-2.1:utt95} and Example~\ref
{ex:d92a-2.1:utt95:a}.
\begin{example}{d92a-2.1 utt95}
okay\trans{tone} uh and that will take a 
$\underbrace{\makebox{total of}}_{\makebox[3em][r]{\em reparandum}
\hfill}$\trans{ip:mod} 
\et{um let's see} total 
$\underbrace{\makebox{of s-}}_{\makebox[2em][r]{\em reparandum
\hfill}}$\trans{ip:mod} of seven hours
\end{example}
As a result of the second instance of ``of'' being part of the
alteration of the first repair and part of the reparandum of the
second, this example has two overlapping modification repairs.  The
winning interpretation, which is the interpretation with the highest
probability, was in fact the correct interpretation.

\subsection{Predicting ``um'' as the Onset of an Editing Term}

Below, we give the probabilities used in predicting the word ``um''
given the correct interpretation of the words ``okay uh and that will
take a total of''.  We show both the probability of the correct tags
(which is part of the winning interpretation) and the probabilities of
the competitors.  For reference, we give a simplified view of the
context that is used for each probability.

For the correction interpretation of the previous words, we first show
the probabilities for the two possible values for $T_6$.  We see that
the correct tag value of {\bf null} is significantly preferred over
the alternative interpretation, namely because boundary tones rarely
follow the preposition ``of''.
\begin{quote}
\begin{tabular}{lll} 
$\Pr(T_6$={\bf null}$|tW$=a total of) &=& 0.9997 \\
$\Pr(T_6$={\bf Tone}$|tW$=a total of) &=& 0.0003 \\ 
\end{tabular}
\end{quote}
For the interpretation with $T_6= {\bf null}$, we now show the possible
values for $E_6$.  Since an editing term is not in progress for this
interpretation, the only possible values for $E_6$ are {\bf Push} and
{\bf null}.
\begin{quote}
\begin{tabular}{lll}
$\Pr(E_6$={\bf Push}$|tW$=a total of) &=& 0.2422 \\
$\Pr(E_6$={\bf null}$|tW$=a total of) &=& 0.7578 \\
\end{tabular}
\end{quote}
For the interpretation with $E_6={\bf Push}$, the only allowable
values for the repair variable $R_6$ is the tag {\bf null}.  Since no
repair has been started, the only possible tag for the reparandum
onset $O_6$ is {\bf null}.  Similarly, since no repair is in progress,
$L_6$, the correspondence licensor, and $C_6$, the type of
correspondence, must both be null.
\begin{quote}
\begin{tabular}{lll}
$\Pr(R_6$={\bf null}$|tW$=a total of $E_6$={\bf Push})   &=& 1.0000 \\
$\Pr(O_6$={\bf null}$|R_6$={\bf null})          &=& 1.0000 \\
$\Pr(L_6$={\bf null}$|$ {\sl no active repair}) &=& 1.0000 \\
$\Pr(C_6$={\bf null}$|L_6$={\bf null})          &=& 1.0000 \\
\end{tabular}
\end{quote}
We now predict $P_6$, the POS tag for the word.  Below we list the
probability for all POS tags, given the correct interpretation, that
are greater than 0.01.  Since we have predicted the start of an
editing term, we see that POS tags associated with the first word of
an editing term are given a high probability, such as {\bf UH\_FP} for
``um'', {\bf AC} for ``okay'', {\bf CC\_D} for ``or'', {\bf UH\_D} for
``well'', and {\bf VB} for ``let's see''.
\begin{quote}
\begin{tabular}{lll}
$\Pr(P_6$={\bf UH\_FP}$|cW$=total of {\bf Push}) &=& 0.7307 \\
$\Pr(P_6$={\bf AC}$|cW$=total of {\bf Push})     &=& 0.1771 \\
$\Pr(P_6$={\bf CC\_D}$|cW$=total of {\bf Push})  &=& 0.0255 \\
$\Pr(P_6$={\bf UH\_D}$|cW$=total of {\bf Push})  &=& 0.0200 \\
$\Pr(P_6$={\bf VB}$|cW$=total of {\bf Push})     &=& 0.0255 \\
\end{tabular}
\end{quote}
For the interpretation with $P_6$ set to {\bf UH\_FP}, we now predict
the actual word, which will be one of ``um'', ``uh'', and ``er''.
\begin{quote}
\begin{tabular}{lll}
$\Pr(W_6$=um$|cW$=total of {\bf Push} {\bf UH\_FP}) &=& 0.5084 \\
$\Pr(W_6$=uh$|cW$=total of {\bf Push} {\bf UH\_FP}) &=& 0.4876 \\
$\Pr(W_6$=er$|cW$=total of {\bf Push} {\bf UH\_FP}) &=& 0.0040 \\
\end{tabular}
\end{quote}
Given the correct interpretation of the previous words, the
probability of the filled pause ``um'' along with the correct POS tag,
boundary tone tag, and repair tags is 0.0898.

\subsection{Predicting ``total'' as the Alteration Onset}

Now consider predicting the second instance of ``total'', which
is the first word of the alteration of the first repair, whose editing
term ``um let's see'', which ends with a boundary tone, has just
finished.  For the tone variable $T_{10}$, we see that the correct
interpretation has a probability of 90.2\%.
\begin{quote}
\begin{tabular}{lll}
$\Pr(T_{10}$={\bf Tone}$|tW$=let's see)&=&0.9018 \\
$\Pr(T_{10}$={\bf null}$|tW$=let's see)&=&0.0982 \\
\end{tabular}
\end{quote}
Given the interpretation with $T_{10}={\bf Tone}$, the probabilities for
the editing term variable $E_{10}$ are given below.  Since an editing
term is in progress, the only possibilities are that it is continued
{\bf ET}, or that the previous word is the end of the editing term
{\bf Pop}.  The correct interpretation of it finishing has a
probability of 83.0\%.
\begin{quote}
\begin{tabular}{lll}
$\Pr(E_{10}$={\bf Pop}$|tW$=let's see $rW$=a total of $T_{10}$={\bf Tone})&=& 0.8303 \\
$\Pr(E_{10}$={\bf ET}$|tW$=let's see $rW$=a total of $T_{10}$={\bf Tone}) &=& 0.1697 \\
\end{tabular}
\end{quote}
For the interpretation in which the editing term has just finished, we
now must decide the type of the repair---whether it is a fresh start,
a modification or an abridged repair.  The probability that it is a
modification repair is given a probability of 22.8\%, which is roughly
a third of the probability of a fresh start.  
\begin{quote}
\begin{tabular}{lll}
$\Pr(R_{10}$={\bf Mod}$|tW$=let's see $rW$=a total of $T_{10}$={\bf Tone} $E_{10}$={\bf Pop})&=&0.2281 \\
$\Pr(R_{10}$={\bf Can}$|tW$=let's see $rW$=a total of $T_{10}$={\bf Tone} $E_{10}$={\bf Pop})&=&0.6436 \\
$\Pr(R_{10}$={\bf Abr}$|tW$=let's see $rW$=a total of $T_{10}$={\bf Tone} $E_{10}$={\bf Pop})&=&0.1283 \\
\end{tabular}
\end{quote}
For the modification repair interpretation, we now examine the
probabilities assigned to the possible values of the reparandum onset.
The nine possibilities are given below, given in terms of $O_{10,X}$
where $X$ represents the proposed reparandum onset.
\begin{quote}
\begin{tabular}{lll}
$\Pr(O_{10,X}$={\bf Onset}$|X$=of $oW$=take a total $R_{10}$={\bf Mod})
&=&0.5891 \\
$\Pr(O_{10,X}$={\bf Onset}$|X$=total $oW$=will take a $R_{10}$={\bf Mod})
&=&0.1262 \\
$\Pr(O_{10,X}$={\bf Onset}$|X$=a $oW$=that will take $R_{10}$={\bf Mod})
&=&0.1446 \\
$\Pr(O_{10,X}$={\bf Onset}$|X$=take $oW$=and that will $R_{10}$={\bf Mod})
&=&0.0232 \\
$\Pr(O_{10,X}$={\bf Onset}$|X$=will $oW$=uh and that $R_{10}$={\bf Mod})
&=&0.0161 \\
$\Pr(O_{10,X}$={\bf Onset}$|X$=that $oW$=\tone uh and $R_{10}$={\bf Mod})
&=&0.0474 \\
$\Pr(O_{10,X}$={\bf Onset}$|X$=and $oW$=okay \tone uh $R_{10}$={\bf Mod})
&=&0.0474 \\
$\Pr(O_{10,X}$={\bf Onset}$|X$=uh $oW$=\turn okay \tone $R_{10}$={\bf Mod})
&=&0.0031 \\
$\Pr(O_{10,X}$={\bf Onset}$|X$=okay $oW$=\turn $R_{10}$={\bf Mod})
&=&0.0031 \\
\end{tabular}
\end{quote}
For the correct interpretation of $O_{10}$ as `total', there are two
possibilities for which word of the reparandum will license the
current word---either the word ``total'' or ``of''.  The correct
choice of ``total'' receives a probability of 97.3\%.
\begin{quote}
\begin{tabular}{lll}
$\Pr(L_{10,X}$={\bf Corr}$|X$=total $R$={\bf Mod}) &=& 0.9730 \\
$\Pr(L_{10,X}$={\bf Corr}$|X$=of $R$={\bf Mod})    &=& 0.0270 \\
\end{tabular}
\end{quote}
Given that ``total'' will license the correspondence to the current word,
we need to decide the type of correspondence: whether it is a word match,
word replacement or otherwise.
\begin{quote}
\begin{tabular}{lll}
$\Pr(C_{10}$={\bf m}$|L_{10}$=total/{\bf NN}) &=& 0.5882 \\
$\Pr(C_{10}$={\bf r}$|L_{10}$=total/{\bf NN}) &=& 0.1790 \\
$\Pr(C_{10}$={\bf x}$|L_{10}$=total/{\bf NN}) &=& 0.2328 \\
\end{tabular}
\end{quote}
For the correct interpretation, the word correspondence is
a word match with the word ``total'' and POS tag {\bf NN}.  Hence, the
POS tag and word identity of the current word are both fixed.
\begin{quote}
\begin{tabular}{lll}
$\Pr(P_{10}$={\bf NN}$|cW$=will take a $L_{10}$=total/{\bf NN} $C_{10}$={\bf m})&=&1.000 \\
$\Pr(W_{10}$=total$|cW$=will take a {\bf NN} $L_{10}$=total/{\bf NN} $C_{10}$={\bf m})   &=&1.000
\end{tabular}
\end{quote}
Given the correct interpretation of the previous words, the
probability of the word ``total'' along with the correct POS tag,
boundary tone tag, and repair tags is 0.0111.  

\begin{table}
\setlength{\tabcolsep}{0.3em}
\renewcommand{\arraystretch}{0.65}
\begin{center}
\begin{tabular}{|l|r|r|r|r|r|r|r|r|l|} \hline
Previous  & $T$    & $E$    & $R$    & $O$   & $L$  & $C$ & $P$ & $W$ & Prob. of \\
Word     &   &    &    &   &   &  &  &  & Interpretation \\ 
\hline \hline
\turn&\bf null&\bf null&\bf null&& & &\bf AC    & okay &       \\
     & 1.000  & 1.000  & 0.999  &     & & &0.480     &0.528 &0.2532 \\ \hline
okay &\bf T   &\bf null&\bf null&     & & &\bf UH\_FP& uh   &       \\
     & 0.977  & 1.000  & 0.999  &     & & &0.067     &0.290 &0.0191 \\ \hline
uh   &\bf null&\bf null&\bf null&     & & &\bf CC\_D & and  &       \\
     & 0.812  & 0.989  & 0.980  &     & & &0.049     &0.424 &0.0163 \\ \hline
and  &\bf null&\bf null&\bf null&     & & &\bf DP    & that &       \\
     & 0.992  & 0.990  & 0.993  &     & & &0.055     &0.939 &0.0506 \\ \hline
that &\bf null&\bf null&\bf null&     & & &\bf MD    & will &       \\
     & 0.999  & 0.998  & 0.990  &     & & &0.315     &0.166 &0.0515 \\ \hline
will &\bf null&\bf null&\bf null&     & & &\bf VB    & take &       \\
     & 1.000  & 0.996  & 0.995  &     & & &0.495     &0.362 &0.1775 \\ \hline
take &\bf null&\bf null&\bf null&     & & &\bf DT    & a    &       \\
     & 0.969  & 0.957  & 0.987  &     & & &0.206     &0.170 &0.0321 \\ \hline
a    &\bf null&\bf null&\bf null&     & & &\bf NN    & total&       \\
     & 0.996  & 0.983  & 0.942  &     & & &0.821     &0.037 &0.0284 \\ \hline
total&\bf null&\bf null&\bf null&     & & &\bf PREP  & of   &       \\
     & 0.912  & 0.999  & 0.998  &     & & &0.769     &0.861 &0.6025 \\ \hline
of   &\bf null&\bf Push&        &     & & &\bf UH\_FP&um    &       \\
     & 0.998  & 0.242  &        &     & & &0.731     &0.508 &0.0898 \\ \hline
um   &\bf null&\bf ET  &        &     & & &\bf VB    & let  &       \\
     & 0.810  & 0.049  &        &     & & &0.283     &0.491 &0.0055 \\ \hline
let  &\bf null&\bf ET  &        &     & & &\bf PRP   & us   &       \\
     & 0.986  & 0.907  &        &     & & &0.689     &0.517 &0.3187 \\ \hline
's   &\bf Tone&\bf ET  &        &     & & &\bf VB    & see  &       \\
     & 0.886  & 0.978  &        &     & & &0.887     &0.785 &0.6030 \\ \hline
see  &\bf Tone&\bf Pop &\bf Mod & total &total &\bf m &\bf NN   & total&     \\
     & 0.902  & 0.830  & 0.228  & 0.126 &0.973 &0.588 &1.000    &1.000&0.0123\\ \hline
total&\bf null&\bf null&\bf null&       & of   &\bf m &\bf PREP & of   &     \\
     & 0.912  & 0.999  & 0.998  &       &1.000 &0.494 &1.000    &1.000&0.4493\\ \hline
of   &\bf null&\bf null&\bf null&       &      &      &\bf FRAG &\frag &    \\
     & 1.000  & 0.986  & 0.998  &       &      &      &0.023    &1.000&0.0226\\ \hline
s-   &\bf null&\bf null&\bf Mod & of    & of   &\bf m &\bf PREP & of   &     \\
     & 1.000  & 0.841  & 0.695  & 0.160 &1.000 &0.507 &1.000    &1.000&0.0473\\ \hline
of   &\bf null&\bf null&\bf null&       &      &      &\bf CD    & seven&     \\
     & 1.000  & 0.986  & 0.998  &       &      &      &0.160    &0.079&0.0125\\ \hline
seven&\bf null&\bf null&\bf null&       &      &      &\bf NNS  & hours&     \\
     & 0.883  & 0.964  & 0.974  &       &      &      &0.519    &0.737&0.3236\\ \hline
hours&\bf Tone&\bf null&\bf null&       &      &      &\bf TURN &\turn & \\
     & 0.682  & 0.996  & 0.996  &       &      &      &0.556    &1.000&0.3761\\
\hline
\end{tabular}
\end{center}
\caption{Interpretation of First Example}
\label{tab:example:1}
\end{table}%
In Table~\ref{tab:example:1}, we give the probabilities for each tag
of each word involved in the correct interpretation.  The first column
gives the previous word.  The following columns give the correct value
for each of the variables and their probability given the correct
interpretation.  Blank entries for the $R$, $O$, $L$, and $C$ tags
indicate that their value is completely determined by the context, and
hence their probability is one.  The last column gives the total
probability for the word and its correct tags given the correct
interpretation of the previous words.  This probability is just the
product of the probabilities given in the previous columns.  Note in
the example that the ``uh'' following the ``okay'' is not interpreted
as being part of an editing term.  Rather, it is correctly interpreted
as an utterance initial filled pause, undoubtedly helped by having the
boundary tone after ``okay'' correctly identified.

\section{Second Example}

Now consider the following example, which has two repairs with the
second one overlapping the first.  The second repair was
hand-annotated as {\bf ip10:can+}, and hence is ambiguous between a
fresh start and a modification repair.
\begin{example}{d93-15.2 utt37}
you have w- {\bf Mod} one {\bf Can} you have two boxcars
\end{example}
The reparandum of the first repair is the word fragment ``w-'', and
the reparandum of the second repair is ``you have one''.  However,
most of the word correspondences that help signal the second repair,
namely the word matches involving ``you have'', also span the
interruption point of the first repair.  Hence, in resolving this
utterance, an algorithm might mistaken attribute the word
correspondences of ``you have'' to the first repair, thus making the
reparandum of the first repair ``you have w-''.  If this happens, then
it will be very difficult for the algorithm to detect the second
repair and correct it.  Approaches to speech repair detection and
correction that can not consider alternative hypotheses, such as an
earlier version of this work \cite {HeemanAllen94:acl}, are
particularly susceptible to this problem.

The above example is actually only part of the speaker's turn. Below, we
repeat the example but in the context of the entire turn.  Here, we see
that there is an intonational boundary tone that occurs before the
reparandum onset of the second repair (which is actually the fourth
repair in this turn).  Identifying the boundary tone should simplify
the problem of determining the reparandum onset because the beginning
of an intonational boundary is a likely candidate for the reparandum
onset.
\begin{example}{d93-15.2 utt37 (Full Turn)}
see w- {\bf Can} see what's going on is you've got at {\bf Push} um
{\bf Pop} {\bf Abr} at three a.m. {\bf Tone} you have w- {\bf Mod} one
{\bf Can} you have two boxcars leaving Corning to Bath {\bf Tone}
\end{example}

To further illustrate how our algorithm works, we contrast two
completing hypotheses for the first repair (which is the third in the
turn). Table~\ref{tab:example:2:correct} gives the winning hypothesis,
starting with the first word after the interruption point of the first
repair.
\begin{table}
\renewcommand{\arraystretch}{0.65}
\setlength{\tabcolsep}{0.3em}
\begin{center}
\begin{tabular}{|l|r|r|r|r|r|r|r|r|l|} \hline
Previous  & $T$    & $E$    & $R$    & $O$   & $L$  & $C$ & $P$ & $W$ & Prob. of \\
Word     &   &    &    &   &   &  &  &  & Interpretation \\ 
\hline \hline
a.m.&\bf Tone&\bf null&\bf null&      &      &     &\bf   PRP&  you&  \\
    &   0.579&   0.997&   0.993&      &      &     &    0.064&0.076& 0.0028 \\ \hline
 you&\bf null&\bf null&\bf null&      &      &     &\bf HAVEP& have&  \\    
    &   0.999&   1.000&   0.996&      &      &     &    0.123&0.840& 0.1030 \\ \hline
have&\bf null&\bf null&\bf null&      &      &     &\bf  FRAG&\frag&  \\
    &   0.986&   0.996&   0.991&      &      &     &    0.013&1.000& 0.0126 \\ \hline
  w-&\bf null&\bf null&\bf  Mod&    w-&      &     &\bf    CD&  one&  \\
    &   1.000&   0.915&   0.676& 0.535&      &     &    0.122&0.162& 0.0065 \\ \hline
 one&\bf null&\bf null&\bf  Mod&   you&   you&\bf M&\bf   PRP&  you&  \\
    &   0.808&   0.877&   0.224& 0.198& 0.957&0.864&    1.000&1.000& 0.0260 \\ \hline
 you&\bf null&\bf null&\bf null&      &  have&\bf M&\bf HAVEP& have&  \\
    &   0.999&   1.000&   0.996&      & 0.973&0.783&    1.000&1.000& 0.7574 \\ \hline
have&\bf null&\bf null&\bf null&      &   one&\bf R&\bf    CD&  two&  \\
    &   0.959&   0.990&   0.950&      & 1.000&0.083&    1.000&0.290& 0.0218 \\ \hline
\end{tabular}
\end{center}
\caption{Correct Interpretation of Second Example}
\label{tab:example:2:correct}
\vspace*{2em}
\begin{center}
\begin{tabular}{|l|r|r|r|r|r|r|r|r|l|} \hline
Previous  & $T$    & $E$    & $R$    & $O$   & $L$  & $C$ & $P$ & $W$ & Prob. of \\
Word     &   &    &    &   &   &  &  &  & Interpretation \\ 
\hline \hline
a.m.&\bf Tone&\bf null&\bf null&     &      &     &\bf   PRP&  you&     \\
    &   0.579&   0.997&   0.993&     &      &     &    0.064&0.076& 0.0028 \\ \hline
 you&\bf null&\bf null&\bf null&     &      &     &\bf HAVEP&have &     \\
    &   0.999&   1.000&   0.996&     &      &     &    0.123&0.840& 0.1030 \\ \hline
have&\bf null&\bf null&\bf null&     &      &     &\bf FRAG&\frag&     \\
    &   0.986&   0.996&   0.991&     &      &     &    0.013&1.000& 0.0126 \\ \hline
  w-&\bf null&\bf null&\bf  Mod&  you&   you&\bf X&\bf    CD&  one&     \\
    &   1.000&   0.915&   0.676&0.367& 0.967&0.123&    0.009&0.327& 0.0001 \\ \hline
 one&\bf null&\bf null&\bf null&     &   you&\bf M&\bf   PRP&  you&     \\
    &   0.918&   0.988&   0.702&     & 0.778&0.092&    1.000&1.000& 0.0456 \\ \hline
 you&\bf null&\bf null&\bf null&     &  have&\bf M&\bf HAVEP& have&    \\
    &   0.878&   0.999&   0.997&     & 1.000&0.783&    1.000&1.000& 0.6851 \\ \hline
have&\bf null&\bf null&\bf null&     &      &     &\bf    CD&  two&    \\
    &   0.959&   0.990&   0.950&     &      &     &    0.122&0.290& 0.0318 \\ \hline
\hline
\end{tabular}
\end{center}
\caption{Incorrect Interpretation of Second Example}
\label{tab:example:2:incorrect}
\end{table}%
Although, the second repair is interpreted as a modification
repair rather than a fresh start, this is not counted as an error
since the repair was labeled as ambiguous.  Furthermore, even if it
was not labeled as ambiguous, it should not be viewed as an error
since the proper correction was made.  In terms of the entire turn,
all four repairs were correctly resolved, as were the two boundary tones.
The only mistake was with the POS tag for the word ``on'', which was
mistaken as a subordinating conjunction ({\bf SC}). 

We contrast the winning interpretation with the interpretation that
the reparandum onset of the first repair is the word ``you'' and the
second repair is not detected, which is the interpretation found by
our earlier version \cite{HeemanAllen94:acl}.  The highest scoring
variant of this is the one in which the first repair is detected as a
modification repair.  The probability scores are given in Table~\ref
{tab:example:2:incorrect}.  This interpretation received a score of
2.9\% of the score of the winning hypothesis at the end of the word
``two''.  Hence, we see that using a model that can evaluate
alternative hypotheses allows this example to be handled without
problem.

\section{Third Example}

The next example further illustrates the ability of our model to deal
with overlapping speech repairs.  The example, given below, has four
repairs.
\begin{example}{d93-14.2 utt30}
and pick up {\bf Push} um {\bf Pop} {\bf Abr} the en- {\bf Push} I
{\bf ET} guess {\bf Pop} {\bf Mod} the entire {\bf Push} um {\bf Pop}
{\bf Mod} pic- {\bf Mod} pick up the load of oranges at Corning
\end{example}
The first repair occurs after ``and pick up'' and is an abridged
repair with editing term ``um''. The second is a modification repair
whose reparandum is ``the en-'' and whose editing term is ``I
guess''. The third is a modification repair whose reparandum is ``and
pick up the entire''. This repair is actually annotated as being
ambiguous between a fresh start and a modification repair, and
ambiguous as to whether the reparandum includes the discourse marker
``and''. The fourth repair, another modification repair, occurs
immediately after the third repair and its reparandum is
``\mbox{pic-}''.

Since our approach predicts speech repairs and their reparandum as
they occur, complicated overlapping repairs such as this do not
cause a problem.  A summary of the probabilities involved in the
winning hypothesis is given in Table~\ref{tab:example:3}.
\begin{table}
\renewcommand{\arraystretch}{0.65}
\setlength{\tabcolsep}{0.3em}
\begin{center}
\begin{tabular}{|l|r|r|r|r|r|r|r|r|l|} \hline
Previous  & $T$    & $E$    & $R$    & $O$   & $L$  & $C$ & $P$ & $W$ & Prob. of \\
Word     &   &    &    &   &   &  &  &  & Interpretation \\ 
\hline \hline
  \turn&\bf null&\bf null&\bf null&     &      &    &\bf CC\_D&    and& \\
       &   0.999&   1.000&   0.999&     &      &     &   0.138&  0.576& 0.0794\\ \hline
    and&\bf null&\bf null&\bf null&     &      &     &\bf   VB&   pick& \\
       &   0.985&   0.965&   0.977&     &      &     &   0.176&  0.109& 0.0177\\ \hline
   pick&\bf null&\bf null&\bf null&     &      &     &\bf   RP&     up& \\
       &   0.998&   0.998&   0.997&     &      &     &   0.924&  0.978& 0.8971\\ \hline
     up&\bf null&\bf Push&        &     &      &   &\bf UH\_FP&     um& \\
       &   0.919&   0.116&        &     &      &     &   0.814&  0.474& 0.0410\\ \hline
     um&\bf null&\bf   Pop&\bf Abr&     &      &     &\bf   DT&    the& \\
       &   0.801&   0.919&   0.757&     &      &     &   0.592&  0.544& 0.1794\\ \hline
    the&\bf null&\bf null&\bf null&     &      &     &\bf FRAG& \frag & \\
       &   0.994&   0.830&   0.853&     &      &     &   0.014&  1.000& 0.0102\\ \hline
    en-&\bf null&\bf Push&        &     &      &     &\bf  PRP&      i& \\
       &   0.996&   0.099&        &     &      &     &   0.021&  0.729& 0.0015\\ \hline
      I&\bf null&\bf   ET&        &     &      &     &\bf  VBP&  guess& \\
       &   0.974&   0.977&        &     &      &     &   0.572&  0.199& 0.1080\\ \hline
  guess&\bf null&\bf  Pop&\bf  Mod&  the&   the&    M&\bf   DT&    the& \\
       &   0.964&   0.390&   0.386&0.347& 1.000&0.663&   1.000&  1.000& 0.0334\\ \hline
    the&\bf null&\bf null&\bf null&     &      &     &\bf   JJ& entire& \\
       &   1.000&   0.988&   0.989&     &      &     &   0.068&  0.000& 0.0000\\ \hline
 entire&\bf null&\bf Push&\bf null&     &      &   &\bf UH\_FP&     um& \\
       &   0.753&   0.108&   1.000&     &      &     &   0.791&  0.524& 0.0337\\ \hline
     um&\bf null&\bf  Pop&\bf  Abr&     &      &     &\bf FRAG&  \frag& \\
       &   0.825&   0.950&   0.747&     &      &     &   0.004&  1.000& 0.0026\\ \hline
   pic-&\bf null&\bf null&\bf  Mod& pick&  pick&    M&\bf   VB&   pick& \\
       &   0.997&   0.803&   0.734&0.036& 0.937&0.673&   1.000&  1.000& 0.0132\\ \hline
   pick&\bf null&\bf null&\bf null&     &    up&    M&\bf   RP&     up& \\
       &   0.998&   0.998&   0.997&     & 0.954&0.603&   1.000&  1.000& 0.5710\\ \hline
     up&\bf null&\bf null&\bf null&     &   the&    M&\bf   DT&    the& \\
       &   0.993&   0.990&   0.995&     & 0.967&0.843&   1.000&  1.000& 0.7982\\ \hline
    the&\bf null&\bf null&\bf null&     &entire&    X&\bf   NN&   load& \\
       &   1.000&   0.988&   0.989&     & 1.000&0.248&   0.227&  0.002& 0.0001\\ \hline
   load&\bf null&\bf null&\bf null&     &entire&    X&\bf PREP&     of& \\
       &   0.859&   0.996&   0.996&     & 1.000&0.797&   0.763&  0.779& 0.4036\\ \hline
     of&\bf null&\bf null&\bf null&     &entire&    X&\bf  NNS&oranges& \\
       &   1.000&   0.992&   0.992&     & 1.000&0.797&   0.311&  0.442& 0.1077\\ \hline
oranges&\bf null&\bf null&\bf null&     &      &     &\bf PREP&     at& \\
       &   0.336&   0.979&   0.986&     &      &     &   0.502&  0.146& 0.0237\\ \hline
     at&\bf null&\bf null&\bf null&     &      &     &\bf  NNP&corning& \\
       &   0.999&   0.989&   0.997&     &      &     &   0.802&  0.537& 0.4242\\ \hline
\end{tabular}
\end{center}
\caption{Interpretation of Third Example}
\label{tab:example:3}
\end{table}
After we have processed the words ``and pick up um the'', the model is
already certain that the ``um'' is part of an abridged repair, making
the context for the subsequent decisions for this interpretation be
``and pick up the''. By the time it sees ``and pick up um the en- I
guess the entire'', the interpretation with ``I guess'' as a cue
phrase for the modification repair with reparandum ``the en-'', is
highly favored. Since we do not interpret fragments, the extent of the
third repair is difficult for our model to determine.  The model is
unable to reason that the alteration onset ``pic-'' matches the
earlier occurrence of ``pick''.  Hence, the winning interpretation for
the third repair interprets it as an abridged repair, rather than a
modification repair whose reparandum is ``and pick up the
entire''. However, the reparandum of the fourth repair is also
incorrectly identified, but as being ``pick up the entire pic-''. Thus
the winning hypothesis makes up for the failure in identifying and
correcting the third repair.  In regards to the initial ``and'', it
was annotated as being ambiguous as to whether it is removed. Hence,
the winning hypotheses, which resolved the utterance to ``and pick up
the load of oranges at Corning'', correctly detected and corrected all
four speech repairs.

\cleardoublepage
\chapter{Results and Comparison}
\label{chapter:results}

In this chapter we present the results of running the statistical
language model on the Trains corpus. The model combines the tasks of
language modeling, POS tagging, identifying discourse markers,
identifying boundary tones, and detecting and correcting speech
repairs. The experiments we run in this chapter not only show the
feasibility of this model, but also support the thesis that these
tasks must be combined in a single model in order to account for the
interactions between the tasks.
In Section~\ref{sec:results:tagging}, we show that by modeling speech
repairs and intonational phrase boundary tones, we improve the
performance on POS tagging, word perplexity and identifying discourse
markers.
Section~\ref {sec:results:tones} demonstrates that the task of
detecting boundary tones benefits from modeling POS tags, discourse
markers, and speech repairs; Section~\ref {sec:results:detection}
shows that the detection of speech repairs is improved by modeling POS
tags, discourse markers, boundary tones and the correction of speech
repairs; and Section~\ref {sec:results:correction} shows that the
correction of speech repairs is facilitated by modeling boundary
tones.
The final experiments, given in Section~\ref
{sec:results:freshstarts}, show that differentiating between fresh
starts and modification repairs leads to better speech repair
modeling, as well as improves boundary tone identification and POS
tagging.
We end this chapter with a comparison with other approaches that have
been proposed for modeling speech repairs (Section~\ref
{sec:results:c:repairs}), boundary tones (Section~\ref
{sec:results:c:tones}) and discourse markers (Section~\ref
{sec:results:c:dm}).

\begin{figure}
\centerline{\psfig{figure=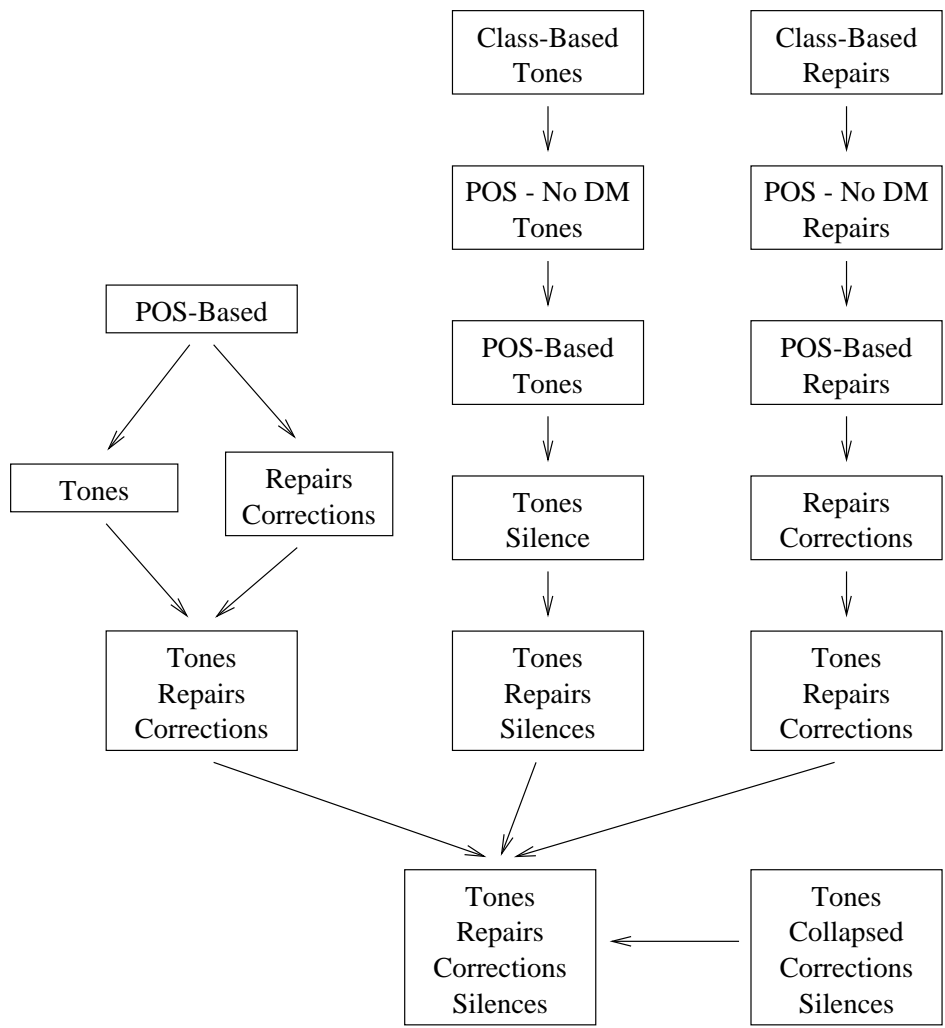}}
\caption{\label{fig:chain}Overview of Experiments}
\end{figure}

In order to show the effect of each part of the model on the other
parts, we start with the language models that we presented in
Chapter~\ref{chapter:model}, and vary which variables of
Chapters~\ref{chapter:detection},~\ref {chapter:correction} and \ref
{chapter:acoustic} we include in the speech recognition problem.
Figure~\ref {fig:chain} gives a diagram of all of the variations that
we test, where the arcs show the comparisons that we make.
We vary whether we model boundary tones by whether we include the
variable $T_i$ of Chapter~\ref{chapter:detection} in the model. We
vary whether we model the detection of speech repairs and their
editing terms by whether we include the variables $R_i$ and $E_i$,
introduced in Chapter~\ref{chapter:detection}. We vary whether we
distinguish between fresh starts and modification repairs by whether
we collapse fresh starts and modification repairs into a single tag
value (which is denoted as {\em collapsed} in Figure~\ref{fig:chain}),
or use two separate tags: {\bf Can} and {\bf Mod}. We vary whether we
model the correction of speech repairs by whether we include
the variables $O_i$, $L_i$, and $C_i$, introduced in
Chapter~\ref{chapter:correction}. Lastly, we vary whether we include
silence information by whether we adjust the tone, editing term, and
repair probabilities as described in Chapter~\ref{chapter:acoustic}.

All results in this chapter were obtained using the six-fold
cross-validation procedure that was described in Section~\ref
{sec:model:setup}, and all results were obtained from the
hand-collected transcripts.  We ran these transcripts through a
word-aligner \cite {Entropic94:aligner}, a speech recognizer
constrained to recognize what was transcribed, in order to
automatically obtain silence durations.  In predicting the end of turn
marker {\bf $<$turn$>$}, we do not use any silence
information.\fnote{Would be nice to give perplexity results of
predicting the end of turn and contrast this to the POS-based model.}

\section{POS Tagging, Perplexity and Discourse Markers}
\label{sec:results:tagging}

The first set of experiments, whose results are given in Table~\ref
{tab:results-pos}, explore how POS tagging, word perplexity, and
discourse marker identification benefit from modeling boundary tones
and speech repairs.\footnote {In Section~\ref {sec:model:comparison},
we showed that perplexity improved by modeling POS tags and discourse
markers.}  The second column gives the results of the POS-based
language model, introduced in Chapter~\ref{chapter:model}.
\newlength{\mcl}%
\settowidth{\mcl}{Correctio}%
\setlength{\tabcolsep}{0.25em}%
\newcommand{\corlen}{\hspace*{\mcl}}%
\begin{table}
\begin{center}\small
\begin{tabular}{|l|r|r|r|r|r|} \hline
&\corlen & \corlen  & \corlen       & \corlen        &\mc{Tones}      \\
&        &          &               &\mc{Tones}      &\mc{Repairs}    \\
&        &          &\mc{Repairs}   &\mc{Repairs}    &\mc{Corrections}\\
&\mc{POS}&\mc{Tones}&\mc{Correction}&\mc{Corrections}&\mc{Silences} \\ 
\hline \hline
{\em POS Tagging}      &       &       &      &       &       \\
\ Errors               &  1711 &  1646 & 1688 &  1652 &  1572 \\
\ Error Rate           &  2.93 &  2.82 & 2.89 &  2.83 &  2.69 \\ \hline
{\em Discourse Markers}&       &       &      &       &       \\
\ Errors               &   630 &   587 &  645 &   611 &   533 \\
\ Error Rate           &  7.61 &  7.09 & 7.79 &  7.38 &  6.43 \\
\ Recall               & 96.75 & 97.01 &96.52 & 96.67 & 97.26 \\
\ Precision            & 95.68 & 95.93 &95.72 & 95.97 & 96.32 \\ \hline
{\em Perplexity}       &       &       &      &       &       \\
\ Word                 & 24.04 & 23.91 &23.17 & 22.96 & 22.35 \\ 
\ Branching            & 26.35 & 30.61 &27.69 & 31.59 & 30.26 \\ \hline
\end{tabular}
\end{center}
\caption{\label{tab:results-pos}POS Tagging and Perplexity}
\end{table}%
The third column adds in boundary tone detection.  This model contains
no additional information, but simply allows the existing training
data to be separated into different contexts based on the occurrence
of the boundary tones in the training data.  We see that adding in
boundary tone modeling reduces the POS error rate by 3.8\%, improves
discourse marker identification by 6.8\%, and reduces perplexity
slightly from 24.04 to 23.91.  These improvements are of course at the
expense of the branching perplexity, which increases from 26.35 to
30.61.

The fourth column gives the results of the POS-based model augmented
with speech repair detection and correction.\footnote{We avoid
comparing the POS-based model to just the speech repair detection
model without correction.  The speech repair detection model on its
own results in a slight degradation in POS tagging (9 extra POS
errors) and discourse marker identification (11 more errors) than the
POS-based model, while only giving a slight reduction in perplexity
(24.04 to 23.74).  As we discussed in Chapter~\ref
{chapter:correction}, speech repair detection and correction need to
be combined into a single model.  Our results with the detection model
on its own lend support to that hypothesis.}  As with adding boundary
tones, we are not adding any further information, but only separating
the training data as to the occurrence and correction of speech
repairs.  We see that modeling repairs results in improved POS tagging
and reduces word perplexity by 3.6\%.  Also note that the branching
perplexity increases much less than it did when we added in boundary
tone identification, increasing from 26.35 to 27.69.\footnote {The
branching perplexity for repair detection and correction is also less
than when just adding repair detection, for which it is 27.90.}
Hence, although we are adding in 5 extra variables into the speech
recognition problem ($R_i$, $E_i$, $O_i$, $L_i$, and $C_i$), most of
the extra ambiguity that arises is resolved by the time the word is
predicted.  Thus, it must be the case that corrections can be
sufficiently resolved by the first word of the alteration.

The fifth column augments the POS-based model with both boundary tone
identification and speech repair detection and correction, and hence
combines the models of columns three and four.  The combined model
results in a further improvement in word perplexity.  The POS tagging
and discourse marker identification do not seem to benefit from
combining the two processes, but both rates remain better than those
obtained from the based model.

Of course, there are other sources of information that give evidence
that a repair or boundary tone occurred.  In column six, we show the
effect of adding silence information.  Silence information is not
directly used to decide the POS tags, the discourse markers, nor what
words are involved.  Rather, it gives evidence as to whether a
boundary tone, speech repair, or editing term occurred.  As the
following sections will show, adding in silence information improves
the performance on these tasks, and this increase translates into a
better language model, resulting in a further decrease in perplexity
from 22.96 to 22.35, giving an overall perplexity reduction of 7.0\%
with respect to the POS-based model.  We also see a significant
improvement in POS tagging with an error rate reduction of 8.1\% over
the POS-based model, and an overall reduction in the discourse marker
error rate of 15.4\%.  As we further improve the modeling of the
user's utterance, we should expect to see further improvements in the
language model.

\section{Boundary Tones}
\label{sec:results:tones}

The experiments summarized in Table~\ref {tab:results-ip} demonstrate
that modeling intonational phrase boundary tones benefits from
modeling POS tags, discourse markers, speech repairs and benefits from
the addition of silence information.
\begin{table}[htb]
\begin{center}\small
\begin{tabular}{|l|r|r|r|r|r|r|} \hline
&\corlen&\corlen&\corlen   &\corlen      & \corlen     &\mc{Tones}       \\ 
&       &       &          &             &\mc{Tones}   &\mc{Repairs}     \\ 
&\mc{Class-Based}&\mc{No DM}
                &          &\mc{Tones}   &\mc{Repairs} &\mc{Corrections} \\ 
&\mc{Tones}      &\mc{Tones}
                &\mc{Tones}&\mc{Silences}&\mc{Silences}&\mc{Silences}    \\ 
\hline\hline
{\em Within Turn}   &        &       &       &       &       &       \\
\ Errors            &   4063 &  3711 &  3585 &  3259 &  3145 &  3199 \\
\ Error Rate        &  73.40 & 67.04 & 64.76 & 58.87 & 56.82 & 57.79 \\
\ Recall            &  58.13 & 63.95 & 65.88 & 70.84 & 71.07 & 71.76 \\
\ Precision         &  64.82 & 67.35 & 68.24 & 70.44 & 71.81 & 70.82 \\ \hline 
{\em End of Turn}   &        &       &       &       &       &       \\
\ Errors            &    796 &   469 &   439 &   439 &   436 &   433 \\
\ Error Rate        &  14.70 &  8.66 &  8.11 &  8.11 &  8.05 &  8.00 \\
\ Recall            &  91.33 & 97.78 & 97.91 & 97.91 & 98.04 & 98.05 \\
\ Precision         &  93.79 & 93.81 & 94.20 & 94.20 & 94.14 & 94.17 \\ \hline 
{\em All Tones}     &        &       &       &       &       &       \\
\ Errors            &   4859 &  4180 &  4024 &  3698 &  3581 &  3632 \\
\ Error Rate        &  44.38 & 38.18 & 36.75 & 33.78 & 32.71 & 33.17 \\
\ Recall            &  74.55 & 80.67 & 81.72 & 84.22 & 84.40 & 84.76 \\
\ Precision         &  79.74 & 81.04 & 81.55 & 82.38 & 83.13 & 82.53 \\ \hline 
\end{tabular}
\end{center}
\caption{\label{tab:results-ip}Detecting Intonational Phrase Boundaries}
\end{table}
As explained in Section~\ref{sec:related:tones}, we report separately on
turn-internal boundaries and end-of-turn boundary tones.

The second, third and fourth columns show the effect on boundary tone
detection of modeling POS tags and discourse markers.  Column two
gives the results of using the class-based model of Section~\ref
{sec:model:c:dt:class}.\footnote {The trigram version of the
class-based model gives better tone detection than the 4-gram version,
but with a slight increase in perplexity, as was found in Section~\ref
{sec:model:c:dt:class}.  Since this section only contrasts the
performance of boundary tone detection, we use the trigram version of
the model.}  Column three gives the results of using the POS-based
model of Section~\ref {sec:model:c:dm}, which does not distinguish
discourse markers.  Column four gives the results of using the full
POS-based model, which models discourse marker usage.  Contrasting
the results in column two with those in column four, we see that using
the full POS-based model results in a reduction in the error rate of
11.8\% for turn-internal boundary tones over the class-based model,
and an reduction of 44.8\% for end-of-turn boundary tones.
Furthermore, as can be seen by contrasting the results of column three
and column four, part of this improvement results from modeling discourse
marker usage, which accounts for a 3.4\% reduction in turn-internal
boundary tones and 6.4\% reduction in end-of-turn boundary tones.
Hence, we see that modeling the POS tags and discourse markers allows
much better modeling of boundary tones than can be achieved with a
class-based model.

The fifth column adds in the use of silence information to the model
given in column four.  We see that this results in a 9.1\% error rate
reduction for turn-internal boundary tones.  As explain in
Section~\ref {sec:acoustic:endofturn}, silence information is not used
at the end of speakers' turns, and hence no improvement is seen for
the end-of-turn boundary tone results.

The sixth column adds in the speech repair detection model (and
editing terms).  Here we see that boundary tone identification
is further improved, with an error rate reduction of 3.5\% for
turn-internal tones.  Hence, modeling the detection of speech repairs
improves boundary tone modeling.

The seventh column adds in speech repair correction.  Curiously, this
actually slightly increases the error rate for detecting boundary
tones.  More work is needed to identify why this is happening, but in
any event, the combined speech repair detection and correction model
(column seven) does result in an improvement in boundary tone
identification versus not modeling speech repairs at all (column
five).

\section{Detecting Speech Repairs}
\label{sec:results:detection}

The experiments in this section demonstrate that detecting speech
repairs benefits from modeling POS tags, discourse markers, the
correction of speech repairs, the detection of boundary tones and from
the use of silence information.  We will be looking at two measures of
speech repair detection.  The first measure, referred to as {\em All
Repairs}, ignores errors that are the result of improperly identifying
the type of repair, and hence scores a repair as correctly detected as
long as it was identified as either an abridged repair, modification
repair or fresh start.  For experiments that include speech repair
correction (columns 5, 6 and 7), we further relax this rule.  When
multiple repairs have contiguous removed speech, we count all repairs
involved (of the hand-annotations) as correct as long as the combined
removed speech is correctly identified.  Hence, for the example below
(given earlier as Example~\ref{ex:d92a-1.4:utt25}), if a single repair
is hypothesized with a reparandum of ``one engine the u-'', both of
the hand-annotated repairs would be counted as correctly identified.
\begin{example}{d92a-1.4 utt25}
\label{ex:d92a-1.4:utt25:a}
\reparandum{one engine}\ip \reparandum{the u-}\ip
the first engine will go back to Dansville 
\end{example}

In Section~\ref {sec:detection:R}, we argued that the proper
identification of the type of the repair is necessary for successful
correction of the repair.  Hence, the second measure, referred to as
{\em Exact Repairs}, counts a repair as being correctly identified
only if the type of the repair is also properly determined.  Under
this measure, a fresh start detected as a modification repair is
counted as a false positive and as a missed repair.  Several
exceptions are made to this rule.  First, modification repairs and
fresh starts that have been hand-labeled as ambiguous (i.e.~labeled as
{\bf ip:mod+} or as {\bf ip:can+}, as explained in Section~\ref
{sec:corpus:repairs:annotations}) are counted as correct as long as
they are identified as either a modification repair or a fresh start.
Second, successful correction of speech repairs is the desired result
of speech repair detection.  Hence, if a modification or a fresh start
is misclassified but successfully corrected, it is still counted as
successfully detected.  Third, as with {\em All Repairs}, when
multiple repairs have contiguous removed speech, we count all repairs
involved (of the hand-annotations) as correct as long as the combined
removed speech is correctly identified.  To illustrate how well we do
on each type of repair, we will also give the results for correctly
identifying each type of repair.  These results are simply a breakdown
of {\em Exact Repairs}.

The results are given in Table~\ref {tab:results:detection}.
\begin{table}
\begin{center}\small
\setlength{\lineskip}{-2em}
\begin{tabular}{|l|r|r|r|r|r|r|} \hline
& \corlen        & \corlen 
         & \corlen    & \corlen        &                &\mc{Tones}      \\
&                &  
         &            &                &\mc{Tones}      &\mc{Repairs}    \\
&\mc{Class-Based}&\mc{No DM} 
         &            &\mc{Repairs}    &\mc{Repairs}    &\mc{Corrections}\\
&\mc{Repairs}    &\mc{Repairs}
         &\mc{Repairs}&\mc{Corrections}&\mc{Corrections}&\mc{Silences}   \\
\hline \hline
{\em All Repairs}   &       &       &       &       &       &       \\
\ Errors            &  1246 &  1129 &  1106 &   982 &   909 &   839 \\
\ Error Rate        & 52.00 & 47.12 & 46.16 & 40.98 & 37.93 & 35.01 \\
\ Recall            & 64.98 & 68.69 & 68.61 & 72.82 & 74.29 & 76.79 \\
\ Precision         & 79.27 & 81.28 & 82.28 & 84.05 & 85.86 & 86.66 \\ \hline 
{\em Exact Repairs} &       &       &       &       &       &       \\
\ Errors            &  1640 &  1533 &  1496 &  1240 &  1185 &  1119 \\
\ Error Rate        & 68.44 & 63.98 & 62.43 & 51.75 & 49.45 & 46.70 \\
\ Recall            & 56.76 & 60.26 & 60.47 & 67.44 & 68.53 & 70.95 \\
\ Precision         & 69.24 & 71.30 & 72.52 & 77.84 & 79.20 & 80.07 \\ \hline 
{\em Abridged}      &       &       &       &       &       &       \\
\ Errors            &   199 &   167 &   161 &   187 &   173 &   170 \\
\ Error Rate        & 47.04 & 39.47 & 38.06 & 44.20 & 40.89 & 40.18 \\
\ Recall            & 77.77 & 81.08 & 81.32 & 76.35 & 76.12 & 75.88 \\
\ Precision         & 75.80 & 79.76 & 80.75 & 78.78 & 81.72 & 82.51 \\ \hline
{\em Modification}  &       &       &       &       &       &       \\
\ Errors            &   796 &   767 &   747 &   512 &   489 &   459 \\
\ Error Rate        & 61.13 & 58.90 & 57.37 & 39.32 & 37.55 & 35.25 \\
\ Recall            & 62.59 & 67.20 & 67.81 & 78.41 & 78.95 & 80.87 \\
\ Precision         & 72.50 & 72.01 & 72.91 & 81.54 & 82.70 & 83.37 \\ \hline
{\em Fresh Starts}  &       &       &       &       &       &       \\
\ Errors            &   645 &   599 &   588 &   541 &   523 &   490 \\
\ Error Rate        & 96.12 & 89.26 & 87.63 & 80.62 & 77.94 & 73.02 \\
\ Recall            & 32.19 & 33.68 & 33.08 & 40.53 & 43.51 & 48.58 \\
\ Precision         & 53.20 & 59.47 & 61.49 & 65.70 & 66.97 & 69.21 \\ \hline
\end{tabular}
\end{center}
\caption{\label{tab:results:detection}Detecting Speech Repairs}
\end{table}
Just as in Section~\ref{sec:results:tones}, the second, third and
fourth columns show the effect of modeling POS tags and discourse
markers on speech repair detection.  Column two gives the results of
using the class-based model of Section~\ref{sec:model:c:dt:class};
column three gives the results of using the POS-based model of
Section~\ref {sec:model:c:dm}, which does not distinguish discourse
markers; and column four gives the results of using the full POS-based
model.  Under every measure, the POS-based model (column four) does
significantly better than the class-based approach (column two).  In
terms of overall detection, the POS-based model reduces the error rate
from 52.0\% to 46.2\%, a reduction of 11.2\%.  This shows that speech
repair detection profits from being able to make use of syntactic
generalizations, which are not available from a class-based approach.
By contrasting column three and column four, we see that part of this
improvement is the result of modeling discourse marker usage in the
POS tagset.

The fifth column gives the results from adding in the correction tags
$O_i$, $L_i$ and $C_i$. Here we see that the error rate for detecting
speech repairs decreases from 46.2\% to 41.0\%, a further reduction of
11.2\%.  Part of this reduction is attributed to the better scoring of
overlapping repairs, as illustrated by Example~\ref
{ex:d92a-1.4:utt25:a}. However, from an analysis of the results, we
found that this could account for at most 32 of the 124 fewer
errors. Hence, a reduction of at least 8.3\% is directly attributed to
incorporating speech repair correction.  Hence, integrating speech
repair correction with speech repair detection improves the detection
of speech repairs. These results are consistent with the results that
we have given in earlier work \cite
{HeemanLokenkimAllen96:icslp,HeemanLokenkim95:ieice}, which used an
earlier version of the model presented in this thesis.

In examining the results for each type of speech repairs, we see that
the biggest impact of adding in correction occurs with the
modification repairs.  This should not be surprising since
modification repairs have strong word correspondences that the
correction model can take advantage of, which translates into improved
detection of these repairs.  There is also an improvement for the
detection of fresh starts, but not as strong as the improvement for
modification repairs.  Note that the model of column four does not
incorporate boundary tone identification, which we feel is an
important element in correcting fresh starts.  Curiously, we see that
the performance in detecting abridged repairs actually declines.  This
is partly a result of the correction model erroneously proposing a
correction for some of the abridged repairs, thus confusing them as
either modification repairs or as fresh starts.

The sixth column gives the results of adding in boundary tone
modeling.  Again, we find a noticeable improvement in speech repair
detection, with the error rate decreasing from 41.0\% to 37.9\%, a
reduction of 7.4\%.  Hence we see that modeling the occurrence of
boundary tones improves speech repair detection.  The final column
adds in silence information, which further reduces the error rate by
7.7\%.  Part of this improvement is probably a result of better
modeling of boundary tones, and partially a result of using silence
information to detect speech repairs.\footnote {We purposely chose to
add silence information after adding in the boundary tone modeling.
We have found that without the boundary tones, it is difficult to take
advantage of the silence information.  This is perhaps should not be
unexpected, since boundary tones occur at a much higher rate than
speech repairs and also tend to be accompanied by pauses, as was shown
in Table~\ref{tab:detection:cues}.}  This gives a final detection
recall rate of 76.8\% and a precision of 86.7\%.

\section{Correcting Speech Repairs}
\label{sec:results:correction}

In this section, we present the results for correcting speech repairs
and examine the role that detecting boundary tones and the use of
silence information has on this task.\footnote{We refrain from
comparing the POS-based model to the Class-based model as we did in
Sections~\ref{sec:results:tones} and~\ref {sec:results:detection}.
Our reason for doing this is that the correction model, as formulated,
is allowed to ask questions specific to the POS tags of the proposed
reparandum; e.g. ``is there an intervening discourse marker, or filled
pause''.  Hence, the comparison would not be fair to the class-based
model.}  Again, we subdivide the repairs by their type in order to
show how well each type is corrected.  Note that if a modification or
a fresh start is misclassified but correctly corrected, it is still
counted as correct.  Also, when multiple repairs have contiguous
removed speech, we count all repairs involved as correct as long as
the combined removed speech is correctly identified.  Note that the
extent of the editing term of a repair needs to be successfully
identified in order for the repair to be counted as correctly
identified.
\begin{table}
\begin{center}\small
\setlength{\lineskip}{-2em}
\begin{tabular}{|l|r|r|r|} \hline
       &                  &                  & \mc{Tones}       \\
       &                  & \mc{Repairs}     & \mc{Repairs}     \\
       & \mc{Repairs}     & \mc{Corrections} & \mc{Corrections} \\
       & \mc{Corrections} & \mc{Tones}       & \mc{Silences}    \\ \hline\hline
{\em All Repairs}         &        &        &        \\
\ Errors                  &   1506 &   1411 &   1363 \\
\ Error Rate              &  62.85 &  58.88 &  56.88 \\
\ Recall                  &  61.89 &  63.81 &  65.85 \\
\ Precision               &  71.43 &  73.75 &  74.32 \\ \hline
{\em Abridged}            &        &        &        \\
\ Errors                  &    187 &    175 &    172 \\
\ Error Rate              &  44.20 &  41.37 &  40.66 \\
\ Recall                  &  76.35 &  75.88 &  75.65 \\
\ Precision               &  78.78 &  81.47 &  82.26 \\ \hline
{\em Modification}        &        &        &        \\
\ Errors                  &    616 &    563 &    535 \\
\ Error Rate              &  47.31 &  43.24 &  41.09 \\
\ Recall                  &  74.42 &  76.11 &  77.95 \\
\ Precision               &  77.39 &  79.72 &  80.36 \\ \hline
{\em Fresh Starts}        &        &        &        \\
\ Errors                  &    703 &    673 &    656 \\
\ Error Rate              & 104.76 & 100.29 &  97.76 \\
\ Recall                  &  28.46 &  32.33 &  36.21 \\
\ Precision               &  46.13 &  49.77 &  51.59 \\ \hline
\end{tabular}
\end{center}
\caption{Correcting Speech Repairs}
\label{tab:results:correction}
\end{table}

The results of the comparison are given in Table~\ref
{tab:results:correction}.  The second column gives the results for
correcting speech repairs using the repair, editing term, and
correction models, but without the boundary tone model nor the silence
information. Here we see that we are able to correct 61.9\% of all
speech repairs with a precision of 71.4\%, giving an error rate of
62.9\%.  Note that abridged and modification repairs are corrected at
roughly the same rate but the correction of fresh starts proves
particularly problematic.  In fact, there are more errors in
correcting fresh starts (703) than the number of fresh starts that
occur in the corpus (671), leading to an error rate above 100\%.

The third column gives the results of adding in boundary tone
modeling.  Just as with speech repair detection, we see that this
results in improvements in correcting each type of repair, with the
overall correction error rate decreasing from 62.9 to 58.9, a
reduction of 6.3\%.  This improvement is partly explained by the
increase in the detection rates.  However, since intonational
boundaries are sometimes the onset of speech repair reparanda, it
might also be explained by better correction of the detected repairs.
In fact, from Table~\ref {tab:results:detection}, we see that only 73
fewer errors were made in detecting repairs, while 95 fewer errors
were made in correcting speech repairs.

For the results of the fourth column, we add in silence information.
Silence information is not directly used in correcting speech repairs,
but it is used in detecting repairs and identifying boundary tones,
and hence impacts correction.  We see that the incorporation of
silence information results in a 3.4\% reduction in the correction
error rate.  The final results of the correction model gives a recall
rate of 65.9\% in comparison to the detection recall rate of 76.8\%,
and a precision rate of 74.3\% in comparison to the detection recall
rate of 86.7\%.  By type of repair, we see that fresh starts are
significantly lagging behind modification and abridged repairs.  The
use of higher level syntactic information as well as better acoustic
information to detect speech repairs and boundary tones should prove
helpful.

\section{Collapsing Repair Distinctions}
\label{sec:results:freshstarts}

Our classification scheme distinguishes between fresh starts,
modification repairs, and abridged repairs.  However, not all
classification schemes distinguish between fresh starts and
modification repairs (e.g.~\cite{Shriberg94:thesis}).  In fact,
because of limited training data, we might not even have enough data
to make this a useful distinction.  Furthermore, since fresh starts
are acoustically signaled as such by the speaker and since the only
acoustic source we currently use is silence, we might not be able to
learn this distinction.  In this section, we compare the full model
with one that collapses modification repairs and fresh starts.  To
ensure a fair comparison, we report detection rates in which we do not
penalize incorrect identification of the repair type (the {\em All
Repairs} metric of Section~\ref {sec:results:detection}).

The results of the comparison are given in Table~\ref
{tab:results:collapse}.
\begin{table}
\begin{center}\small
\setlength{\lineskip}{-2em}
\begin{tabular}{|l|r|r|} \hline
                     & \mc{Collapsed}&\mc{Distinct}    \\ \hline \hline
{\em Speech Repairs} &           &          \\ \hline
{\em Detection}      &           &          \\
\ Errors             &      902  &     839  \\ 
\ Error Rate         &    37.64  &   35.01  \\
\ Recall             &    76.25  &   76.79  \\
\ Precision          &    84.58  &   86.66  \\ \hline
{\em Correction}     &           &          \\
\ Errors             &     1460  &    1363  \\ 
\ Error Rate         &    60.93  &   56.88  \\
\ Recall             &    64.60  &   65.85  \\
\ Precision          &    71.66  &   74.32  \\ \hline \hline
{\em Boundary Tones} &           &          \\ \hline
{\em Within Turn}    &           &          \\ 
\ Errors             &     3260  &    3199  \\ 
\ Error Rate         &    58.89  &   57.79  \\
\ Recall             &    71.32  &   71.76  \\
\ Precision          &    70.23  &   70.82  \\ \hline \hline
POS Errors           &     1572  &    1563  \\ 
POS Error Rate       &     2.69  &    2.68  \\ 
Word Perplexity      &    22.32  &   22.35  \\
Branching Perplexity &    30.08  &   30.26  \\ \hline 
\end{tabular}
\end{center}
\caption{Effect of Collapsing Modification Repairs and Fresh Starts}
\label{tab:results:collapse}
\end{table}%
The second column gives the results of collapsing fresh starts and
modification repairs, and the third column gives the results of the
full model, in which fresh starts and modification repairs are treated
separately.  We find that distinguishing fresh starts and modification
repairs results in a 7.0\% improvement in speech repair detection (as
measured by reduction in error rate) and a 6.6\% improvement in speech
repair correction.  Hence, the two types of repairs differ enough both
in how they are signaled and the manner in which they are corrected
that it is worthwhile to model them separately.  Interestingly, we
also see that distinguishing between fresh starts and modification
repairs improves boundary tone detection by 1.9\%.  The improved
boundary tone detection is undoubtedly attributable to the fact that
the reparandum onset of fresh starts interacts more strongly with
boundary tones than does the reparandum onset of modification repairs.

\section{Comparison to Other Work}

Comparing the performance of our model to others that have been
proposed in the literature is very difficult. First, there is the
problem of differences in corpora. The Trains corpus is a collection
of dialogs between two people, both of which realize that they are
talking to another person. The ATIS corpus \cite{Madcow92:snlp}, on
the other hand, is a collection of queries to a speech recognition
system, and hence the speech is very different. The rate of speech
repair occurrence is much lower in this corpus, and almost all speaker
turns consists of just one contribution. A comparison to the
Switchboard corpus \cite {Godfrey-etal92:icassp}, which is a corpus of
human-human dialogs, is also problematic, since those dialogs are much
less constrained and are about a much wider domain. Even more extreme
are differences that result from using read speech rather than
spontaneous speech.

The second problem is that the various proposals have employed
different input criteria. For instance, does the input include POS
tags, some form of utterance segmentation, or hand transcriptions of
the words that were uttered. A third problem is that different
approaches might employ different algorithms to account for aspects
that are not the focus of the comparison. But yet these differences
might explain some of the differences. For instance, in Section~\ref
{sec:model:c:bo:word}, we found that part of the improvement of our
POS model lies in how unknown words are handled. In light of these
problems, we will tread cautiously in comparing our model to others
that have been proposed.

Before proceeding with the comparison, we also note that this work is
the first proposal for combining the detection and correction of
speech repairs, with the identification of boundary tones, discourse
markers and POS tagging in a framework that is amenable to speech
recognition. Hence our comparison will be to systems that address only
part of this problem.  We start with a comparison of the speech repair
results, then the identification of boundary tones and utterance units,
and then the identification of discourse markers.

\subsection{Speech Repairs}
\label{sec:results:c:repairs}

We start with the detection and correction of speech repairs, in which
we obtain an overall correction recall rate of 64.4\% and precision of
74.1\%.  The full results are given in Table~\ref
{tab:results:sum:repairs}.  We also report the results for each type
of repair using the {\em Exact Repair} metric.  To facilitate
comparisons with approaches that distinguish between abridged repairs
but not between modification repairs and fresh starts, we give the
results for detecting and correcting modification repairs and fresh
starts where we do not count errors that result from a confusion
between the two types.
\begin{table}[htb]
\begin{center}\small
\setlength{\lineskip}{-2em}
\begin{tabular}{|l|r|r|r|} \hline
                     & \mc{Recall}& Precision & Error Rate \\ \hline \hline
{\em All Repairs}    & \corlen  &           &            \\ 
\ Detection          &   76.79  &   86.66   &   35.01    \\ 
\ Correction         &   65.85  &   74.32   &   56.88    \\ \hline
{\em Abridged}       &          &           &            \\ 
\ Detection          &   75.88  &   82.51   &   40.18    \\
\ Correction         &   75.65  &   82.26   &   40.66    \\ \hline
{\em Modification}   &          &           &            \\
\ Detection          &   80.87  &   83.37   &   35.25    \\
\ Correction         &   77.95  &   80.36   &   41.09    \\ \hline
{\em Fresh Starts}   &          &           &            \\
\ Detection          &   48.58  &   69.21   &   73.02    \\
\ Correction         &   36.21  &   51.59   &   97.76    \\ \hline
{\em Modification \& Fresh Starts}
                     &          &           &            \\
\ Detection          &   73.69  &   83.85   &   40.49    \\
\ Correction         &   63.76  &   72.54   &   60.36    \\ \hline
\end{tabular}
\end{center}
\caption{Summary of Speech Repair Detection and Correction Results}
\label{tab:results:sum:repairs}
\end{table}

We avoid comparing ourself to models that focus only on correction
(e.g.~\cite{Hindle83:acl,KikuiMorimoto94:icslp}). Such models assume
that speech repairs have already been identified, and so do not
address this problem.  Furthermore, as we demonstrated in Section~\ref
{sec:results:detection}, speech repair detection profits from
combining detection and correction.

Of relevance to our work is the work by Bear \etal~\shortcite
{Bear-etal92:acl} and Dowding \etal~\shortcite
{Dowding-etal93:acl}. This work was done on the ATIS corpus. Bear
\etal used a simple pattern matching approach on the word
transcriptions and obtained a correction recall rate of 43\% and a
precision of 50\% on a corpus from which they removed repairs
consisting of just a filled pause or word fragment. Although word
fragments indicate a repair, they do not indicate the extent of the
repair. Also, our rates are not based on assuming that all filled
pauses should be treated equally, but are based on classifying them as
abridged repairs only if they are mid-utterance.  Dowding
\etal~\shortcite{Dowding-etal93:acl} used a similar setup for their
data. In this experiment they used a parser-first approach in which
the pattern matching routines are only applied if the parser
fails. Using this approach they obtained a correction recall rate of
30\% and a precision of 62\%.

Nakatani and Hirschberg \shortcite{NakataniHirschberg94:jasa} examined
how speech repairs can be detected using a variety of information,
including acoustic, lexical, presence of word matchings, and POS tags.
Using these cues they were able to train a decision tree that
achieved a recall rate of 86.1\% and a precision of 92.1\% on a subset
of the ATIS corpus.  The cues they found most useful were pauses,
presence of word fragments, and lexical matching.  Note that in their
corpus 73.3\% of the repairs were accompanied by a word fragment, as
opposed to 32\% of the modification repairs and fresh starts in the
Trains corpus.  Hence, word fragments are a stronger indicator of
speech repairs in their corpus than in the Trains corpus.  Also note
that since their training set and test sets only included turns with
speech repairs; hence ``[the] findings should be seen more as indicative of
the relative importance of various predictors of [speech repair]
location than as a true test of repair site location'' (pg.~1612).

Stolcke and Shriberg \shortcite {StolckeShriberg96:icassp} modeled
simple types of speech repairs in a language model, and find that it
actually makes their perplexity worse.  They attribute this problem to
not having a linguistic segmentation available, which would allow
utterance-initial filled pauses to be treated separately from
utterance-medial filled pauses.  As we mentioned in Section~\ref
{sec:intro:classification}, our annotation scheme distinguishes
between utterance-medial filled pauses and utterance-initial ones by
only treating the utterance-medial ones as abridged repairs.  Hence,
our model distinguishes automatically between these two types of
filled pauses.  Furthermore, especially for distinguishing
utterance-medial filled pauses, one needs to also model the occurrence
of boundary tones and discourser markers, as well as incorporate
syntactic disambiguation.

\subsection{Utterance Units and Boundary Tones}
\label{sec:results:c:tones}

In this section, we contrast our results in identifying boundary
tones with the results of other researchers in identifying boundary
tones, or other definitions of utterance units.  Table~\ref
{tab:results:sum:tones} gives our performance.
\begin{table}[htb]
\begin{center}\small
\setlength{\lineskip}{-2em}
\begin{tabular}{|l|r|r|r|} \hline
     & \makebox[\mcl][c]{Recall}& Precision & Error Rate \\ \hline \hline
Within Turn        & 71.76    & 70.82     & 57.79     \\
End of Turn        & 98.05    & 94.17     &  8.00     \\
All Tones          & 84.76    & 82.53     & 33.17     \\ \hline
\end{tabular}
\end{center}
\caption{Summary of Boundary Tone Identification Results}
\label{tab:results:sum:tones}
\end{table}
Note especially the difference in results when we factor in turn-final
tones.  Almost all turns in the Trains corpus end in a turn, and hence
when comparing our results, we will try to account for such
tones.\footnote {See Traum and Heeman~\shortcite
{TraumHeeman97:chapter} for an analysis of turns that do not end with
a boundary tone.}

For detecting boundary tones, the model of Wightman and Ostendorf
\shortcite {WightmanOstendorf94:ieee} performs very well.  They achieve 
a recall rate of 78.1\% and a precision of 76.8\%, in contrast to our
turn-internal recall of 70.5\% and precision rate of 69.4\%.  This difference
is partly attributed to their better acoustic modeling, which is
speaker dependent.  However, their model was trained and tested on
professionally read speech, and it is unclear how their model will be
able to deal with spontaneous speech, especially since a number of
the cues they use for detecting boundaries are the same cues that
signal speech repairs.

Wang and Hirschberg~\shortcite{WangHirschberg92:csl} did employ
spontaneous speech, in fact they used the ATIS corpus.  For
turn-internal boundary tones, they achieved a recall rate 72.2\% and a
precision of 76.2\% using a decision tree approach that combined both
textual features, such as POS tags, and syntactic constituents with
intonational features, namely observed pitch accents.  These results
are difficult to compare to our results because they are from a
decision tree that classifies disfluencies as boundary tones.  In
their corpus, there were 424 disfluencies and 405 turn-internal
boundary tones.  The recall rate of the decision tree that does not
classify disfluencies as boundary tones is significantly worse.
However, these results were achieved using approximately one-tenth the
amount of data that is in the Trains corpus.  Our approach differs
from theirs since their decision trees are used to classify each data
point independently of the next.  Our decision trees are used to
provide a probability estimate for the tone given the previous
context, while other trees predict the likelihood of future events,
including the occurrence of speech repairs and discourse markers,
based on the presence or absence of a tone in the context.  This might
lead to a much richer model from which to predict boundary tones.  In
addition, our model provides a basis upon which boundary tone
detection can be directly incorporated into a speech recognition
model (cf.~\cite{Hirschberg91:eurospeech}).

The models of Kompe \etal~\shortcite{Kompe-etal94:icassp} and
Mast \etal~\shortcite{Mast-etal96:icslp} are the most similar to our
model in terms of incorporating a language model.  Mast \etal~achieve
a recall rate of 85.0\% and a precision of 53.1\%.  Given the skew in
their results towards recall it is difficult to compare these results
to our own.  In terms of error rates, their model achieves an error
rate of 90.1\%, in comparison to our error rate of 60.5\%.  However,
their task was dialog act segmentation on a German corpus, so again it
is unclear how valuable a comparison of results is.  Their model does
employ a much more fine grained acoustic analysis, however, it does
not account for other aspects of utterance modeling, such as speech
repairs.

\subsection{Discourse Marker Identification}
\label{sec:results:c:dm}

Table~\ref {tab:results:sum:dm} gives the results of our full model in 
identifying discourse markers.
\begin{table}[htb]
\begin{center}\small
\setlength{\lineskip}{-2em}
\begin{tabular}{|l|r|} \hline
Errors     &   533 \\ 
Error Rate &  6.43 \\ 
Recall     & 97.26 \\ 
Precision  & 96.32 \\ \hline
\end{tabular}
\end{center}
\caption{Discourse Marker Identification}
\label{tab:results:sum:dm}
\end{table}
The only other work in automatically identifying discourse markers is
the work of Hirschberg and Litman \shortcite{HirschbergLitman93:cl}
and Litman \shortcite{Litman96:jair}.  As explained in
Section~\ref{sec:related:dm}, Litman improves on the results of
Hirschberg and Litman by using machine learning techniques to
automatically build algorithms for classifying ambiguous lexical items
as to whether they are being used as a discourse markers.  The
features that the learning algorithm can query are intonational
features, namely information about the phrase accents (which mark the
end of intermediate phrases) boundary tones, and the lexical item
under consideration.  She also explored other features, such as the
POS tag of the word and whether the word has a pitch accent, but these
features were not used in the best model.  With this approach, she was
able to achieve an error rate of 37.3\% in identifying discourse
markers.  

Direct comparisons with our results are problematic since our corpus
is approximately five times as large.  Further, we use task-oriented
human-human dialogs rather than a monologue, and hence our corpus
includes a lot of turn-initial discourse markers for co-ordinating
mutual belief.  However, our results are based on automatically
identifying intonational boundaries, rather than including these as
part of the input.  In any event, the work of Litman and the earlier
work with Hirschberg indicate that our results can be further improved
by also modeling intermediate phrase boundaries (phrase accents), and
word accents, and by improving our modeling of these events, perhaps
by using more acoustic cues.  Conversely, we feel that our approach,
which integrates discourse marker identification with speech
recognition along with POS tagging, boundary tone identification and
the resolution of speech repairs, allows different interpretations to
be explored in parallel, rather than forcing individual decisions to
be made about each ambiguous token.  This allows interactions between
these problems to be modeled, which we feel accounts for some of the
improvement between our results and the results reported by Litman.

\cleardoublepage
\chapter{Conclusion and Future Work}
\label{chapter:conclusion}

This thesis concerns modeling speakers' utterances.  In spoken dialog,
speakers often make more than one contribution or utterance in a turn.
Speech repairs complicate this since some of the words are not even
intended to be part of the utterance.  In order to understand the
speaker's utterance, we need to segment the turn into utterance units
and resolve all speech repairs that occur.  Discourse markers and
boundary tones are devices that speakers use to help indicate this
segmentation; as well, discourse markers play a role in signaling
speech repairs.  In the introduction, we argued that these three
problems are intertwined, and are also intertwined with the problem of
determining the syntactic role (or POS tag) of each word in the turn
as well as the speech recognition problem of predicting the next word
given the previous context.

In this thesis, we proposed a model that can detect and correct speech
repairs, including their editing terms, and identify boundary tones
and discourse markers.  This model is based on a statistical language
model that also determines the POS tag for each word involved.  The
model was derived by redefining the speech recognition problem.
Rather than just predicting the next word, the model also predicts the
POS tags, discourse markers, boundary tones and speech repairs.  Thus
the model can account for the interactions that exist between these
phenomena.  The model also allows these problems to be resolved using
local context without bringing to bear full syntactic and semantic
analysis.  This means that these tasks can be done prior to parsing
and semantic interpretation, thus separating these modules from the
complications that these problems would otherwise introduce.

Constraining the language model to the hand transcription of the
dialogs, our model is able to identify 71.8\% of all turn-internal
intonational boundaries with a precision rate of 70.8\%, and we are
able to detect and correct 65.9\% of all speech repairs with a
precision of 74.3\%.  These results are partially attributable to
accounting for the interaction between these two tasks, as well as the
interaction between detecting speech repairs and correcting them.  In
Section~\ref {sec:results:tones}, we showed that modeling speech
repair detection results in a 3.5\% improvement in modeling
turn-internal boundary tones.  Section~\ref {sec:results:detection}
showed that modeling boundary tones results in a 7.4\% improvement in
detecting speech repairs, while modeling the correction of speech
repairs results in an 11.2\% improvement in detecting speech repairs.
We also see that modeling boundary tones results in a 6.3\%
improvement in correcting speech repairs.

Our model also identifies discourse marker usage by using special POS
tags.  Our full model is able to identify 97.3\% of all discourse
markers with a precision of 96.3\%.  The thesis argued that discourse
marker identification is intertwined with resolving speech repairs and
identifying boundary tones.  Section~\ref {sec:results:tones} and \ref
{sec:results:detection} demonstrated that modeling discourse markers
improves our ability to detect speech repairs and boundary tones.
Conversely, Section~\ref {sec:results:tagging} demonstrated that
discourse marker identification improves by 15.4\% by modeling speech
repairs and boundary tones.

Our thesis also claimed that POS tagging was interrelated to discourse
marker, speech repair and boundary tone modeling.  Section~\ref
{sec:model:c:dm} demonstrated that distinguishing discourse marker
usage results in a small improvement in POS tagging and Section~\ref
{sec:results:tagging} demonstrated that modeling speech repairs and
boundary tones results in a 8.1\% reduction in the POS error rate.
Conversely, Section~\ref {sec:results:tones} demonstrated that using a
POS-based model instead of a class-based model results in a 11.8\%
improvement for detecting turn-internal boundary tones and
Section~\ref {sec:results:detection} demonstrated it results in an
11.2\% improvement in detecting speech repairs.

Since our model is a statistical language model, it means that the
tasks of detecting and correcting speech repairs, identifying
intonational boundary tones, discourse markers and POS tags can be
done in conjunction with speech recognition, with the model serving as
the language model that the speech recognizer uses to prune acoustic
alternatives.  This approach is attractive because speech repairs and
boundary tones present discontinuities that traditional speech
recognition language models have difficulty modeling.  Just as
modeling speech repairs, intonational boundaries and discourse markers
improves POS tagging, the same holds for the speech recognition task
of predicting the next word given the previous context.  In terms of
perplexity, a measure used to determine the ability of a language
model to predict the next word, our results reveal an improvement from
26.1 for a word-based trigram backoff model to 22.4 using our model
that accounts for the user's utterances, and the discourse phenomena
that occur in them.  In comparison to a POS-based language model built
using the same decision tree technology for estimating the probability
distributions as is used for the full model, we see a perplexity
improvement from 24.0 to 22.4, a reduction of 7.0\%.

The results of this thesis show that tasks long viewed as the domain
of discourse processing, such as identifying discourse markers,
determining utterance segmentation, and resolving speech repairs need
to be modeled very early on in the processing stream, and that by
doing this we can improve the robustness of actually determining what
the speaker said.  Hence, this thesis is helping to build a bridge
between discourse processing and speech recognition.

This thesis has made use of a number of techniques in order to
estimate the probability distributions needed by the statistical
language model.  One of the most important was the use of decision
trees, which can decide what features of the context to use in
estimating the probability distributions.  Using decision trees made
it possible for us to expand beyond traditional POS technology, which
ignores a lot of critical features of the context as demonstrated in
Section~\ref {sec:model:r:richer}.  Although these extra features,
namely word identities, only reduce the POS tagging rate by 3.8, they
do result in a POS tagging model that is usable as a language model,
improving perplexity from 43.2 to 24.0, which is even better than the
perplexity of a word-based backoff model trained on the same data,
which gave a perplexity of 26.1.

In using word and POS information in a decision tree, we advocated
building word and POS classification trees so as to allow the decision
tree to ask more meaningful questions and generalize about similar
words and POS tags.  In Section~\ref {sec:model:r:ctrees}, we
demonstrated that word information can be viewed as a further
refinement of the POS tags.  This means that POS and word information
do not have to be viewed as two completing sources of information
about the context, and allows a better quality word classification
tree to be learned from the training data, as well as significantly
speeding up the training procedure, as discussed in Section~\ref
{sec:model:ctrees} and Section~\ref {sec:model:grow}.

Using the Trains corpus has limited the amount of data that we can use
for training the language model.  Rather than relying on the order of
a million words or so to build the model, we use approximately 50,000
words of data.  Hence, one of the issues that we were faced with in
this work was to make maximum use of the limited amount of training
data that we had.  This was a factor in our use of the word identities
as a further refinement of the POS tags.  It was also a factor in
determining what questions the decision tree could ask about the
context when modeling the occurrence of speech repairs and boundary
tones, so as to allow appropriate generalizations between instances
with speech repairs and boundary tones and instances without.

There are many directions that this research work can be pursued.
First, with the exception of silence durations between words, we do
not consider acoustic cues.  This is an area that we are currently
exploring and will undoubtedly have the most impact on detecting fresh
starts and boundary tones.  It will also improve our ability to
determine the onset of the reparanda of fresh starts.  In our corpus
of spoken dialogs, speakers sometimes make several contributions in a
turn, and the previous intonation phrase boundary is a likely
candidate for the onset of the reparandum.  By simply including the
silence duration between words, we found that the error rate for
boundary tones improved by 9.1\%.  Acoustic modeling is also needed in
order to help identify word fragments, which were labeled as fragments
in the input for the experiments in this thesis, as explained in
Section~\ref {sec:model:setup}.

The second area that we have not delved into is using higher level
syntactic and semantic knowledge.  Having access to partial syntactic
and even semantic interpretation would give a richer context for
modeling a speaker's contribution, especially in detecting the
ill-formedness that often occurs at the interruption point of speech
repairs.  It would also help in finding higher level correspondences
between the reparandum and alteration.  For instance, we can not
currently account for the replacement of a noun phrase with a pronoun,
as in the following example.
\begin{example}{d93-14.3 utt27}
\reparandum{the engine can take as many}\ip
\et{um} 
\alteration{it can take} up to three loaded boxcars
\end{example}
Given recent work in statistical parsing \cite
{Magerman94:thesis,JoshiSrinivas94:coling}, it should be possible to
incorporate and make use of such information.

A third area that we are interested in exploring is the use of our
model with other languages.  Since the modeling of boundary tones,
speech repairs and discourse markers is completely learned from a
training corpus (unlike the modeling of speech repair correction in
Heeman and Allen \shortcite{HeemanAllen94:acl}), it should be possible
to apply this model to corpora in other languages.  Preliminary work
on the Artimis-AGS corpus \cite
{Sadek-etal96:icslp,Sadek-etal97:ijcai}, a corpus of human-computer
dialogs where the human queries the system about information services
available through France T\'el\'ecom, indicates that the model is not
English specific nor specific to human-human corpora \cite
{Heeman97:cnet}.

The fourth and probably the most important area that we need to
further explore is tying our model with a speech recognizer.  Our
modeling of intonational boundary tones, discourse markers and speech
repair detection and correction is ideally suited for this task.  Our
perplexity improvements indicate that our model should improve speech
recognition results.  However, as we pointed out in Section~\ref
{sec:model:c:bo:word}, our improvement over a word-based model occurs
with the lower probability words.  Hence, if the word error rate of a
word-based approach is above a certain threshold, then the improvement
that will be gained from the richer language modeling will not be as
large as would be expected.

\cleardoublepage


\begin{thebibliography}{}

\bibitem[\protect\citename{Allen {\em et~al.}, }1996]{Allen-etal96:acl}
J.~F. Allen, B.~W. Miller, E.~K. Ringger, and T.~Sikorski,
\newblock ``A Robust System for Natural Spoken Dialogue,''
\newblock In {\em Proceedings of the 34$^{th}$ Annual Meeting of the
  Association for Computational Linguistics}, June 1996.

\bibitem[\protect\citename{Allen and Perrault, }1980]{AllenPerrault80:ai}
James~F. Allen and C.~Raymond Perrault,
\newblock ``Analyzing Intention in Utterances,''
\newblock {\em Artificial Intelligence}, 15:143--178, 1980,
\newblock Reprinted in \cite{readings-in-nlp:86}, pages~441--458.

\bibitem[\protect\citename{Allen {\em et~al.}, }1995]{Allen-etal95:jetai}
James~F. Allen, Lenhart~K. Schubert, George Ferguson, Peter Heeman, Chung~Hee
  Hwang, Tsuneaki Kato, Marc Light, Nathaniel Martin, Bradford Miller, Massimo
  Poesio, and David~R. Traum,
\newblock ``The {T}rains Project: A case study in building a conversational
  planning agent,''
\newblock {\em Journal of Experimental and Theoretical AI}, 7:7--48, 1995,
\newblock Also published as Trains TN 94-3 and TR 532, Computer Science Dept.,
  U. Rochester, September 1994.

\bibitem[\protect\citename{Anderson {\em et~al.}, }1991]{Anderson-etal91:ls}
Anne~H. Anderson, Miles Bader, Ellen~Curman Bard, Elizabeth Boyle, Gwyneth
  Doherty, Simon Garrod, Stephen Isard, Jacqueline Kowtko, Jan McAllister, Jim
  Miller, Catherine Sotillo, Henry Thompson, and Regina Weinert,
\newblock ``The {HCRC} Map Task Corpus,''
\newblock {\em Language and Speech}, 34(4):351--366, 1991.

\bibitem[\protect\citename{Austin, }1962]{Austin62:book}
J.~L. Austin,
\newblock {\em How to do things with words},
\newblock Oxford University Press, New York, 1962.

\bibitem[\protect\citename{Bahl {\em et~al.}, }1977]{Bahl-etal77:asa}
L.~R. Bahl, J.~K. Baker, F.~Jelinek, and R.~L. Mercer,
\newblock ``Perplexity---A Measure of the Difficulty of Speech Recognition
  Tasks,''
\newblock In {\em Proceedings of the 94th Meeting of the Acoustical Society of
  America}, 1977.

\bibitem[\protect\citename{Bahl {\em et~al.}, }1989]{Bahl-etal89:tassp}
L.~R. Bahl, P.~F. Brown, P.~V. deSouza, and R.~L. Mercer,
\newblock ``A Tree-Based Statistical Language Model for Natural Language Speech
  Recognition,''
\newblock {\em {IEEE} Transactions on Acoustics, Speech, and Signal
  Processing}, 36(7):1001--1008, 1989.

\bibitem[\protect\citename{Beach, }1991]{Beach91:jml}
Cheryl~M. Beach,
\newblock ``The Interpretation of Prosodic Patterns at Points of Syntactic
  Structure Ambiguity: Evidence for Cue Trading Relations,''
\newblock {\em Journal of Memory and Language}, 30(6):644--663, 1991.

\bibitem[\protect\citename{Bear {\em et~al.}, }1992]{Bear-etal92:acl}
John Bear, John Dowding, and Elizabeth Shriberg,
\newblock ``Integrating Multiple Knowledge Sources for Detection and Correction
  of Repairs in Human-Computer Dialog,''
\newblock In {\em Proceedings of the 30$^{th}$ Annual Meeting of the
  Association for Computational Linguistics}, pages 56--63, 1992.

\bibitem[\protect\citename{Bear {\em et~al.}, }1993]{Bear-etal93:tr}
John Bear, John Dowding, Elizabeth Shriberg, and Patti Price,
\newblock ``A System for Labeling Self-Repairs in Speech,''
\newblock Technical Note 522, SRI International, February 1993.

\bibitem[\protect\citename{Bear and Price, }1990]{BearPrice90:acl}
John Bear and Patti Price,
\newblock ``Prosody, Syntax, and Parsing,''
\newblock In {\em Proceedings of the 28$^{th}$ Annual Meeting of the
  Association for Computational Linguistics}, pages 17--22, Pittsburgh, June
  1990.

\bibitem[\protect\citename{Beckman and Ayers, }1994]{BeckmanAyers94:tr}
Mary~E. Beckman and Gayle~M. Ayers,
\newblock ``Guidelines for {ToBI} Labelling, version 2.0,''
\newblock Manuscript and accompanying speech materials, Ohio State University,
  (obtain by writing to tobi@ling.ohio-state.edu), 1994.

\bibitem[\protect\citename{Beckman and Hirschberg,
  }1994]{BeckmanHirschberg94:tr}
Mary~E. Beckman and Julia Hirschberg,
\newblock ``The {ToBI} Annotation Conventions,''
\newblock Manuscript, Ohio State University, (obtain by writing to
  tobi@ling.ohio-state.edu), 1994.

\bibitem[\protect\citename{Black {\em et~al.}, }1992a]{Black-etal92:darpa:hbg}
Ezra Black, Fred Jelinek, John Lafferty, David Magerman, Robert Mercer, and
  Salim Roukos,
\newblock ``Towards History-based Grammars: Using Richer Models for
  Probabilistic Parsing,''
\newblock In {\em Proceedings of the DARPA Speech and Natural Language
  Workshop}, pages 134--139. Morgan Kaufman, 1992.

\bibitem[\protect\citename{Black {\em et~al.}, }1992b]{Black-etal92:darpa:pos}
Ezra Black, Fred Jelinek, John Lafferty, Robert Mercer, and Salim Roukos,
\newblock ``Decision Tree Models Applied to the Labeling of Text with
  Parts-of-Speech,''
\newblock In {\em Proceedings of the DARPA Speech and Natural Language
  Workshop}, pages 117--121. Morgan Kaufman, 1992.

\bibitem[\protect\citename{Blackmer and Mitton, }1991]{BlackmerMitton91:cog}
E.~R. Blackmer and J.~L. Mitton,
\newblock ``Theories of Monitoring and the Timing of Repairs in Spontaneous
  Speech,''
\newblock {\em Cognition}, 39:173--194, 1991.

\bibitem[\protect\citename{Bloomfield, }1926]{Bloomfield26}
Leonard Bloomfield,
\newblock ``A set of Postulates for the Science of Language,''
\newblock {\em Language}, 2:153--164, 1926.

\bibitem[\protect\citename{Breiman {\em et~al.}, }1984]{Breiman-etal84:book}
Leo Breiman, Jerome~H. Friedman, Richard~A. Olshen, and Charles~J. Stone,
\newblock {\em Classification and Regression Trees},
\newblock Wadsworth \& Brooks, Monterrey, CA, 1984.

\bibitem[\protect\citename{Brill, }1995]{Brill95:cl}
Eric Brill,
\newblock ``Transformation-Based Error-Driven Learning and Natural Language
  Processing: A Case Study in Part of Speech Tagging,''
\newblock {\em Computational Linguistics}, 21(4), 1995.

\bibitem[\protect\citename{Brown and Yule, }1983]{BrownYule83:book}
Gillian Brown and George Yule,
\newblock {\em Discourse Analysis},
\newblock Cambridge University Press, Cambridge, 1983.

\bibitem[\protect\citename{Brown {\em et~al.}, }1992]{Brown-etal92:cl}
Peter~F. Brown, Vincent~J. {Della Pietra}, Peter~V. deSouza, Jenifer~C. Lai,
  and Robert~L. Mercer,
\newblock ``Class-Based $n$-gram Models of Natural Language,''
\newblock {\em Computational Linguistics}, 18(4):467--479, 1992.

\bibitem[\protect\citename{Byron and Heeman, }1997]{ByronHeeman97:eurospeech}
Donna~K. Byron and Peter~A. Heeman,
\newblock ``Discourse Marker Use in Task-Oriented Spoken Dialog,''
\newblock In {\em Proceedings of the 5$^{th}$ European Conference on Speech
  Communication and Technology (Eurospeech)}, Rhodes, Greece, September 1997.

\bibitem[\protect\citename{Charniak {\em et~al.}, }1993]{Charniak-etal93:aaai}
E.~Charniak, C.~Hendrickson, N.~Jacobson, and M.~Perkowitz,
\newblock ``Equations for Part-of-Speech Tagging,''
\newblock In {\em Proceedings of the National Conference on Artificial
  Intelligence (AAAI~'93)}, 1993.

\bibitem[\protect\citename{Charniak, }1993]{Charniak93:book}
Eugene Charniak,
\newblock {\em Statistical Language Learning},
\newblock MIT Press, Cambridge, Massachusetts, 1993.

\bibitem[\protect\citename{Chen and Goodman, }1996]{ChenGoodman96:acl}
Stanley~F. Chen and Joshua~T. Goodman,
\newblock ``An Empirical Study of Smoothing Techniques for Language Modeling,''
\newblock In {\em Proceedings of the 34$^{th}$ Annual Meeting of the
  Association for Computational Linguistics}, 1996.

\bibitem[\protect\citename{Chow and Schwartz, }1989]{ChowSchwartz89:darpa}
Y.S. Chow and R.~Schwartz,
\newblock ``The N-Best Algorithm: An Efficient Procedure for Finding Top N
  Sentence Hypotheses,''
\newblock In {\em Proceedings of the DARPA Speech and Natural Language
  Workshop}, pages 199--202, San Mateo, California, October 1989. Morgan
  Kaufman.

\bibitem[\protect\citename{Church, }1988]{Church88:anlp}
K.~Church,
\newblock ``A Stochastic Parts Program and Noun Phrase Parser for Unrestricted
  Text,''
\newblock In {\em Proceedings of the 2nd Conference on Applied Natural Language
  Processing}, pages 136--143, Febuary 1988.

\bibitem[\protect\citename{Clark, }1996]{Clark96:book}
Herbert~H. Clark,
\newblock {\em Using Language},
\newblock Cambridge University Press, Cambridge, 1996.

\bibitem[\protect\citename{Cohen and Perrault, }1979]{CohenPerrault79:cs}
Philip~R. Cohen and C.~Raymond Perrault,
\newblock ``Elements of a Plan-Based Theory of Speech Acts,''
\newblock {\em Cognitive Science}, 3(3):177--212, 1979,
\newblock Reprinted in \cite{readings-in-nlp:86}, pages~423--440.

\bibitem[\protect\citename{Cohen, }1984]{RCohen84:coling}
Robin Cohen,
\newblock ``A Computational Theory of the Function of Clue Words in Argument
  Understanding,''
\newblock In {\em Proceedings of the 10$^{th}$ International Conference on
  Computational Linguistics (COLING)}, pages 251--255, 1984.

\bibitem[\protect\citename{Cohen, }1992]{WCohen92:ml}
W.~W. Cohen,
\newblock ``Compiling Knowledge into an Explicit Bias,''
\newblock In {\em Proceedings of the Ninth International Conference on Machine
  Learning}, 1992.

\bibitem[\protect\citename{Cohen, }1993]{WCohen93:ijcai}
W.~W. Cohen,
\newblock ``Efficient Pruning Methods for Separate-and-Conquer Rule Learning
  Systems,''
\newblock In {\em Proceedings of the International Joint Conference on
  Artificial Intelligence (IJCAI~'93)}, 1993.

\bibitem[\protect\citename{Cole {\em et~al.}, }1994]{Cole-etal94:icslp}
R.~A. Cole, D.~G. Novick, M.~Fanty, P.~Vermeulen, S.~Sutton, D.~Burnett, and
  J.~Schalkwyk,
\newblock ``A Prototype Voice-Response Questionaire for the U.S. Cenus,''
\newblock In {\em Proceedings of the 3rd International Conference on Spoken
  Language Processing (ICSLP-94)}, pages 683--686, Yokohama, Japan, 1994.

\bibitem[\protect\citename{Cruttenden, }1986]{Cruttenden86:book}
Alan Cruttenden,
\newblock {\em Intonation},
\newblock Cambridge University Press, Cambridge, 1986.

\bibitem[\protect\citename{Crystal, }1980]{Crystal80}
D.~Crystal,
\newblock ``Neglected Grammatical Factors in Conversational {E}nglish,''
\newblock In S.~Greenbaum, G.~Leech, and J.~Svartvik, editors, {\em Studies in
  English Linguistics}. Longman, 1980.

\bibitem[\protect\citename{Dermatas and Kokkinakis,
  }1995]{DermatasKokkinakis95:cl}
Evangelos Dermatas and George Kokkinakis,
\newblock ``Automatic Stochastic Tagging of Natural Language Texts,''
\newblock {\em Computational Linguistics}, 21(2):137--163, June 1995.

\bibitem[\protect\citename{DeRose, }1988]{DeRose88:cl}
Steven~J DeRose,
\newblock ``Grammatical Category Disambiguation by Statistical Optimization,''
\newblock {\em Computational Linguistics}, 14(1):31--39, 1988.

\bibitem[\protect\citename{Dowding {\em et~al.}, }1993]{Dowding-etal93:acl}
John Dowding, Jean~Mark Gawron, Doug Appelt, John Bear, Lynn Cherny, Robert
  Moore, and Douglas Moran,
\newblock ``Gemini: A Natural Language System for Spoken-Language
  Understanding,''
\newblock In {\em Proceedings of the 31$^{th}$ Annual Meeting of the
  Association for Computational Linguistics}, pages 54--61, 1993.

\bibitem[\protect\citename{Duda and Hart, }1973]{DudaHart73:book}
R.~O. Duda and P.~E. Hart,
\newblock {\em Pattern Classification and Scene Analysis},
\newblock Wiley, New York, 1973.

\bibitem[\protect\citename{Entropic, }1993]{Entropic93:waves}
Entropic Research Laboratory, Inc.,
\newblock {\em WAVES+ Reference Manual}, 1993,
\newblock Version 5.0.

\bibitem[\protect\citename{Entropic, }1994]{Entropic94:aligner}
Entropic Research Laboratory, Inc.,
\newblock {\em Aligner Reference Manual}, 1994,
\newblock Version 1.3.

\bibitem[\protect\citename{Finkler, }1997a]{Finkler97:thesis}
Wolfgang Finkler,
\newblock ``Automatische Selbstkorrektur bei der inkrementellen Generierung
  gesprochener Sprache unter Realzeitbedingungen: Ein empirisch-simulativer
  Ansatz unter Verwendung eines Begr{\"u}ndungsverwaltungssystems,''
\newblock Doctoral dissertation, Technische Fakult{\"a}t, Universit{\"a}t
  Saarbr{\"u}cken, Germany, January 1997.

\bibitem[\protect\citename{Finkler, }1997b]{Finkler97}
Wolfgang Finkler,
\newblock ``Nonmonotonic Aspects of Incremental Natural Language Production:
  Performing Self-Corrections in a Situated Generator,''
\newblock In {\em DFKI Workshop on Natural Language Generation}. German
  Research Center for Artificial Intelligence (DFKI), Saarbr{\"u}cken, Germany,
  1997.

\bibitem[\protect\citename{Ford and Thompson, }1991]{FordThompson91}
Cecelia Ford and Sandra Thompson,
\newblock ``On Projectability in Conversation: Grammar, Intonation, and
  Semantics,''
\newblock Presented at the {\em Second International Cognitive Linguistics
  Association Conference}, August 1991.

\bibitem[\protect\citename{Gee and Grosjean, }1983]{GeeGrosjean83:cp}
James~Paul Gee and Francois Grosjean,
\newblock ``Saying What You Mean in Dialogue: A Study in Conceptual and
  Semantic Co-ordination,''
\newblock {\em Cognitive Psychology}, 15(3):411--458, 1983.

\bibitem[\protect\citename{Godfrey {\em et~al.}, }1992]{Godfrey-etal92:icassp}
J.~J. Godfrey, E.~C. Holliman, and J.~McDaniel,
\newblock ``{SWITCHBOARD}: {T}elephone Speech Corpus for Research and
  Development,''
\newblock In {\em Proceedings of the International Conference on Audio, Speech
  and Signal Processing (ICASSP)}, pages 517--520, 1992.

\bibitem[\protect\citename{Good and Butterworth, }1980]{GoodButterworth80}
D.~A. Good and B.~L. Butterworth,
\newblock ``Hesitancy as a Conversational Resource: {S}ome Methodological
  Implications,''
\newblock In H.~W. Dechert and M.~Raupach, editors, {\em Termporal Variables in
  Speech: Studies in Honour of Frieda Goldman-Eisler}. Mounton, The Hague,
  1980.

\bibitem[\protect\citename{Goodwin, }1991]{Goodwin81:book}
C.~Goodwin,
\newblock {\em Conversational Organization: {I}nteraction between Speakers and
  Hearers},
\newblock Academic Press, New York, 1991.

\bibitem[\protect\citename{Greene and Rubin, }1981]{Greene-Rubin81:brown}
B.~B. Greene and G.~M. Rubin,
\newblock ``Automatic Grammatical Tagging of English,''
\newblock Department of Linguistics, Brown University, Providence, R.I., 1981.

\bibitem[\protect\citename{Grice, }1957]{Grice57}
H.~P. Grice,
\newblock ``Meaning,''
\newblock {\em Philosophical Review}, 66:377--388, 1957.

\bibitem[\protect\citename{Grosjean, }1983]{Grosjean83:l}
Fran\c{c}ois Grosjean,
\newblock ``How Long is the Sentence? {P}redicting Prosody in the On-line
  Processing of Language,''
\newblock {\em Linguistics}, 21(3):501--529, 1983.

\bibitem[\protect\citename{Gross {\em et~al.}, }1993]{Gross-etal93:tr}
Derek Gross, James Allen, and David Traum,
\newblock ``The {Trains} 91 Dialogues,''
\newblock Trains Technical Note 92-1, Department of Computer Science,
  University of Rochester, June 1993.

\bibitem[\protect\citename{Grosz and Hirschberg,
  }1992]{GroszHirschberg92:icslp}
Barbara Grosz and Julia Hirschberg,
\newblock ``Some Intonational Characteristics of Discourse Structure,''
\newblock In {\em Proceedings of the 2nd International Conference on Spoken
  Language Processing (ICSLP-92)}, pages 429--432, October 1992.

\bibitem[\protect\citename{Grosz and Sidner, }1986]{GroszSidner86:cl}
Barbara~J. Grosz and Candace~L. Sidner,
\newblock ``Attention, Intentions, and the Structure of Discourse,''
\newblock {\em Computational Linguistics}, 12(3):175--204, 1986.

\bibitem[\protect\citename{Grosz {\em et~al.}, }1986]{readings-in-nlp:86}
Barbara~J. Grosz, Karen {Sparck Jones}, and Bonnie~Lynn Webber, editors,
\newblock {\em Readings in Natural Language Processing},
\newblock Morgan Kaufmann Publishers, 1986.

\bibitem[\protect\citename{Halliday, }1967]{Halliday67:jl}
M.~A. Halliday,
\newblock ``Notes on Transitivity and Theme in {E}nglish: {P}art 2,''
\newblock {\em Journal of Linguistics}, 3:199--244, 1967.

\bibitem[\protect\citename{Heeman and Allen, }1994a]{HeemanAllen94:acl}
Peter Heeman and James Allen,
\newblock ``Detecting and Correcting Speech Repairs,''
\newblock In {\em Proceedings of the 32$^{th}$ Annual Meeting of the
  Association for Computational Linguistics}, pages 295--302, Las Cruces, New
  Mexico, June 1994.

\bibitem[\protect\citename{Heeman and Allen, }1994b]{HeemanAllen94:arpa}
Peter Heeman and James Allen,
\newblock ``Tagging Speech Repairs,''
\newblock In {\em Proceedings of the ARPA Human Language Technology Workshop},
  pages 187--192, Princeton, March 1994.

\bibitem[\protect\citename{Heeman, }1997]{Heeman97:cnet}
Peter~A. Heeman,
\newblock ``Spontaneous Speech Modeling of French,''
\newblock Technical report, France T\'el\'ecom -- Centre National d'Etudes
  T\'el\'ecommunications, Lannion, France, July 1997.

\bibitem[\protect\citename{Heeman and Allen, }1995a]{HeemanAllen94:tn-tools}
Peter~A. Heeman and James Allen,
\newblock ``Dialogue Transcription Tools,''
\newblock Trains Technical Note 94-1, Department of Computer Science,
  University of Rochester, March 1995,
\newblock Revised.

\bibitem[\protect\citename{Heeman and Allen, }1995b]{HeemanAllen95:tn-dialogs}
Peter~A. Heeman and James Allen,
\newblock ``The {T}rains 93 Dialogues,''
\newblock Trains Technical Note 94-2, Department of Computer Science,
  University of Rochester, March 1995.

\bibitem[\protect\citename{Heeman and Allen, }1995c]{HeemanAllen95:cdrom}
Peter~A. Heeman and James~F. Allen,
\newblock ``The {T}rains Spoken Dialog Corpus,''
\newblock {CD-ROM}, Linguistics Data Consortium, April 1995.

\bibitem[\protect\citename{Heeman and Allen, }1997a]{HeemanAllen97:eurospeech}
Peter~A. Heeman and James~F. Allen,
\newblock ``Incorporating {POS} Tagging into Language Modeling,''
\newblock In {\em Proceedings of the 5$^{th}$ European Conference on Speech
  Communication and Technology (Eurospeech)}, Rhodes, Greece, September 1997.

\bibitem[\protect\citename{Heeman and Allen, }1997b]{HeemanAllen97:acl}
Peter~A. Heeman and James~F. Allen,
\newblock ``Intonational Boundaries, Speech Repairs, and Discourse Markers:
  Modeling Spoken Dialog,''
\newblock In {\em Proceedings of the 35$^{th}$ Annual Meeting of the
  Association for Computational Linguistics}, pages 254--261, Madrid, July
  1997.

\bibitem[\protect\citename{Heeman and Damnati, }1997]{HeemanDamnati97:asru}
Peter~A. Heeman and {G\'eraldine} Damnati,
\newblock ``Deriving Phrase-based Language Models,''
\newblock In {\em {IEEE} Workshop on Speech Recognition and Understanding},
  Santa Barbara, California, December 1997.

\bibitem[\protect\citename{Heeman and Hirst, }1995]{HeemanHirst95:cl}
Peter~A. Heeman and Graeme Hirst,
\newblock ``Collaborating on Referring Expressions,''
\newblock {\em Computational Linguistics}, 21(3):351--382, 1995,
\newblock Also published as Revised TR 435, Computer Science Dept., U.
  Rochester, April 1995.

\bibitem[\protect\citename{Heeman and Loken-Kim, }1995]{HeemanLokenkim95:ieice}
Peter~A. Heeman and {Kyung-ho} Loken-Kim,
\newblock ``Using Structural Information to Detect Speech Repairs,''
\newblock In {\em Institute of Electronics, Information and Communication
  Engineers (IEICE), TR SP95-91}, Japan, December 1995.

\bibitem[\protect\citename{Heeman {\em et~al.},
  }1996]{HeemanLokenkimAllen96:icslp}
Peter~A. Heeman, {Kyung-ho} Loken-Kim, and James~F. Allen,
\newblock ``Combining the Detection and Correction of Speech Repairs,''
\newblock In {\em Proceedings of the 4rd International Conference on Spoken
  Language Processing (ICSLP-96)}, pages 358--361, Philadephia, October 1996,
\newblock Also appears in {\em International Symposium on Spoken Dialogue},
  1996, pages 133-136.

\bibitem[\protect\citename{Hindle, }1983]{Hindle83:acl}
Donald Hindle,
\newblock ``Deterministic Parsing of Syntactic Non-fluencies,''
\newblock In {\em Proceedings of the 21$^{st}$ Annual Meeting of the
  Association for Computational Linguistics}, pages 123--128, 1983.

\bibitem[\protect\citename{Hirschberg, }1991]{Hirschberg91:eurospeech}
Julia Hirschberg,
\newblock ``Using Text Analysis to Predict Intonational Boundaries,''
\newblock In {\em Proceedings of the 2nd European Conference on Speech
  Communication and Technology (Eurospeech)}, 1991.

\bibitem[\protect\citename{Hirschberg and Litman,
  }1987]{HirschbergLitman87:acl}
Julia Hirschberg and Diane Litman,
\newblock ``Now Let's Talk about {\em Now}: Identifying Cue Phrases
  Intonationally,''
\newblock In {\em Proceedings of the 25$^{th}$ Annual Meeting of the
  Association for Computational Linguistics}, pages 163--171, Stanford,
  California, 1987.

\bibitem[\protect\citename{Hirschberg and Litman, }1993]{HirschbergLitman93:cl}
Julia Hirschberg and Diane Litman,
\newblock ``Empirical Studies on the Disambiguation of Cue Phrases,''
\newblock {\em Computational Linguistics}, 19(3):501--530, 1993.

\bibitem[\protect\citename{Hirschberg and Pierreumbert,
  }1986]{HirschbergPierrehumbert86:acl}
Julia Hirschberg and Janet Pierreumbert,
\newblock ``The Intonational Structuring of Discourse,''
\newblock In {\em Proceedings of the 24$^{th}$ Annual Meeting of the
  Association for Computational Linguistics}, pages 136--144, 1986.

\bibitem[\protect\citename{Jelinek, }1985]{Jelinek85}
F.~Jelinek,
\newblock ``Self-organized Language Modeling for Speech Recognition,''
\newblock Technical report, IBM T.J.~Watson Research Center, Continuous Speech
  Recognition Group, Yorktown Heights, NY, 1985.

\bibitem[\protect\citename{Jelinek and Mercer, }1980]{JelinekMercer80}
F.~Jelinek and R.~L. Mercer,
\newblock ``Interpolated Estimation of Markov Source Paramaters from Sparse
  Data,''
\newblock In {\em Proceedings, Workshop on Pattern Recognition in Practice},
  pages 381--397, Amsterdam, 1980.

\bibitem[\protect\citename{Johansson {\em et~al.}, }1986]{Johansson-etal86:lob}
S.~Johansson, E.~Atwell, R.~Garside, and G.~Leech,
\newblock ``The Tagged LOB Corpus: Users' manual,''
\newblock ICAME, The Norwegian Computing Centre for the Humanities, Bergen
  University, Norway, 1986.

\bibitem[\protect\citename{Joshi and Srinivas, }1994]{JoshiSrinivas94:coling}
Aravind Joshi and B.~Srinivas,
\newblock ``Disambiguation of Super Parts of Speech (or Supertags),''
\newblock In {\em Proceedings of the 15$^{th}$ International Conference on
  Computational Linguistics (COLING)}, Kyoto, Japan, 1994.

\bibitem[\protect\citename{Junkawitsch {\em et~al.},
  }1996]{Junkawitsch-etal96:icslp}
J.~Junkawitsch, L.~Neubauer, H.~H{\"oge}, and G.~Ruske,
\newblock ``A New Keyword Spotting Algorithm with Pre-calculated Optimal
  Thresholds,''
\newblock In {\em Proceedings of the 4rd International Conference on Spoken
  Language Processing (ICSLP-96)}, Philadelphia, October 1996.

\bibitem[\protect\citename{Katz, }1987]{Katz87:assp}
Slava~M. Katz,
\newblock ``Estimation of Probabilities from Sparse Data for the Language Model
  Component of a Speech Recognizer,''
\newblock {\em {IEEE} Transactions on Acoustics, Speech, and Signal
  Processing}, pages 400--401, March 1987.

\bibitem[\protect\citename{Kikui and Morimoto, }1994]{KikuiMorimoto94:icslp}
Gen-ichiro Kikui and Tsuyoshi Morimoto,
\newblock ``Similarity-based Identification of Repairs in Japanese Spoken
  Language,''
\newblock In {\em Proceedings of the 3rd International Conference on Spoken
  Language Processing (ICSLP-94)}, pages 915--918, 1994.

\bibitem[\protect\citename{Kneser and Ney, }1993]{KneserNey93:eurospeech}
Reinhard Kneser and Hermann Ney,
\newblock ``Improved Clustering Techniques for Class-Based Statistical Language
  Modelling,''
\newblock In {\em Proceedings of the 3rd European Conference on Speech
  Communication and Technology (Eurospeech)}, pages 973--976, 1993.

\bibitem[\protect\citename{Kompe {\em et~al.}, }1994]{Kompe-etal94:icassp}
R.~Kompe, A.~Batliner, A.~Kie{\ss}ing, U.~Kilian, H.~Niemann, E.~{N\"{o}th},
  and P.~Regel-Brietzmann,
\newblock ``Automatic Classification of Prosodically Marked Phrase Boundaries
  in German,''
\newblock In {\em Proceedings of the International Conference on Audio, Speech
  and Signal Processing (ICASSP)}, pages 173--176, Adelaide, 1994.

\bibitem[\protect\citename{Kompe {\em et~al.}, }1995]{Kompe-etal95:eurospeech}
R.~Kompe, A.~Kie{\ss}ling, H.~Niemann, E.~{N\"{o}th}, E.~G.
  Schukat-Talamazzini, A.~Zottmann, and A.~Batliner,
\newblock ``Prosodic Scoring of Word Hypotheses Graphs,''
\newblock In {\em Proceedings of the 4$^{th}$ European Conference on Speech
  Communication and Technology (Eurospeech)}, pages 1333--1336, Madrid, 1995.

\bibitem[\protect\citename{Kurohashi and Nagao, }1992]{KurohashiNagao92:coling}
Sadao Kurohashi and Makoto Nagao,
\newblock ``Dynamic Programming Method for Analyzing Conjunctive Structures in
  {J}apanese,''
\newblock In {\em Proceedings of the 14$^{th}$ International Conference on
  Computational Linguistics (COLING)}, 1992.

\bibitem[\protect\citename{Labov, }1966]{Labov66}
William Labov,
\newblock ``On the Grammaticality of Everyday Speech,''
\newblock Paper presented at the Linguistic Society of America Annual Meeting,
  1966.

\bibitem[\protect\citename{Lavie {\em et~al.}, }1997]{Lavie-etal97}
Alon Lavie, Donna Gates, Noah Coccaro, and Lori Levin,
\newblock ``Input Segmentation of Spontaneous Speech in {JANUS}: a
  Speech-to-Speech Translation System,''
\newblock In Elisabeth Maier, Marion Mast, and Susann LuperFoy, editors, {\em
  Dialogue Processing in Spoken Language Systems}, Lecture Notes in Artificial
  Intelligence. Springer-Verlag, Heidelberg, 1997.

\bibitem[\protect\citename{Lea, }1980]{Lea80}
W.~Lea,
\newblock ``Prosodic Aids to Speech Recognition,''
\newblock In W.~Lea, editor, {\em Trends in Speech Recognition}. Prentice-Hall,
  Englewood Cliffs, NJ, 1980.

\bibitem[\protect\citename{Levelt, }1983]{Levelt83:cog}
Willem J.~M. Levelt,
\newblock ``Monitoring and Self-Repair in Speech,''
\newblock {\em Cognition}, 14:41--104, 1983.

\bibitem[\protect\citename{Lickley and Bard, }1992]{LickleyBard92:icslp}
R.~J. Lickley and E.~G. Bard,
\newblock ``Processing Disfluent Speech: Recognizing Disfluency before Lexical
  Access,''
\newblock In {\em Proceedings of the 2nd International Conference on Spoken
  Language Processing (ICSLP-92)}, pages 935--938, October 1992.

\bibitem[\protect\citename{Lickley {\em et~al.},
  }1991]{Lickley-etal91:eurospeech}
R.~J. Lickley, R.~C. Shillcock, and E.~G. Bard,
\newblock ``Processing Disfluent Speech: How and When are Disfluencies
  Found?,''
\newblock In {\em Proceedings of the 2nd European Conference on Speech
  Communication and Technology (Eurospeech)}, pages 1499--1502, Genova, Italy,
  September 1991.

\bibitem[\protect\citename{Lickley and Bard, }1996]{LickleyBard96:icslp}
Robin.~J. Lickley and Ellen~Gurman Bard,
\newblock ``On not Recognizing Disfluencies in Dialogue,''
\newblock In {\em Proceedings of the 4rd International Conference on Spoken
  Language Processing (ICSLP-96)}, pages 1876--1879, October 1996.

\bibitem[\protect\citename{Litman, }1996]{Litman96:jair}
Diane~J. Litman,
\newblock ``Cue Phrase Classification Using Machine Learning,''
\newblock {\em Journal of Artificial Intelligence Research}, 5:53--94, 1996.

\bibitem[\protect\citename{Litman and Allen, }1987]{LitmanAllen87}
Diane~J. Litman and James~F. Allen,
\newblock ``A Plan Recognition Model for Subdialogues in Conversations,''
\newblock {\em Cognitive Science}, 11(2):163--200, April--June 1987.

\bibitem[\protect\citename{MADCOW, }1992]{Madcow92:snlp}
MADCOW,
\newblock ``Multi-site Data Collection for a Spoken Language Corpus,''
\newblock In {\em Proceedings of the DARPA Workshop on Speech and Natural
  Language Processing}, pages 7--14, February 1992.

\bibitem[\protect\citename{Magerman, }1994]{Magerman94:thesis}
David~M. Magerman,
\newblock ``Natural Language Parsing as Statistical Pattern Recognition,''
\newblock Doctoral dissertation, Department of Computer Science, Stanford
  University, 1994.

\bibitem[\protect\citename{Marcus and Hindle, }1990]{MarcusHindle90:cmsp}
Mitchell Marcus and Donald Hindle,
\newblock ``Description Theory and Intonation Boundaries,''
\newblock In Gerry T.~M. Altmann, editor, {\em Cognitive Models of Speech
  Processing}, pages 483--512. MIT Press, 1990.

\bibitem[\protect\citename{Marcus {\em et~al.}, }1993]{Marcus-etal93:cl}
Mitchell~P. Marcus, Beatrice Santorini, and Mary~Ann Marcinkiewicz,
\newblock ``Building a Large Annotated Corpus of English: The {P}enn
  {T}reebank,''
\newblock {\em Computational Linguistics}, 19(2):313--330, 1993.

\bibitem[\protect\citename{Martin and Strange, }1968]{MartinStrange68}
J.~G. Martin and W.~Strange,
\newblock ``The Perception of Hestitation in Spontaneous Speech,''
\newblock {\em Perception and Psychophysics}, 53:1--15, 1968.

\bibitem[\protect\citename{Mast {\em et~al.}, }1996]{Mast-etal96:icslp}
M.~Mast, R.~Kompe, S.~Harbeck, A.~Kie{\ss}ling, H.~Niemann, E.~{N\"{o}th},
  E.~G. Schukat-Talamazzini, and V.~Warnke,
\newblock ``Dialog Act Classification with the Help of Prosody,''
\newblock In {\em Proceedings of the 4rd International Conference on Spoken
  Language Processing (ICSLP-96)}, Philadelphia, October 1996.

\bibitem[\protect\citename{Meteer and Iyer, }1996]{MeteerIyer96:emnl}
M.~Meteer and R.~Iyer,
\newblock ``Modeling Conversational Speech for Speech Recognition,''
\newblock In {\em Proceedings of the Conference on Emphirical Methods in
  Natural Language Processing}, Philadelphia, May 1996.

\bibitem[\protect\citename{Mikheev, }1996]{Mikheev96:acl}
Andrei Mikheev,
\newblock ``Unsupervised Learning of Word-Category Guessing Rules,''
\newblock In {\em Proceedings of the 34$^{th}$ Annual Meeting of the
  Association for Computational Linguistics}, June 1996.

\bibitem[\protect\citename{Nakajima and Allen,
  }1993]{NakajimaAllen93:phonetica}
Shin'ya Nakajima and James~F. Allen,
\newblock ``A Study on Prosody and Discourse Structure in Cooperative
  Dialogues,''
\newblock {\em Phonetica}, 50(3):197--210, 1993.

\bibitem[\protect\citename{Nakatani and Hirschberg,
  }1994]{NakataniHirschberg94:jasa}
Chistine~H. Nakatani and Julia Hirschberg,
\newblock ``A Corpus-based Study of Repair Cues in Spontaneous Speech,''
\newblock {\em Journal of the Acoustical Society of America}, 95(3):1603--1616,
  1994.

\bibitem[\protect\citename{Niesler and Woodland,
  }1996]{NieslerWoodland96:icassp}
T.~R. Niesler and P.~C. Woodland,
\newblock ``A Variable-Length Category-based n-gram Language Model,''
\newblock In {\em Proceedings of the International Conference on Audio, Speech
  and Signal Processing (ICASSP)}, pages 164--167, 1996.

\bibitem[\protect\citename{Nooteboom, }1980]{Nooteboom80}
S.~G. Nooteboom,
\newblock ``Speaking and Unspeaking: Detection and Correction of Phonological
  and Lexial Errors,''
\newblock In Victoria~A. Fromkin, editor, {\em Errors in Linguistic
  Performance}. Academic Press, New York, 1980.

\bibitem[\protect\citename{O'Shaughnessy, }1992]{Oshaughnessy92:icslp}
Douglas O'Shaughnessy,
\newblock ``Analysis of False Starts in Spontaneous Speech,''
\newblock In {\em Proceedings of the 2nd International Conference on Spoken
  Language Processing (ICSLP-92)}, pages 931--934, October 1992.

\bibitem[\protect\citename{Ostendorf {\em et~al.}, }1993]{Ostendorf-etal93:csl}
M.~Ostendorf, C.~Wightman, and N.~Veilleux,
\newblock ``Parse Scoring with Prosodic Information: an Analysis/Synthesis
  Approach,''
\newblock {\em Computer Speech and Language}, 7(2), 1993.

\bibitem[\protect\citename{Oviatt, }1995]{Oviatt95:csl}
Sharon Oviatt,
\newblock ``Predicting Spoken Disfluencies during Human-Computer Interaction,''
\newblock {\em Computer Speech and Language}, 9:19--35, 1995.

\bibitem[\protect\citename{Pierrehumbert, }1980]{Pierrehumbert80:thesis}
J.~B. Pierrehumbert,
\newblock ``The Phonology and Phonetics of English Intonation,''
\newblock Doctoral dissertation, Massachusetts Institute of Technology, 1980.

\bibitem[\protect\citename{Pierrehumbert and Hirschberg,
  }1990]{PierrehumbertHirschberg90:iic}
Janet Pierrehumbert and Julia Hirschberg,
\newblock ``The Meaning of Intonational Contours in the Interpretation of
  Discourse,''
\newblock In Philip~R. Cohen, Jerry Morgan, and Martha~E. Pollack, editors,
  {\em Intentions in Communication}, SDF Benchmark Series, pages 271--311. MIT
  Press, 1990.

\bibitem[\protect\citename{Pitrelli {\em et~al.}, }1994]{Pitrelli-etal94:icslp}
John~F. Pitrelli, Mary~E. Beckman, and Julia Hirschberg,
\newblock ``Evaluation of Prosodic Transcription Labeling Reliability in the
  {ToBI} Framework,''
\newblock In {\em Proceedings of the 3rd International Conference on Spoken
  Language Processing (ICSLP-94)}, Yokohama, September 1994.

\bibitem[\protect\citename{Pollack {\em et~al.},
  }1982]{PollackHirschbergWebber82:tr}
M.~E. Pollack, J.~Hirschberg, and B.~Webber,
\newblock ``User Participation in the Reasoning Processes of Expert Systems,''
\newblock Technical Note MS-CIS-82-9, University of Pennsylvannia, July 1982.

\bibitem[\protect\citename{Price {\em et~al.}, }1991]{Price-etal91:jasa}
P.~J. Price, M.~Ostendorf, S.~Shattuck-Hufnagel, and C.~Fong,
\newblock ``The Use of Prosody in Syntactic Disambiguation,''
\newblock {\em Journal of the Acoustical Society of America}, 90(6):2956--2970,
  December 1991.

\bibitem[\protect\citename{Quinlan, }1993]{Quinlan93:book}
J.~R. Quinlan,
\newblock {\em C4.5: Programs for Machine Learning},
\newblock Morgan Kaufman, San Mateo, California, 1993.

\bibitem[\protect\citename{Rabiner and Juang, }1993]{RabinerJuang93}
Lawrence~R. Rabiner and Biing-Hwang Juang,
\newblock {\em Fundamentals of Speech Recognition},
\newblock Prentice Hall, Englewood Cliffs, NJ, 1993.

\bibitem[\protect\citename{Reichman-Adar, }1984]{Reichmanadar84:ai}
Rachel Reichman-Adar,
\newblock ``Extended Person-Machine Interface,''
\newblock {\em Artificial Intelligence}, 22:157--218, 1984.

\bibitem[\protect\citename{Rosenfeld, }1995]{Rosenfeld95:arpa}
R.~Rosenfeld,
\newblock ``The {CMU} Statistical Language Modeling Toolkit and its Use in the
  1994 {ARPA} {CSR} Evaluation,''
\newblock In {\em Proceedings of the {ARPA} Spoken Language Systems Technology
  Workshop}, San Mateo, California, 1995. Morgan Kaufmann.

\bibitem[\protect\citename{Rosenfeld {\em et~al.},
  }1996]{Rosenfeld-etal96:icslp}
Roni Rosenfeld, Rajeev Agarwal, Bill Byrne, Rukmini Iyer, Mark Liberman, Liz
  Shriberg, Jack Unverferth, Dimitra Vergyri, and Enrique Vidal,
\newblock ``Error Analysis and Disfluency Modeling in the Switchboard Domain:
  Project Team Report,''
\newblock In {\em Proceedings of the 4rd International Conference on Spoken
  Language Processing (ICSLP-96)}, 1996.

\bibitem[\protect\citename{Sadek {\em et~al.}, }1997]{Sadek-etal97:ijcai}
M.~D. Sadek, P.~Bretier, and F.~Panaget,
\newblock ``{ARTIMIS}: Natural Dialogue Meets Rational Agency,''
\newblock In {\em Proceedings of the International Joint Conference on
  Artificial Intelligence (IJCAI~'97)}, Japan, August 1997.

\bibitem[\protect\citename{Sadek {\em et~al.}, }1996]{Sadek-etal96:icslp}
M.~D. Sadek, A.~Ferrieux, A.~Cozannet, P.~Bretier, F.~Panaget, and J.~Simonin,
\newblock ``Effective Human-Computer Cooperative Spoken Dialogue: The {AGS}
  Decomonstrator,''
\newblock In {\em Proceedings of the 4rd International Conference on Spoken
  Language Processing (ICSLP-96)}, Philadelphia, October 1996.

\bibitem[\protect\citename{Sagawa {\em et~al.}, }1994]{Sagawa-etal94:coling}
Yuji Sagawa, Noboru Ohnishi, and Noboru Sugie,
\newblock ``A Parser Coping with Self-Repaired Japanese Utterances and Large
  Corpu-Based Evaluation,''
\newblock In {\em Proceedings of the 15$^{th}$ International Conference on
  Computational Linguistics (COLING)}, pages 593--597, 1994.

\bibitem[\protect\citename{Santorini, }1990]{Santorini90:tr}
Beatrice Santorini,
\newblock ``Part-of-Speech Tagging Guidelines for the {P}enn Treebank
  Project,''
\newblock Technical report ms-cis-90-47, Department of Computer and Information
  Science, University of Pennsylvania, 1990.

\bibitem[\protect\citename{Schegloff {\em et~al.},
  }1977]{Schegloff-etal77:lang}
Emanuel~A. Schegloff, Gail Jefferson, and Harvey Sacks,
\newblock ``The Preference for Self-Correction in the Organization of Repair in
  Conversation,''
\newblock {\em Language}, 53:361--382, 1977.

\bibitem[\protect\citename{Schiffrin, }1987]{Schiffrin87:book}
Deborah Schiffrin,
\newblock {\em Discourse Markers},
\newblock Cambridge University Press, New York, 1987.

\bibitem[\protect\citename{Searle, }1969]{Searle69:book}
J.~R. Searle,
\newblock {\em Speech Acts: An Essay in the Philosophy of Language},
\newblock Cambridge University Press, Cambridge, 1969.

\bibitem[\protect\citename{Seligman {\em et~al.},
  }1997]{SeligmanHosakaSinger97}
Mark Seligman, Junko Hosaka, and Harald Singer,
\newblock ```Pause Units' and Analysis of Spontaneous Japanese Dialogues:
  Preliminary Studies,''
\newblock In Elisabeth Maier, Marion Mast, and Susann LuperFoy, editors, {\em
  Dialogue Processing in Spoken Language Systems}, Lecture Notes in Artificial
  Intelligence. Springer-Verlag, Heidelberg, 1997.

\bibitem[\protect\citename{Shabes {\em et~al.}, }1988]{Shabes-etal88:coling}
Yves Shabes, Anne Abeill{\'e}, and Aravind~K. Joshi,
\newblock ``Parsing Strategies with `Lexicalized' Grammars: Application to Tree
  Adjoining Grammars,''
\newblock In {\em Proceedings of the 12$^{th}$ International Conference on
  Computational Linguistics (COLING)}, Budapest, August 1988.

\bibitem[\protect\citename{Shriberg and Lickley,
  }1993]{ShribergLickley93:phonetica}
Elizabeth~E. Shriberg and Robin~J. Lickley,
\newblock ``Intonation of Clause-Internal Filled Pauses,''
\newblock {\em Phonetica}, 50(3):172--179, 1993.

\bibitem[\protect\citename{Shriberg, }1994]{Shriberg94:thesis}
Elizabeth~Ellen Shriberg,
\newblock ``Preliminaries to a Theory of Speech Disfluencies,''
\newblock Doctoral dissertion, University of California at Berkeley, 1994.

\bibitem[\protect\citename{Sidner, }1985]{Sidner85:ci}
Candace~L. Sidner,
\newblock ``Plan Parsing for Intended Response Recognition in Discourse,''
\newblock {\em Computational Intelligence}, 1(1):1--10, 1985.

\bibitem[\protect\citename{Silverman {\em et~al.},
  }1992]{Silverman-etal92:icslp}
K.~Silverman, M.~Beckman, J.~Pitrelli, M.~Ostendorf, C.~Wightman, P.~Price,
  J.~Pierrehumbert, and J.~Hirschberg,
\newblock ``{ToBI}: A Standard for Labelling {E}nglish Prosody,''
\newblock In {\em Proceedings of the 2nd International Conference on Spoken
  Language Processing (ICSLP-92)}, pages 867--870, 1992.

\bibitem[\protect\citename{Siu and Ostendorf, }1996]{SiuOstendorf96:icslp}
{Man-hung} Siu and Mari Ostendorf,
\newblock ``Modeling Disfluencies in Conversational Speech,''
\newblock In {\em Proceedings of the 4rd International Conference on Spoken
  Language Processing (ICSLP-96)}, pages 382--391, 1996.

\bibitem[\protect\citename{Srinivas, }1996]{Srinivas96:icslp}
B.~Srinivas,
\newblock ````{A}lmost Parsing'' Techniques for Language Modeling,''
\newblock In {\em Proceedings of the 4rd International Conference on Spoken
  Language Processing (ICSLP-96)}, pages 1169--1172, 1996.

\bibitem[\protect\citename{Steedman, }1990]{Steedman90:cmsp}
Mark~J. Steedman,
\newblock ``Syntax and Intonational Structure in a Combinatory Grammar,''
\newblock In Gerry T.~M. Altmann, editor, {\em Cognitive Models of Speech
  Processing}, pages 457--482. MIT Press, 1990.

\bibitem[\protect\citename{Stolcke and Shriberg,
  }1996a]{StolckeShriberg96:icslp}
Andreas Stolcke and Elizabeth Shriberg,
\newblock ``Automatic Linguistic Segmentation of Conversational Speech,''
\newblock In {\em Proceedings of the 4rd International Conference on Spoken
  Language Processing (ICSLP-96)}, October 1996.

\bibitem[\protect\citename{Stolcke and Shriberg,
  }1996b]{StolckeShriberg96:icassp}
Andreas Stolcke and Elizabeth Shriberg,
\newblock ``Statistical Language Modeling for Speech Disfluencies,''
\newblock In {\em Proceedings of the International Conference on Audio, Speech
  and Signal Processing (ICASSP)}, May 1996.

\bibitem[\protect\citename{Takagi and Itahashi, }1996]{TakagiItahashi96:icslp}
Kazuyuki Takagi and Shuichi Itahashi,
\newblock ``Segmentation of Spoken Dialogue by Interjection, Disfluent
  Utterances and Pauses,''
\newblock In {\em Proceedings of the 4rd International Conference on Spoken
  Language Processing (ICSLP-96)}, pages 693--697, Philadelphia, October 1996.

\bibitem[\protect\citename{Traum and Heeman, }1997]{TraumHeeman97:chapter}
David~R. Traum and Peter~A. Heeman,
\newblock ``Utterance Units in Spoken Dialogue,''
\newblock In Elisabeth Maier, Marion Mast, and Susann LuperFoy, editors, {\em
  Dialogue Processing in Spoken Language Systems}, Lecture Notes in Artificial
  Intelligence, pages 125--140. Springer-Verlag, Heidelberg, 1997.

\bibitem[\protect\citename{Traum {\em et~al.}, }1996]{Traum-etal96:jes}
David~R. Traum, L.~K. Schubert, M.~Poesio, N.~G. Martin, M.~Light, C.~H. Hwang,
  P.~Heeman, G.~Ferguson, and J.~F. Allen,
\newblock ``Knowledge Representation in the {TRAINS}-93 Conversation System,''
\newblock {\em International Journal of Expert Systems}, 9(1):173--223, 1996,
\newblock Also published as Trains TN 96-4 and TR 633, Computer Science Dept.,
  U. Rochester, August 1996.

\bibitem[\protect\citename{Wahlster, }1993]{Wahlster93}
W.~Wahlster,
\newblock ``Verbmobil --- Translation of Face-to-Face Dialogs,''
\newblock In {\em Proceedings of the 3rd European Conference on Speech
  Communication and Technology (Eurospeech)}, pages 29--38, Berlin, 1993.

\bibitem[\protect\citename{Walker, }1993]{Walker93:thesis}
Marilyn~A. Walker,
\newblock ``Informational Redundancy and Resource Bounds in Dialogue,''
\newblock Doctoral dissertion, {Institute for Research in Cognitive Science
  report IRCS-93-45}, University of Pennsylvania, December 1993.

\bibitem[\protect\citename{Wang and Hirschberg, }1992]{WangHirschberg92:csl}
Michelle~Q. Wang and Julia Hirschberg,
\newblock ``Automatic Classification of Intonational Phrase Boundaries,''
\newblock {\em Computer Speech and Language}, 6:175--196, 1992.

\bibitem[\protect\citename{Ward, }1991]{Ward91:icassp}
W.~Ward,
\newblock ``Understanding Spontaneous Speech: the {P}hoenix System,''
\newblock In {\em Proceedings of the International Conference on Audio, Speech
  and Signal Processing (ICASSP)}, pages 365--367, 1991.

\bibitem[\protect\citename{Weischedel {\em et~al.},
  }1993]{Weischedel-etal93:cl}
Ralph Weischedel, Marie Meteer, Richard Schwartz, Lance Ramshaw, and Jeff
  Palmucci,
\newblock ``Coping with Ambiguity and Unknown Words through Probabilistic
  Models,''
\newblock {\em Computational Linguistics}, 19(2):359--382, 1993.

\bibitem[\protect\citename{Wightman and Ostendorf,
  }1994]{WightmanOstendorf94:ieee}
Colin~W. Wightman and Mari Ostendorf,
\newblock ``Automatic Labeling of Prosodic Patterns,''
\newblock {\em {IEEE} Transactions on Speech and Audio Processing}, October
  1994.

\bibitem[\protect\citename{Wightman {\em et~al.}, }1992]{Wightman-etal92:jasa}
Colin~W. Wightman, Stefanie Shattuck-Hufnagel, Mari Ostendorf, and Patti~J.
  Price,
\newblock ``Segmental Durations in the Vicinity of Prosodic Phrase
  Boundaries,''
\newblock {\em Journal of the Acoustical Society of America}, 91(3):1707--1717,
  March 1992.

\bibitem[\protect\citename{Young and Matessa, }1991]{YoungMatessa91:eurospeech}
Sheryl~R. Young and Michael Matessa,
\newblock ``Using Pragmatic and Semantic Knowledge to Correct Parsing of Spoken
  Language Utterances,''
\newblock In {\em Proceedings of the 2nd European Conference on Speech
  Communication and Technology (Eurospeech)}, pages 223--227, Genova, Italy,
  September 1991.

\bibitem[\protect\citename{Zeppenfeld {\em et~al.},
  }1997]{Zeppenfeld-etal97:icassp}
Torsten Zeppenfeld, Michael Finke, Klaus Ries, Martin Westphal, and Alex
  Waibel,
\newblock ``Recognition of Conversational Telephone Speech Using the Janus
  Speech Engine,''
\newblock In {\em Proceedings of the International Conference on Audio, Speech
  and Signal Processing (ICASSP)}, Munich, April 1997.

\end{thebibliography}
\end{document}